\documentclass[conference]{IEEEtran}
\usepackage{times}
\usepackage{amsmath}
\usepackage{tikz}
\usepackage{listings}
\usepackage{xcolor}
\usepackage{algorithm}
\usepackage{algorithmic}
\usepackage{svg}
\usetikzlibrary{shapes.geometric, arrows.meta, positioning, fit, backgrounds, calc}

\usepackage[numbers]{natbib}
\usepackage{multicol}
\usepackage[bookmarks=true]{hyperref}
\usepackage{booktabs}
\usepackage{multirow}
\usepackage{graphicx}
\usepackage{subcaption}
\usepackage[font=footnotesize]{caption}
\usepackage{float} 
\usepackage{grffile}
\usepackage{amssymb}
\usepackage{placeins}

  \usepackage{xcolor}     % for colored checkmarks and crosses
  \usepackage[table]{xcolor}
  \usepackage{amssymb}    % for \checkmark symbol

\usepackage{pifont}
\usepackage{xcolor}

\usepackage{float}

\newif\ifnotes
\notestrue   % set \notesfalse to hide all notes
\ifnotes
  \newcommand{\authnote}[3]{\textcolor{#2}{[#1: #3]}}
\else
  \newcommand{\authnote}[3]{}
\fi

\definecolor{good}{RGB}{0,120,0}
\definecolor{bad}{RGB}{160,0,0}

\newcommand{\yes}{\textcolor{good}{\ding{51}}} % ✓
\newcommand{\no}{\textcolor{bad}{\ding{55}}}  % ✗

\newcommand{\methodname}{RoboPlayground}

\usepackage{titlesec}
\titleformat{\subsubsection}
{\normalfont\bfseries}
{\thesubsubsection}
{1em}
{}
\usepackage{etoc}
\usepackage{hyperref}
\IEEEoverridecommandlockouts
\begin{document}

% paper title
% \title{\includegraphics[width=2.1cm]{assets/logo.jpeg} \methodname: Structured Physical Domains for Robotic Manipulation}
\title{\methodname: Democratizing Robotic Evaluation through Structured Physical Domains}
% You will get a Paper-ID when submitting a pdf file to the conference system
% \author{Author Names Omitted for Anonymous Review. Paper-ID 97}

\author{
\thanks{Correspondence to: yiruwang@cs.washington.edu, carterung@gmail.com}
\authorblockN{
Yi Ru Wang\authorrefmark{1}\authorrefmark{3}\authorrefmark{4},
Carter Ung\authorrefmark{1}\authorrefmark{3},
Evan Gubarev\authorrefmark{3},
Christopher Tan\authorrefmark{3},
Siddhartha Srinivasa\authorrefmark{2}\authorrefmark{3},
Dieter Fox\authorrefmark{2}\authorrefmark{3}\authorrefmark{4}
}

\authorblockA{
\authorrefmark{1}Equal contribution \hspace{1em}
\authorrefmark{2}Equal advising \\
\authorrefmark{3}University of Washington, Seattle, WA, USA \\
\authorrefmark{4}Allen Institute for Artificial Intelligence, Seattle, WA, USA
}
}

% avoiding spaces at the end of the author lines is not a problem with
% conference papers because we don't use \thanks or \IEEEmembership

% for over three affiliations, or if they all won't fit within the width
% of the page, use this alternative format:
% 
%\author{\authorblockN{Michael Shell\authorrefmark{1},
%Homer Simpson\authorrefmark{2},
%James Kirk\authorrefmark{3}, 
%Montgomery Scott\authorrefmark{3} and
%Eldon Tyrell\authorrefmark{4}}
%\authorblockA{\authorrefmark{1}School of Electrical and Computer Engineering\\
%Georgia Institute of Technology,
%Atlanta, Georgia 30332--0250\\ Email: mshell@ece.gatech.edu}
%\authorblockA{\authorrefmark{2}Twentieth Century Fox, Springfield, USA\\
%Email: homer@thesimpsons.com}
%\authorblockA{\authorrefmark{3}Starfleet Academy, San Francisco, California 96678-2391\\
%Telephone: (800) 555--1212, Fax: (888) 555--1212}
%\authorblockA{\authorrefmark{4}Tyrell Inc., 123 Replicant Street, Los Angeles, California 90210--4321}}

\maketitle

% \begin{abstract}
% Developing robot policies that understand and execute semantically meaningful manipulation tasks requires large, diverse, and precisely specified task suites, yet generating such tasks is still largely manual. We propose a pipeline that transforms natural-language instructions into executable block manipulation tasks with grounded scenes and programmatically derived semantic and geometric success conditions. The pipeline defines a structured task space that captures colors, letters, symbolic relations, multi-step goals, and combinatorial variations, which enables both human-authored and programmatically generated task families. Through a series of demonstrations, we show that the system reliably interprets language instructions, generates diverse and complex scenes, and interfaces with a range of manipulation policies for controlled evaluation. By decoupling task specification from manual environment design, this work enables scalable experimentation in structured tabletop manipulation and creates a foundation for research on semantic evaluation, generalization, and sim-to-real transfer.
% \end{abstract}

\begin{abstract}
% \ssnote{Concerns we need to mitigate - 0) why not use a structured representation like PDDL with a language front-end? why do we need to re-invent a bunch of things that are already known at the risk of making mistakes? 1) there's a lot of work that uses language instead of code. what's our unique take? 2) if we rely heavily on LLMs how do we address concerns around reproducibility, determinism, and verification?, 3) our evaluation domain is too narrow.}
Evaluation of robotic manipulation systems has largely relied on fixed benchmarks authored by a small number of experts, where task instances, constraints, and success criteria are predefined and difficult to extend. This paradigm limits who can shape evaluation and obscures how policies respond to user-authored variations in task intent, constraints, and notions of success. We argue that evaluating modern manipulation policies requires reframing evaluation as a language-driven process over structured physical domains. We present \methodname, a framework that enables users to author executable manipulation tasks using natural language within a structured physical domain. Natural language instructions are compiled into reproducible task specifications with explicit asset definitions, initialization distributions, and success predicates. Each instruction defines a structured family of related tasks, enabling controlled semantic and behavioral variation while preserving executability and comparability. We instantiate \methodname~in a structured block manipulation domain and evaluate it along three axes. A user study shows that the language-driven interface is easier to use and imposes lower cognitive workload than programming-based and code-assist baselines. Evaluating learned policies on language-defined task families reveals generalization failures that are not apparent under fixed benchmark evaluations. Finally, we show that task diversity scales with contributor diversity rather than task count alone, enabling evaluation spaces to grow continuously through crowd-authored contributions. Project Page: \href{https://roboplayground.github.io}{roboplayground.github.io}
\end{abstract}

\IEEEpeerreviewmaketitle

\section{Introduction}
\begin{figure*}[t]
    \centering
    \includegraphics[width=\textwidth]{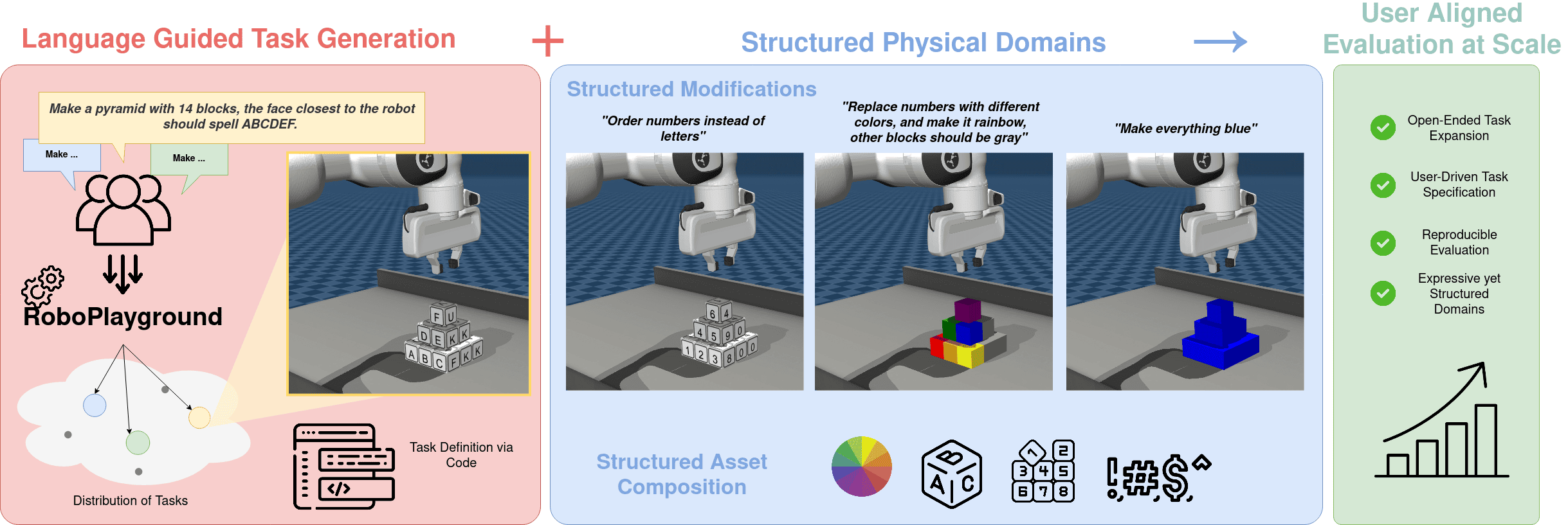}
    \caption{
    \textbf{Language-Guided Task Generation in Structured Physical Domains.}
    Natural language instructions are compiled into executable task templates within \textsc{RoboPlayground}, enabling open-ended task generation while preserving structure and control.
    Structured physical domains support systematic steering through controlled variations in task semantics, constraints, and asset composition (e.g., symbols, colors, and ordering).
    This combination enables user-aligned evaluation at scale that is expressive, reproducible, and continuously extensible.
    }
    \label{fig:roboplayground_overview}
\end{figure*}
Who gets to decide what it means for a robot to be competent? Today, robotic manipulation systems are evaluated almost exclusively through benchmarks designed by a small number of experts. These benchmarks specify fixed task instances, success conditions, and evaluation protocols, implicitly encoding which behaviors matter and which variations are worth testing. While this paradigm has driven substantial progress, it centralizes control over evaluation and limits both who can define evaluation tasks and what questions can be asked about a system’s behavior.

Outside of benchmarks, competence is rarely assessed through a single task instance. Understanding is revealed through exploration and variation: tightening constraints, rephrasing goals, or modifying what counts as success after observing an execution \cite{lake2016buildingmachineslearnthink}. Language plays a central role in this process. It provides a natural interface for expressing task intent and probing variations, where changes in wording often correspond to differences in spatial relations, constraints, or success criteria. As such, language offers a powerful handle for exploring structured variation in manipulation tasks.

However, while language has been used to generate tasks or guide robot behavior, existing benchmarks do not support language-driven exploration as a first-class evaluation interface, where users can iteratively vary task intent, constraints, and success definitions in a reproducible and comparable manner \cite{wang2023gensim, wang2024robogenunleashinginfinitedata, ahn2022icanisay, liang2023codepolicieslanguagemodel}. Evaluation tasks are typically realized as fixed environment configurations with success conditions encoded procedurally in code, while natural language serves only as informal documentation or as input to the policy itself \cite{james2019rlbenchrobotlearningbenchmark, liu2023liberobenchmarkingknowledgetransfer, nasiriany2024robocasalargescalesimulationeveryday, li2024behavior1khumancenteredembodiedai, mu2021maniskillgeneralizablemanipulationskill, li2024evaluatingrealworldrobotmanipulation}. As a result, introducing new task variations or alternative notions of success requires direct intervention at the level of benchmark implementation, placing meaningful control over evaluation in the hands of domain experts and limiting accessibility for broader users.

Similar limitations have appeared in other domains. In visual reasoning, diagnostic datasets such as CLEVR reframed evaluation around controlled task generation within a structured domain \cite{johnson2016clevrdiagnosticdatasetcompositional}. By focusing on interpretable primitives rather than maximal realism, CLEVR shifted evaluation from static instances to structured families of tasks, enabling clearer attribution of failure modes. In contrast, work in natural language processing has emphasized the dynamics of evaluation over time. Dynamic benchmarking efforts such as Dynabench treat evaluation as a human-in-the-loop process, where users iteratively generate and refine examples to surface model weaknesses \cite{kiela2021dynabenchrethinkingbenchmarkingnlp}. Together, these efforts highlight two complementary principles for informative evaluation: structured task spaces that make variation interpretable, and participatory mechanisms that allow evaluation to grow continuously beyond its initial design.

Evaluating modern manipulation policies requires embracing both principles. An effective evaluation system should satisfy four key desiderata. First, it must be \textbf{accessible}, allowing users to express task intent, constraints, and success criteria using natural language without expertise in simulation internals or benchmark-specific code. Second, it should support \textbf{continuous growth}, enabling the evaluation space to expand over time through contributions from many users rather than remaining fixed. Third, it must ensure \textbf{reproducibility}, so that tasks can be precisely re-executed across models and evaluations, enabling fair comparison as the evaluation space grows. Finally, it should provide \textbf{structured control}, constraining user-authored instructions to remain executable and interpretable while enabling systematic variation and meaningful attribution of failure modes.

\textbf{We propose reframing robotic manipulation evaluation as a language-driven, user-authored process over structured physical domains, shifting evaluation from static expert-defined benchmarks to an accessible, reproducible, and continuously expanding task space.}

In this work, we present \methodname, a language-driven framework for defining robotic manipulation tasks in a structured physical domain. Natural language serves as the primary authoring interface, allowing users to specify task intent, constraints, and success conditions without interacting with benchmark-specific code. Each instruction is compiled into an executable task specification with explicit definitions of assets, initialization distributions, and success predicates, enabling tasks to be precisely re-executed across models and evaluations. Rather than yielding a single fixed task instance, each instruction defines a family of related tasks, whose systematic variations are authored and controlled by users through language within the domain’s physical structure, allowing the evaluation space to grow continuously over time while preserving controlled, comparable structure.

We instantiate this framework in a structured manipulation domain that make language-defined tasks executable and systematically variable, rather than free-form descriptions. Within this setting, we demonstrate that language-driven task variation uncovers meaningful behavioral differences and failure modes that are not exposed by fixed benchmark tasks, including sensitivity to constraint changes and brittleness with respect to success definitions.

Our evaluation demonstrates that \methodname\ achieves these goals in practice.
Through a controlled user study, we show that the language-driven task authoring interface is substantially easier to use and imposes lower cognitive workload than both programming-centric and code-assist baselines, indicating that non-expert users can reliably construct valid manipulation tasks.
Evaluating learned policies on structured families of language-defined tasks reveals systematic generalization failures that are not apparent when testing only on fixed training-distribution benchmarks, including sensitivity to semantic changes and brittleness under altered success definitions.
Finally, we show that the evaluation space defined by \methodname\ scales through contributor diversity rather than task count alone, with crowd-authored task sets exhibiting significantly greater semantic and structural coverage than tasks generated by individual authors. Together, these results suggest that language-driven, structured task generation enables evaluation that is not only more accessible, but also more diagnostic and more representative of the space of behaviors we wish manipulation policies to master.

In summary, this paper makes three contributions. (1) We introduce a language-driven evaluation framework that democratizes manipulation evaluation by making task specification accessible to non-expert users while maintaining reproducibility.
(2) We empirically demonstrate that language-defined task families reveal policy behaviors and limitations that are missed by conventional instance-based benchmarks.
(3) We show that evaluation spaces can grow continuously through user- and model-authored instructions without sacrificing comparability or scientific rigor.
\section{Related Works}

\subsection{Evaluation and Benchmarking for Robotic Manipulation}

Robotic manipulation systems are most commonly evaluated using fixed benchmark suites composed of predefined task instances, environments, and success criteria, such as RLBench \cite{james2019rlbenchrobotlearningbenchmark}, LIBERO \cite{liu2023liberobenchmarkingknowledgetransfer}, RoboCasa \cite{nasiriany2024robocasalargescalesimulationeveryday}, Behaviour-1k \cite{li2024behavior1khumancenteredembodiedai}, ManiSkill \cite{mu2021maniskillgeneralizablemanipulationskill}, Colosseum \cite{pumacay2024colosseumbenchmarkevaluatinggeneralization},  Simpler \cite{li2024evaluatingrealworldrobotmanipulation}, and RoboEval \cite{wang2025roboevalroboticmanipulationmeets}. While these benchmarks have been instrumental in standardizing evaluation, they define evaluation over a fixed and finite set of expert-authored tasks, with task structure, constraints, and success criteria encoded procedurally and not exposed for user modification. Although some benchmarks include natural language annotations or language-conditioned tasks, language is typically treated as documentation or policy input rather than as part of the executable task specification, making it difficult to introduce task variations or alternative notions of success without modifying benchmark code. Recent work has explored complementary directions for scaling evaluation: RoboArena \cite{atreya2025roboarenadistributedrealworldevaluation} democratizes who evaluates and where evaluation occurs through crowd-sourced, double-blind pairwise comparisons over unconstrained real-world tasks, while Polaris \cite{jain2025polarisscalablerealtosimevaluations} improves the fidelity and scalability of simulation-based evaluation via real-to-sim scene reconstruction, but retains fixed, expert-authored tasks and success criteria. In contrast, our work focuses on democratizing what is evaluated by enabling users to author, modify, and refine executable task specifications through language, while preserving structure, reproducibility, and comparability.

\subsection{LLM-Based Task and Environment Generation}

Recent work has explored using large language models to generate robotic tasks, environments, rewards, or curricula at scale. Systems such as GenSim \cite{wang2023gensim}, Gen2Sim \cite{katara2023gen2simscalingrobotlearning}, RoboGen \cite{wang2024robogenunleashinginfinitedata}, and AnyTask \cite{gong2026anytaskautomatedtaskdata} leverage LLMs to synthesize tasks or simulation assets, while Eureka \cite{ma2024eurekahumanlevelrewarddesign} and Eurekaverse \cite{liang2024eurekaverseenvironmentcurriculumgeneration} use language models to automatically generate reward functions or learning curricula. These approaches primarily target data generation and training diversity, rather than evaluation itself: generated tasks are treated as inputs to learning pipelines, and the task space is not exposed to users as a controllable or interpretable evaluation interface. Task generation is typically decoupled from mechanisms for enforcing reproducibility, tracking task lineage, or systematically relating task variations to evaluation outcomes. Language has also been used to guide robot behavior at execution time, as in SayCan \cite{ahn2022icanisay} and Code as Policies \cite{liang2023codepolicieslanguagemodel}, where it serves as a high-level planning or control signal while task definitions and success criteria remain fixed and externally specified. In contrast, our work treats language as a first-class interface for evaluation: natural language instructions are compiled into structured, executable task specifications with explicit asset definitions, initialization distributions, and success predicates, enabling users to author reproducible families of semantically related evaluation tasks that support controlled variation and systematic comparison.

% \subsection{Evaluation and Benchmarking}
% Typical evaluation frameworks:
% RoboTwin
% RLBench
% RoboEval
% RoboCasa
% Libero
% Behaviour-1k
% ManipulationNet
% Colloseum
% RoboArena
% Polaris
% Simpler

% \subsection{LLM-based Task Generation}
% GenSim, GenSim2
% AnyTask
% Evaluating Gemini Robotics Policies in a Veo World Simulator
% Scaling Up and Distilling Down
% Gen2Sim
% RoboGen
% SteerableScene Generation with Post Training and Inference Time Search
% BrickGPT
% BloxNet
% EurekaVerse
% Ctrl-Net

\section{Methods}

This section describes how the framework operationalizes the core desiderata: accessibility, continuous growth, reproducibility, and structured control. We first define the design principles and formalize the task representation in Section \ref{sec:task_representation}. We then describe the task orchestration process that make language-defined manipulation tasks executable and interpretable in Section \ref{sec:task_orchestration}. We then describe how natural language instructions are compiled into concrete, reproducible task artifacts through a validation pipeline that enforces physical realizability and consistency Section \ref{sec:validation}. Finally, we introduce a context-aware steering mechanism that enables users to systematically vary tasks and expand the evaluation space over time while preserving explicit comparability between task variants in Section \ref{sec:task_variation}.

% The design of the system is guided by four principles that mirror the desiderata for effective manipulation evaluation. 
% \emph{Accessibility} is achieved by allowing users to specify task intent, constraints, and success conditions directly in natural language, without interacting with benchmark-specific code. 
% \emph{Continuous growth} is supported by treating each instruction as defining a family of related tasks, which can be refined and extended through iterative modification. 
% \emph{Reproducibility} is enforced by compiling language into concrete task artifacts that fully specify task structure and execution, rather than relying on underspecified descriptions. 
% Finally, \emph{grounding in a structured physical domain} constrains language-defined tasks to remain executable and interpretable, enabling systematic variation and meaningful attribution of failure modes.

% Our framework is designed to address the limitations of instance-based manipulation benchmarks by satisfying four desiderata: accessibility through natural language task specification, continuous growth through user-authored variation, reproducibility through explicit and shareable task realizations, and grounding in a structured physical domain that preserves executability and interpretability.

\begin{figure*}[t]
    \centering
    \includegraphics[width=\textwidth]{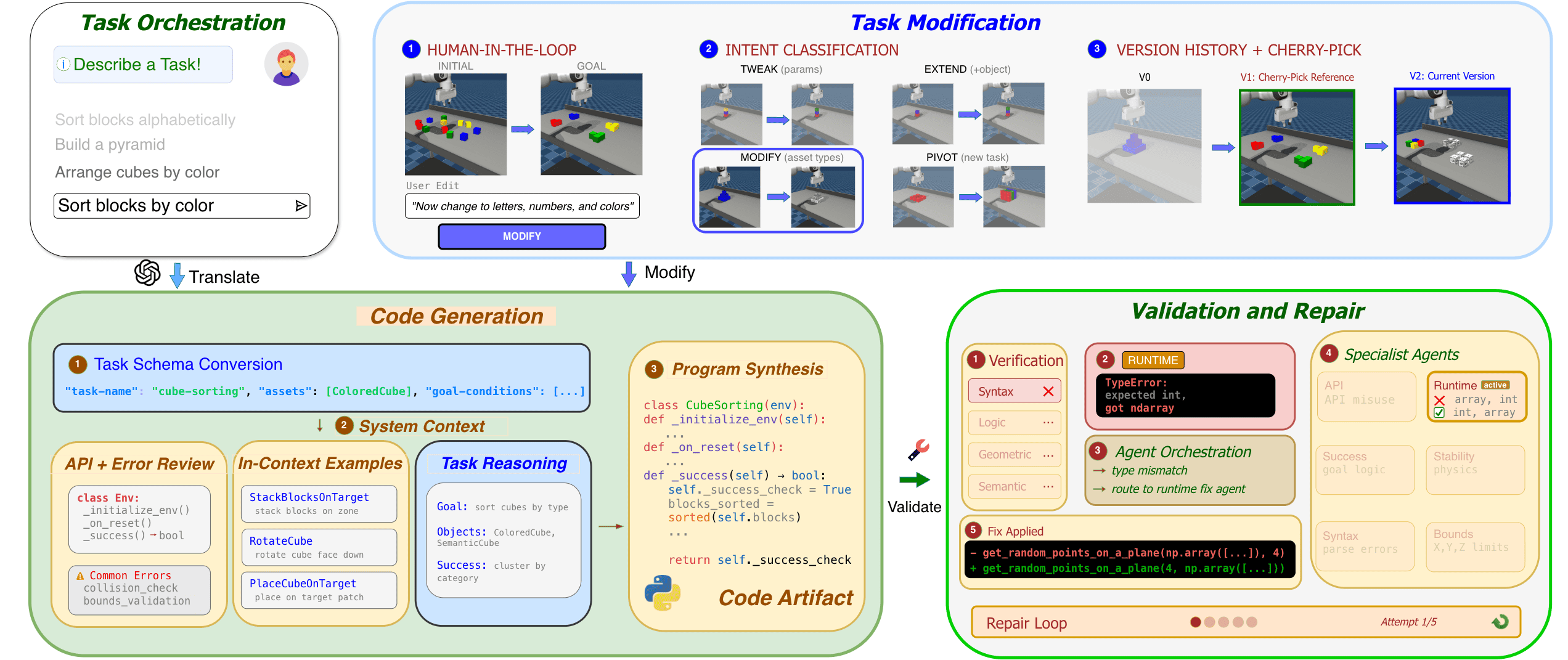}
    \vspace{-1mm}
    \caption{
\textbf{System overview.}
The framework compiles natural language instructions into executable manipulation tasks and supports their controlled evolution over time.
(\emph{Left}) A user provides a natural language description of a task, which is translated into a structured task schema specifying assets, initialization logic, and success conditions.
Conditioned on this schema and retrieved context (APIs, prior tasks, and error patterns), the system synthesizes an executable task implementation.
(\emph{Bottom right}) The generated task is admitted only after passing a multi-stage validation and repair pipeline that enforces software correctness, API consistency, and physical realizability.
(\emph{Top right}) Once a validated task artifact is established, users may issue modification requests that are interpreted through context-aware steering.
Edits are classified into structured intent categories (e.g., \textit{tweak}, \textit{extend}, \textit{modify}, \textit{pivot}), with all variants tracked via explicit version history and cherry-picking to preserve lineage and reference tasks.
The result is a versioned family of reproducible task artifacts that enables systematic task variation while maintaining explicit comparability across evaluations.
}
    \label{fig:system_overview}
\end{figure*}

\begin{figure*}[t]
    \centering
    \includegraphics[width=\textwidth]{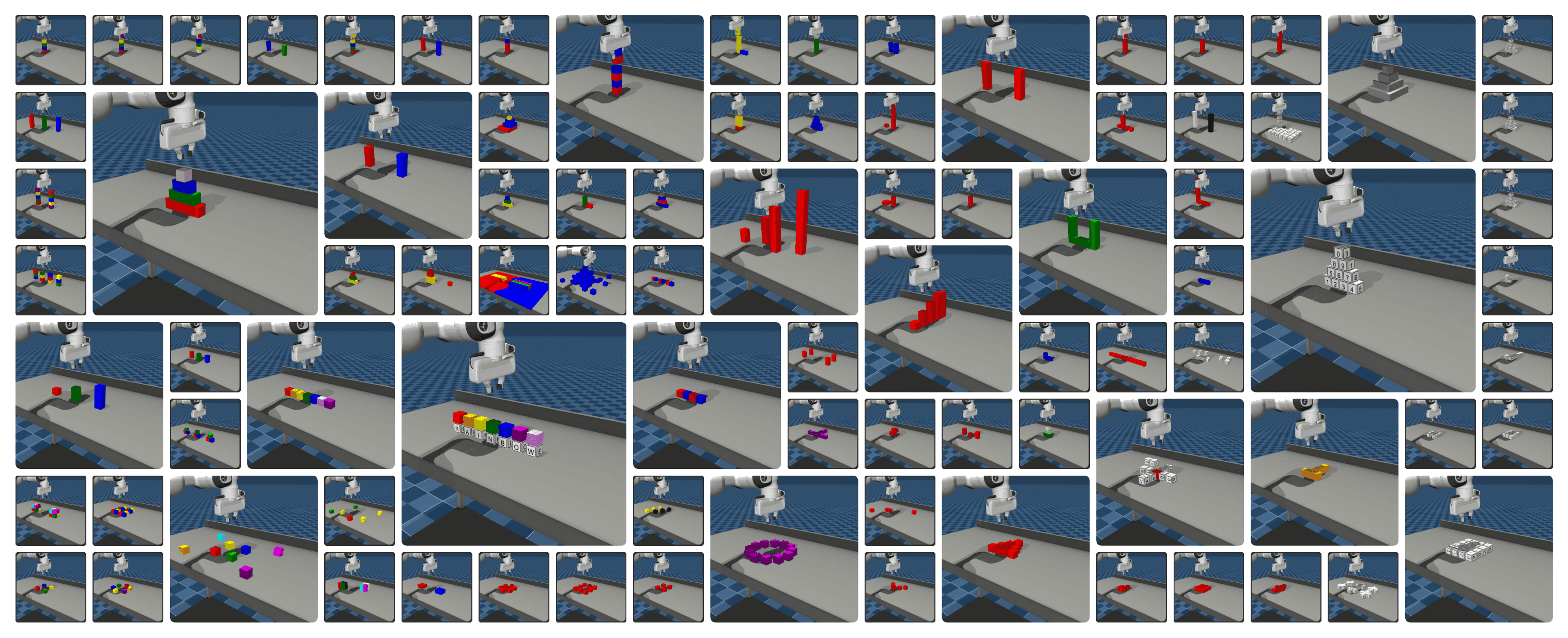}
    \caption{
\textbf{Sample of Language-Defined Manipulation Tasks.}
We visualize a subset of executable manipulation tasks generated by our framework, spanning geometric constructions, spatial alignment, and semantically constrained object arrangements.
Tasks are organized by proximity in a learned task-embedding space, yielding coherent clusters that correspond to families of manipulation problems with shared object attributes, spatial relations, and success definitions (e.g., stacking, ordering, grouping, and alignment).
Each image represents an instantiated task instance, demonstrating that continuous language representations induce discrete, interpretable structure over a diverse space of manipulation tasks.
}
    \label{fig:semantic_grid}
\end{figure*}

\subsection{Task Representation and Design Principles}
\label{sec:task_representation}
Our framework treats natural language as an executable interface for task specification. User instructions are compiled into concrete task realizations that fully determine assets, initialization logic, and success conditions. Rather than producing isolated benchmark instances, the system yields reusable task artifacts that can be shared, re-executed, and systematically varied, enabling evaluation spaces to grow through user contribution without sacrificing scientific rigor. Figure~\ref{fig:system_overview} provides an overview of this process. Natural language instructions are compiled into structured task proposals, synthesized into executable implementations, and admitted as task artifacts only after passing a multi-stage validation pipeline. Validated artifacts can then be iteratively refined through context-aware steering, enabling controlled task variation while preserving reproducibility and explicit lineage.

Formally, we define a manipulation task as a tuple
\begin{equation}
T = (\mathcal{A}, \rho_0, G, \ell, \mathcal{V}),
\end{equation}
where $\mathcal{A}$ denotes the set of task assets, $\rho_0$ is a distribution over initial states, $G : \mathcal{S} \rightarrow \{0,1\}$ is a success predicate over simulator states, $\ell$ is the canonical natural language instruction, and $\mathcal{V}$ is a set of paraphrases used for robustness testing.

This decomposition reflects a deliberate design choice. Logical equivalence at the level of language does not imply equivalence of task realization: differences in tolerances, reset distribution, or success-check timing can lead to divergent evaluation outcomes even when tasks are described identically. Language alone is therefore insufficient as a unit of evaluation.

% For this reason, the fundamental unit produced by the system is not a task description, but a \emph{task artifact}
% \begin{equation}
% T_{\text{artifact}} = (\texttt{TaskProposal}, \texttt{Code}, \texttt{ValidationLog}),
% \end{equation}
% where \texttt{TaskProposal} is a structured specification derived from language, \texttt{Code} is an executable implementation, and \texttt{ValidationLog} records the outcomes of all verification checks. Sharing $T_{\text{artifact}}$ ensures that policies are evaluated against the same concrete task realization rather than underspecified linguistic intent.

% By elevating task artifacts as the object of evaluation, we decouple evaluation from benchmark-specific codebases and enable evaluation spaces that can grow continuously through user-authored instructions, while remaining reproducible and scientifically grounded.

\subsection{Task Orchestration Through Language}
\label{sec:task_orchestration}
Given a natural language description $u$, task construction begins by translating language into a structured representation of task intent. Specifically, the system infers and populates a fixed \texttt{TaskSchema} that explicitly specifies the task name, relevant assets, goal conditions, and initialization logic. The use of a fixed schema ensures that all task-relevant fields are present and disambiguated before execution, enabling complete interpretation of the instruction and preventing underspecified task definitions.

% Before any executable code is synthesized, the filled schema undergoes feasibility pre-validation to enforce domain constraints. These checks include asset compatibility, label availability, and consistency between semantic requirements and geometric realizability. When violations are detected, structured repair rules are applied to minimally modify the schema while preserving the original task intent. Proposals that cannot be repaired are rejected prior to synthesis, ensuring that only feasible task specifications are passed downstream. \hwnote{This is not reflected in Figure 2, either exclude or include in Figure 2 for consistency.}

Conditioned on the validated task schema, the system then synthesizes an executable task implementation. This process leverages an LLM with access to relevant environment APIs, prior task implementations, and diagnostic error information retrieved based on structural similarity to the proposed task. The LLM produces an intermediate natural-language task specification that articulates the intended objects, goal configuration, and success criteria, which is subsequently compiled into executable code.

Executable tasks are implemented as classes that extend a fixed environment interface. Each task defines methods for environment initialization, reset-time sampling from $\rho_0$, and success evaluation corresponding to $G$. This constrained interface enforces uniform structure across task implementations and limits variation arising from authoring style, ensuring that differences in evaluation outcomes reflect task content rather than implementation artifacts.

\subsection{Validation and Physical Realizability}
\label{sec:validation}
Language-defined tasks are only meaningful if they are both executable and physically realizable. Each synthesized task implementation is therefore subjected to a multi-stage validation pipeline before being admitted as a task artifact.

\textbf{Basic validation.}
Basic validation enforces software correctness independent of physics simulation. Generated code is subjected to static analysis to detect syntactic errors and forbidden patterns, compiled in an isolated execution environment to detect import and definition errors, and instantiated to detect runtime failures during object creation or reset-time sampling.

\textbf{Goal-state verification.}
To enforce physical realizability of the success predicate, tasks are instantiated directly in the goal configuration and simulated forward under zero action to allow contacts to settle. The success predicate $G$ must evaluate to true after settling and remain true over an extended horizon, ensuring that the goal configuration is both achievable and stable under the simulator’s physics model.

\textbf{Iterative repair.}
When validation fails, the failure is classified according to its source (e.g., syntax, API usage, runtime instantiation, goal satisfaction, or physical instability), and a corresponding repair operator proposes a localized modification to the task implementation. Repairs may adjust object placements, relax geometric constraints, or rewrite components of the success predicate, depending on the failure type. Validation is then re-run on the repaired implementation. This process repeats until all checks pass or a fixed retry budget is exhausted, ensuring that admitted tasks satisfy executability and physical consistency while remaining faithful to the original language intent.

\subsection{Controlled Task Modifications}
\label{sec:task_variation}
A validated task artifact defines a reference task instance from which a family of related tasks can be derived. To enable systematic task variation without sacrificing comparability, the framework provides a context-aware steering mechanism that interprets user modification requests and constrains how tasks may evolve.

Given a modification request, the system first interprets the intent and extracts structured parameters such as dimensional changes, ordering constraints, or asset-type substitutions. It then classifies the request into one of five steering categories: \emph{Tweak}, \emph{Extend}, \emph{Modify}, \emph{Pivot}, or \emph{Fresh}. Each category specifies explicit preservation guarantees over the task components $(\mathcal{A}, \rho_0, G)$. For example, \emph{Tweak} and \emph{Extend} preserve the original task structure and success predicate, enabling direct comparability with the reference task, while \emph{Modify} and \emph{Pivot} permit progressively broader semantic or structural changes when required by the user intent.

Task evolution is tracked through versioned snapshots that record structured summaries of assets, goals, and code hashes. When a modification requires asset types incompatible with the current version, the system selects a compatible prior snapshot as the reference. This allows coherent multi-step refinement without manual bookkeeping. Each validated variant produces a new snapshot, yielding version-controlled task families with explicit lineage suitable for systematic evaluation and controlled analysis of task variation.

\section{Results}

We evaluate \textsc{\methodname}~along three axes that are central to its role as a democratized evaluation framework for robotic manipulation: 
(i) the usability of its task authoring interface, 
(ii) the diagnostic value of the resulting task set for assessing policy generalization, and 
(iii) the scalability of task creation under open-world, crowd-driven use.
Across all experiments, we focus on whether \textsc{\methodname}~ enables task specifications that are both easier to author and more informative for evaluation than existing alternatives.

\subsection{Usability of the Task Authoring Interface}

\begin{table}[t]
\centering
\caption{User study results (N=26).
Values are reported as mean $\pm$ 95\% confidence interval margin across participants.
SUS is scored on a 0--100 scale.
TLX workload is the unweighted mean of five NASA-TLX subscales
(Mental Demand, Temporal Demand, Effort, Frustration, and reversed Performance),
normalized to 0--100.
Usability rank is mean rank (1 = best) across systems.
Preferred (\%) is the share of participants who named that system as their overall preference.}
\label{tab:usability_results}
\footnotesize
\resizebox{\columnwidth}{!}{%
\begin{tabular}{lcccc}
\toprule
System & SUS $\uparrow$ & TLX workload $\downarrow$ & Usability rank $\downarrow$ & Preferred (\%) \\
\midrule
GenSim & 52.5$\pm$9.3 & 41.8$\pm$9.0 & 2.7$\pm$0.2 & 8 \\
Cursor & 68.8$\pm$7.8 & 36.7$\pm$10.4 & 2.0$\pm$0.2 & 23 \\
RoboPlayground & \textbf{83.4$\pm$6.9} & \textbf{18.6$\pm$7.7} & \textbf{1.3$\pm$0.3} & \textbf{69} \\
\bottomrule
\end{tabular}%
}
\end{table}

\textbf{Experimental setting.}
We evaluate the usability of \methodname\ in comparison to two baseline task authoring interfaces, GenSim \cite{wang2023gensim} and Cursor \cite{cursor_ai}, using a within-subjects user study ($N=26$).
Participants were asked to construct an identical manipulation task (build a 3D structure using blocks under various constraints) using each system.
All participants interacted with all three systems, enabling paired comparisons of perceived usability, cognitive workload, and user preference.
We measure usability using the System Usability Scale (SUS) \cite{brooke1996sus}, cognitive workload using NASA-TLX subscales \cite{hart1988development}, and overall preference through usability and forced-choice rankings.
Results are summarized in Table~\ref{tab:usability_results}.

\textbf{\methodname\ achieves higher perceived usability than baselines.}
Across participants ($N{=}26$), \methodname\ attains the highest System Usability Scale (SUS) score ($83.4 \pm 6.9$; mean $\pm$ 95\% confidence interval margin), well above the conventional acceptability threshold of 68.
GenSim and Cursor achieve substantially lower mean SUS ($52.5 \pm 9.3$ and $68.8 \pm 7.8$, respectively).
Paired Wilcoxon signed-rank tests confirm that \methodname\ significantly outperforms both GenSim ($p{<}0.001$) and Cursor ($p{=}0.0017$), so the advantage is not limited to the weakest baseline: \methodname\ is rated more usable than a strong general-purpose assistant interface as well.
The interval for GenSim is the widest of the three, consistent with more heterogeneous experiences in that condition, whereas \methodname\ shows the tightest margin among systems, indicating comparatively consistent high ratings.

\textbf{\methodname\ reduces perceived cognitive workload relative to baselines.}
Cognitive workload is summarized as the unweighted mean of five NASA-TLX subscales (Mental Demand, Temporal Demand, Effort, Frustration, and reversed Performance), each normalized to 0--100 and oriented so that lower is better.
\methodname\ yields the lowest mean composite score ($18.6 \pm 7.7$; mean $\pm$ 95\% confidence interval margin), compared to $41.8 \pm 9.0$ for GenSim and $36.7 \pm 10.4$ for Cursor.
Paired Wilcoxon signed-rank tests show that \methodname\ significantly reduces perceived workload relative to both GenSim ($p{=}0.0007$) and Cursor ($p{=}0.0019$).
GenSim and Cursor do not differ significantly from each other on this composite ($p{=}0.22$), whereas \methodname\ separates clearly from each baseline; Cursor also exhibits the widest TLX margin among the three, indicating somewhat more spread in workload ratings even though the paired comparison to \methodname\ remains significant.

\textbf{Participants consistently prefer \methodname\ over baseline interfaces.}
Subjective measures reinforce the quantitative usability and workload results.
\methodname\ achieves the best mean usability rank ($1.3 \pm 0.3$; lower is better), with GenSim and Cursor at $2.7 \pm 0.2$ and $2.0 \pm 0.2$, respectively.
A Friedman test shows strong differences in rankings across systems ($p{<}0.001$), and post-hoc paired Wilcoxon tests confirm that \methodname\ is ranked significantly better than both GenSim ($p{=}0.0001$) and Cursor ($p{=}0.0078$).
In forced-choice overall preference, $69\%$ of participants select \methodname, compared to $23\%$ for Cursor and $8\%$ for GenSim.
A chi-square goodness-of-fit test rejects a uniform split across the three options ($p{=}0.0003$), consistent with concentration of preference on \methodname.
Together, the ranking and preference distributions indicate a stable, statistically supported tilt toward \methodname\ over both baselines.

\subsection{Evaluating Policies on Training and Generated Generalization Tasks}
\label{sec:static_tasks}

\begin{table*}[t]
\centering
\scriptsize

\caption{\textbf{Results on in-distribution and generalization tasks.} Success rates (\%) with standard errors across six policies evaluated on training (in-distribution) tasks (top) and held-out generalization tasks (bottom). The best result per task is shown in \textbf{bold}.
Generalization tasks are constructed by perturbing training tasks along one or more axes: semantic (S), denoting language perturbations, visual (V), denoting visual appearance differences in the initial state, and behavioural (B), denoting changes in the required behaviour, adhering to the definitions in \cite{gao2026taxonomy}.
The perturbation type for each generalization task is shown in the row labeled \emph{Perturbation}, with semantic perturbations indicated in \textcolor{blue}{blue}, visual perturbations in \textcolor{green}{green}, and behavioural perturbations in \textcolor{red}{red}.}
\label{tab:eval_results}

\resizebox{\textwidth}{!}{
\begin{tabular}{lcccccccccc}
\toprule
\textbf{Method} & \multicolumn{1}{c}{\begin{tabular}[c]{@{}c@{}}\texttt{Red Block} \\\texttt{Front of Yellow}\end{tabular}} & \multicolumn{1}{c}{\begin{tabular}[c]{@{}c@{}}\texttt{Red Block} \\\texttt{Right Placement}\end{tabular}} & \multicolumn{1}{c}{\begin{tabular}[c]{@{}c@{}}\texttt{Red Behind} \\\texttt{Yellow}\end{tabular}} & \multicolumn{1}{c}{\begin{tabular}[c]{@{}c@{}}\texttt{Red Block} \\\texttt{Stacking}\end{tabular}} & \multicolumn{1}{c}{\begin{tabular}[c]{@{}c@{}}\texttt{Three Block} \\\texttt{Color Stacking}\end{tabular}} & \multicolumn{1}{c}{\begin{tabular}[c]{@{}c@{}}\texttt{Red Block} \\\texttt{Left Placement}\end{tabular}} & \multicolumn{1}{c}{\begin{tabular}[c]{@{}c@{}}\texttt{Red on} \\\texttt{Yellow Stack}\end{tabular}} & \multicolumn{1}{c}{\begin{tabular}[c]{@{}c@{}}\texttt{Yellow on} \\\texttt{Red Stack}\end{tabular}} & \multicolumn{1}{c}{\begin{tabular}[c]{@{}c@{}}\texttt{Color Block} \\\texttt{Alignment}\end{tabular}} & \multicolumn{1}{c}{\begin{tabular}[c]{@{}c@{}}\texttt{Place Two} \\\texttt{Blocks on Patch}\end{tabular}} \\
\cmidrule(lr){2-2} \cmidrule(lr){3-3} \cmidrule(lr){4-4} \cmidrule(lr){5-5} \cmidrule(lr){6-6} \cmidrule(lr){7-7} \cmidrule(lr){8-8} \cmidrule(lr){9-9} \cmidrule(lr){10-10} \cmidrule(lr){11-11}
\\[-13pt]
\\
\texttt{Pi-0.5} & 72.0 $\pm$ 6.3 & 68.0 $\pm$ 6.6 & 58.0 $\pm$ 7.0 & 0.0 $\pm$ 0.0 & 6.0 $\pm$ 3.4 & 64.0 $\pm$ 6.8 & 46.0 $\pm$ 7.0 & 62.0 $\pm$ 6.9 & 0.0 $\pm$ 0.0 & 74.0 $\pm$ 6.2 \\
\texttt{Pi-0.5 (LoRA)} & 32.0 $\pm$ 6.6 & 16.0 $\pm$ 5.2 & 48.0 $\pm$ 7.1 & 0.0 $\pm$ 0.0 & 0.0 $\pm$ 0.0 & 12.0 $\pm$ 4.6 & 12.0 $\pm$ 4.6 & 36.0 $\pm$ 6.8 & 0.0 $\pm$ 0.0 & 28.0 $\pm$ 6.3 \\
\texttt{Adapter} & 66.0 $\pm$ 6.7 & 64.0 $\pm$ 6.8 & 60.0 $\pm$ 6.9 & 0.0 $\pm$ 0.0 & 2.0 $\pm$ 2.0 & 26.0 $\pm$ 6.2 & 56.0 $\pm$ 7.0 & 54.0 $\pm$ 7.0 & 2.0 $\pm$ 2.0 & 78.0 $\pm$ 5.9 \\
\texttt{Dual} & \textbf{88.0 $\pm$ 4.6} & 76.0 $\pm$ 6.0 & \textbf{74.0 $\pm$ 6.2} & 2.0 $\pm$ 2.0 & 12.0 $\pm$ 4.6 & \textbf{84.0 $\pm$ 5.2} & 64.0 $\pm$ 6.8 & 54.0 $\pm$ 7.0 & 6.0 $\pm$ 3.4 & 84.0 $\pm$ 5.2 \\
\texttt{GR00T} & 82.0 $\pm$ 5.4 & 84.0 $\pm$ 5.2 & \textbf{74.0 $\pm$ 6.2} & \textbf{10.0 $\pm$ 4.2} & \textbf{22.0 $\pm$ 5.9} & 68.0 $\pm$ 6.6 & 66.0 $\pm$ 6.7 & 56.0 $\pm$ 7.0 & \textbf{12.0 $\pm$ 4.6} & \textbf{96.0 $\pm$ 2.8} \\
\texttt{Qwen-OFT} & 82.0 $\pm$ 5.4 & \textbf{86.0 $\pm$ 4.9} & 68.0 $\pm$ 6.6 & 2.0 $\pm$ 2.0 & 14.0 $\pm$ 4.9 & \textbf{84.0 $\pm$ 5.2} & \textbf{76.0 $\pm$ 6.0} & \textbf{68.0 $\pm$ 6.6} & 2.0 $\pm$ 2.0 & 78.0 $\pm$ 5.9 \\
\bottomrule
\end{tabular}
}

\vspace{0.1cm}
\resizebox{\textwidth}{!}{
\begin{tabular}{lcccccccccccc}
\toprule
\\[-6pt]
\textbf{Perturbation}
& {\color{blue}\textbf{S}}{\color{black}\textbf{+}}{\color{red}\textbf{B}}
& {\color{green}\textbf{V}}
& {\color{blue}\textbf{S}}
& {\color{blue}\textbf{S}}{\color{black}\textbf{+}}{\color{green}\textbf{V}}
& {\color{green}\textbf{V}}{\color{black}\textbf{+}}{\color{red}\textbf{B}}
& {\color{blue}\textbf{S}}
& {\color{green}\textbf{V}}
& {\color{blue}\textbf{S}}{\color{black}\textbf{+}}{\color{red}\textbf{B}}
& {\color{green}\textbf{V}}
& {\color{green}\textbf{V}}
& {\color{blue}\textbf{S}}{\color{black}\textbf{+}}{\color{red}\textbf{B}}
& {\color{green}\textbf{V}} \\
% \\[-4pt]
\midrule
\textbf{Method} & \multicolumn{1}{c}{\begin{tabular}[c]{@{}c@{}}\texttt{Red Block} \\\texttt{Two Tower}\end{tabular}} & \multicolumn{1}{c}{\begin{tabular}[c]{@{}c@{}}\texttt{Blue Block} \\\texttt{Stacking}\end{tabular}} & \multicolumn{1}{c}{\begin{tabular}[c]{@{}c@{}}\texttt{Yellow Block} \\\texttt{Left Placement}\end{tabular}} & \multicolumn{1}{c}{\begin{tabular}[c]{@{}c@{}}\texttt{Red Block Left} \\\texttt{of Blue}\end{tabular}} & \multicolumn{1}{c}{\begin{tabular}[c]{@{}c@{}}\texttt{Yellow on Red} \\\texttt{Unstack Restack}\end{tabular}} & \multicolumn{1}{c}{\begin{tabular}[c]{@{}c@{}}\texttt{Green on} \\\texttt{Blue Stack}\end{tabular}} & \multicolumn{1}{c}{\begin{tabular}[c]{@{}c@{}}\texttt{Three Block} \\\texttt{Perturbed}\end{tabular}} & \multicolumn{1}{c}{\begin{tabular}[c]{@{}c@{}}\texttt{Three Block} \\\texttt{Beside}\end{tabular}} & \multicolumn{1}{c}{\begin{tabular}[c]{@{}c@{}}\texttt{Place Two Blue} \\\texttt{Blocks on Patch}\end{tabular}} & \multicolumn{1}{c}{\begin{tabular}[c]{@{}c@{}}\texttt{Place Two Blocks} \\\texttt{on Green Patch}\end{tabular}} & \multicolumn{1}{c}{\begin{tabular}[c]{@{}c@{}}\texttt{Stack Two} \\\texttt{Blocks on Patch}\end{tabular}} & \multicolumn{1}{c}{\begin{tabular}[c]{@{}c@{}}\texttt{Place Two Blocks} \\\texttt{on Long Patch}\end{tabular}} \\
\cmidrule(lr){2-2} \cmidrule(lr){3-3} \cmidrule(lr){4-4} \cmidrule(lr){5-5} \cmidrule(lr){6-6} \cmidrule(lr){7-7} \cmidrule(lr){8-8} \cmidrule(lr){9-9} \cmidrule(lr){10-10} \cmidrule(lr){11-11} \cmidrule(lr){12-12} \cmidrule(lr){13-13}
\\[-13pt]
\\
\texttt{Pi-0.5}  & 4.0 $\pm$ 2.8 & 0.0 $\pm$ 0.0 & 76.0 $\pm$ 6.0 & 50.0 $\pm$ 7.1 & 0.0 $\pm$ 0.0 & 60.0 $\pm$ 6.9 & 10.0 $\pm$ 4.2 & 12.0 $\pm$ 4.6 & 46.0 $\pm$ 7.0 & 80.0 $\pm$ 5.7 & 0.0 $\pm$ 0.0 & \textbf{68.0 $\pm$ 6.6} \\
\texttt{Pi-0.5 (LoRA)} & 2.0 $\pm$ 2.0 & 0.0 $\pm$ 0.0 & 10.0 $\pm$ 4.2 & 32.0 $\pm$ 6.6 & 0.0 $\pm$ 0.0 & 0.0 $\pm$ 0.0 & 2.0 $\pm$ 2.0 & 0.0 $\pm$ 0.0 & 10.0 $\pm$ 4.2 & 30.0 $\pm$ 6.5 & \textbf{2.0 $\pm$ 2.0} & 28.0 $\pm$ 6.3 \\
\texttt{Adapter} & 0.0 $\pm$ 0.0 & 0.0 $\pm$ 0.0 & 4.0 $\pm$ 2.8 & 36.0 $\pm$ 6.8 & 0.0 $\pm$ 0.0 & 0.0 $\pm$ 0.0 & 8.0 $\pm$ 3.8 & 0.0 $\pm$ 0.0 & 46.0 $\pm$ 7.0 & 38.0 $\pm$ 6.9 & 0.0 $\pm$ 0.0 & 26.0 $\pm$ 6.2 \\
\texttt{Dual} & 8.0 $\pm$ 3.8 & 0.0 $\pm$ 0.0 & 74.0 $\pm$ 6.2 & \textbf{60.0 $\pm$ 6.9} & 0.0 $\pm$ 0.0 & 56.0 $\pm$ 7.0 & 8.0 $\pm$ 3.8 & 10.0 $\pm$ 4.2 & 80.0 $\pm$ 5.7 & 72.0 $\pm$ 6.3 & \textbf{2.0 $\pm$ 2.0} & 54.0 $\pm$ 7.0 \\
\texttt{GR00T} & \textbf{10.0 $\pm$ 4.2} & 0.0 $\pm$ 0.0 & \textbf{78.0 $\pm$ 5.9} & 52.0 $\pm$ 7.1 & \textbf{2.0 $\pm$ 2.0} & 62.0 $\pm$ 6.9 & \textbf{20.0 $\pm$ 5.7} & 4.0 $\pm$ 2.8 & \textbf{86.0 $\pm$ 4.9} & \textbf{90.0 $\pm$ 4.2} & 0.0 $\pm$ 0.0 & \textbf{68.0 $\pm$ 6.6} \\
\texttt{Qwen-OFT} & 6.0 $\pm$ 3.4 & 0.0 $\pm$ 0.0 & 54.0 $\pm$ 7.0 & 48.0 $\pm$ 7.1 & 0.0 $\pm$ 0.0 & \textbf{66.0 $\pm$ 6.7} & 14.0 $\pm$ 4.9 & \textbf{14.0 $\pm$ 4.9} & 54.0 $\pm$ 7.0 & 66.0 $\pm$ 6.7 & 0.0 $\pm$ 0.0 & 56.0 $\pm$ 7.0 \\
\bottomrule
\end{tabular}
}
\vspace{2pt}
\end{table*}

\textbf{Tasks.} All policies are trained on a shared set of base manipulation tasks covering spatial relations, stacking, alignment, semantic disambiguation, and targeted placement. For each training task, we generate successful demonstration trajectories using CuTAMP~\cite{shen2024differentiable} and hold the resulting dataset fixed across policies, ensuring that performance differences reflect policy behavior rather than differences in supervision. Evaluation is performed on (i) tasks drawn from the training distribution, and (ii) user generated generalization tasks that require adaptation beyond the training distribution.
Generalization tasks are constructed by applying controlled modifications to base tasks using \methodname, including semantic changes to language instructions, visual changes to object attributes and initial configurations (e.g.\ partially stacked versus scattered on table), and behavioral changes that alter the required action sequences or temporal structure.
These transformations follow the task taxonomy described in \cite{gao2026taxonomy}, and are designed to isolate specific dimensions of generalization while preserving task validity.

\textbf{Models.} We evaluate six policies spanning different action head architectures and fine-tuning strategies. Four are built on a shared Qwen3-VL-4B-Instruct~\cite{bai2025qwen3vltechnicalreport} vision-language backbone and differ in their action decoding mechanism:
\texttt{Adapter} appends learnable action query tokens to the VLM sequence and decodes actions via an MLP-ResNet regression head with an L1 objective;
\texttt{GR00T} conditions a flow-matching diffusion transformer (DiT) on the VLM's final hidden states, adopting a dual-system architecture inspired by GR00T N1.5~\cite{nvidia2025gr00tn1openfoundation};
\texttt{Dual} extends this flow-matching action head with a secondary DINOv2~\cite{oquab2024dinov2learningrobustvisual} visual encoder whose patch features are concatenated with the VLM hidden states before conditioning;
and \texttt{Qwen-OFT} regresses actions from special action-token positions via an MLP head, following the OpenVLA-OFT design~\cite{kim2025finetuningvisionlanguageactionmodelsoptimizing}.
As external baselines, we include \texttt{Pi-0.5}~\cite{intelligence2025pi05visionlanguageactionmodelopenworld}, a proprietary VLA with flow-matching action generation, evaluated both with full finetuning and with low-rank adaptation (\texttt{Pi-0.5 (LoRA)}).
All models are trained end-to-end on identical demonstration data.
Details of training configurations and generalization tasks are outlined in the appendix.

\textbf{Training-distribution performance.}
We first report performance on tasks drawn from the training distribution (top section of Table~\ref{tab:eval_results}) to provide context for subsequent generalization results.
All models achieve moderate to high success on tasks involving simple spatial relations and single-object placement, with \texttt{GR00T}, \texttt{Dual}, and \texttt{Qwen-OFT} consistently outperforming \texttt{Pi-0.5} and \texttt{Adapter} on most placement and relational tasks. \texttt{GR00T} achieves the highest overall in-distribution performance, reaching 96\% on \texttt{Place Two Blocks on Patch} and leading on stacking-related tasks. However, all models exhibit a consistent difficulty gradient: performance degrades sharply on training tasks requiring greater compositional structure or longer-horizon execution, such as multi-block stacking and color block alignment, where even the strongest model does not exceed 22\%. \texttt{Adapter} shows notably uneven in-distribution performance, achieving competitive results on some tasks (e.g., 78\% on patch placement) while lagging substantially on others (e.g., 26\% on left placement). The \texttt{Pi-0.5 (LoRA)} variant underperforms all other models across nearly every training task, suggesting that low-rank adaptation alone is insufficient to retain the base model's capabilities in this setting.

\textbf{Generalization results reveal a clear asymmetry across perturbation types.}
Across held-out evaluation tasks, all models generalize unevenly across perturbation axes, though the degree of degradation varies by architecture. Performance remains relatively strong under visual perturbations that alter perceptual attributes while preserving execution structure: \texttt{GR00T} achieves 90\% on \texttt{Place Two Blocks on Green Patch} and 86\% on \texttt{Place Two Blue Blocks on Patch}, and \texttt{Dual} similarly transfers well on these tasks (72\% and 80\%, respectively). Semantic perturbations alone yield mixed but non-zero success for most models, with \texttt{GR00T} reaching 78\% on \texttt{Yellow Block Left Placement} and 62\% on \texttt{Green on Blue Stack}, and \texttt{Dual} achieving 74\% and 56\% on the same tasks, indicating meaningful robustness to relational re-specification. \texttt{Adapter}, however, largely fails under semantic perturbation (e.g., 4\% on \texttt{Yellow Block Left Placement}, 0\% on \texttt{Green on Blue Stack}), suggesting that adapter-based finetuning may overfit to surface-level task features. In contrast, tasks involving behavioural perturbations consistently expose severe failure modes across \emph{all} architectures. Tasks requiring multi-stage execution, non-monotonic progress (e.g., unstack-restack), or compositional sequencing yield near-zero success universally---no model exceeds 2\% on \texttt{Yellow on Red Unstack Restack} or \texttt{Stack Two Blocks on Patch}, and \texttt{Blue Block Stacking} elicits 0\% across the board. These failures occur despite reasonable performance on simpler stacking or placement tasks in isolation, suggesting limited procedural and compositional generalization rather than a lack of basic manipulation competence. \texttt{Pi-0.5 (LoRA)} further degrades performance across nearly all perturbation axes, confirming increased sensitivity to deviations from training task structure under constrained adaptation. Together, these results establish behavioural perturbations as the dominant source of generalization failure \emph{independent of model architecture} and motivate evaluation protocols that explicitly probe execution structure.
% In contrast, evaluation on the generated generalization tasks (bottom portion of Table~\ref{tab:eval_results}) reveals a substantial drop in performance.
% Tasks that introduce new object attributes, unseen compositions of spatial relations, or altered spatial constraints lead to pronounced failures, with several tasks exhibiting near-zero success despite their close structural relationship to training tasks.
% Notably, generalization failures are not uniform.
% Tasks involving relatively simple semantic substitutions, such as changes in object color with otherwise similar spatial structure, retain partial performance, whereas tasks that require new multi-stage behaviors or combine semantic and behavioral changes are significantly more challenging.

% \textbf{Diagnosing adaptation through structured task generation.}
% Comparing \texttt{Pi-0.5} with its frozen counterpart further highlights the role of adaptation.
% While freezing degrades performance across both task sets, the performance gap widens markedly on generated generalization tasks.
% This indicates that task-conditioned adaptation is particularly important when policies must extrapolate beyond memorized task structure and respond to systematic changes in task specification.
% These effects are difficult to observe using a fixed benchmark, but emerge clearly when evaluating over structured families of generated tasks.

\textbf{Implications.}
Together, these results demonstrate that success on a fixed set of training-distribution tasks can substantially overestimate a policy’s robustness.
By enabling the generation of creative task variants that probe specific semantic, visual, and behavioral dimensions of generalization, \textsc{\methodname} enables fine-grained diagnosis of policy capabilities and failure modes that are obscured by static task definitions.

\subsection{Scalability and Diversity of \methodname}
\begin{figure*}[t]
    \centering
    \begin{subfigure}[t]{0.32\linewidth}
        \centering
        \includegraphics[width=\linewidth]{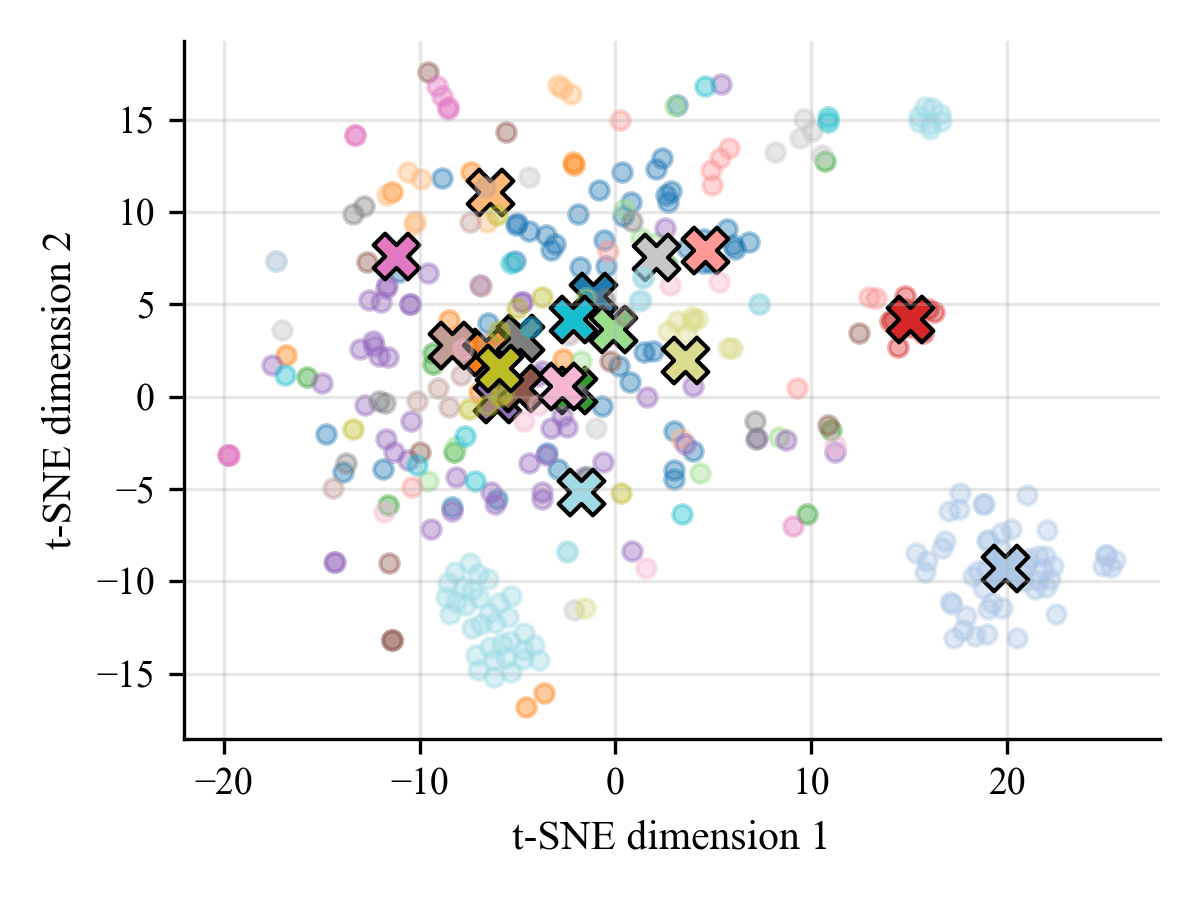}
        \caption{t-SNE visualization of task embeddings grouped by user.}
        \label{fig:tsne_users}
    \end{subfigure}
    \hfill
    \begin{subfigure}[t]{0.32\linewidth}
        \centering
        \includegraphics[width=\linewidth]{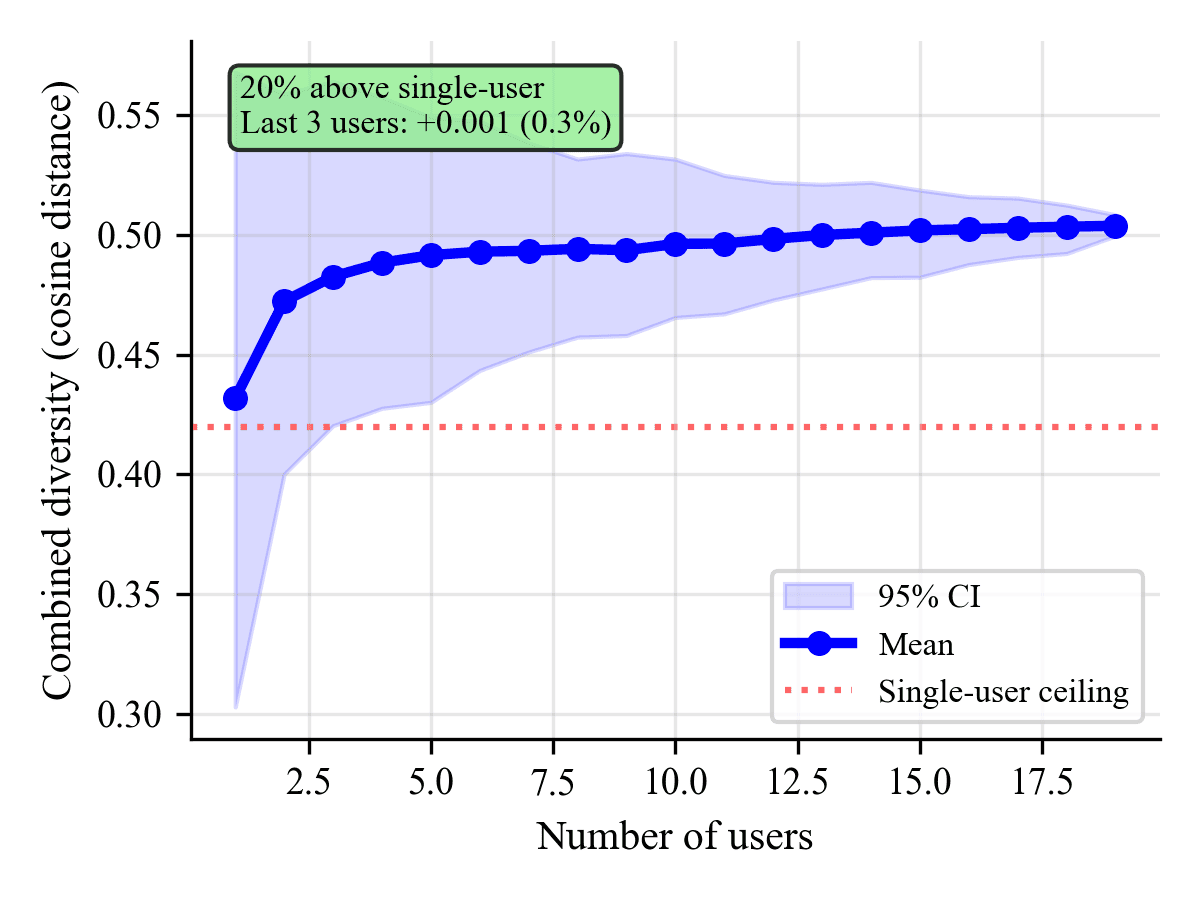}
        \caption{Diversity as a function of number of users.}
        \label{fig:diversity_users}
    \end{subfigure}
    \hfill
    \begin{subfigure}[t]{0.32\linewidth}
        \centering
        \includegraphics[width=\linewidth]{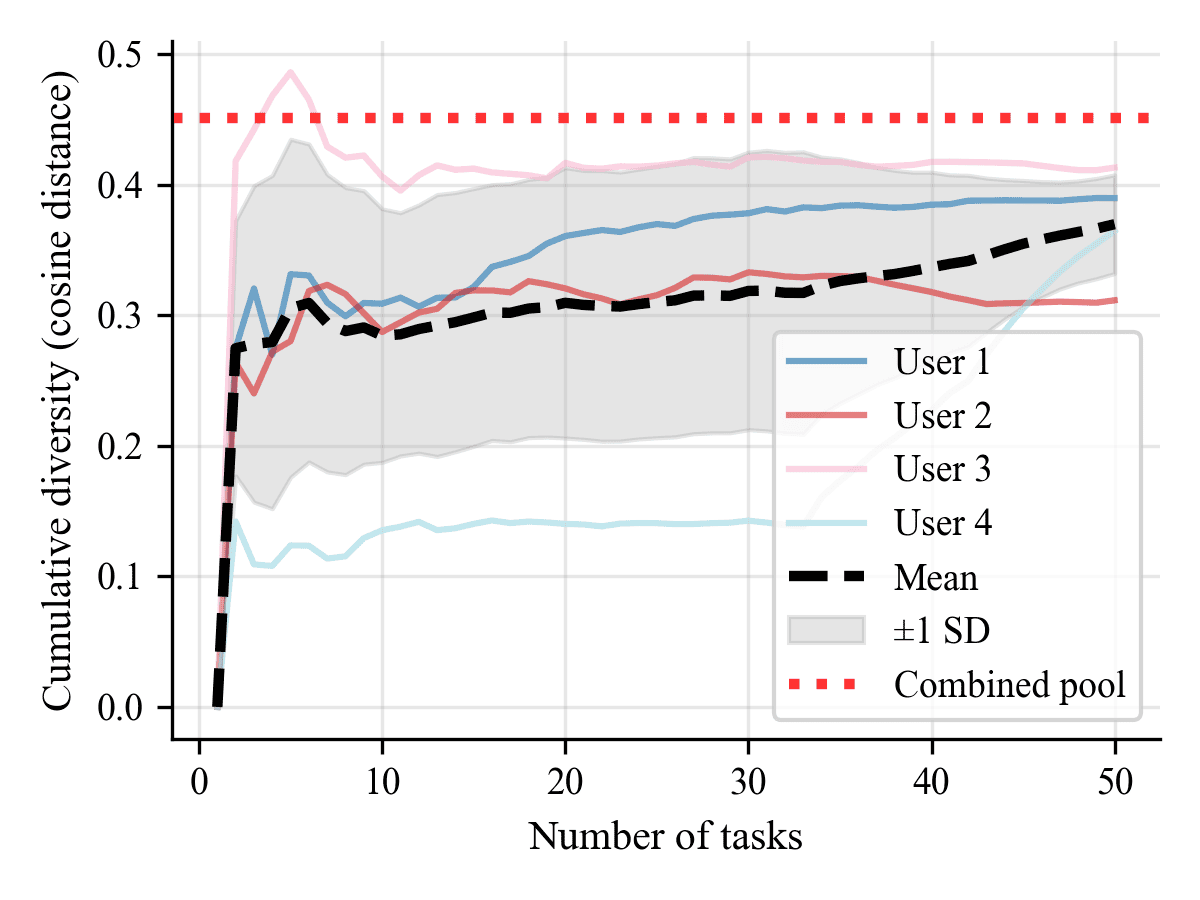}
        \caption{Cumulative diversity as tasks are added per user.}
        \label{fig:diversity_plateau}
    \end{subfigure}

    \caption{
    \textbf{Inter-user and intra-user diversity of natural-language manipulation tasks.}
    \textbf{(Left)} A t-SNE projection of sentence embeddings for all tasks, colored by user, with crosses indicating per-user centroids. Some tasks cluster by author, indicating systematic differences in how users conceptualize and describe manipulation goals.
    \textbf{(Middle)} Mean pairwise diversity (cosine distance) as tasks from an increasing number of users are pooled. Diversity increases monotonically with the number of users, with shaded regions indicating 95\% confidence intervals.
    \textbf{(Right)} Cumulative diversity as additional tasks are added from a single user. Individual users exhibit diminishing returns, while the combined pool (red dashed line) achieves substantially higher diversity, highlighting the complementary contribution of multiple authors.
    }
    \label{fig:user_diversity}
\end{figure*}

For evaluation, diversity matters not as raw task count, but as coverage of distinct task intents and constraint combinations that probe different policy behaviors. We evaluate how task diversity in \textsc{\methodname} scales with the number of contributors and the number of authored tasks. Each contributor authors up to 50 valid manipulation tasks in the blocks domain using the same interface and asset set. To quantify diversity, we compute average pairwise distances between task representations using semantic sentence embeddings, and analyze both pooled task sets across contributors and cumulative task sets authored by individuals. In the appendix, we report additional analyses using alternative diversity measures, which show consistent trends.

\textbf{Inter-user diversity scales with contributors.}
Figure~\ref{fig:user_diversity}(a) shows the distribution of tasks authored by individual users. A t-SNE projection reveals that some users occupy distinct regions of the embedding space, while others exhibit substantial overlap, suggesting systematic differences in how contributors conceptualize and describe manipulation goals within the domain. When tasks are pooled across users (10 tasks per user), the mean pairwise diversity increases monotonically with the number of contributors (Fig.~\ref{fig:user_diversity}(b)). Even after substantial saturation, adding the final three contributors yields a consistent, non-zero increase in diversity, indicating that new contributors continue to introduce semantically novel task formulations. 

\textbf{Intra-user diversity exhibits diminishing returns.}
In contrast, Figure~\ref{fig:user_diversity}(c) shows that when tasks are added incrementally from a single user, cumulative diversity grows rapidly at first but quickly plateaus. This behavior is consistent across most users, suggesting that individual authors often explore a limited region of the task space, potentially shaped by their preferred abstractions, phrasing, and constraint patterns. Even prolific contributors produce increasingly redundant task variations over time.

\textbf{Complementarity of multiple contributors.}
Notably, the combined task pool outperforms any individual contributor in terms of cumulative diversity, as showin in Figure \ref{fig:user_diversity}(c). This gap highlights the complementary nature of crowd-authored task generation: different users introduce distinct semantic concepts, compositional structures, and constraint combinations that are rarely discovered by a single author alone. Qualitative inspection of task clusters confirms the presence of novel formulations of spatial relations, multi-object constraints, and success conditions that are absent from single-author collections.

Overall, these results show that \textsc{\methodname} scales not merely by increasing task count, but by expanding coverage of the underlying task space through contributor diversity. By expanding coverage across task intent and constraint structure, \textsc{\methodname} enables evaluation to reveal brittleness to even seemingly minor semantic variations.

% Overall, these results show that \textsc{\methodname} scales not merely by increasing task count, but by expanding coverage of the underlying task space through contributor diversity. This property is critical for evaluation settings, where closed, expert-designed task sets risk imposing structural bottlenecks that obscure meaningful differences in policy behavior.

\section{Ablative Studies}
\begin{table*}[t]
\centering
\scriptsize
\caption{\textbf{Ablation Study.} We evaluate the contribution of each pipeline component through cumulative addition. Starting with all gates disabled for that module, we progressively enable components to measure their incremental impact. All success metrics are percentages ($n=26$); LLM Alignment is scored out of 100. Green values with $\uparrow$ indicate improvement from the previous row.}
\label{tab:ablation}

\resizebox{\textwidth}{!}{
\begin{tabular}{llcccccc}
\toprule
\textbf{Module} & \textbf{Configuration} & \textbf{Task Succ.} & \textbf{Compile} & \textbf{Smoke Test} & \textbf{Human-Ver.} & \textbf{LLM Align.} \\
\midrule
\multirow{3}{*}{\textbf{Task Proposal}} 
& None (all disabled) & 100.0 & 100.0 & 100.0 & 88.5 & 74.0 \\
& \quad + asset inference & 100.0 & 100.0 & 100.0 & 92.3 {\scriptsize\textcolor{green!50!black}{($\uparrow$3.8)}} & 71.5 {\scriptsize\textcolor{red!70!black}{($\downarrow$2.5)}} \\
& \quad + feasibility checking (all gates on) & \textbf{100.0} & \textbf{100.0} & \textbf{100.0} & \textbf{100.0} {\scriptsize\textcolor{green!50!black}{($\uparrow$7.7)}} & \textbf{73.6} {\scriptsize\textcolor{green!50!black}{($\uparrow$2.1)}} \\
\midrule
\multirow{4}{*}{\textbf{Code Generation}} 
& None (all disabled) & 100.0 & 100.0 & 100.0 & 96.2 & 70.8 \\
& \quad + API review & 96.2 {\scriptsize\textcolor{red!70!black}{($\downarrow$3.8)}} & 100.0 & 100.0 & 100.0 {\scriptsize\textcolor{green!50!black}{($\uparrow$3.8)}} & 70.5 {\scriptsize\textcolor{red!70!black}{($\downarrow$0.3)}} \\
& \quad + common errors review & 100.0 {\scriptsize\textcolor{green!50!black}{($\uparrow$3.8)}} & 100.0 & 100.0 & 96.2 {\scriptsize\textcolor{red!70!black}{($\downarrow$3.8)}} & 72.4 {\scriptsize\textcolor{green!50!black}{($\uparrow$1.9)}} \\
& \quad + in-context examples (all gates on) & \textbf{100.0} & \textbf{100.0} & \textbf{100.0} & \textbf{100.0} {\scriptsize\textcolor{green!50!black}{($\uparrow$3.8)}} & \textbf{73.6} {\scriptsize\textcolor{green!50!black}{($\uparrow$1.2)}} \\
\midrule
\multirow{7}{*}{\textbf{Validation}} 
& None (all disabled) & 12.0 & 96.0 & 12.0 & 96.2 & 71.1 \\
& \quad + text validation & 96.2 {\scriptsize\textcolor{green!50!black}{($\uparrow$84.2)}} & 100.0 {\scriptsize\textcolor{green!50!black}{($\uparrow$4)}} & 100.0 {\scriptsize\textcolor{green!50!black}{($\uparrow$88)}} & 96.2 & 73.8 {\scriptsize\textcolor{green!50!black}{($\uparrow$2.7)}} \\
& \quad + compilation & 100.0 {\scriptsize\textcolor{green!50!black}{($\uparrow$3.8)}} & 100.0 & 100.0 & 100.0 {\scriptsize\textcolor{green!50!black}{($\uparrow$3.8)}} & 71.5 {\scriptsize\textcolor{red!70!black}{($\downarrow$2.3)}} \\
& \quad + instantiation runtime & 100.0 & 100.0 & 100.0 & 96.2 {\scriptsize\textcolor{red!70!black}{($\downarrow$3.8)}} & 72.3 {\scriptsize\textcolor{green!50!black}{($\uparrow$0.8)}} \\
& \quad + success checking & 96.2 {\scriptsize\textcolor{red!70!black}{($\downarrow$3.8)}} & 100.0 & 96.2 {\scriptsize\textcolor{red!70!black}{($\downarrow$3.8)}} & 96.2 & 73.0 {\scriptsize\textcolor{green!50!black}{($\uparrow$0.7)}} \\
& \quad + bounds checking & 100.0 {\scriptsize\textcolor{green!50!black}{($\uparrow$3.8)}} & 100.0 & 100.0 {\scriptsize\textcolor{green!50!black}{($\uparrow$3.8)}} & 92.3 {\scriptsize\textcolor{red!70!black}{($\downarrow$3.9)}} & 73.1 {\scriptsize\textcolor{green!50!black}{($\uparrow$0.1)}} \\
& \quad + specialist agents (all gates on) & \textbf{100.0} & \textbf{100.0} & \textbf{100.0} & \textbf{100.0} {\scriptsize\textcolor{green!50!black}{($\uparrow$7.7)}} & \textbf{73.6} {\scriptsize\textcolor{green!50!black}{($\uparrow$0.5)}} \\
\midrule
\multirow{5}{*}{\textbf{Context Steering}} 
& None (all disabled) & 95.7 & 100.0 & 100.0 & 92.3 & 73.7 \\
& \quad + intent interpretation & 96.2 {\scriptsize\textcolor{green!50!black}{($\uparrow$0.5)}} & 100.0 & 96.2 {\scriptsize\textcolor{red!70!black}{($\downarrow$3.8)}} & 100.0 {\scriptsize\textcolor{green!50!black}{($\uparrow$7.7)}} & 73.8 {\scriptsize\textcolor{green!50!black}{($\uparrow$0.1)}} \\
& \quad + routing classification & 96.2 & 100.0 & 96.2 & 76.9 {\scriptsize\textcolor{red!70!black}{($\downarrow$23.1)}} & 72.5 {\scriptsize\textcolor{red!70!black}{($\downarrow$1.3)}} \\
& \quad + version history tracking & 96.2 & 100.0 & 96.2 & 100.0 {\scriptsize\textcolor{green!50!black}{($\uparrow$23.1)}} & 73.6 {\scriptsize\textcolor{green!50!black}{($\uparrow$1.1)}} \\
& \quad + reference selection (all gates on) & \textbf{100.0} {\scriptsize\textcolor{green!50!black}{($\uparrow$3.8)}} & \textbf{100.0} & \textbf{100.0} {\scriptsize\textcolor{green!50!black}{($\uparrow$3.8)}} & \textbf{100.0} & \textbf{73.6} \\
\bottomrule
\end{tabular}
}
\end{table*}

\label{sec:ablation}

We conduct a cumulative ablation study to quantify the functional contribution of each component in the task generation pipeline. For each module, we begin with all components disabled and progressively enable individual gates. This design disentangles changes in semantic task specification from improvements in robustness and correctness under session-level evaluation (Table~\ref{tab:ablation}).

\textbf{Metrics.}
We report complementary metrics capturing distinct failure modes. \textit{Compile} and \textit{Smoke Test} measure code correctness and execution stability; \textit{Task Success} measures end-to-end satisfaction of the success predicate; \textit{Human Verification} evaluates perceived task validity; and \textit{LLM Alignment} measures consistency between the natural language instruction and the implemented success condition.

\textbf{Evaluation setting.}
Ablations are evaluated on ten benchmark testcases, each consisting of multiple related tasks evaluated as a single session. Some testcases involve multi-stage task refinement via context-aware steering; additional details are provided in the Appendix.

\textbf{Task Proposal.}
Task proposal components primarily affect semantic grounding rather than executability. Enabling asset inference improves Human Verification (88.5 to 92.3) but slightly reduces LLM Alignment (74.0 to 71.5), while leaving execution metrics unchanged. Adding feasibility checking improves both Human Verification (92.3 to 96.2) and LLM Alignment (71.5 to 73.6) without affecting executability.

\textbf{Code Generation.}
Code generation components primarily improve robustness to systematic implementation errors. Across ablations, compilation and smoke test success remain near-perfect. API review, error checks, and in-context examples incrementally improve LLM Alignment (70.8 to 73.6) while preserving end-to-end executability.

\textbf{Validation.}
Validation is the dominant determinant of task correctness. With validation disabled, Task Success drops to 12.0 despite high compilation rates. Text-level validation alone recovers Task Success to 96.2, while the full validation stack achieves perfect Task Success, Compile, Smoke Test, and Human Verification.

\textbf{Context Steering.}
Context steering influences semantic coherence across multi-step task sessions. Intent interpretation and version history tracking improve Task Success, Human Verification, and LLM Alignment, while routing without history degrades semantic consistency. With full context steering enabled, execution metrics remain perfect.

\textbf{Summary.}
Overall, task proposal and context steering shape semantic intent and coherence, code generation improves robustness, and validation enforces correctness. Improvements in Task Success do not monotonically track alignment metrics, motivating a modular, gated design that balances expressiveness and executability.

\section{Discussions}
This work explores how robotic manipulation evaluation changes when task specification is opened to a broader set of contributors. By treating language as an executable interface, \methodname~allows users to express task intent, constraints, and success criteria directly, rather than relying on fixed, expert-authored benchmarks. In doing so, it reframes evaluation as a process shaped not only by models and metrics, but by the people defining what is being tested.

Language-driven evaluation becomes meaningful when grounded in a shared physical structure. Compiling language into explicit assets, initialization logic, and success predicates enables users to author and vary tasks in ways that remain reproducible and comparable. Within this structure, semantic differences in task descriptions translate into controlled differences in evaluation, allowing policies to be assessed across families of related tasks rather than isolated instances.

Lowering the barrier to task authoring also changes how evaluation spaces grow. Our results show that task diversity scales more strongly with contributor diversity than with task count alone, indicating that opening task specification to many users leads to broader and more complementary coverage of the task space. In this sense, \methodname~democratizes not only access to evaluation, but influence over what behaviors are examined.

We instantiate the framework in a deliberately constrained block manipulation domain to emphasize interpretability and control. Extending structured, language-driven evaluation to richer domains will require careful design, but the underlying principle remains: scalable evaluation benefits from being both structured and open to user-driven contribution.

\section*{Acknowledgments}

Yi Ru Wang is supported by the Natural Sciences and Engineering Research Council of Canada Postgraduate Scholarships – Doctoral program (NSERC-PGSD). This work was partially supported
by the National Science Foundation NRI program (\#2132848),
DARPA RACER (\#HR0011-21-C-0171), the Office of Naval Research (\#N00014-24-S-B001 and \#2022-016-01 UW), and the DEVCOM Army Research Laboratory (Award: W911NF-24-2-0191).
We gratefully acknowledge support from Amazon and the Allen
Institute for Artificial Intelligence (AI2), as well as gifts from
Collaborative Robotics, Cruise, and other industry partners.

%% Use plainnat to work nicely with natbib. 

\bibliographystyle{plainnat}
\bibliography{references}

\newpage

\clearpage
\onecolumn  
\appendix
\setcounter{tocdepth}{4}
\section*{}
\addcontentsline{toc}{section}{Appendix}

\localtableofcontents
\clearpage

\setlength{\parindent}{0pt}
% \AddToHook{cmd/subsection/after}{\noindent}
\setlength{\parskip}{1em}

% \setcounter{section}{0}

% \setcounter{tocdepth}{4}
% \addcontentsline{toc}{section}{Appendix}
\subsection{System Explanation Expanded}
\label{sec:appendix-system}
\subsubsection{System Modules}
\label{sec:appendix-modules}

In this section, we elaborate on the module components and visualize the relationship flow of our multi-stage task generation system based on Figure~\ref{fig:system_overview}.

\begin{figure}[H]
\centering
\makebox[\textwidth][c]{\includegraphics[width=\textwidth]{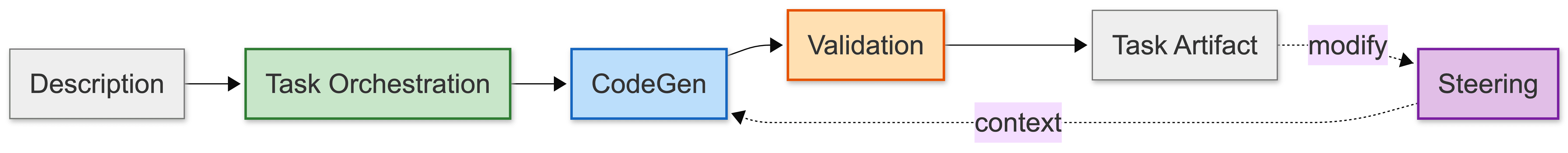}}
\caption{\textbf{Overall Pipeline.} Task descriptions flow through Task Orchestration, CodeGen, and Validation to produce Task Artifacts. User modifications trigger the Steering module, which provides context for regeneration.}
\label{fig:overall-pipeline}
\end{figure}

% ----------------------------------------------------------------------------
% Overall Pipeline
% ----------------------------------------------------------------------------
\textbf{Overall Pipeline.} Figure~\ref{fig:overall-pipeline} shows a high-level flow diagram of the task and scene generation system. The user-created task descriptions flow through four main modules: Task Orchestration, Code Generation, Validation, and Steering. 

\begin{figure}[H]
\centering
\makebox[\textwidth][c]{\includegraphics[width=\textwidth]{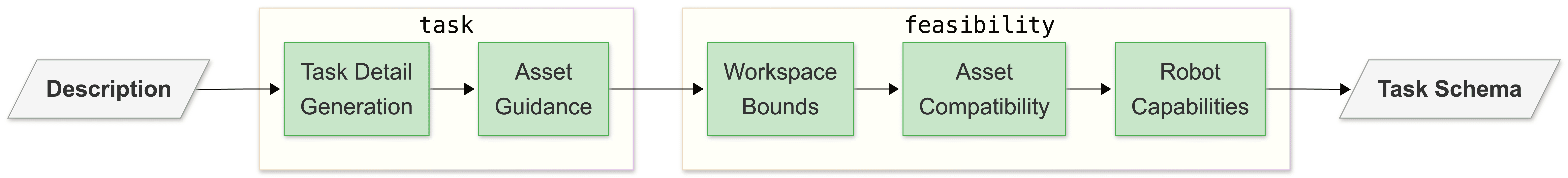}}
\caption{\textbf{Task Orchestration.} Two components: \texttt{task} (detail generation and asset guidance) and \texttt{feasibility} (workspace, asset, and robot capability checks).}
\label{fig:task-orchestration}
\end{figure}

% ----------------------------------------------------------------------------
% Task Orchestration Module
% ----------------------------------------------------------------------------
\textbf{Task Orchestration.} The Task Orchestration module (Figure~\ref{fig:task-orchestration}) transforms natural language descriptions into a structured task schema that is parsed and injected into the Code Generation agent. It consists of two components: \texttt{task} (task curation and asset guidance) and \texttt{feasibility} (workspace bounds, asset compatibility, and robot capabilities checking).

\begin{figure}[H]
\centering
\makebox[\textwidth][c]{\includegraphics[width=\textwidth]{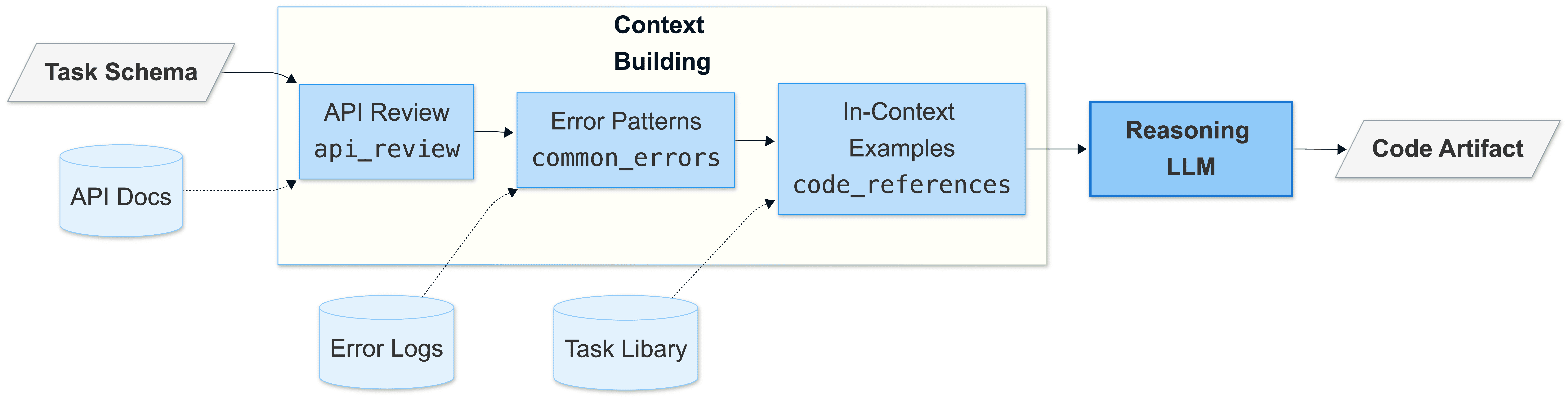}}
\caption{\textbf{Program Synthesis Module.} Three context components feed the Reasoning LLM: \texttt{api\_review} (API Docs), \texttt{common\_errors} (Error Logs), and \texttt{code\_references} (Task Library).}
\label{fig:program-synthesis}
\end{figure}

% ----------------------------------------------------------------------------
% Program Synthesis (CodeGen) Module
% ----------------------------------------------------------------------------
\textbf{Code Generation.} The code generation module is designed to (Figure~\ref{fig:program-synthesis}) produce task-aligned executable MuJoCo task code. Code synthesis is grounded using three context sources: API documentation, a catalog of common error patterns, and reference implementations selected based on semantic similarity and manipulation structure. Structured analyses injected into synthesis prompts include spatial analysis for face visibility, capability analysis for asset-specific APIs, and geometric constraints such as object dimensions and workspace bounds.

\begin{figure}[H]
\centering
\makebox[\textwidth][c]{\includegraphics[width=\textwidth]{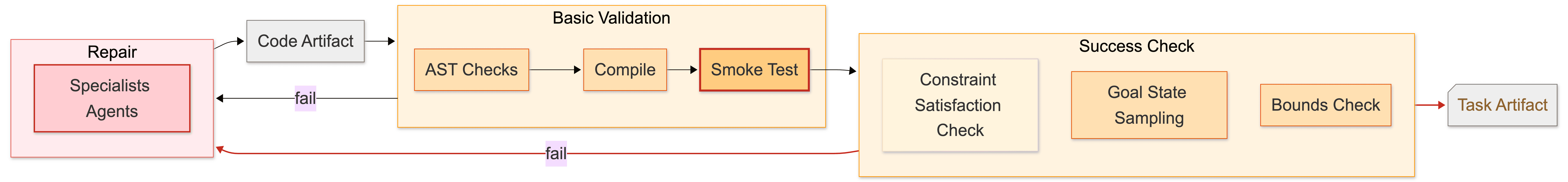}}
\caption{\textbf{Validation Module.} Sequential checks across Basic Validation (AST, Compile, Smoke Test) and Success Check (Constraint Satisfaction, Goal State Sampling, Bounds). Failures route to Specialist Agents for repair.}
\label{fig:validation}
\end{figure}

% ----------------------------------------------------------------------------
% Validation Module
% ----------------------------------------------------------------------------
\textbf{Validation.} The Validation module (Figure~\ref{fig:validation}) performs sequential checks across Basic Validation and Success Check stages. Basic Validation includes Abstract-Syntax Tree (AST) Checks, Compilation, and smoke tests by instantiating and stepping through the environment. Success Check verifies logical correctness with Constraint Satisfaction and Goal State Sampling, and valid geometric areas with a Bounds Check that verifies the blocks are in our defined workspace. Failures at any stage passes information to an Agent Orchestrator that routes targeted validation error information (object poses, subgoals passed, traceback messages) to Specialist Fix Agents for task code repair.

\begin{figure}[H]
\centering
\makebox[\textwidth][c]{\includegraphics[width=\textwidth]{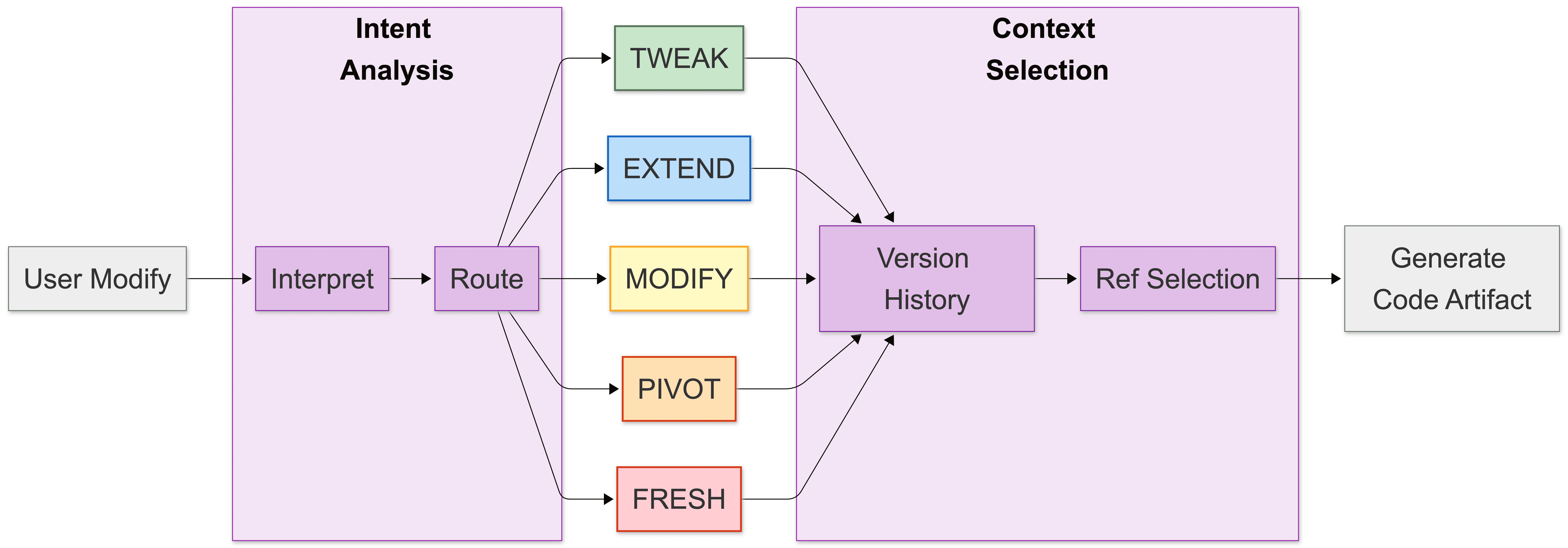}}
\caption{\textbf{Context Steering Module.} Intent Analysis classifies user modifications into five categories (Tweak, Extend, Modify, Pivot, Fresh). Context Selection uses Version History and Reference Selection to generate updated Code Artifacts.}
\label{fig:context-steering}
\end{figure}

% ----------------------------------------------------------------------------
% Context Steering Module
% ----------------------------------------------------------------------------
\textbf{Context Steering.} The Context Steering module (Figure~\ref{fig:context-steering}) 
handles human-in-the-loop capabilities for task modification and refinement through text-based instructions. Intent Analysis classifies user requests into five categories (Tweak, Extend, Modify, Pivot, Fresh), determining whether to 
build upon an existing task or generate a new one. Context Selection then identifies 
the appropriate version from the task history to reference for code generation.

\subsubsection{Simulation Configuration}
\label{app:simulation}

All tasks execute in MuJoCo using a standardized configuration to ensure reproducibility across evaluation sites. The table surface is fixed at height $z = 0.95$\,m. Objects are spawned within workspace bounds
\[
x \in [0.40, 0.70]\,\text{m}, \quad y \in [-0.25, 0.25]\,\text{m}.
\]
A fixed camera viewpoint is used for all tasks to ensure consistent observations and deterministic visibility checks. Physics parameters, including gravity, friction, and solver settings, are fixed globally and are not task-dependent.

\subsubsection{Geometric Visibility and Face Readability}
\label{app:visibility}

SemanticCube tasks require verifying both geometric visibility and perceptual readability of symbolic labels.

\textbf{Geometric Visibility.} In our structured physical domain, three-dimensional configurations may contain faces that are occluded by other blocks. Since occluded faces cannot be perceived, we restrict semantic verification to exterior faces visible from the camera. For example, a task requiring a 3D tower with alphabetically ordered blocks is verified only on visible exterior faces; occluded faces contribute solely to spatial and physical constraints without semantic requirements.

Visibility is determined via ray casting from the camera to five sample points on each cube face: the face center and four corner points offset at 80\% of the face extent. A face is considered visible if at least three of the five rays are unobstructed. 

For coplanar arrangements, all TOP faces are considered visible. For vertical stacks, BACK faces are considered visible by construction. These analytical shortcuts avoid unnecessary ray casting for common configurations.

\textbf{Face Readability.} Readable faces must satisfy two constraints: (i) face normal alignment with the expected viewing direction with cosine similarity at least 0.97 (approximately $14^\circ$), and (ii) in-plane glyph orientation alignment with cosine similarity at least 0.94 (approximately $20^\circ$). These thresholds were tuned empirically to balance false rejections against visually ambiguous acceptances.

\subsubsection{Task Proposal Feasibility Reasoning}
\label{app:feasibility}

Before code synthesis, task proposals undergo feasibility pre-validation. The feasibility agent performs four steps: asset analysis, requirement extraction, feasibility checking, and repair suggestion.

Asset analysis enforces limits such as at most 26 letters and 10 digits for SemanticCubes. Feasibility checking evaluates whether ordering, spatial, and visibility constraints can be satisfied simultaneously. When violations occur, the agent proposes structured repairs including \textsc{Reduce\_Labels\_Preserve\_Geometry}, \textsc{Switch\_Ordering\_Pattern}, and \textsc{Switch\_To\_Spatial\_Only}.

\subsubsection{Validation Parameters and Stability Checks}
\label{app:validation}

\textbf{Physics Settling.} During goal-state verification, tasks are simulated forward for 50 steps with zero action to allow contacts to settle. The success predicate must evaluate to true after settling.

\textbf{Extended Stability.} An additional 50 simulation steps are used to verify stability. Objects must not drift more than 1\,cm from their settled positions. Vertical displacements exceeding 2\,cm indicate toppling and result in validation failure.

\subsubsection{Validating Task Goal Logic}        \label{app:csp}                                               
  Before physics simulation, goal predicates extracted from the \texttt{\_success()} method undergo constraint satisfaction analysis to detect logically infeasible configurations. The validator constructs a        
  directed graph from support relationships (e.g., \texttt{On(block\_A, block\_B)}) and performs three checks.                                                           
  First, cycle detection identifies circular dependencies such as object A supports B, B supports C, and C supports A -- a physically impossible configuration. Second, support completeness verifies that every movable   
  object has a valid support path terminating at a fixed surface object (table). Third, transitivity analysis flags redundant predicates that can be inferred from existing relationships, reducing unnecessary       
  constraint complexity. Base surfaces, including \texttt{table}, \texttt{floor}, and \texttt{ground} are treated as grounded nodes in the graph. When the analysis detects infeasibility, the error is routed to a specialist constraint repair agent, which proposes predicate modifications such as reordering the support chain or removing conflicting relationships. This lightweight graph-based check quickly catches task logic errors before    
  incurring the time and computing cost of continuously sampling object goal positions in physics simulation.  

\subsubsection{Iterative Repair and Agent Taxonomy}
\label{app:repair}

Validation failures are routed to specialized repair agents: \texttt{SyntaxFixAgent}, \texttt{APIUsageFixAgent}, \texttt{RuntimeFixAgent}, \texttt{SuccessCheckFixAgent}, \texttt{StructureStabilityFixAgent}, and \texttt{GeometricBoundsFixAgent}. The orchestrator allows up to five validation--repair cycles, with up to three attempts per agent.

Agents receive previous failed strategies as negative examples to prevent oscillatory fixes. Algorithm~1 in the main text describes the orchestration logic.

\subsubsection{Version History Representation}
\label{app:versioning}

Multi-turn task refinement requires tracking task evolution to enable context-aware modifications. Without version history, each steering request would be interpreted in isolation, losing accumulated constraints and preventing users from referencing prior task states (e.g., ``go back to the 3-block version''). We address this with a structured versioning system comprising two dataclasses: \texttt{TaskSnapshot} for immutable version records and \texttt{SteeringMeta} for evolution tracking.

\paragraph{TaskSnapshot Fields.}
Each validated task version is captured as a \texttt{TaskSnapshot} with five fields:
\begin{itemize}
    \item \textbf{version\_id}: Sequential integer (0 = base task) enabling explicit back-references
    \item \textbf{description}: Human-readable summary of what changed (e.g., ``Added green block on top'')
    \item \textbf{assets\_used}: List of asset class names (e.g., \texttt{["ColoredCube", "SemanticCube"]}) for compatibility checking
    \item \textbf{goal\_summary}: Natural language description of terminal success conditions for semantic matching
    \item \textbf{code\_hash}: SHA-256 hash of generated code enabling deduplication and cache lookups
\end{itemize}

\paragraph{Asset Compatibility.}
Asset incompatibility arises when a steering request requires capabilities absent from the current asset type. For example, ``sort alphabetically'' requires visible labels (SemanticCube), while the current task uses ColoredCube (visible colors only). When such incompatibility is detected, the router searches prior snapshots for a version using compatible assets, enabling requests like ``go back to the letter version and sort it.''

\paragraph{Reference Resolution Algorithm.}
Given a user request and task history, the SteeringRouter determines the reference version through the following process:

\begin{enumerate}
    \item \textbf{Intent Classification}: An LLM classifies the request into one of five categories:
    \begin{itemize}
        \item \textsc{Tweak}: Parameter changes only (colors, counts) $\rightarrow$ reference current version
        \item \textsc{Extend}: Additive modifications $\rightarrow$ reference current version
        \item \textsc{Modify}: Property changes within existing structure $\rightarrow$ reference current version
        \item \textsc{Pivot}: Structural rewrites $\rightarrow$ reference current version but overwrite goals
        \item \textsc{Fresh}: Unrelated new task $\rightarrow$ no reference
    \end{itemize}

    \item \textbf{Explicit Reference Detection}: The LLM identifies explicit back-references in the request:
    \begin{itemize}
        \item Object references: ``the red block,'' ``the previous pyramid''
        \item Version references: ``go back to,'' ``the earlier version''
        \item Additive language: ``also,'' ``add to it'' $\rightarrow$ implies current version
    \end{itemize}

    \item \textbf{Snapshot Search}: If the request references a non-current state, the router iterates through \texttt{task\_history} matching:
    \begin{itemize}
        \item Asset types mentioned in the request against \texttt{assets\_used}
        \item Goal descriptions against \texttt{goal\_summary} via LLM-based semantic comparison
        \item Explicit version numbers if provided (``version 2'')
    \end{itemize}

    \item \textbf{Preservation Flags}: Based on intent, the router sets boolean flags (\texttt{preserve\_assets}, \texttt{preserve\_positions}, \texttt{preserve\_goals}) that guide code generation.
\end{enumerate}

\paragraph{Code Hash Utility.}
The \texttt{code\_hash} field enables two optimizations: (a) cache lookups to avoid regenerating identical tasks, and (b) detecting when a steering sequence loops back to a prior state, allowing the system to warn users or reuse cached validation results.de hash. When incompatible assets are requested, prior snapshots are searched to identify a suitable reference.

% Carter: System component Figures -> Should model ablation study components  
% Task Orchestration + Grounding + Task Schema
% Code Generation: API Review, Task Retrieval (In-Context Examples), Common Errors
% Validation Pipeline: Basic Checks, Syntax, bounds checks, Runtime, goal sampling, CSP cyclic checks
% Steering: Intent Classification, Preserving/Replacing Assets
%
\subsubsection{In-Context Examples}
We curate a library of 10 human-authored reference tasks spanning the taxonomy dimensions, as shown in Figure \ref{fig:task_overview_1} and Figure \ref{fig:task_overview_2}. For each generation request, we retrieve 2-3 relevant examples using LLM-based selection across four criteria: a) asset class, b) reasoning type, c) complexity level, and d) task action primitives.             

\textbf{Asset class.} Asset class matching ensures the retrieved examples use the same object types as the target task, preventing API hallucination errors that arise when the model references methods unavailable on the target asset. 

\textbf{Reasoning.} Reasoning type distinguishes semantic tasks, which verify symbolic state such as label orientation and sequence validity, from spatial tasks that check geometric predicates like position tolerance and alignment. 

\textbf{Complexity.} Complexity level captures whether the task requires atomic single-predicate success conditions, compositional multi-predicate conjunctions, or hierarchical staged sub-goals with progress tracking. 

\textbf{Task Actions.} Manipulation primitive matching selects examples with similar physical actions—stacking, arranging, rotating, or pushing—to provide correct goal state geometry patterns.                                           

These criteria target the three primary failure modes observed during development: a) API hallucination from mismatched asset examples, b) incorrect success predicates from reasoning type confusion, and c) malformed goal states from inappropriate manipulation patterns. The selected implementations are injected as reference code to guide stylistic consistency and reduce domain drift.  

% Insert Figure list of Examples Tasks in Goal State, Task Description
% Place \newcommand in preamble or before \begin{document}
% \newcommand{\taskcard}[4]{...}
% If already defined, this will avoid errors:
\providecommand{\taskcard}[4]{%
  \begin{minipage}[t]{0.48\textwidth}
    \centering
    \begin{tabular}{@{}c@{\hskip 2pt}c@{\hskip 2pt}c@{}}
      \includegraphics[width=0.44\linewidth,trim=80 80 80 80,clip]{#3} &
      \raisebox{1.8cm}{\Large$\boldsymbol{\Rightarrow}$} &
      \includegraphics[width=0.44\linewidth,trim=80 80 80 80,clip]{#4}\\[-2pt]
      {\scriptsize Initial State} & & {\scriptsize Goal State}
    \end{tabular}\\[3pt]
    \textbf{#1}\\[1pt]
    {\small\textit{#2}}
  \end{minipage}%
}

% ──────────────────────────────────────────────
% Page 1: Tasks 1–6
% ──────────────────────────────────────────────
\begin{figure*}[htbp]
\centering

% ── Row 1 ──
\taskcard{Stack Blocks}{Stack two blocks on top of each other.}
  {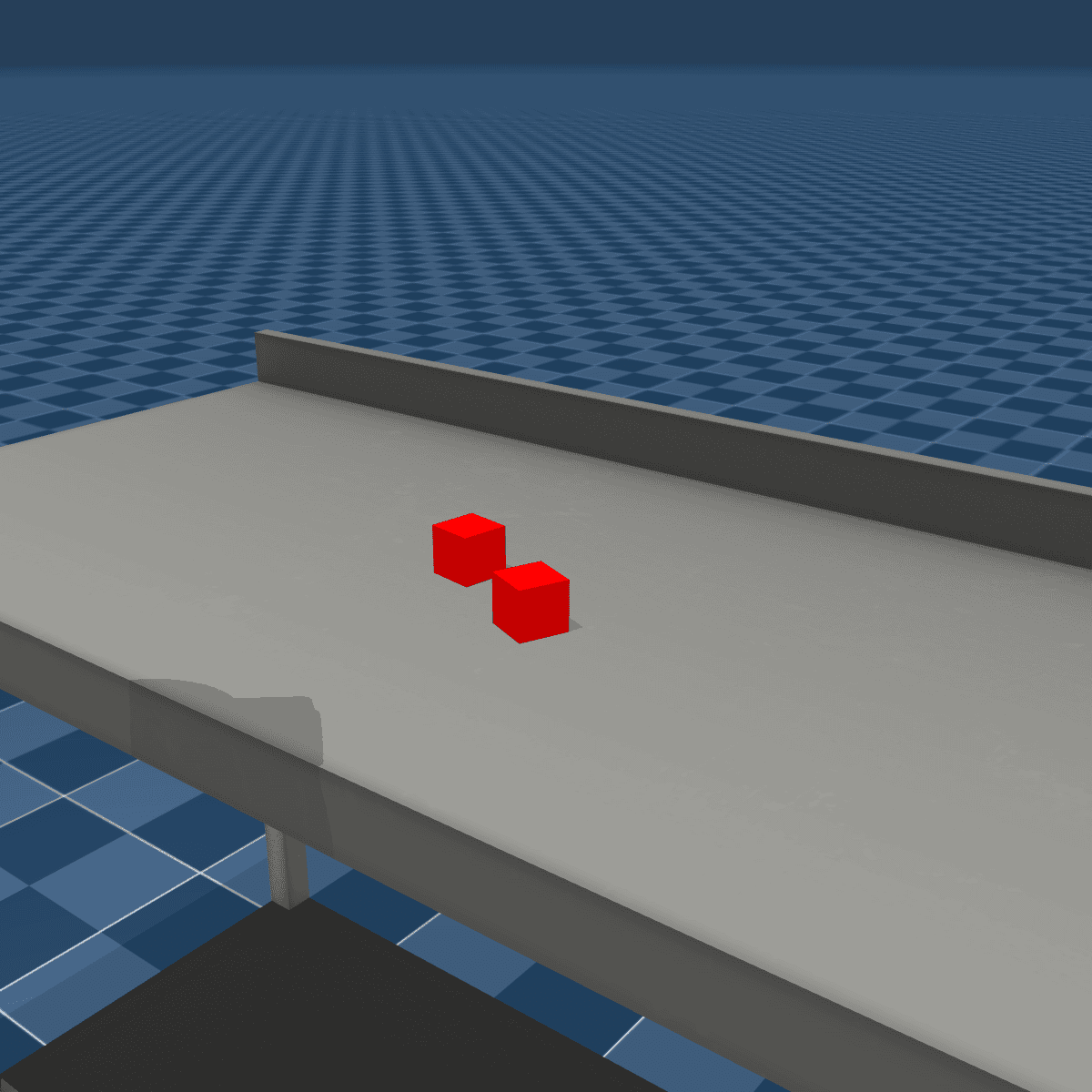}
  {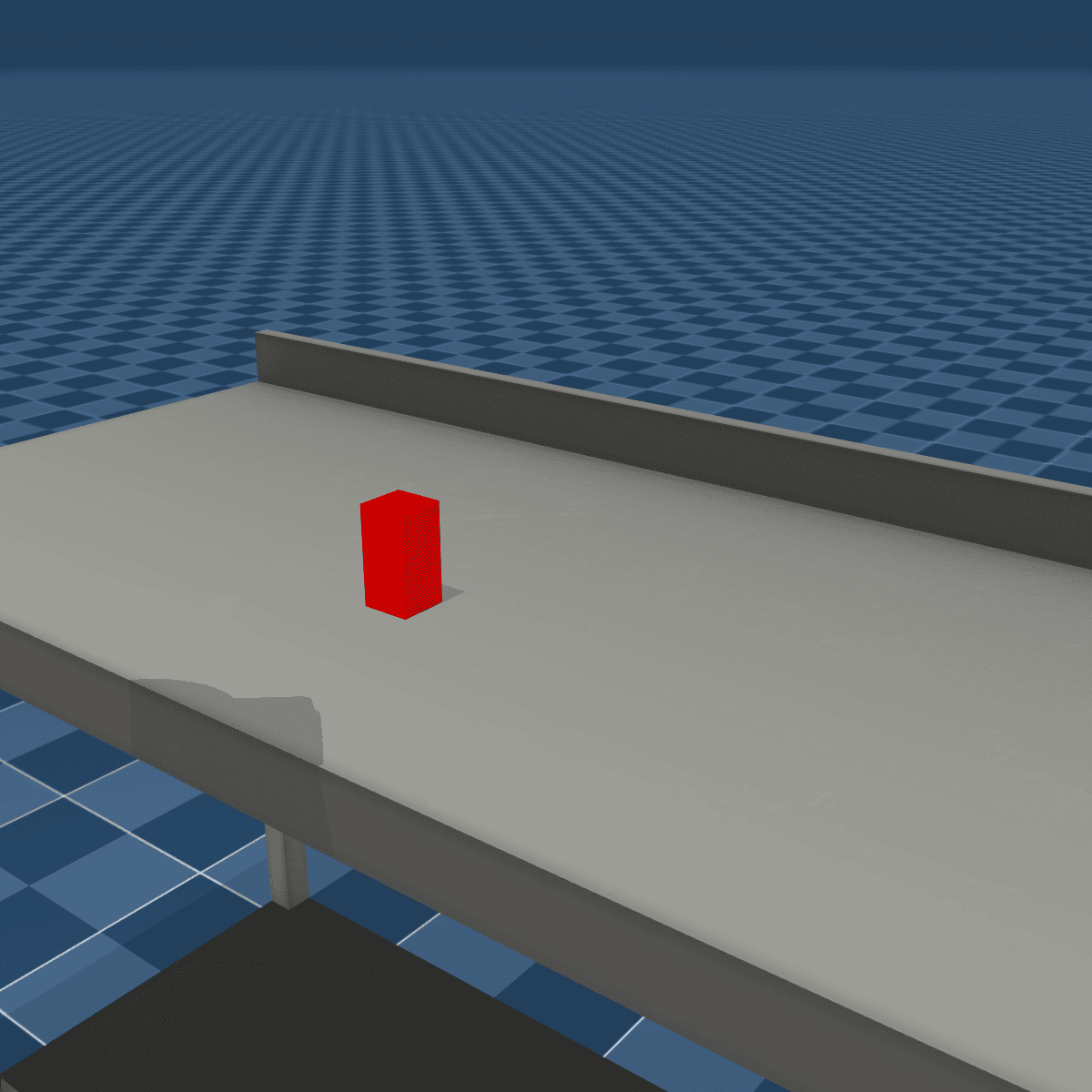}
\hfill
\taskcard{Stack Blocks on Target}{Stack blocks on the target square patch designated on the table.}
  {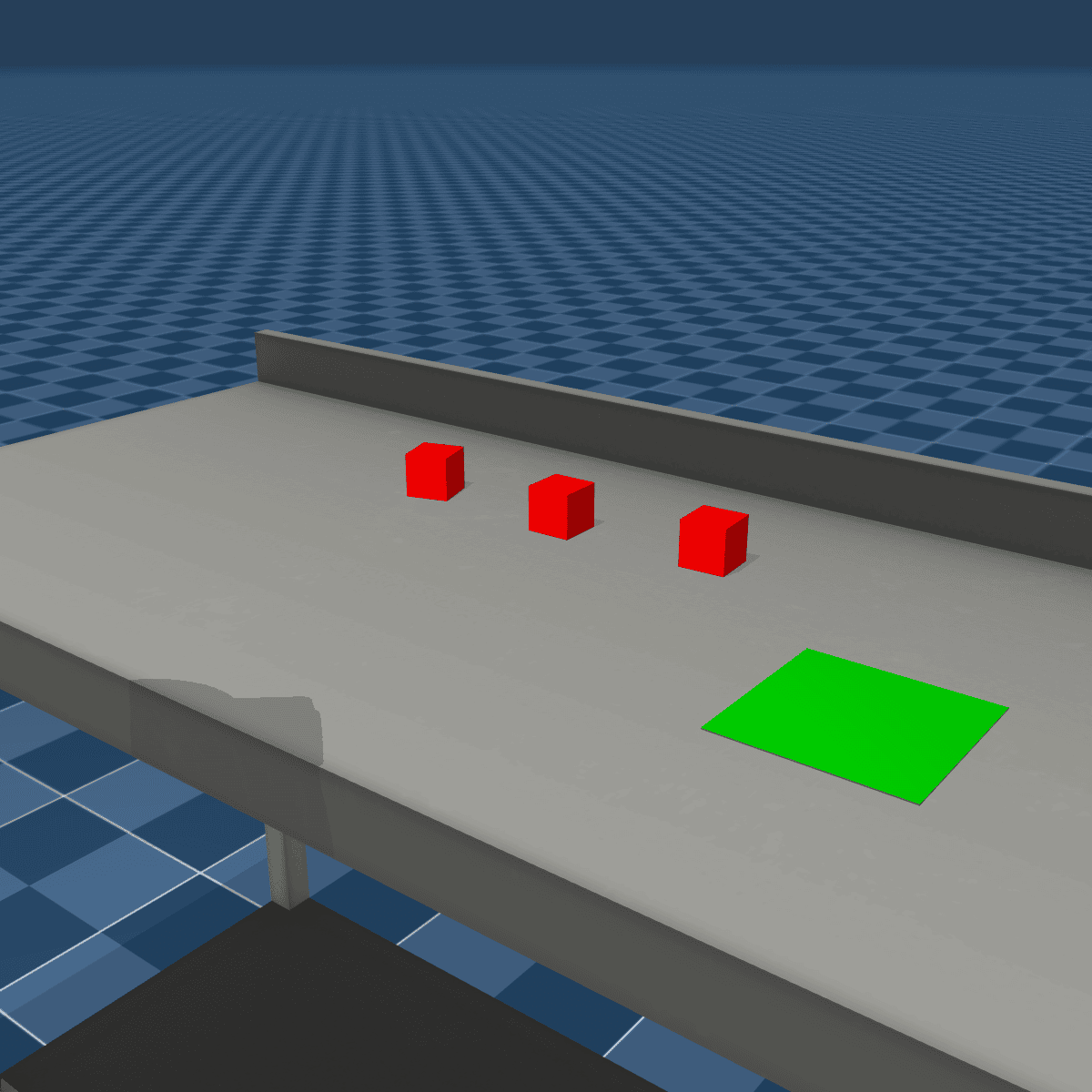}
  {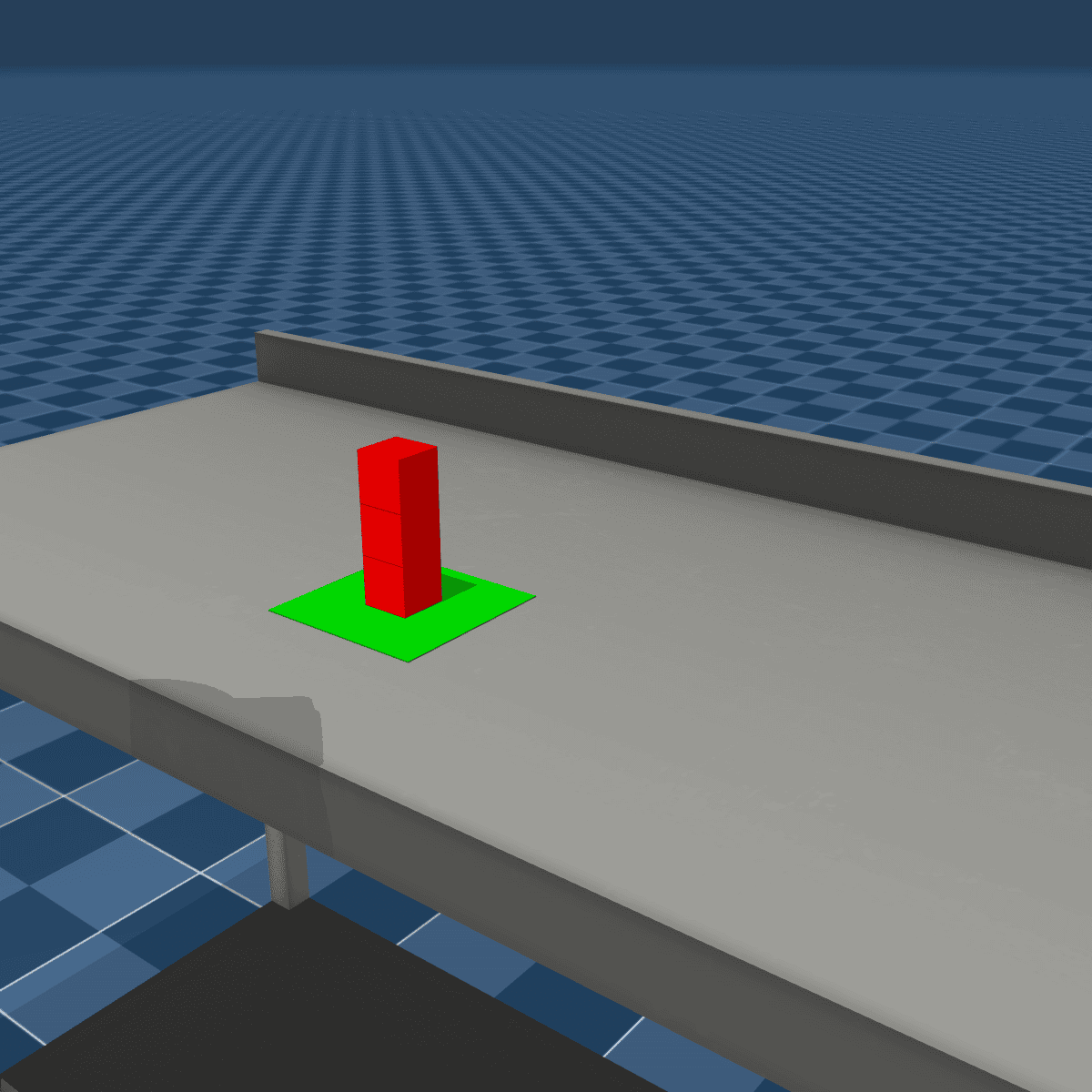}

\vspace{8pt}

% ── Row 2 ──
\taskcard{Align Blocks}{Align N cubic blocks in a single straight horizontal line.}
  {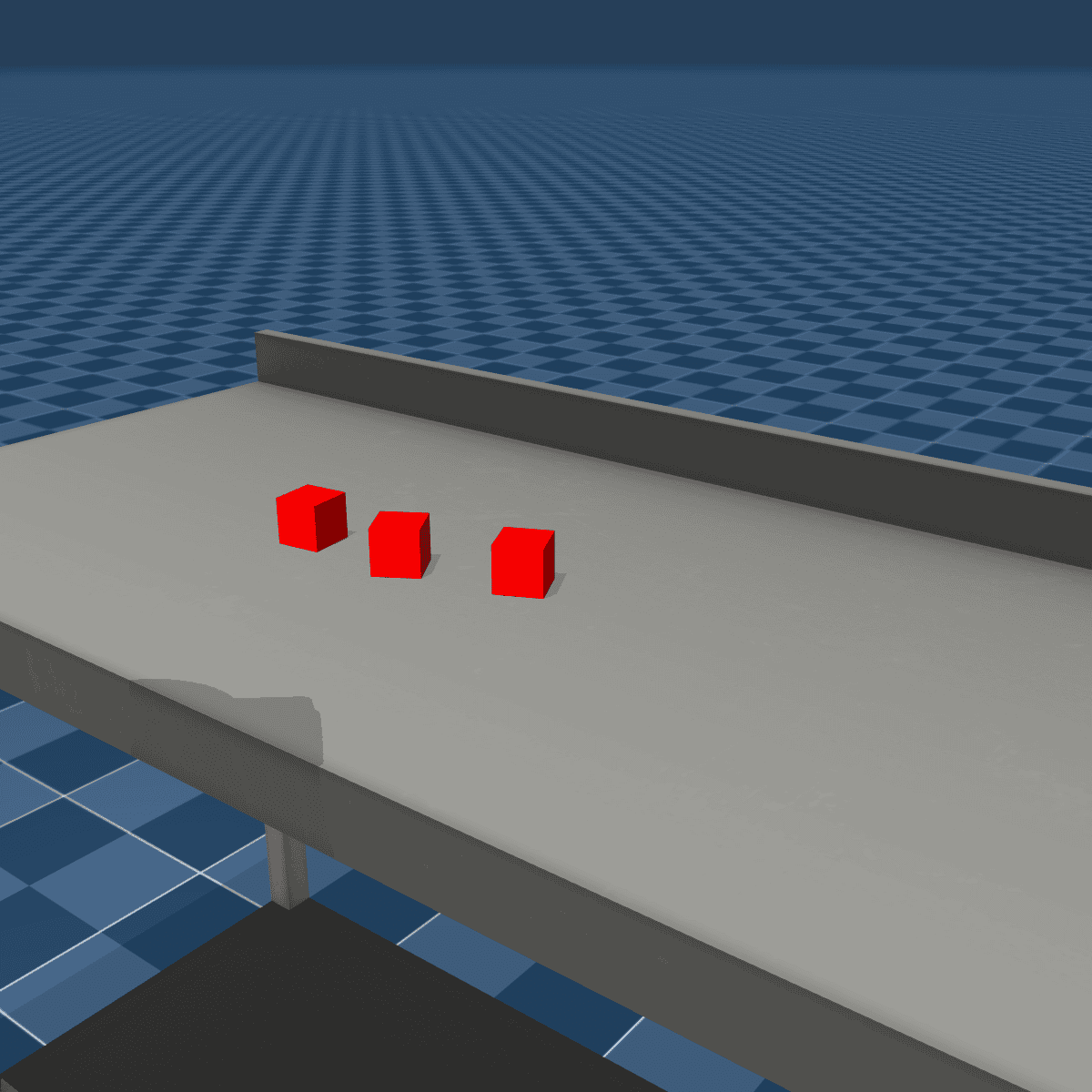}
  {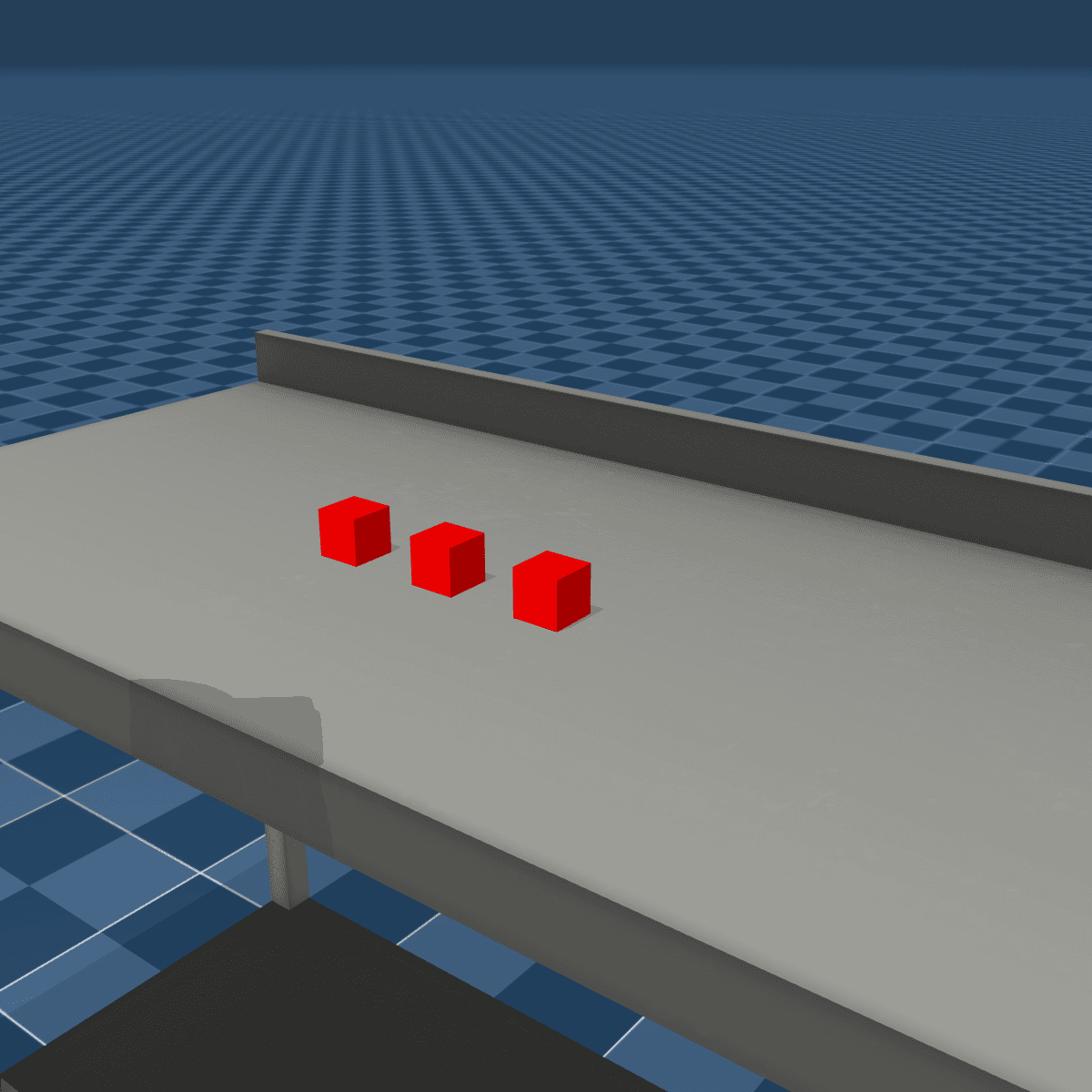}
\hfill
\taskcard{Arrange Letters}{Arrange cubes so that their up-faces read A, B, C\ldots{} in alphabetical order.}
  {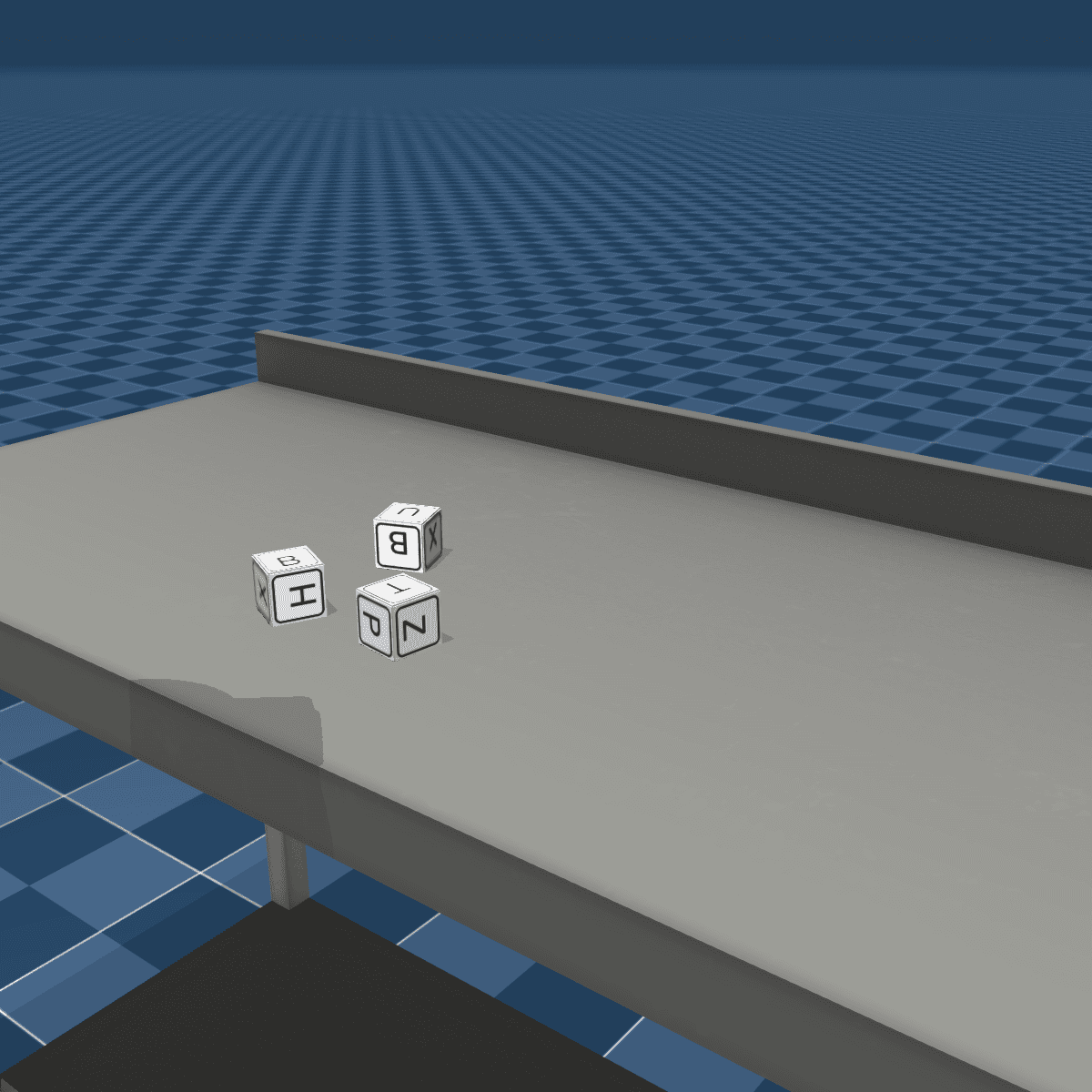}
  {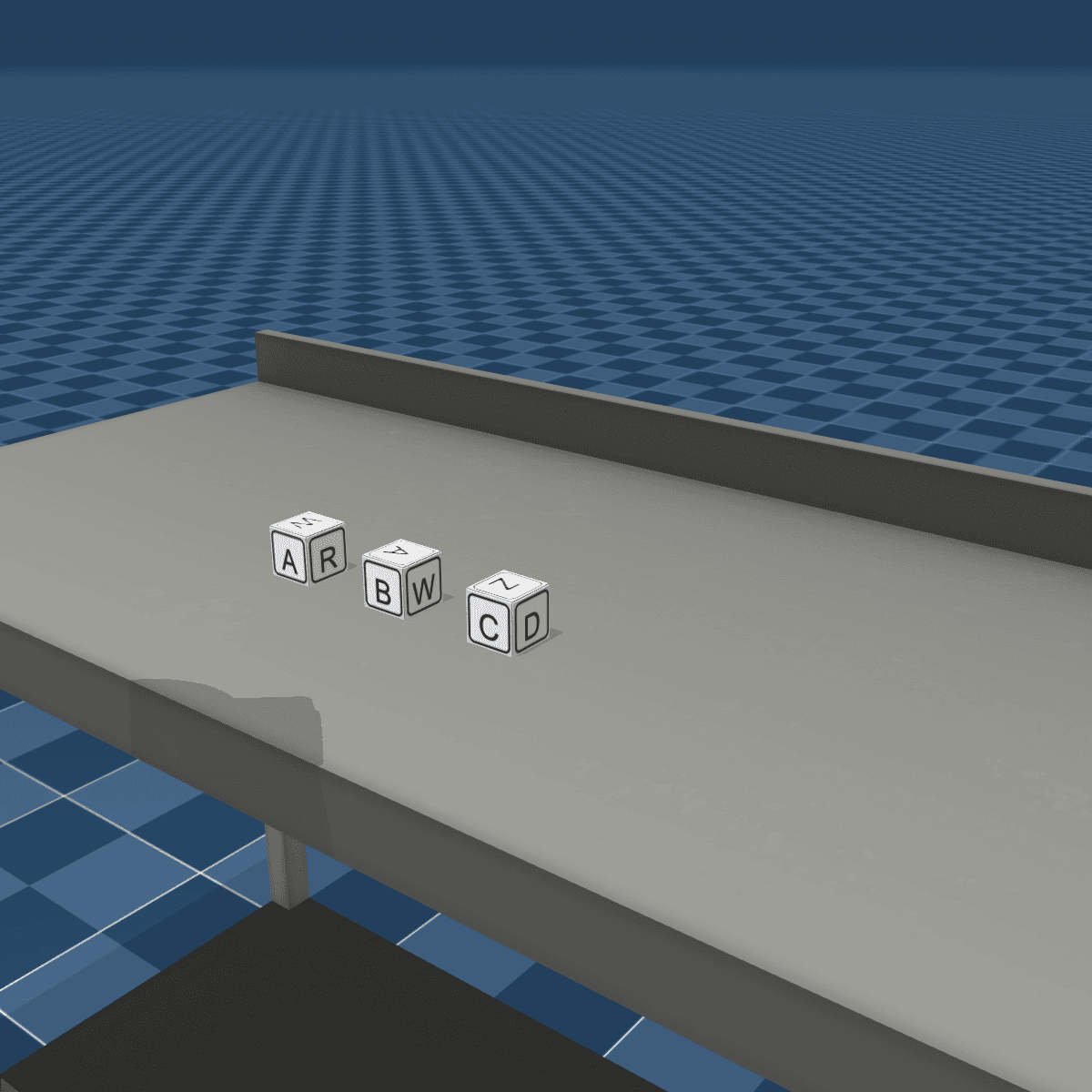}

\vspace{8pt}

% ── Row 3 ──
\taskcard{Arrange Word}{Arrange cubes so that they spell a word from left to right.}
  {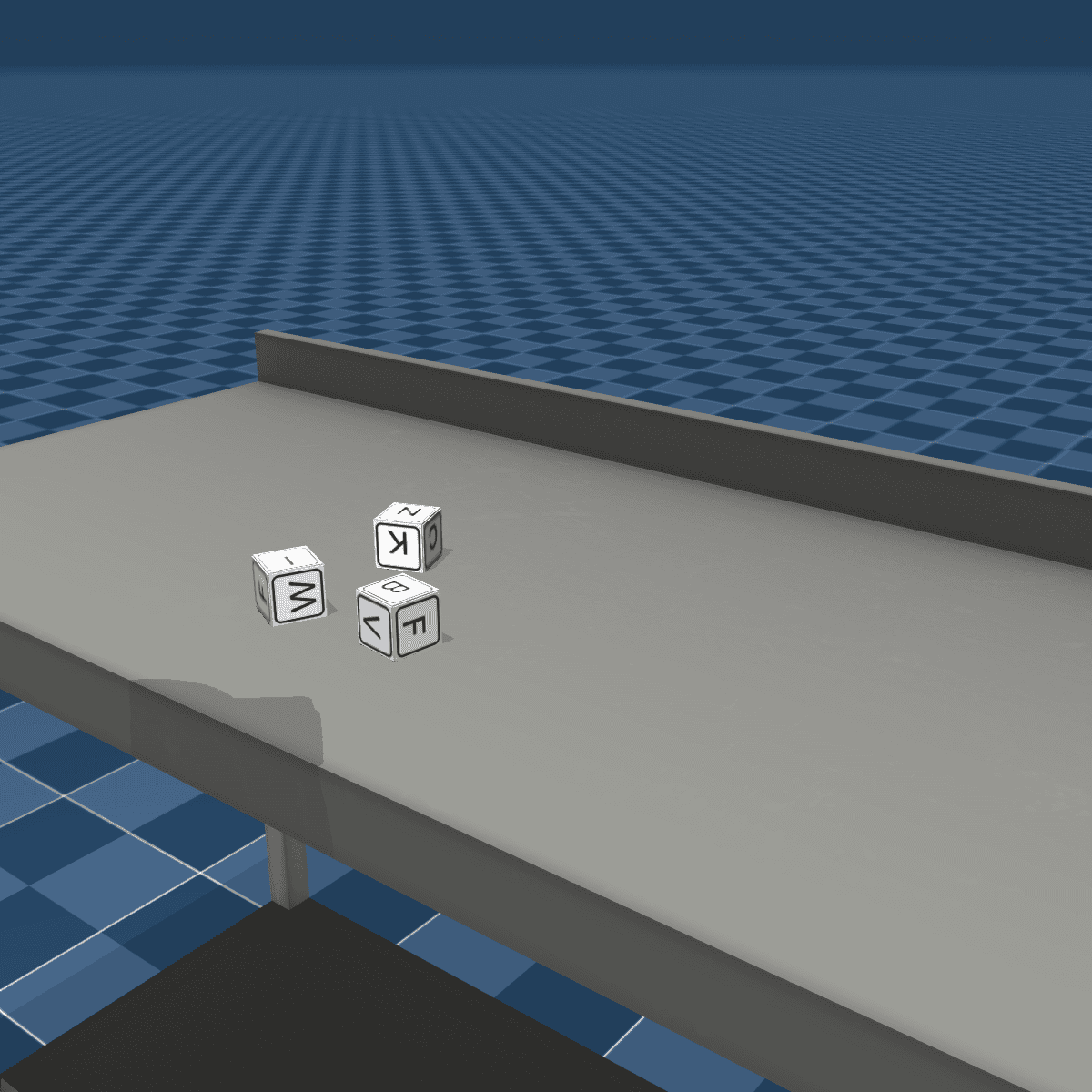}
  {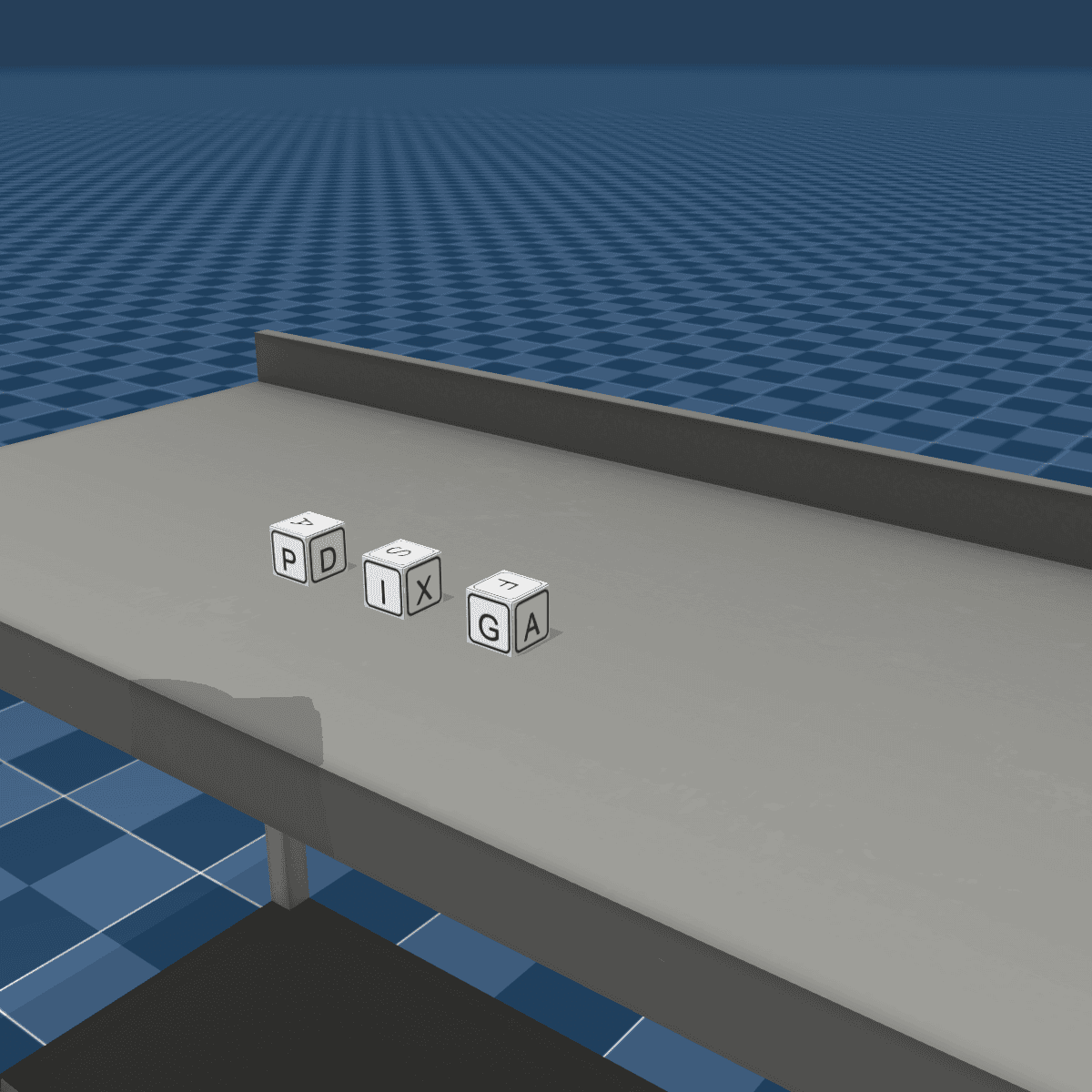}
\hfill
\taskcard{Arrange Numbers}{Arrange cubes in strictly increasing numerical order.}
  {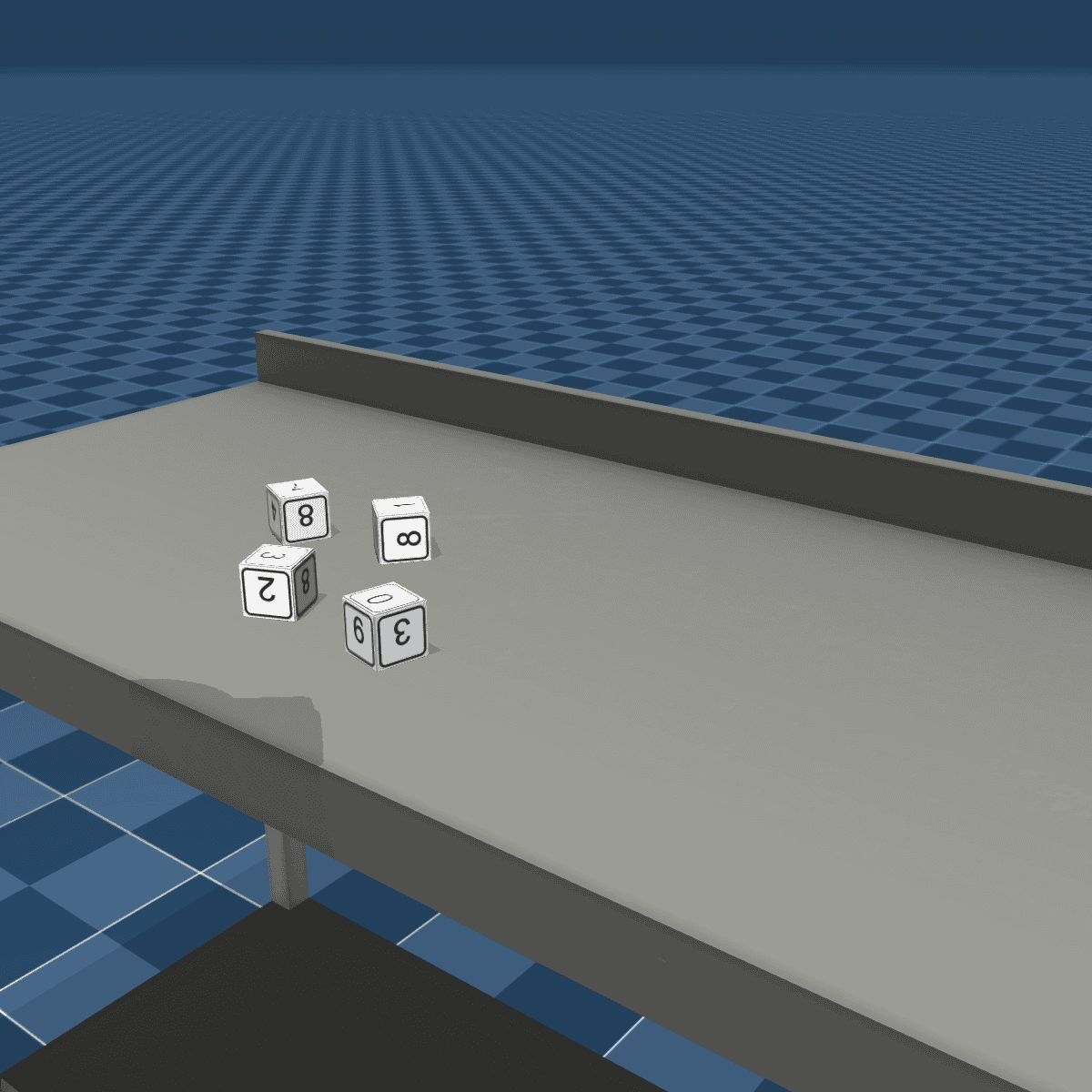}
  {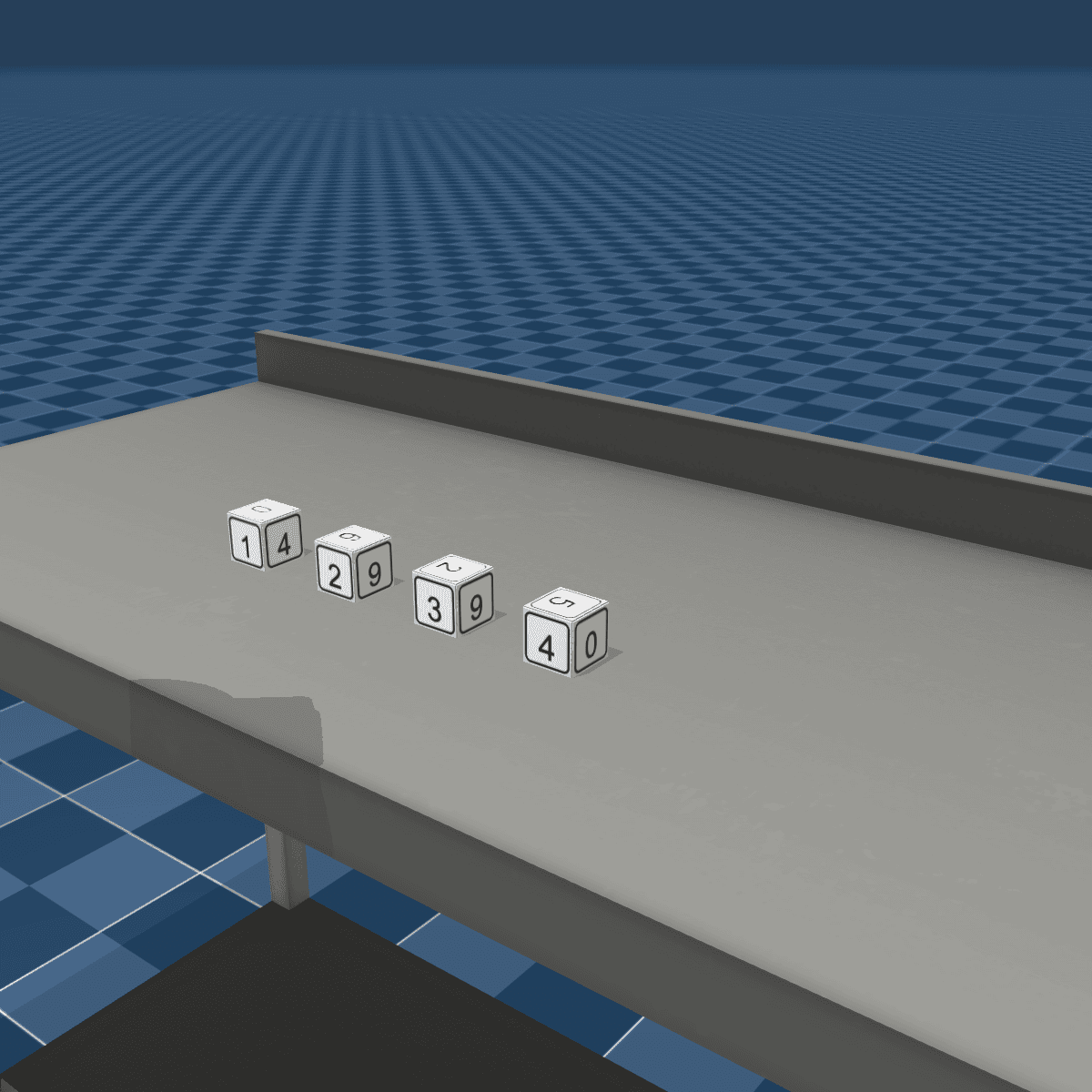}

\caption{\textbf{In-Context Reference Task Suite (1/2).} Each pair shows the initial randomized state (left) and goal configuration (right). Tasks shown: pick-and-place (\textit{Stack Blocks}), target-driven placement (\textit{Stack Blocks on Target}), spatial alignment (\textit{Align Blocks}), and semantic ordering (\textit{Arrange Letters}, \textit{Arrange Word}, \textit{Arrange Numbers}).}
\label{fig:task_overview_1}
\end{figure*}

% ──────────────────────────────────────────────
% Page 2: Tasks 7–10
% ──────────────────────────────────────────────
\begin{figure*}[htbp]
\centering

% ── Row 4 ──
\taskcard{Arrange Equation}{Arrange cubes into a correct single-digit math equation left to right.}
  {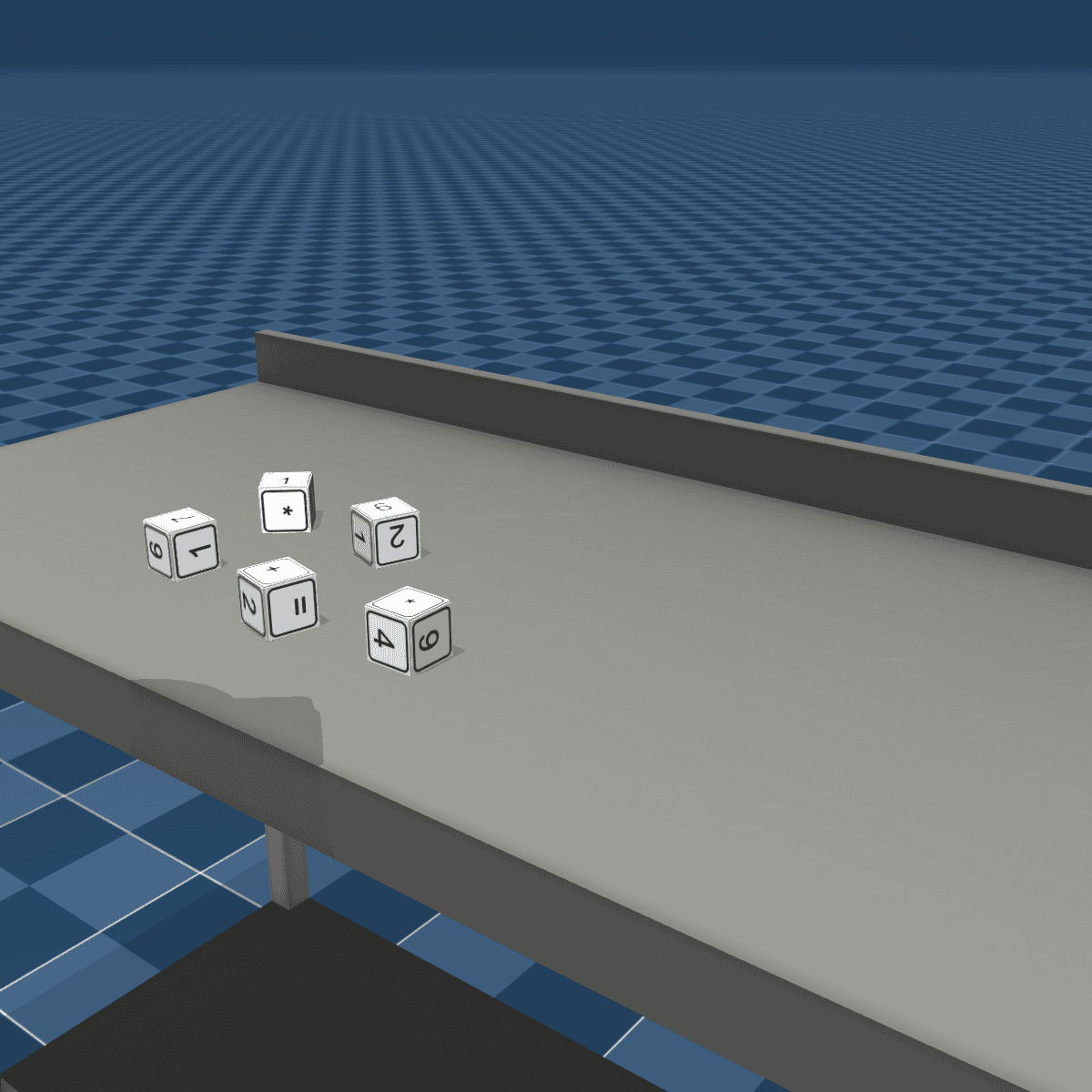}
  {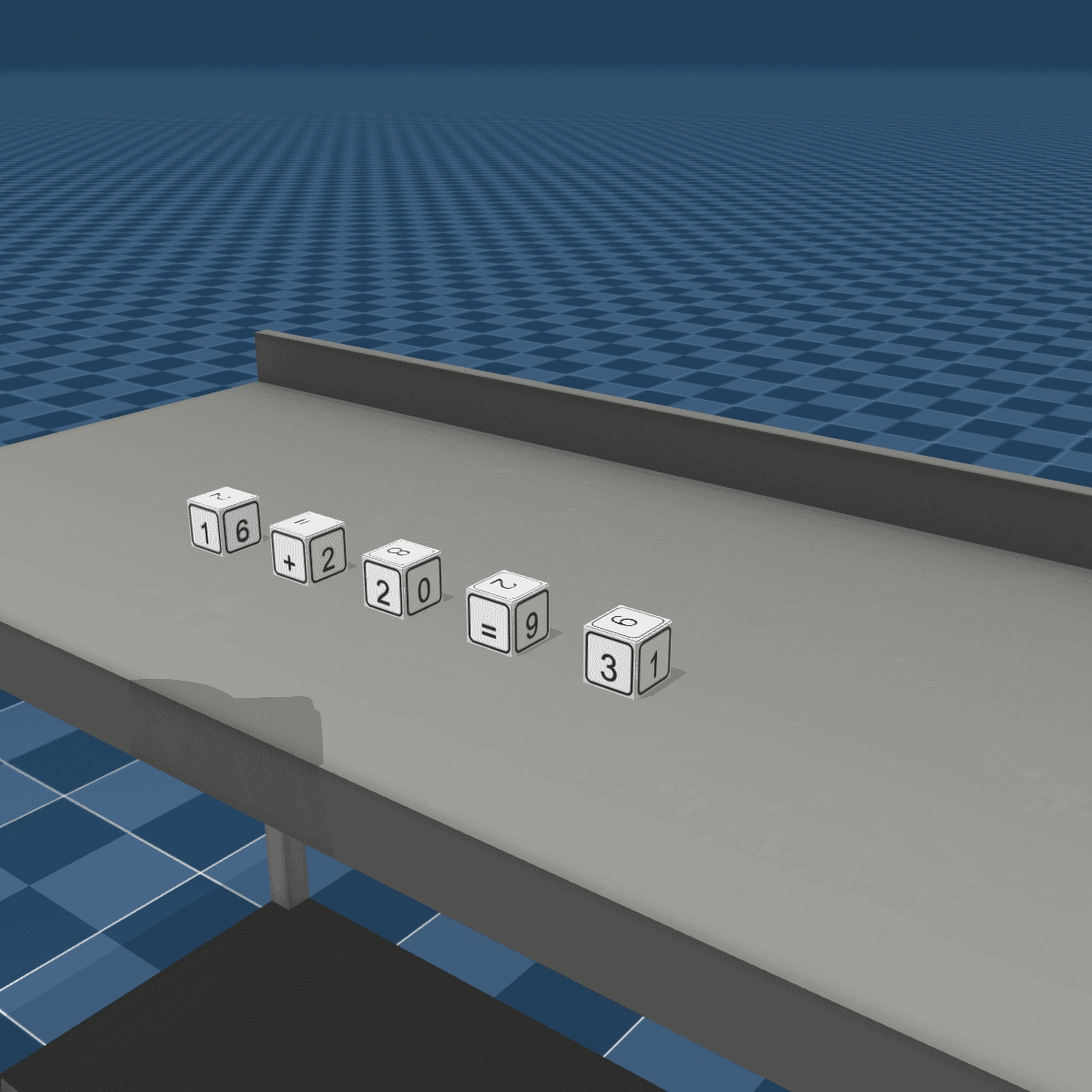}
\hfill
\taskcard{Arrange Shapes}{Place cubes in a circular arrangement.}
  {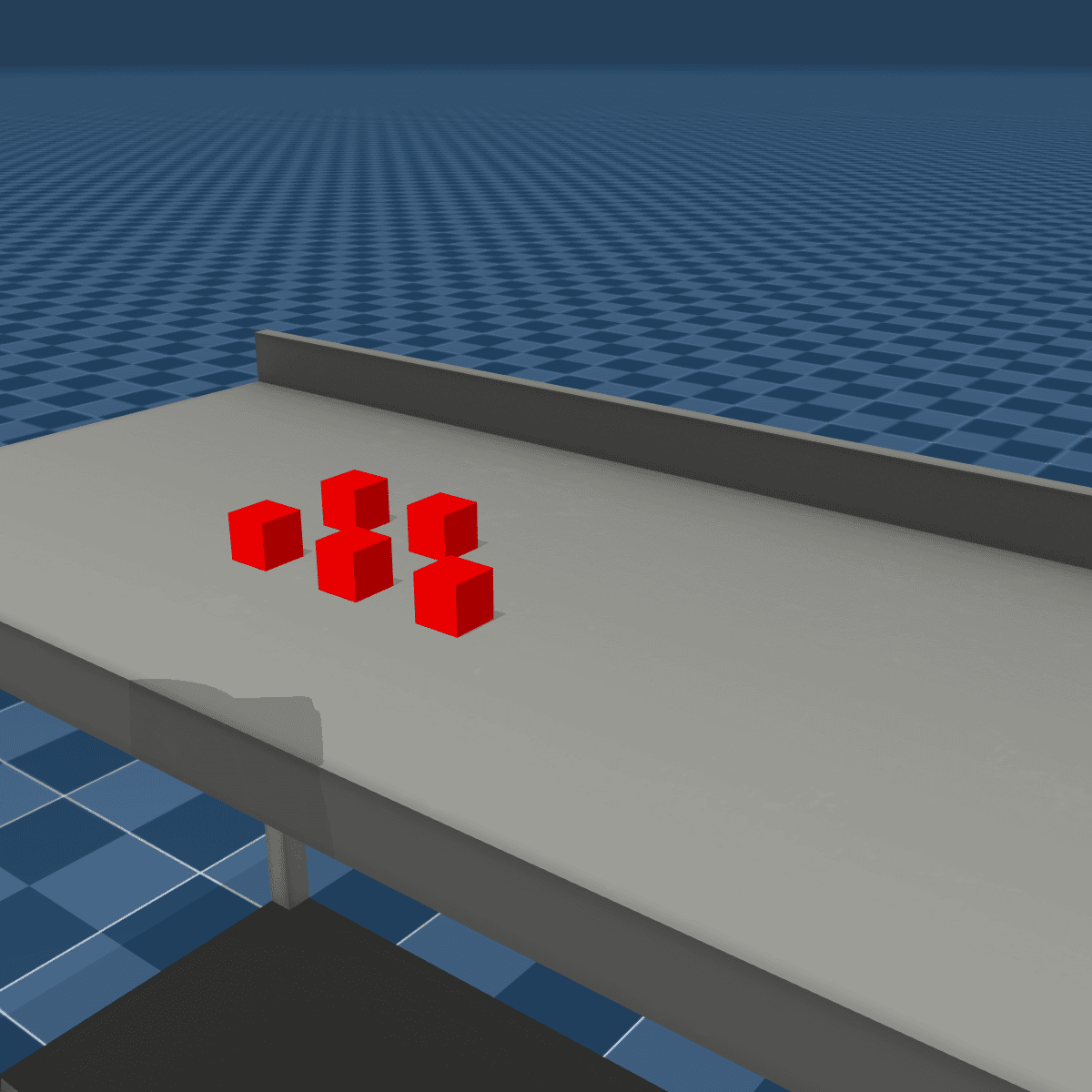}
  {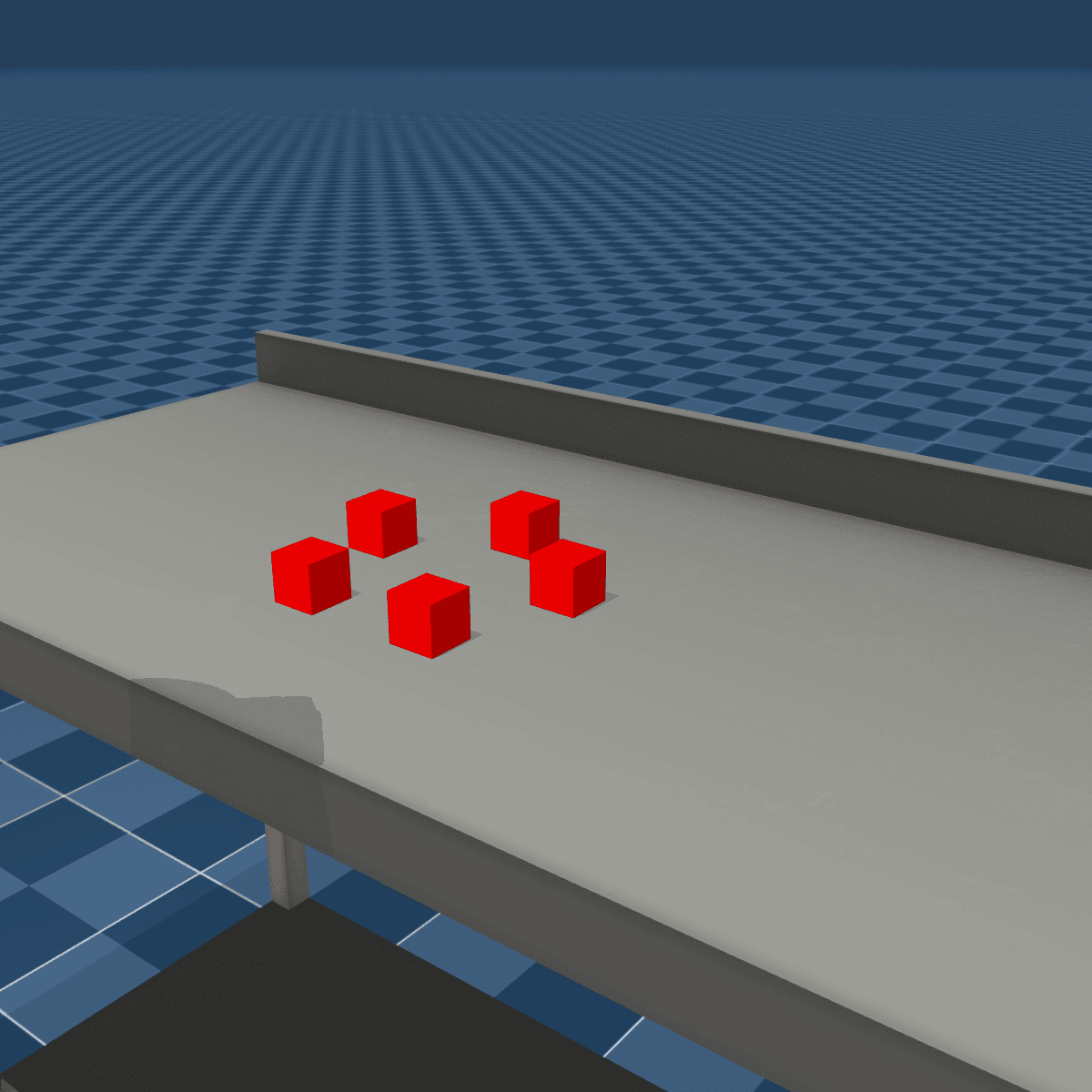}

\vspace{8pt}

% ── Row 5 ──
\taskcard{Sequential Place on Target}{Place each colored block on its matching colored target patch in exact temporal order (blue first, then red).}
  {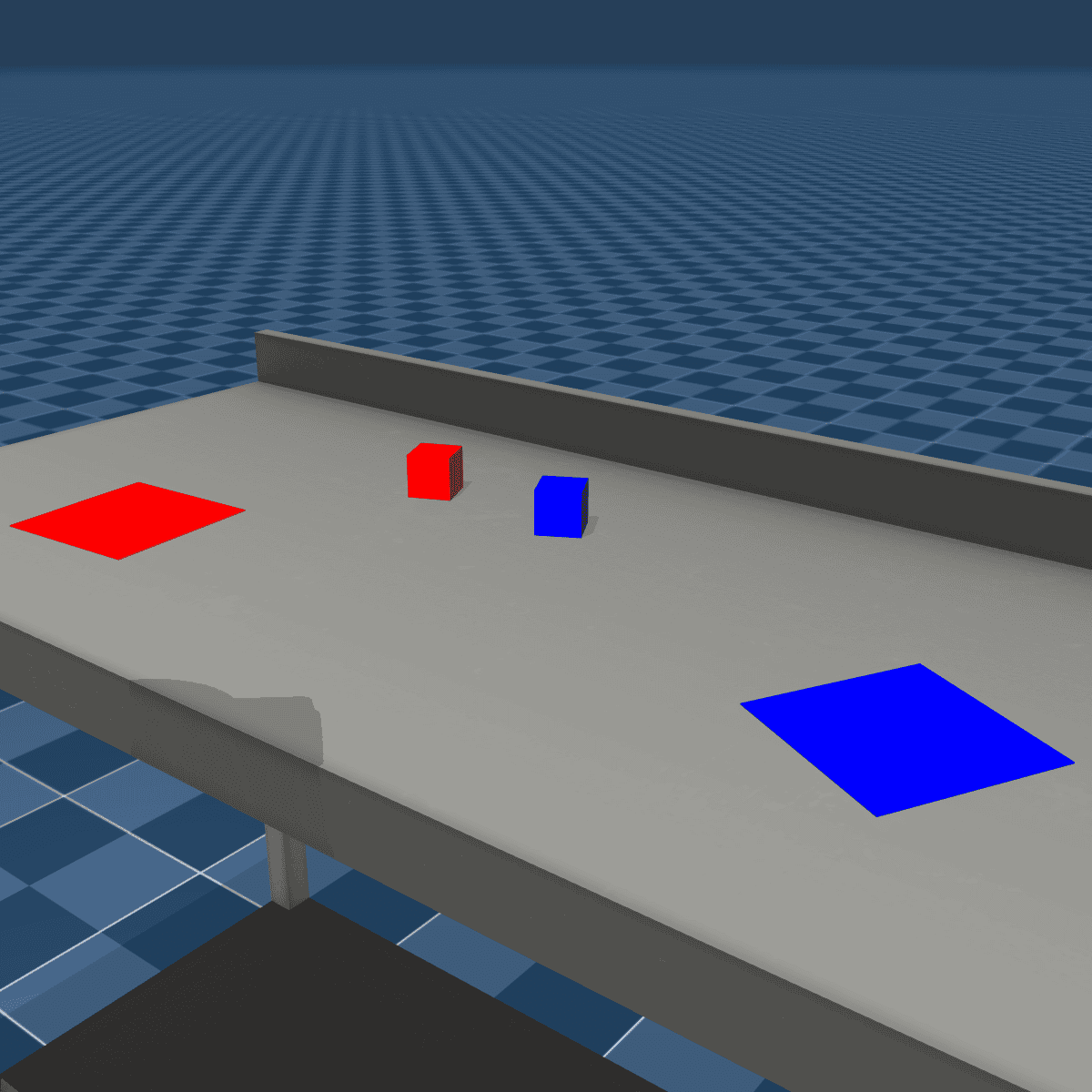}
  {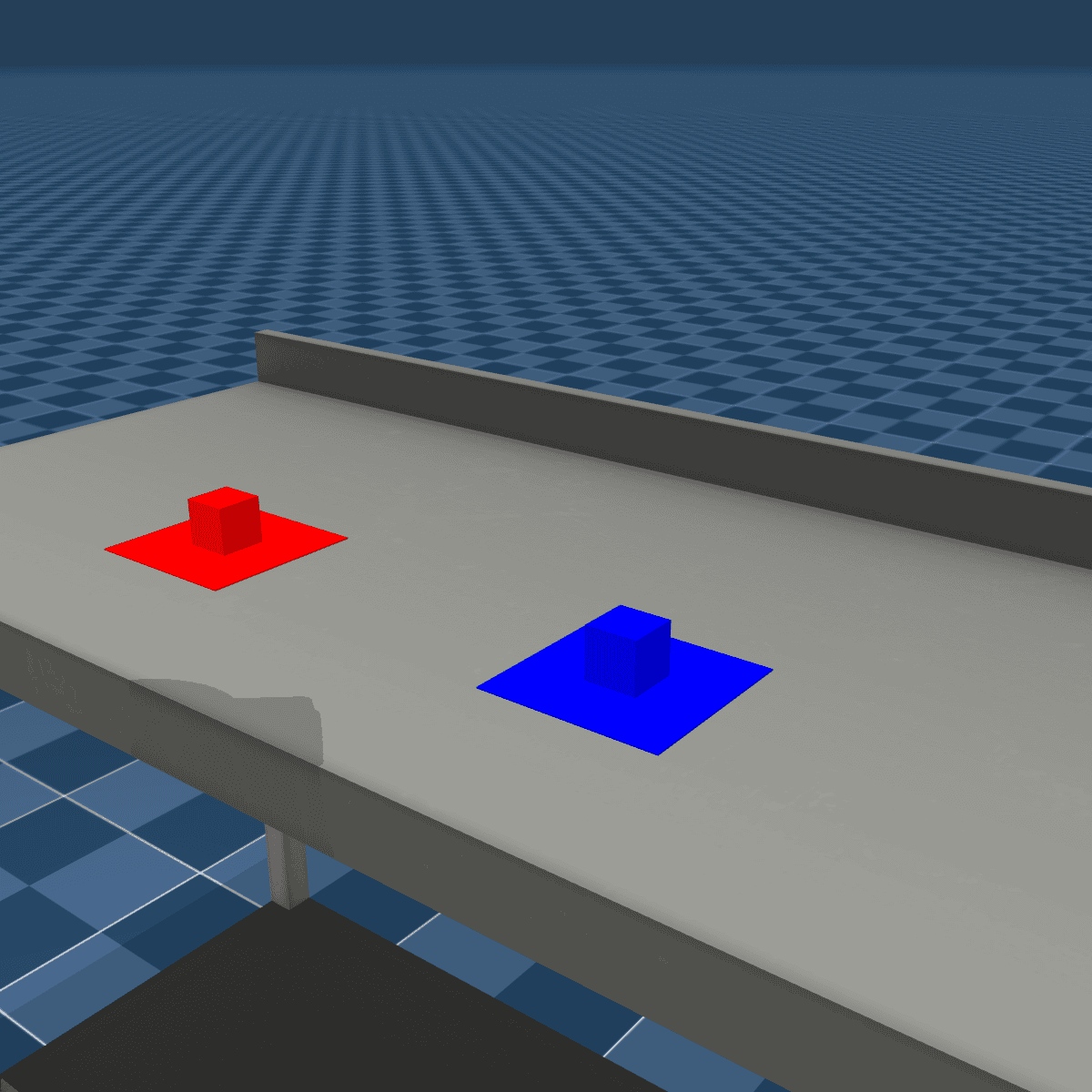}
\hfill
\taskcard{Rotate Cube}{Rotate a block so that the target letter is in the target orientation.}
  {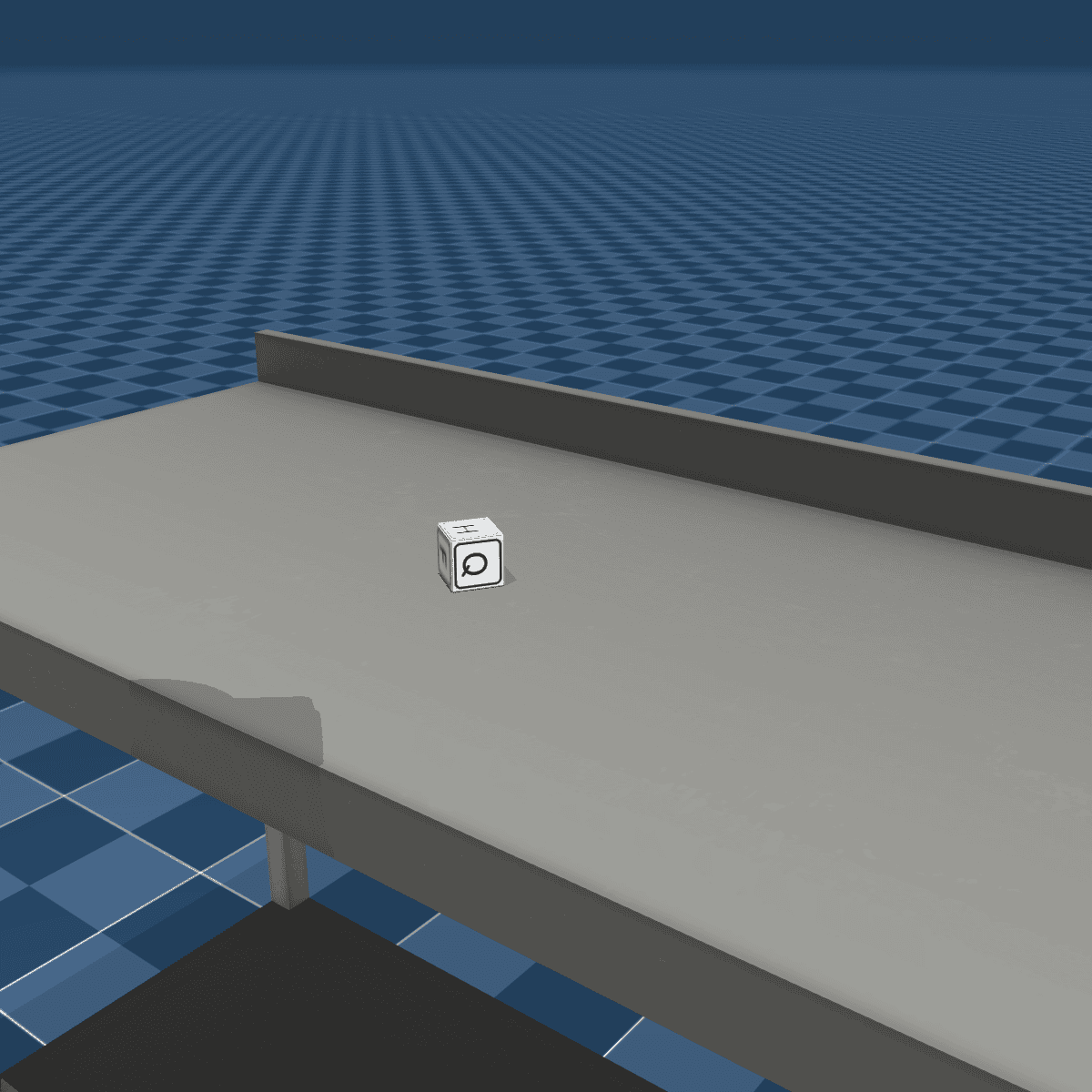}
  {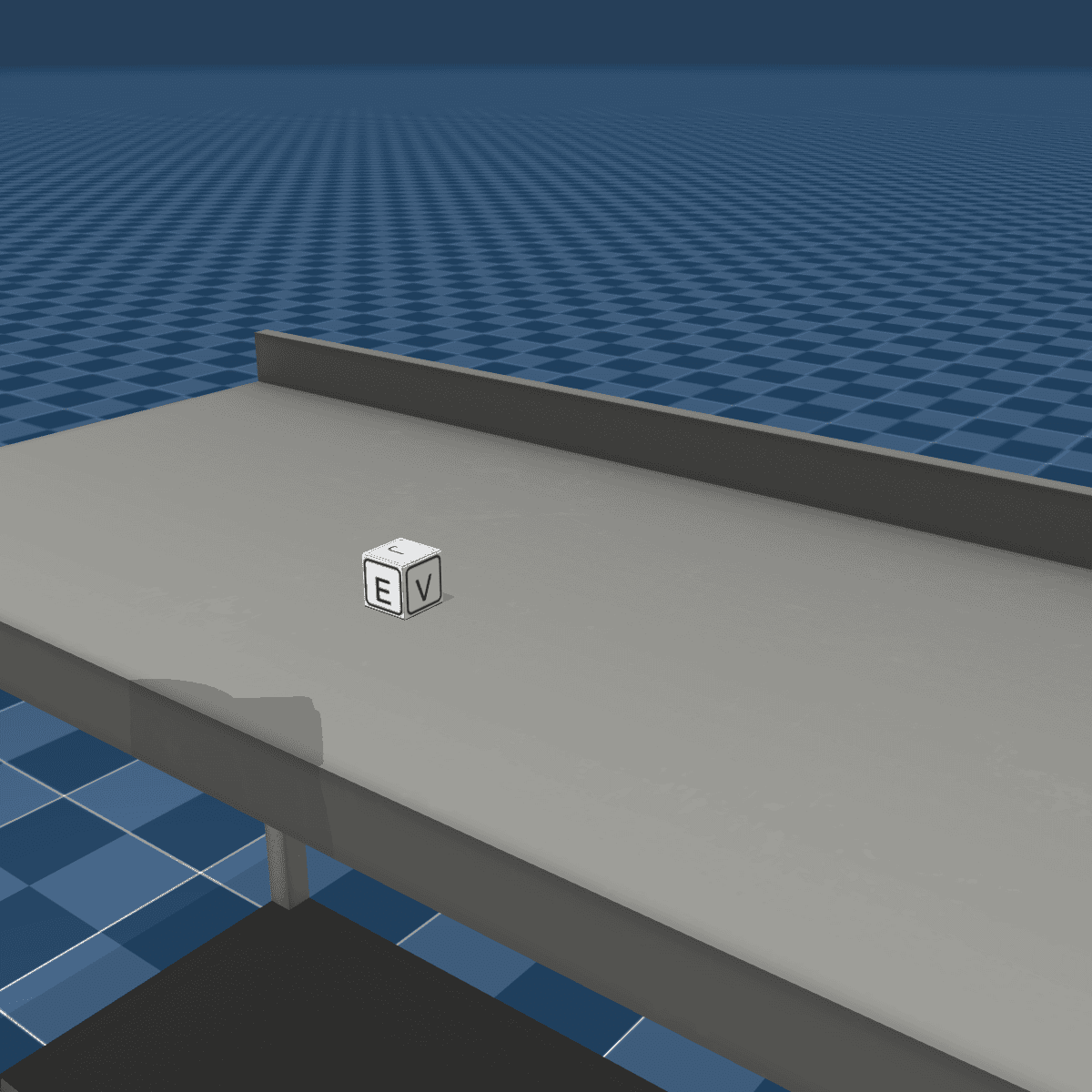}

\caption{\textbf{In-Context Reference Task Suite (2/2).} Continued: semantic reasoning (\textit{Arrange Equation}), spatial pattern formation (\textit{Arrange Shapes}), sequential multi-object target placement (\textit{Sequential Place on Target}), and precise reorientation (\textit{Rotate Cube}).}
\label{fig:task_overview_2}
\end{figure*}

\clearpage
\subsection{Experiments and Results Expanded}
\label{sec:experiments-results}
\subsubsection{Comparison with Baselines}

\textbf{Baseline Details.}
We compare \methodname to two strong baselines: GenSim \cite{wang2023gensim}, and Cursor \cite{cursor_ai}. The GenSim baseline follows the original implementation, where the user interacts with a command-line interface, describing the task they intend to construct via the task name. The Cursor baseline involves a clean version of the base code that RoboPlayground is instantiated with, without any code related to agentic task construction and validation modules. The cursor agent has access to a README detailing the stucture of the repository, including examples of task structure, available assets, and available apis.

\textbf{Results Analysis.}
We detail results in Table 
\ref{tab:task_authoring_quality}, where we compare the quality of our method in comparison to baselines across 5 metrics. Detailed explanations on what the metrics entail are described in Appendix \ref{app:ablation_metrics}.
Table~\ref{tab:task_authoring_quality} compares the quality of task specifications produced by three task authoring systems across successive stages of validity, from syntactic correctness to semantic intent alignment and final human validation. \methodname~consistently outperforms both Cursor and GenSim at every stage of the pipeline, indicating higher reliability and better preservation of user intent throughout task generation.
All systems achieve perfect performance on the initial \textit{Test Case} metric when valid outputs are produced, but differences emerge immediately at compilation and execution. \methodname~achieves a $100\%$ success rate with zero variance for both \textit{Compile} and \textit{Smoke Test}, demonstrating that its structured task representation reliably produces executable and stable task specifications. In contrast, both Cursor and GenSim exhibit substantial failure rates and high variance, suggesting brittle generation behavior and sensitivity to prompt or task formulation.
The largest performance gap appears in \textit{LLM Intent Alignment}, which measures whether the generated success condition correctly reflects the natural language instruction. \methodname~achieves significantly higher alignment ($73.5\%$) than Cursor ($61.1\%$) and GenSim ($38.8\%$), highlighting the advantage of constraining language driven task authoring within a structured physical domain. Notably, GenSim’s low alignment score indicates that while tasks may be executable, their success conditions often fail to capture the intended semantics of the instruction.
These differences propagate to \textit{Human Verification}, where \methodname~achieves unanimous approval across all test cases, while Cursor and GenSim exhibit both lower approval rates and higher variance. This result suggests that failures in earlier stages, particularly semantic misalignment, translate directly into user visible errors that require manual correction or rejection.
Overall, these results show that \methodname~not only improves syntactic and execution level validity, but more importantly, preserves semantic intent from language to executable task specification. This consistency across stages is critical for scalable, user authored evaluation, where both correctness and interpretability must be maintained without expert intervention.

\begin{table}[ht]
	\centering
	\small
    \caption{Comparison of task specification quality across three task authoring systems.
We report mean $\pm$ standard deviation (in percentage) over all evaluated test cases.
Metrics capture successive stages of task validity, including compilation success, basic executability (smoke test), alignment between the generated success condition and the natural language intent, and final human verification.
RoboPlayground achieves consistently higher validity and intent alignment across all stages.}
\label{tab:task_authoring_quality}
	\begin{tabular}{lccc}
		\toprule
		\textbf{Metric}      & \textbf{RoboPlayground (Ours)} & \textbf{Cursor} & \textbf{GenSim} \\
		\midrule
		Test Case            & $100 \pm 0$             & $100 \pm 0$     & $80.0 \pm 24.8$ \\
		Compile              & $100 \pm 0$             & $80.0 \pm 24.8$ & $80.0 \pm 24.8$ \\
		Smoke Test           & $100 \pm 0$             & $80.0 \pm 24.8$ & $60.0 \pm 30.4$ \\
		LLM Intent Alignment & $73.5 \pm 11.1$         & $61.1 \pm 14.0$ & $38.8 \pm 14.0$ \\
		Human Verification   & $100 \pm 0$             & $80 \pm 24.8$   & $70 \pm 28.4$   \\
		\bottomrule
	\end{tabular}
\end{table}

\subsubsection{System Usability Study Details} 
\par
\begin{figure*}[t]
\centering

% First row - 4 figures
\begin{subfigure}[b]{0.24\textwidth}
    \centering
    \includegraphics[width=\textwidth]{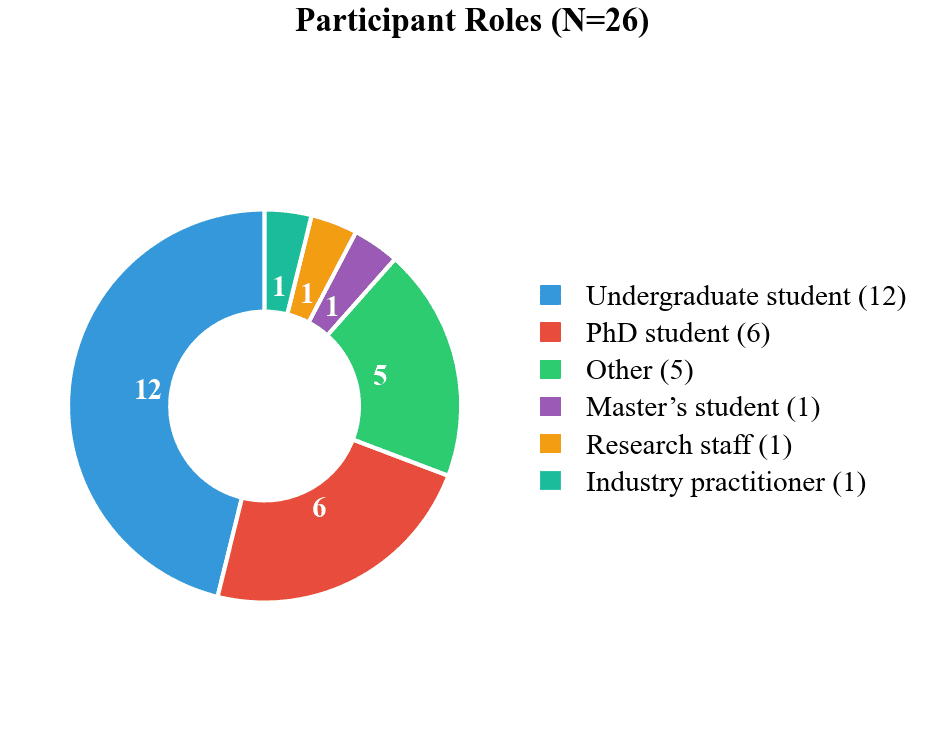}
    \caption{Participant roles}
    \label{fig:demo_role}
\end{subfigure}
\hfill
\begin{subfigure}[b]{0.24\textwidth}
    \centering
    \includegraphics[width=\textwidth]{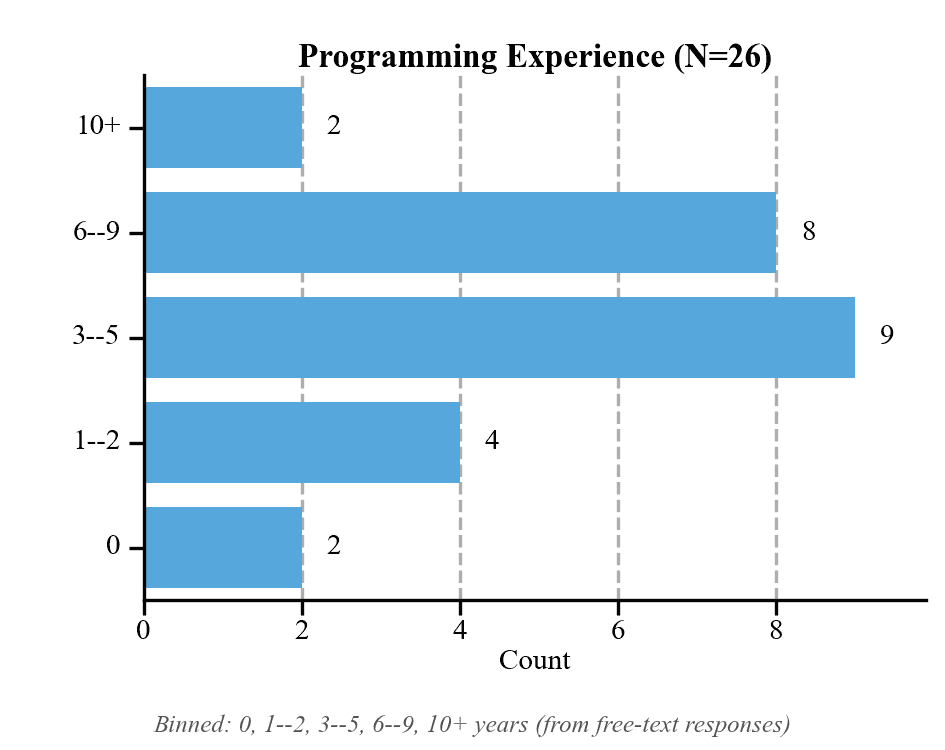}
    \caption{Programming experience}
    \label{fig:demo_prog_exp}
\end{subfigure}
\hfill
\begin{subfigure}[b]{0.24\textwidth}
    \centering
    \includegraphics[width=\textwidth]{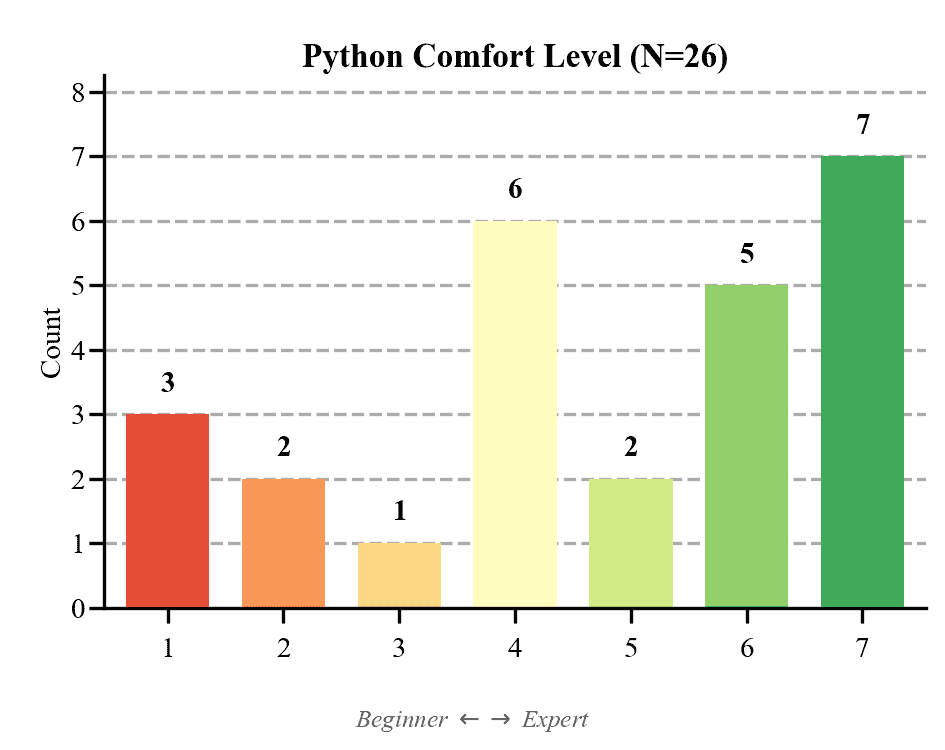}
    \caption{Python comfort level}
    \label{fig:demo_python}
\end{subfigure}
\hfill
\begin{subfigure}[b]{0.24\textwidth}
    \centering
    \includegraphics[width=\textwidth]{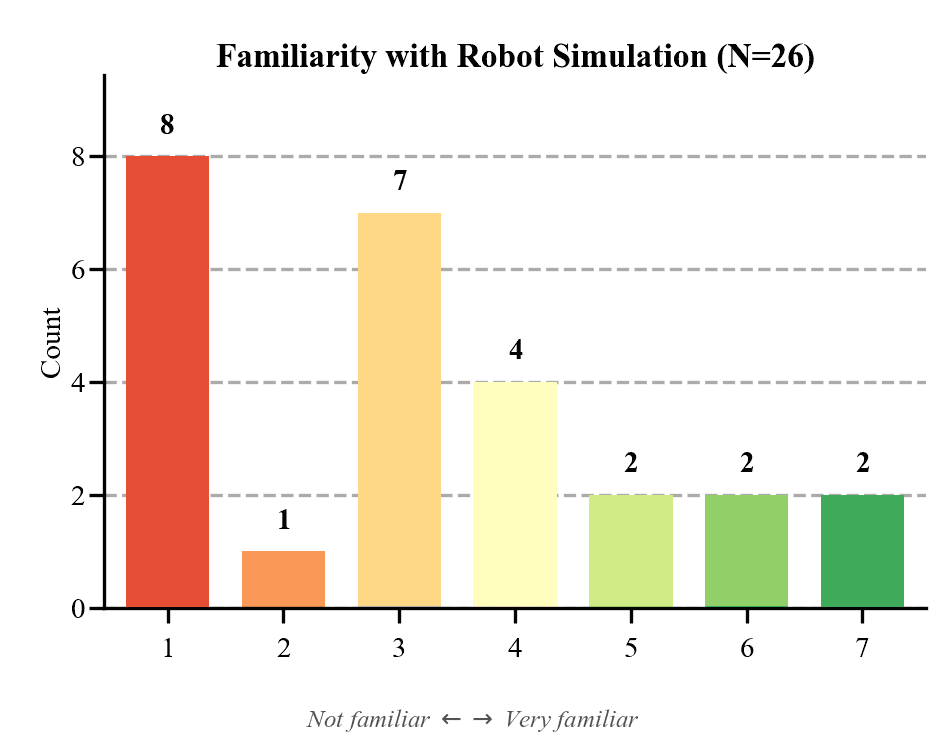}
    \caption{Robot simulation familiarity}
    \label{fig:demo_robot_sim}
\end{subfigure}

\vspace{0.5em}

% Second row - 4 figures
\begin{subfigure}[b]{0.24\textwidth}
    \centering
    \includegraphics[width=\textwidth]{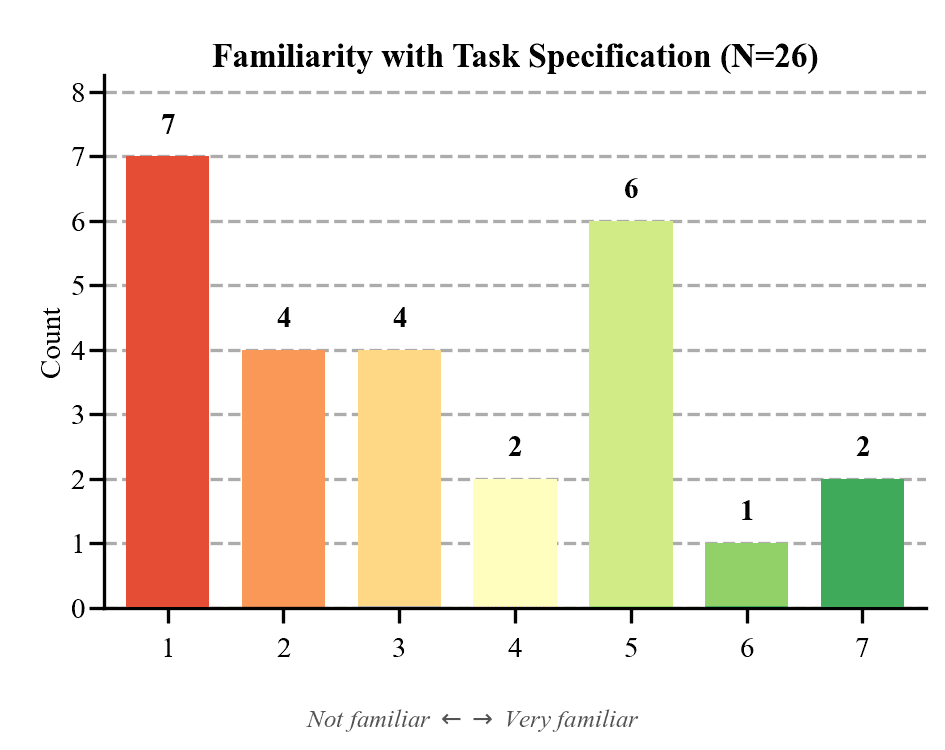}
    \caption{Task specification familiarity}
    \label{fig:demo_task_spec}
\end{subfigure}
\hfill
\begin{subfigure}[b]{0.24\textwidth}
    \centering
    \includegraphics[width=\textwidth]{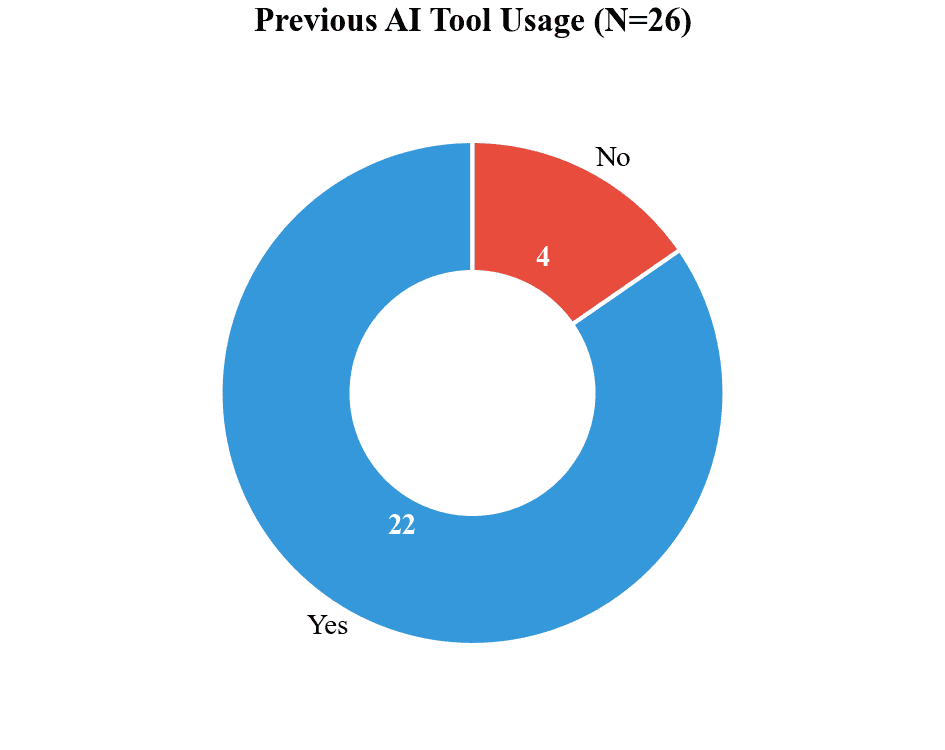}
    \caption{Previous AI tool usage}
    \label{fig:demo_ai_tools}
\end{subfigure}
\hfill
\begin{subfigure}[b]{0.24\textwidth}
    \centering
    \includegraphics[width=\textwidth]{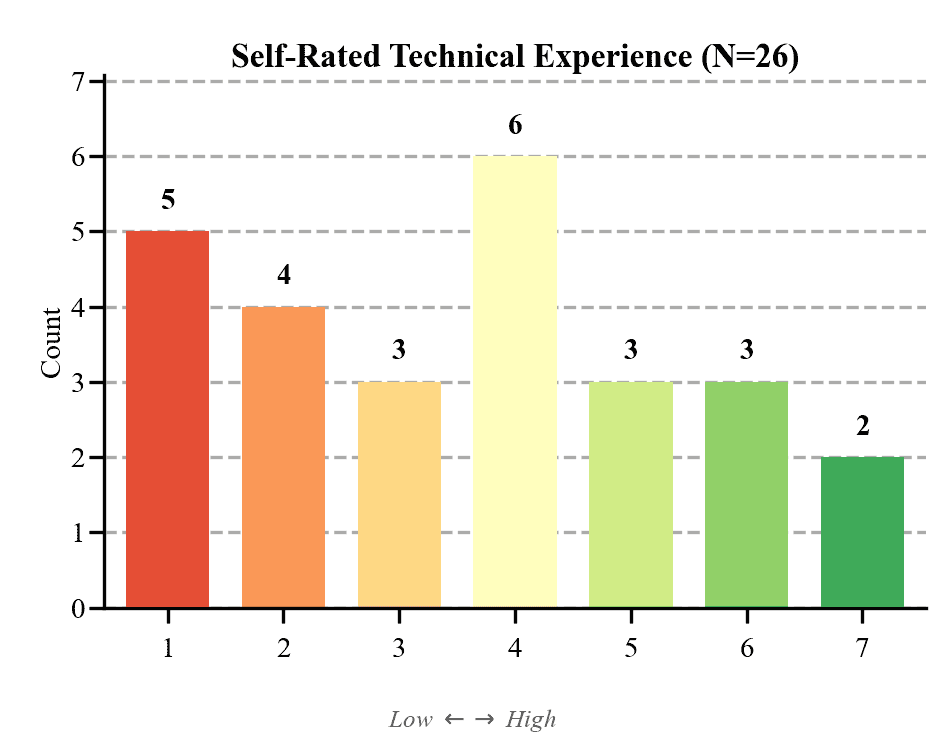}
    \caption{Self-rated technical experience}
    \label{fig:demo_tech_exp}
\end{subfigure}
\hfill
\begin{subfigure}[b]{0.24\textwidth}
    \centering
    \includegraphics[width=\textwidth]{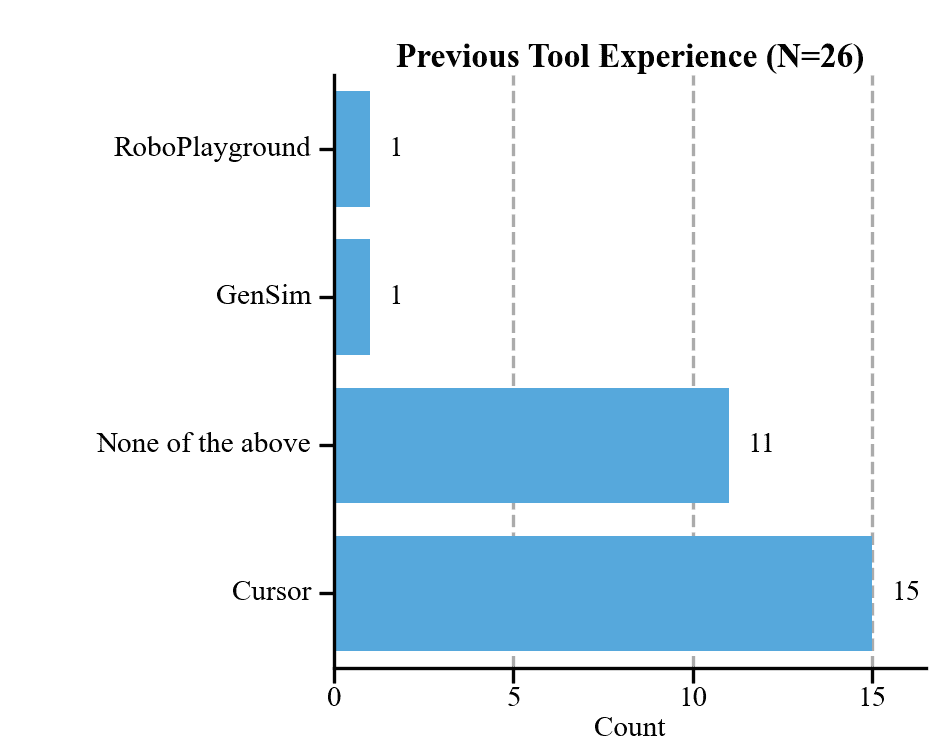}
    \caption{Previous tool experience}
    \label{fig:demo_tools}
\end{subfigure}

\caption{\textbf{Participant demographics ($N{=}26$).}
Participants spanned varied roles and experience (all percentages are shares of $N{=}26$ except programming bins, which use the $n{=}25$ non-missing year responses).
(a) Roles: undergraduate 12 (46.2\%), PhD 6 (23.1\%), other 5 (19.2\%), master's, research staff, and industry practitioner 1 each (3.8\% each).
(b) Programming years (one missing): parsed free text into bins $0$, $1$--$2$, $3$--$5$, $6$--$9$, $10+$; among 25 responses, counts 2, 4, 9, 8, 2 (8.0\%, 16.0\%, 36.0\%, 32.0\%, 8.0\%); reported values ranged from 0 to 14 years.
(c) Python comfort (1--7): counts by level were 3, 2, 1, 6, 2, 5, 7 (11.5\%, 7.7\%, 3.8\%, 23.1\%, 7.7\%, 19.2\%, 26.9\%); combined levels 1--2: 5 (19.2\%).
(d) Robot simulation familiarity: level~1: 8 (30.8\%), 2: 1 (3.8\%), 3: 7 (26.9\%), 4: 4 (15.4\%), 5--7: 2 each (7.7\% each).
(e) Task specification familiarity: level~1: 7 (26.9\%), 2--3: 4 each (15.4\% each), 4: 2 (7.7\%), 5: 6 (23.1\%), 6: 1 (3.8\%), 7: 2 (7.7\%).
(f) Prior AI-assisted tools for task/program specification: Yes 22 (84.6\%), No 4 (15.4\%).
(g) Self-rated technical experience (1--7): level~4 was most frequent with 6 (23.1\%), then 1: 5 (19.2\%), 2: 4 (15.4\%), 3 and 5: 3 each (11.5\% each), 6: 3 (11.5\%), 7: 2 (7.7\%).
(h) Prior tools (multi-select; categories not mutually exclusive): Cursor selected in 15 rows (57.7\%), ``None of the above'' in 11 (42.3\%), RoboPlayground and GenSim each in 1 (3.8\%).}
\label{fig:demographics}
\end{figure*}

\textbf{User Demographics.} We recruited 26 participants (Figure~\ref{fig:demographics}). The largest role group was undergraduate students ($n{=}12$, 46\%), followed by PhD students ($n{=}6$, 23\%), participants who selected ``Other'' ($n{=}5$, 19\%), and one each in a Master's program, research staff, and industry practice. Participants reported programming experience from 0 to 14 years among 25 respondents who provided a parseable numeric answer (median 4 years; one missing response). The most common binned category was 3--5 years ($n{=}9$), followed by 6--9 years ($n{=}8$).

On a 1--7 scale, Python comfort was spread across levels 1--7, with the largest counts at levels 7 ($n{=}7$), 4 ($n{=}6$), and 6 ($n{=}5$), indicating moderate to strong Python familiarity overall. Familiarity with robot simulation or task authoring skewed toward the lower half of the same scale (median 3): for example, 8 participants chose level~1 and 7 chose level~3. Familiarity with task specification or programming-based interfaces had the same median (3) but somewhat more mass at mid-to-high levels (9 participants at levels 5--7 versus 6 at those levels for robot simulation).

Twenty-two participants (85\%) reported prior use of AI-assisted or language-based tools to specify tasks or programs. For the specific tools listed in the questionnaire, 15 participants (58\%) indicated prior use of Cursor, one (4\%) had used RoboPlayground, one (4\%) had used GenSim, and 11 (42\%) selected ``None of the above'' (multi-select was allowed, so categories are not mutually exclusive). Self-rated technical experience relevant to the study ranged from 1 to 7 with a modal rating of 4 ($n{=}6$). Overall, the sample combines general programming competence with limited prior exposure to the specialized authoring tools under study, as detailed in the figure.

\textbf{User Study Details.}
The user study employed a within-subjects design where each participant used all three systems (GenSim, Cursor, and RoboPlayground) to complete task generation exercises. The study was conducted in a controlled lab environment and consisted of the following stages:

\textbf{Study Procedure.}
The study followed a structured protocol:
\begin{enumerate}
    \item \textit{Pre-study questionnaire:} Participants completed demographic questions and rated their experience with programming, Python, robot simulation, task specification, and AI-assisted tools (see Appendix for demographics).
    \item \textit{System training:} For each system, participants received a brief tutorial and demonstration of the interface and capabilities.
    \item \textit{Task completion:} Participants completed two task generation exercises per system (Task 1: basic pyramid stacking, Task 2: constrained pyramid with specific colors/ordering). Task completion times were recorded, and tasks marked as DNF (Did Not Finish) if participants could not complete them within the allotted time or abandoned the task.
    \item \textit{Post-system evaluation:} After using each system, participants completed the System Usability Scale (SUS) questionnaire, NASA Task Load Index (TLX) assessment, and additional task-specific evaluation questions.
    \item \textit{Final comparative evaluation:} After using all three systems, participants ranked the systems, indicated their overall preference, and answered questions about future usage scenarios.
\end{enumerate}

The order of system presentation was counterbalanced across participants to mitigate learning and order effects.

\textbf{Evaluation Metrics.}
We collected both quantitative and qualitative data across multiple dimensions:

\textit{Task Performance Metrics:}
\begin{itemize}
    \item Task completion time (minutes:seconds)
    \item Did Not Finish (DNF) rate (percentage of tasks not completed)
\end{itemize}

\textit{System Usability Scale (SUS):} 
The SUS is a widely-used 10-item Likert scale questionnaire that provides a global measure of system usability \cite{brooke1996sus}. Responses are given on a 5-point scale from ``Strongly Disagree'' (1) to ``Strongly Agree'' (5), and the final SUS score ranges from 0 to 100, with higher scores indicating better usability. A score above 68 is considered above average. Table~\ref{tab:sus_questions} lists the SUS items.

\begin{table}[h]
\centering
\footnotesize
\begin{tabular}{p{0.9\columnwidth}}
\toprule
\textbf{System Usability Scale (SUS) Items} \\
\midrule
1. I think that I would like to use this system frequently. \\
2. I found the system unnecessarily complex. \\
3. I thought the system was easy to use. \\
4. I think that I would need the support of a technical person to be able to use this system. \\
5. I found the various functions in this system were well integrated. \\
6. I thought there was too much inconsistency in this system. \\
7. I would imagine that most people would learn to use this system very quickly. \\
8. I found the system very cumbersome to use. \\
9. I felt very confident using the system. \\
10. I needed to learn a lot of things before I could get going with this system. \\
\bottomrule
\end{tabular}
\caption{System Usability Scale (SUS) questionnaire items. Items 2, 4, 6, 8, and 10 are reverse-scored.}
\label{tab:sus_questions}
\end{table}

\textit{NASA Task Load Index (TLX):}
The NASA-TLX is a multidimensional assessment tool for measuring perceived workload \cite{hart1988development}. We used an unweighted version with five subscales rated on 7-point Likert scales, with responses normalized to 0--100 for analysis. Lower TLX scores indicate lower cognitive workload (better). Table~\ref{tab:tlx_dimensions} describes the dimensions.

\begin{table}[h]
\centering
\footnotesize
\begin{tabular}{lp{0.7\columnwidth}}
\toprule
\textbf{Dimension} & \textbf{Question} \\
\midrule
Mental Demand & How mentally demanding was the task? \\
Temporal Demand & How hurried or rushed did you feel while performing the task? \\
Effort & How hard did you have to work to accomplish your level of performance? \\
Frustration & How frustrated, irritated, or stressed did you feel while using the system? \\
Performance & How successful do you think you were in accomplishing the task goals? (reversed) \\
\bottomrule
\end{tabular}
\caption{NASA-TLX dimensions used in the study. The Performance dimension is reverse-scored such that higher values indicate worse performance.}
\label{tab:tlx_dimensions}
\end{table}

\textit{Task-Specific Evaluation Questions:}
In addition to SUS and TLX, we asked participants to rate their agreement (on 5-point Likert scales) with statements about specific aspects of task generation:

\begin{table}[h]
\centering
\footnotesize
\begin{tabular}{p{0.9\columnwidth}}
\toprule
\textbf{Task Generation Evaluation Statements} \\
\midrule
The generated task accurately reflected my intended task structure. \\
I felt that I had precise control over how the task was specified. \\
It was easy to specify constraints such as block color and ordering. \\
It was easy to revise or refine the task when the initial result was not correct. \\
Small changes to my input resulted in predictable changes to the generated task. \\
It was easy to identify and fix mistakes in the task specification. \\
The system allowed me to express the task at an appropriate level of abstraction. \\
I was able to make meaningful progress toward a correct task quickly. \\
\bottomrule
\end{tabular}
\caption{Task-specific evaluation statements for assessing task generation quality and control.}
\label{tab:task_questions}
\end{table}

\textit{Comparative Evaluation:}
After experiencing all three systems, participants provided:
\begin{itemize}
    \item \textit{Usability ranking:} Rank the three systems from 1 (best) to 3 (worst) in terms of overall usability
    \item \textit{Overall preference:} ``Overall, which system did you prefer for task generation?''
    \item \textit{Future usage preference:} ``Which system would you prefer to use to generate new evaluation tasks in the future?''
    \item \textit{Scalability assessment:} ``Which system do you believe would scale best to more complex tasks than the pyramid task?''
\end{itemize}

\textit{Qualitative Feedback:}
For each system, participants answered two open-ended questions:
\begin{itemize}
    \item ``What aspects of this system were most helpful for task generation?''
    \item ``What aspects of this system were most frustrating or limiting?''
\end{itemize}

These responses provided insights into specific usability issues and design strengths that complemented the quantitative metrics.

\textbf{Statistical Analysis.}
Given the small sample size (N=26) and non-parametric nature of the data, we used appropriate non-parametric statistical tests for comparisons. For paired comparisons of continuous metrics (task completion times, SUS scores, TLX scores), we employed the Wilcoxon signed-rank test. For comparing usability rankings across all three systems, we used the Friedman test followed by post-hoc pairwise Wilcoxon tests. For binary outcomes (DNF rates), we used McNemar's test. For system preference, we used a chi-square goodness-of-fit test to determine if preferences differed from a uniform distribution. Statistical significance was assessed at $\alpha = 0.05$. We present results of statistical significance tests as follows:
\begin{itemize}
    \item \textbf{Task completion times:} Wilcoxon signed-rank tests comparing pairwise completion times between systems for Task 1 and Task 2 (Table~\ref{tab:wilcoxon_tests}). RoboPlayground was significantly faster than Cursor for Task 1 (p=0.0312).
    
    \item \textbf{Did Not Finish (DNF) rates:} McNemar's tests comparing the proportion of tasks participants failed to complete between each pair of systems (Table~\ref{tab:mcnemar_tests}). No significant differences were found, though RoboPlayground showed trends toward fewer DNFs compared to GenSim (p=0.0736).
    
    \item \textbf{System Usability Scale (SUS) scores:} Wilcoxon signed-rank tests comparing perceived usability between systems (Table~\ref{tab:wilcoxon_sus_tests}). RoboPlayground had significantly higher SUS scores than GenSim (p=0.0078).
    
    \item \textbf{NASA-TLX workload scores:} Wilcoxon signed-rank tests comparing perceived cognitive workload between systems (Table~\ref{tab:wilcoxon_tlx_tests}). RoboPlayground had significantly lower workload than GenSim (p=0.0156), indicating lower cognitive demand.
    
    \item \textbf{System preference:} Chi-square goodness-of-fit test examining whether participants' stated preferences differed from a uniform distribution (Table~\ref{tab:chi2_preference_test}). Preferences were significantly non-uniform (p=0.0302), with 75\% of participants preferring RoboPlayground.
    
    \item \textbf{Usability rankings:} Friedman test with post-hoc pairwise Wilcoxon tests comparing how participants ranked the three systems (Table~\ref{tab:friedman_rank_tests}). Overall rankings differed significantly (p=0.0046), with RoboPlayground ranked significantly better than GenSim (p=0.0078).
\end{itemize}

All statistical tests used non-parametric methods appropriate for the small sample size (N=26) and ordinal data. Significance was assessed at $\alpha = 0.05$.

% \textbf{Results Analysis Expanded.} 

% Statistical significance calculation, etc.

\subsubsection{Policy Evaluation Details}
\textbf{Training and Evaluation Task Generation Details.} To demonstrate the utility of \methodname~as a benchmarking tool, we leverage the framework to automatically generate a series of training and evaluation tasks. The tasks for training are detailed in Table \ref{tab:dataset_stats}, and the tasks for evaluation are detailed in Table \ref{tab:eval_tasks_perturbations}. Note that for all the evaluation tasks, we used the training tasks as base tasks and used the pipeline's modification feature to automatically generate the evaluation variants.

\textbf{Training Data Generation Details.}
To generate the finetuning data to enable the policies to adapt to the skills and action distribution of our training tasks, we leverage CuTAMP~\cite{shen2024differentiable} as a tool to automatically curate demonstrations. The statistics are detailed in Table \ref{tab:dataset_stats}.

\textbf{Model Training Details.}
We evaluate six policies grouped into two families: four StarVLA variants built on a shared vision-language backbone \cite{starvla2025}, and two variants of the external \texttt{Pi-0.5} baseline.

\paragraph{StarVLA variants.}
All four StarVLA models (\texttt{Adapter}, \texttt{Dual}, \texttt{GR00T}, \texttt{Qwen-OFT}) share a Qwen3-VL-4B-Instruct~\cite{bai2025qwen3vltechnicalreport} vision-language backbone and predict 7-dimensional end-effector delta actions (6D pose + 1D gripper) with an action horizon of 16 steps. All models are trained end-to-end using AdamW ($\beta_1 = 0.9$, $\beta_2 = 0.95$) with gradient clipping (max norm 1.0), mixed-precision training, and a per-GPU effective batch size of 16. Training runs for up to 100{,}000 steps with 5{,}000 warmup steps. The models differ in their action decoding architecture:
\begin{itemize}
    \item \texttt{Adapter} appends 64 learnable action query tokens to the VLM sequence and decodes actions through a 24-block MLP-ResNet with an L1 regression objective, using a constant learning rate schedule.
    \item \texttt{GR00T} conditions a 16-layer flow-matching DiT on the VLM's final hidden states, using 8 diffusion steps during training and 4 during inference. It uses a cosine learning rate schedule with separate rates for the VLM ($1 \times 10^{-5}$) and action model ($1 \times 10^{-4}$).
    \item \texttt{Dual} extends \texttt{GR00T} with a secondary DINOv2~\cite{oquab2024dinov2learningrobustvisual} (ViT-S/14) encoder whose features are concatenated with the VLM hidden states before conditioning the DiT. All other settings match \texttt{GR00T}.
    \item \texttt{Qwen-OFT} injects special action tokens into the VLM vocabulary and regresses actions from their hidden-state positions via a lightweight MLP head with an L1 objective, following the OpenVLA-OFT design~\cite{kim2025finetuningvisionlanguageactionmodelsoptimizing}.
\end{itemize}
\paragraph{Pi-0.5 variants.}
Both \texttt{Pi-0.5} variants were initialized from the Pi0.5-DROID checkpoint~\cite{intelligence2025pi05visionlanguageactionmodelopenworld}, which pairs a PaliGemma 2B vision-language backbone with a Gemma 300M action expert. The action space matches the StarVLA models (7D end-effector delta), discretized into 32 action tokens with an action horizon of 10 steps.
Both variants were trained for 30{,}000 steps with a batch size of 24, using AdamW with cosine decay (peak learning rate $5 \times 10^{-5}$, 10{,}000 warmup steps) and EMA with decay rate 0.999.
The \texttt{Pi-0.5 (LoRA)} variant applies Low-Rank Adaptation to both the vision-language backbone and action expert while freezing all other parameters; the full fine-tuning variant updates all parameters without constraints.

% \textbf{Results Analysis Expanded.}

% \subsubsection{Scalability and Diversity Experiments}

% \textbf{Diversity Metrics.}
% \textbf{}

% \subsubsection{Comparison with Baselines}

\begin{table}[h]
\centering
\resizebox{0.5\linewidth}{!}{%
\begin{tabular}{lcc}
\toprule
Task & Total Trajectories & Avg Steps / Trajectory \\
\midrule
Color Block Alignment & 280 & 332.7 \\
Place Two Blocks on Patch & 1,959 & 216.8 \\
Red Behind Yellow & 1,716 & 135.7 \\
Red in Front of Yellow & 3,331 & 133.2 \\
Red Block Left Placement & 2,277 & 133.2 \\
Red Block Right Placement & 3,549 & 130.1 \\
Red Block Stacking & 1,589 & 366.0 \\
Red on Yellow Stack & 3,381 & 132.2 \\
Three Block Color Stacking & 2,551 & 231.5 \\
Yellow on Red Stack & 3,484 & 134.1 \\
\bottomrule
\end{tabular}%
}
\caption{Dataset statistics for training tasks.}
\label{tab:dataset_stats}
\end{table}

\begin{table}[h]
\centering
\resizebox{0.5\columnwidth}{!}{%
\begin{tabular}{lccc}
\toprule
\textbf{Task} & \textbf{Semantic} & \textbf{Visual} & \textbf{Behavioural} \\
\midrule
Blue Block Stacking                     & \no  & \yes & \no  \\
Green on Blue Stack                     & \yes & \no  & \no  \\
Place Two Blocks on Green Patch         & \no  & \yes & \no  \\
Place Two Blocks on Long Patch          & \no  & \yes & \no  \\
Place Two Blue Blocks on Patch          & \no  & \yes & \no  \\
Red Block Left of Blue                  & \yes & \yes & \no  \\
Red Block Two Tower Stacking            & \yes & \no  & \yes \\
Stack Two Blocks on Patch               & \yes & \no  & \yes \\
Three Block Color Stacking Beside        & \yes & \no  & \yes \\
Three Block Color Stacking Perturbed    & \no  & \yes & \no  \\
Yellow Block Left Placement             & \yes & \no  & \no  \\
Yellow on Red Unstack Restack           & \no  & \yes & \yes \\
\bottomrule
\end{tabular}%
}
\caption{Perturbation axes exercised by each evaluation task. Semantic perturbations modify task specifications or relations, visual perturbations alter perceptual attributes, and behavioural perturbations require multi-stage or non-monotonic execution.}
\label{tab:eval_tasks_perturbations}
\end{table}

% Training Tasks - include this in your main document
% Required packages in preamble: graphicx, grffile, amsmath, amssymb

% Single-image task card for training/example tasks
\providecommand{\trainingcard}[3]{%
  \begin{minipage}[t]{0.48\textwidth}
    \centering
    \includegraphics[width=0.55\linewidth]{#3}\\[3pt]
    \textbf{#1}\\[1pt]
    {\small\textit{#2}}
  \end{minipage}%
}

% ────────────────────────────────────────────────────────────
% TRAINING TASKS (Page 1: Tasks 1–6)
% ────────────────────────────────────────────────────────────
\begin{figure*}[htbp]
\centering

% ── Row 1 ──
\trainingcard{Arrange Blocks}
  {Arrange the blocks in order red, yellow, green, blue from left to right}
  {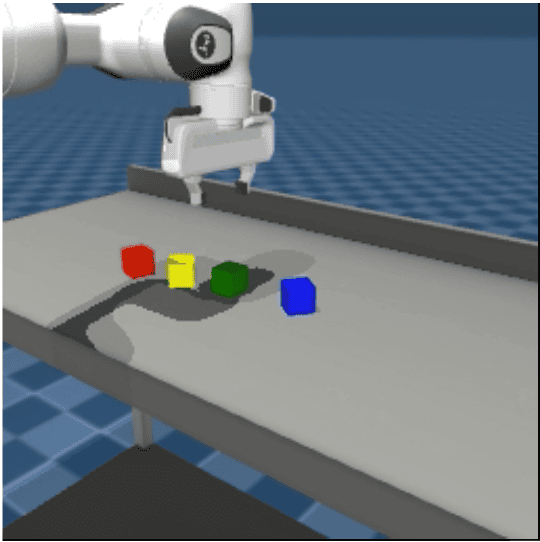}
\hfill
\trainingcard{Place on Goal Patch}
  {Place both blocks onto the goal patch}
  {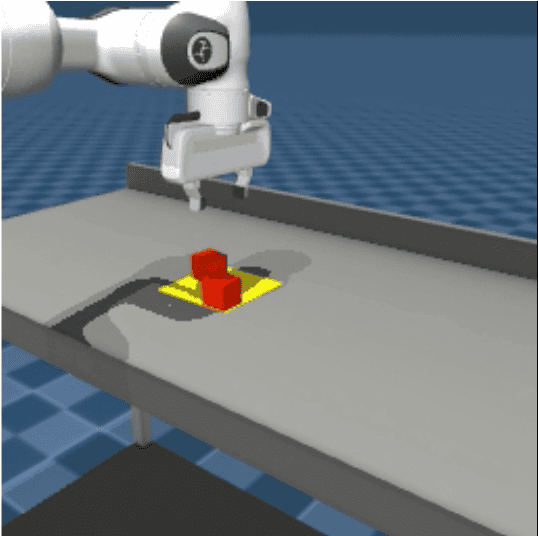}

\vspace{8pt}

% ── Row 2 ──
\trainingcard{Place Behind}
  {Place the red block behind the yellow block}
  {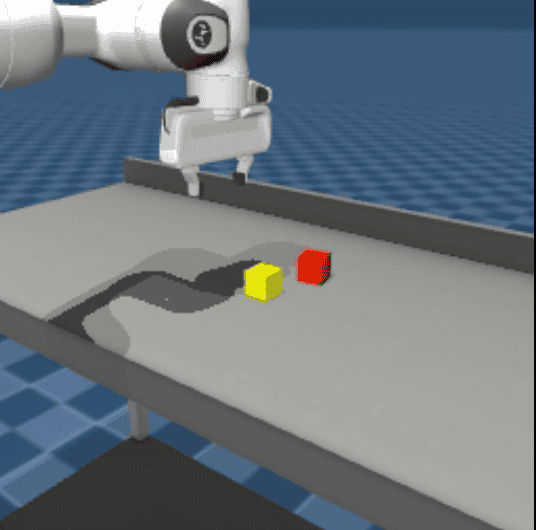}
\hfill
\trainingcard{Place In Front}
  {Place the red block in front of the yellow block}
  {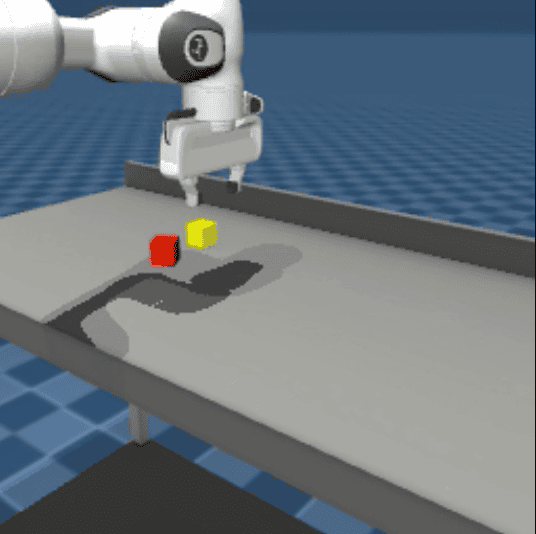}

\vspace{8pt}

% ── Row 3 ──
\trainingcard{Stack Red on Yellow}
  {Stack the red block on top of the yellow block}
  {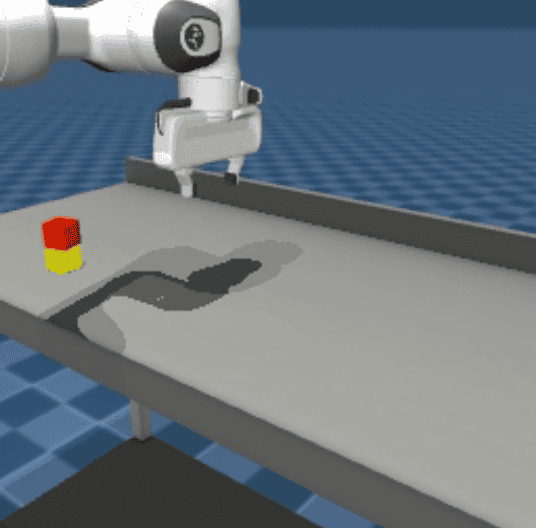}
\hfill
\trainingcard{Place Left}
  {Place the red block on the left side of the yellow block}
  {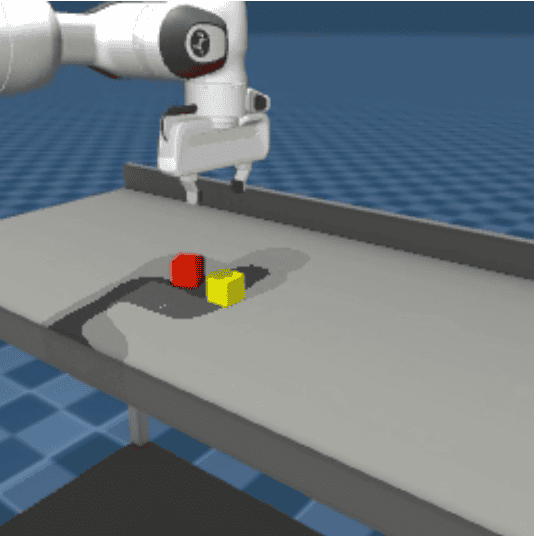}

\caption{\textbf{Training Tasks (1/2).} Goal-state snapshots for training tasks used as in-context examples. Tasks shown: color-ordered arrangement (\textit{Arrange Blocks}), target-driven placement (\textit{Place on Goal Patch}), and relative spatial positioning (\textit{Place Behind}, \textit{Place In Front}, \textit{Stack Red on Yellow}, \textit{Place Left}).}
\label{fig:training_tasks_1}
\end{figure*}

% ────────────────────────────────────────────────────────────
% TRAINING TASKS (Page 2: Tasks 7–10)
% ────────────────────────────────────────────────────────────
\begin{figure*}[htbp]
\centering

% ── Row 4 ──
\trainingcard{Place Right}
  {Place the red block on the right side of the yellow block}
  {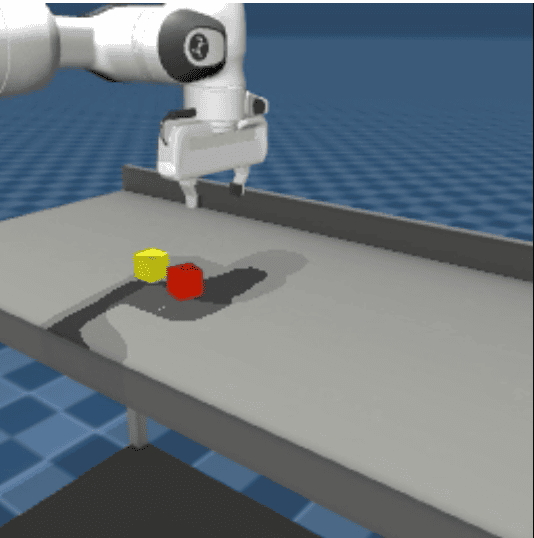}
\hfill
\trainingcard{Stack Yellow on Red}
  {Stack the yellow block on top of the red block}
  {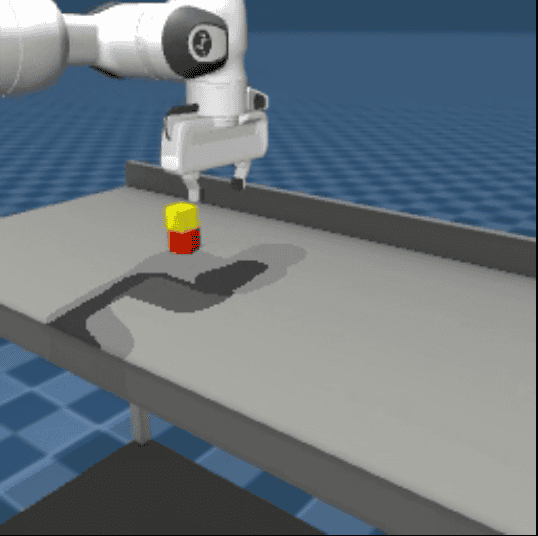}

\vspace{8pt}

% ── Row 5 ──
\trainingcard{Three-Block Stack}
  {Stack the green block on the red block then the yellow block on the green block}
  {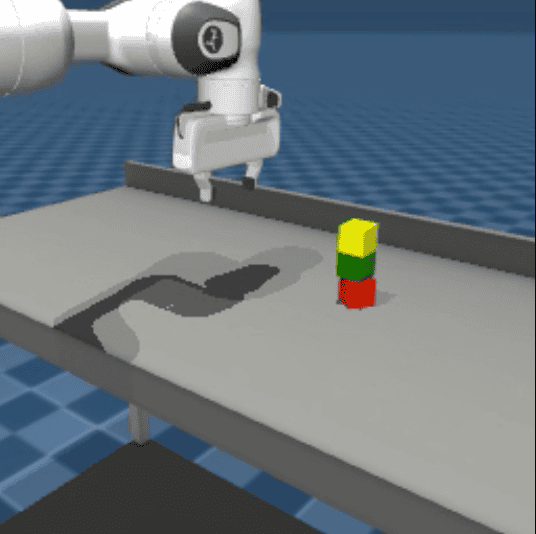}
\hfill
\trainingcard{Stack Same-Color Blocks}
  {Stack the red blocks on top of each other}
  {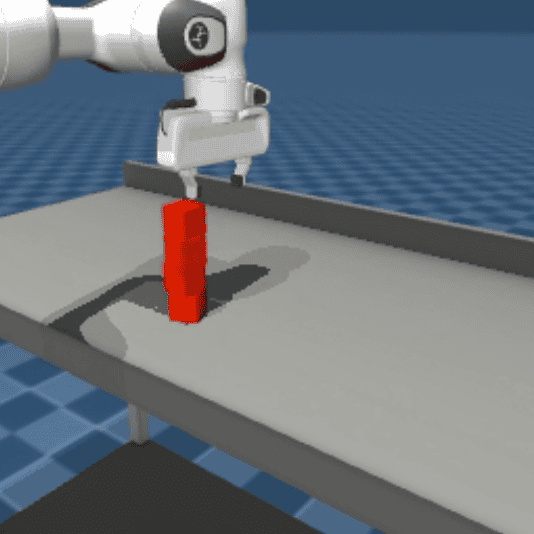}

\caption{\textbf{Training Tasks (2/2).} Continued: relative spatial positioning (\textit{Place Right}, \textit{Stack Yellow on Red}), multi-block sequential stacking (\textit{Three-Block Stack}), and same-color block stacking (\textit{Stack Same-Color Blocks}).}
\label{fig:training_tasks_2}
\end{figure*}

% Evaluation Tasks - include this in your main document
% Required packages in preamble: graphicx, grffile, amsmath, amssymb

% Single-image task card for evaluation tasks (reuse trainingcard if already defined)
\providecommand{\evalcard}[3]{%
  \begin{minipage}[t]{0.48\textwidth}
    \centering
    \includegraphics[width=0.55\linewidth]{#3}\\[3pt]
    \textbf{#1}\\[1pt]
    {\small\textit{#2}}
  \end{minipage}%
}

% ────────────────────────────────────────────────────────────
% EVALUATION TASKS (Page 1: Tasks 1–6)
% ────────────────────────────────────────────────────────────
\begin{figure*}[htbp]
\centering

% ── Row 1 ──
\evalcard{Stack on Goal Patch}
  {Stack the blocks on the goal patch}
  {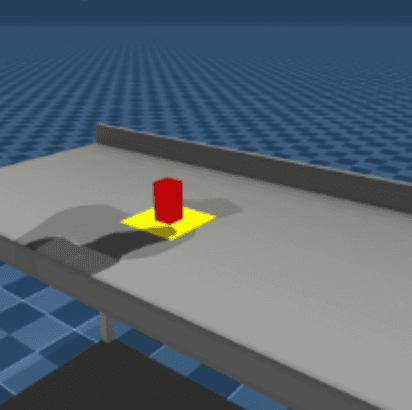}
\hfill
\evalcard{Place on Goal Patch (2 blocks, A)}
  {Place both blocks onto the goal patch}
  {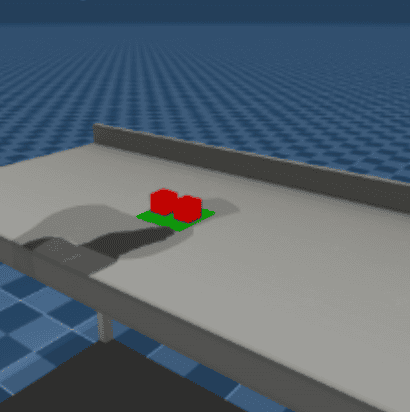}

\vspace{8pt}

% ── Row 2 ──
\evalcard{Place on Goal Patch (2 blocks, B)}
  {Place both blocks onto the goal patch}
  {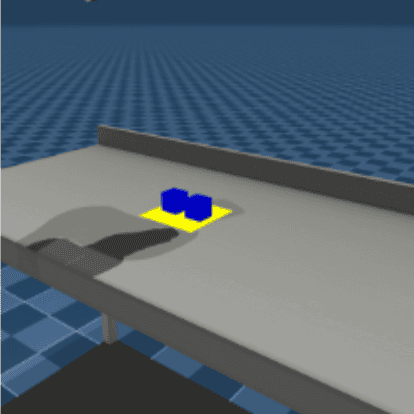}
\hfill
\evalcard{Place on Goal Patch (2 blocks, C)}
  {Place both blocks onto the goal patch}
  {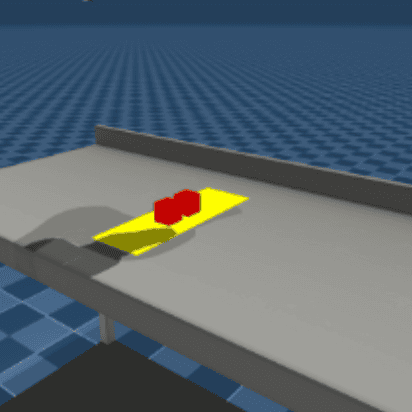}

\vspace{8pt}

% ── Row 3 ──
\evalcard{Stack Beside}
  {Stack the green block on the yellow block beside the red block}
  {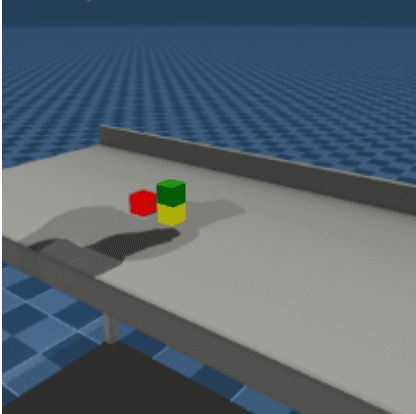}
\hfill
\evalcard{Three-Block Stack}
  {Stack the green block on the yellow block then the red block on the green block}
  {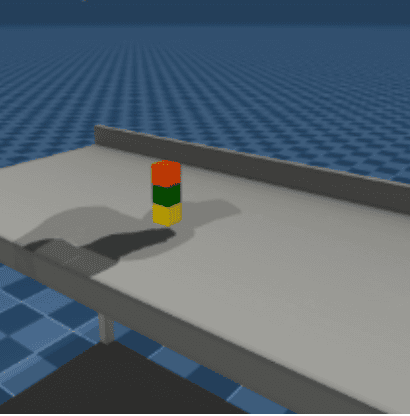}

\caption{\textbf{Evaluation Tasks (1/2).} Goal-state snapshots for evaluation tasks. Tasks shown: goal-patch stacking (\textit{Stack on Goal Patch}), target-driven placement with varying block configurations (\textit{Place on Goal Patch A--C}), adjacent stacking (\textit{Stack Beside}), and multi-block sequential stacking (\textit{Three-Block Stack}).}
\label{fig:eval_tasks_1}
\end{figure*}

% ────────────────────────────────────────────────────────────
% EVALUATION TASKS (Page 2: Tasks 7–12)
% ────────────────────────────────────────────────────────────
\begin{figure*}[htbp]
\centering

% ── Row 4 ──
\evalcard{Stack Green on Blue}
  {Stack the green block on top of the blue block}
  {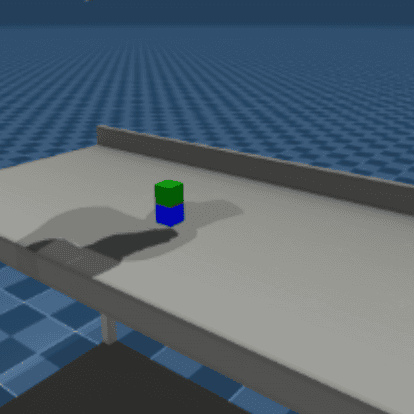}
\hfill
\evalcard{Stack Red on Yellow}
  {Stack the red block on top of the yellow block}
  {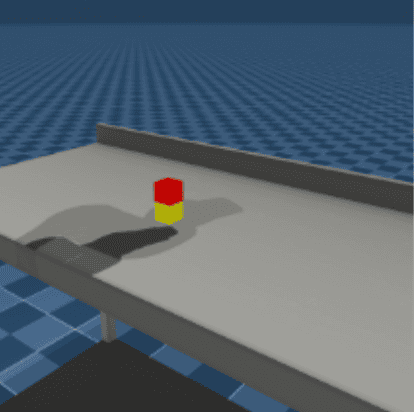}

\vspace{8pt}

% ── Row 5 ──
\evalcard{Place Red Left of Blue}
  {Place the red block on the left side of the blue block}
  {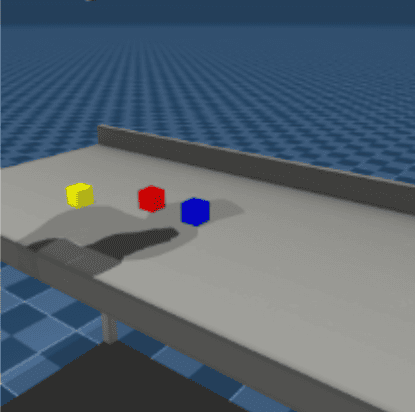}
\hfill
\evalcard{Place Yellow Left of Red}
  {Place the yellow block on the left side of the red block}
  {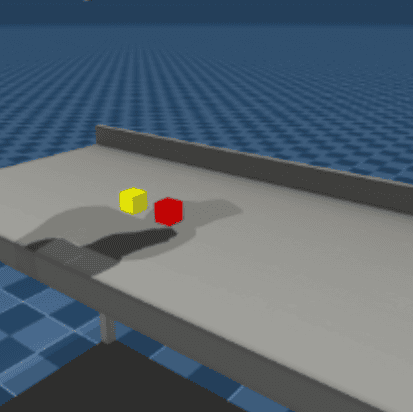}

\vspace{8pt}

% ── Row 6 ──
\evalcard{Stack Same-Color (Blue)}
  {Stack the blue blocks on top of each other}
  {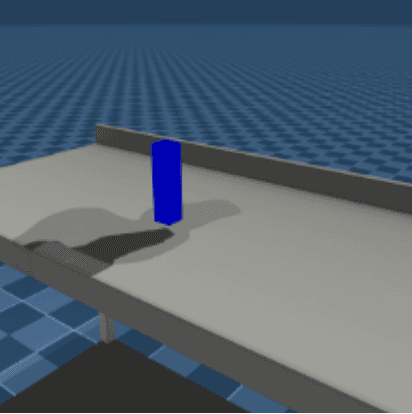}
\hfill
\evalcard{Two Red Stacks}
  {Make two stacks of red blocks.}
  {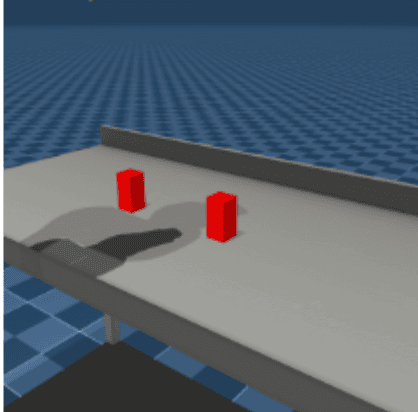}

\caption{\textbf{Evaluation Tasks (2/2).} Continued: color-specific stacking (\textit{Stack Green on Blue}, \textit{Stack Red on Yellow}), relative spatial positioning (\textit{Place Red Left of Blue}, \textit{Place Yellow Left of Red}), same-color stacking (\textit{Stack Same-Color Blue}), and multi-structure construction (\textit{Two Red Stacks}).}
\label{fig:eval_tasks_2}
\end{figure*}

\clearpage

\begin{table}[t]
\centering
\caption{Task completion times and DNF rates (N=26).
Times are reported as mean completion time per task in minutes:seconds.
DNF percentage is the proportion of attempted tasks that were not finished.
Darker green cells indicate better performance (faster times and lower DNF).}
\label{tab:task_times_dnf}
\footnotesize
\setlength{\tabcolsep}{3pt}
\begin{tabular}{lccc}
\toprule
System & Average time for Task 1 & Average time for Task 2 & DNF percentage \\
\midrule
GenSim & \colorbox{green!17}{7:45 $\pm$ 0:56} & \colorbox{green!6}{9:14 $\pm$ 0:43} & \colorbox{green!35}{56\% $\pm$ 14\%} \\
Cursor & \colorbox{green!27}{6:30 $\pm$ 0:55} & \colorbox{green!15}{8:03 $\pm$ 0:54} & \colorbox{green!54}{32\% $\pm$ 13\%} \\
RoboPlayground & \colorbox{green!48}{3:53 $\pm$ 0:37} & \colorbox{green!24}{6:59 $\pm$ 0:53} & \colorbox{green!68}{14\% $\pm$ 10\%} \\
\bottomrule
\end{tabular}
\end{table}

\begin{table}[t]
\centering
\caption{Wilcoxon signed-rank tests comparing task completion times between systems (paired, non-parametric).
Significance levels: $^{***}$ $p<0.001$, $^{**}$ $p<0.01$, $^{*}$ $p<0.05$.
$^a$Faster system indicated in parentheses when significant difference found.}
\label{tab:wilcoxon_tests}
\footnotesize
\setlength{\tabcolsep}{3pt}
\begin{tabular}{llccc}
\toprule
Task & Comparison$^a$ & W statistic & p-value & n \\
\midrule
Task 1 & GenSim vs Cursor (Cursor faster) & 19.00 & 0.0354* & 15 \\
Task 1 & GenSim vs RoboPlayground (RoboPlayground faster) & 0.00 & 0.0000*** & 16 \\
Task 1 & Cursor vs RoboPlayground (RoboPlayground faster) & 15.00 & 0.0010** & 18 \\
Task 2 & GenSim vs Cursor & -- & -- & 0 \\
Task 2 & GenSim vs RoboPlayground (RoboPlayground faster) & 0.00 & 1.0000 & 1 \\
Task 2 & Cursor vs RoboPlayground (RoboPlayground faster) & 12.00 & 0.2500 & 9 \\
\bottomrule
\end{tabular}
\end{table}

\begin{table}[t]
\centering
\caption{McNemar's tests comparing DNF rates between systems (paired binary outcomes).
Significance levels: $^{***}$ $p<0.001$, $^{**}$ $p<0.01$, $^{*}$ $p<0.05$.
$^a$System with fewer DNFs indicated in parentheses.
$^b$Contingency table format: completed/completed, completed/DNF, DNF/completed, DNF/DNF.}
\label{tab:mcnemar_tests}
\footnotesize
\setlength{\tabcolsep}{3pt}
\begin{tabular}{llcc}
\toprule
Comparison$^a$ & Contingency table$^b$ & $\chi^2$ statistic & p-value \\
\midrule
GenSim vs Cursor (Cursor fewer DNFs) & 5/2/11/8 & 4.9231 & 0.0265* \\
GenSim vs RoboPlayground (RoboPlayground fewer DNFs) & 6/1/14/5 & 9.6000 & 0.0019** \\
Cursor vs RoboPlayground (RoboPlayground fewer DNFs) & 14/2/6/4 & 1.1250 & 0.2888 \\
\bottomrule
\end{tabular}
\end{table}

\begin{table}[t]
\centering
\caption{Wilcoxon signed-rank tests comparing SUS scores between systems (paired, non-parametric).
Significance levels: $^{***}$ $p<0.001$, $^{**}$ $p<0.01$, $^{*}$ $p<0.05$.
$^a$System with higher SUS score (better usability) indicated in parentheses when significant difference found.}
\label{tab:wilcoxon_sus_tests}
\footnotesize
\setlength{\tabcolsep}{3pt}
\begin{tabular}{lccc}
\toprule
Comparison$^a$ & W statistic & p-value & n \\
\midrule
GenSim vs Cursor (Cursor higher) & 60.50 & 0.0105* & 26 \\
GenSim vs RoboPlayground (RoboPlayground higher) & 8.50 & 0.0000*** & 26 \\
Cursor vs RoboPlayground (RoboPlayground higher) & 52.00 & 0.0017** & 26 \\
\bottomrule
\end{tabular}
\end{table}

\begin{table}[t]
\centering
\caption{Wilcoxon signed-rank tests comparing TLX workload scores between systems (paired, non-parametric).
Significance levels: $^{***}$ $p<0.001$, $^{**}$ $p<0.01$, $^{*}$ $p<0.05$.
$^a$System with lower TLX score (lower workload, better) indicated in parentheses when significant difference found.}
\label{tab:wilcoxon_tlx_tests}
\footnotesize
\setlength{\tabcolsep}{3pt}
\begin{tabular}{lccc}
\toprule
Comparison$^a$ & W statistic & p-value & n \\
\midrule
GenSim vs Cursor (Cursor lower) & 126.00 & 0.2173 & 26 \\
GenSim vs RoboPlayground (RoboPlayground lower) & 37.00 & 0.0007*** & 26 \\
Cursor vs RoboPlayground (RoboPlayground lower) & 41.50 & 0.0019** & 26 \\
\bottomrule
\end{tabular}
\end{table}

\begin{table}[t]
\centering
\caption{Chi-square goodness-of-fit test for system preference (N=26).
Tests whether the distribution of preferences differs significantly from a uniform distribution.
Significance levels: $^{***}$ $p<0.001$, $^{**}$ $p<0.01$, $^{*}$ $p<0.05$.}
\label{tab:chi2_preference_test}
\footnotesize
\setlength{\tabcolsep}{3pt}
\begin{tabular}{lcc}
\toprule
System & Observed & Expected (uniform) \\
\midrule
GenSim & 2 & 8.7 \\
Cursor & 6 & 8.7 \\
RoboPlayground & 18 & 8.7 \\
\midrule
\multicolumn{3}{l}{$\chi^2$ = 16.0000, p-value = 0.0003***} \\
\bottomrule
\end{tabular}
\end{table}

\begin{table}[t]
\centering
\caption{Friedman test and post-hoc pairwise Wilcoxon tests for usability rankings.
The Friedman test examines whether there are significant differences in ranks across the three systems.
Post-hoc tests compare each pair of systems.
Significance levels: $^{***}$ $p<0.001$, $^{**}$ $p<0.01$, $^{*}$ $p<0.05$.
$^a$System with better rank (lower is better) indicated in parentheses when significant difference found.}
\label{tab:friedman_rank_tests}
\footnotesize
\setlength{\tabcolsep}{3pt}
\begin{tabular}{lccc}
\toprule
\multicolumn{4}{l}{\textbf{Friedman Test (Overall)}} \\
\multicolumn{4}{l}{$\chi^2$ = 23.6154, p-value = 0.0000***, n = 26} \\
\midrule
\multicolumn{4}{l}{\textbf{Post-hoc Pairwise Comparisons (Wilcoxon)}} \\
Comparison$^a$ & W statistic & p-value & n \\
\midrule
GenSim vs Cursor (Cursor better) & 75.00 & 0.0047** & 26 \\
GenSim vs RoboPlayground (RoboPlayground better) & 28.00 & 0.0001*** & 26 \\
Cursor vs RoboPlayground (RoboPlayground better) & 76.00 & 0.0078** & 26 \\
\bottomrule
\end{tabular}
\end{table}

% \subsubsection{Suitability for Evaluation}
% \subsubsection{Scaling with Diversity}
\clearpage
\subsection{Ablative Studies Expanded}
\subsubsection{Metrics}
\label{app:ablation_metrics}
We evaluate each ablation configuration using five complementary metrics that assess different aspects of task generation quality. All success metrics are computed as percentages over $n=26$ benchmark test cases.
  
  \paragraph{Task Success (\%).}                                                                                                              
  The percentage of generated tasks where both the initial state code (\texttt{code.py}) and goal state code (\texttt{goal\_state\_code.py}) pass all validation stages. A task is considered successful only if it compiles, instantiates without runtime errors, runs for at least 10 simulation steps without crashing, and the \texttt{\_success()} method returns \texttt{True} when evaluated at the goal state. This metric captures end-to-end generation quality. 
  
  \paragraph{Compile (\%).}                                              The percentage of tasks that compile without syntax errors. We perform post-hoc independent validation by parsing the generated code with Python's \texttt{ast.parse()} followed by \texttt{compile()} to detect syntax errors. This is computed independently of the pipeline's internal validation to ensure unbiased measurement.    
  
  \paragraph{Smoke Test (\%).}                                            The percentage of tasks that can be instantiated and run for 10 simulation steps without crashing. This metric tests runtime correctness beyond syntax validity: the environment must initialize properly           
  (\texttt{env.reset()}), accept zero-action inputs, and execute physics steps without exceptions. Tasks that pass smoke testing are structurally sound even if semantically incorrect.                      
  
  \paragraph{Human-Verified (\%).}                                        The percentage of tasks verified as semantically correct through manual inspection of rendered snapshots. For each task, we generate PNG images of both the initial state (randomized object positions) and goal    
  state (objects in target configuration). Human evaluators assess whether the visual scene matches the task description and whether the goal configuration is physically plausible. This metric captures semantic correctness that automated metrics may miss.                   \paragraph{LLM Alignment (0--100).} An LLM-based score evaluating alignment between the \texttt{\_success()} method implementation and the task description. We extract the success method source code via AST parsing, then prompt an LLM to analyze   
  whether the implementation correctly captures the described success conditions. The LLM returns a score from 0 (no alignment) to 100 (perfect alignment) along with identified strengths and weaknesses. This metric assesses whether the generated success logic matches the intended task semantics without requiring simulation execution.

\begin{figure}[t]
\centering
\begin{minipage}{\linewidth}
\begin{lstlisting}[
    language=Python,
    basicstyle=\ttfamily\scriptsize,
    keywordstyle=\color{blue}\bfseries,
    stringstyle=\color{orange!80!black},
    commentstyle=\color{green!50!black}\itshape,
    showstringspaces=false,
    breaklines=true,
    frame=single,
    framerule=0.5pt,
    xleftmargin=4pt,
    xrightmargin=4pt,
    aboveskip=6pt,
    belowskip=6pt,
    columns=flexible,
    keepspaces=true
]
You are a code review expert. Review whether 
this _success() method correctly validates the task intent.

## Task Description
{task_description}

## Generated _success() Method
{success_method}

## Review Criteria
1. Does _success() check ALL conditions implied by the task?
2. Are there missing checks (e.g., ordering, colors, positions)?
3. Are there extra checks not required by the task?
4. Is the logic sound (correct comparisons, thresholds)?

## Output Format
Respond with ONLY a valid JSON object:
{
    "alignment_score": <0-100 integer>,
    "aligned": <true if score >= 80, false otherwise>,
    "missing_checks": ["list of missing checks"],
    "extra_checks": ["list of unnecessary checks"],
    "logic_issues": ["list of logical problems"],
    "reasoning": "brief explanation of the analysis",
    "success_conditions_breakdown": {
        "final_condition": "line where _success_check is assigned",
        "boolean_variables": [{
            "variable_name": "e.g., positions_ok",
            "code_definition": "exact definition line",
            "human_explanation": "what this checks",
            "dependencies": ["sub-variables it depends on"]
        }],
        "condition_tree": "e.g., Success requires: (1) red 
            on table AND (2) blue lifted AND (3) red released"
    }
}
\end{lstlisting}
\end{minipage}
\caption{\textbf{LLM Alignment Evaluation Prompt.} The prompt template used to assess whether a generated \texttt{\_success()} method correctly captures the task intent. The LLM reviews the method against the original task description and returns a structured JSON response containing an alignment score (0--100), identified gaps (missing or extraneous checks), and a decomposition of the success condition logic.}
\label{fig:llm-alignment-prompt}
\end{figure}
\subsubsection{Test Tasks}
% Benchmark Task Results - include this in your main document
% Required packages in preamble: graphicx, grffile, amsmath, amssymb

% Task card command for benchmark results

Figures~\ref{fig:benchmark_add_5_block_alignment} through \ref{fig:benchmark_add_5_stack_and_modify} illustrate the ten benchmark tasks used to evaluate our system, spanning spatial arrangement, structural construction, semantic reasoning, and long-horizon task refinement.
Several tasks test single-shot spatial reasoning under a fixed prompt, such as linear color alignment (Fig.~\ref{fig:benchmark_add_5_block_alignment}), circular arrangements (Fig.~\ref{fig:benchmark_add_5_circle_test}), semantic spelling (Fig.~\ref{fig:benchmark_add_5_semantic_spelling}), and parallel multi-structure construction (Fig.~\ref{fig:benchmark_add_5_multiple_stacks}).
Other tasks emphasize progressive steering and task evolution, where users iteratively modify goals over multiple turns, including pyramid construction with semantic substitutions (Fig.~\ref{fig:benchmark_add_5_pyramid_construction}), long-horizon tower modification (Fig.~\ref{fig:benchmark_add_5_progressive_tower}), and reverting to earlier task states before branching in new directions (Fig.~\ref{fig:benchmark_add_5_evolution_revert_and_extend}).
We additionally include tasks that require semantic categorization and controlled perturbations, such as pile sorting with mixed block types and deliberate outliers (Fig.~\ref{fig:benchmark_add_5_pile_sorting}), as well as transitions between two-dimensional and three-dimensional structures (Fig.~\ref{fig:benchmark_add_5_stack_and_modify}).
Together, these benchmarks probe not only execution accuracy, but also the system’s ability to interpret abstract language, maintain task context over time, and adapt to non-linear refinements within a structured physical domain.

\providecommand{\benchmarkcard}[4]{%
  \begin{minipage}[t]{0.48\textwidth}
    \centering
    \begin{tabular}{@{}c@{\hskip 2pt}c@{\hskip 2pt}c@{}}
      \includegraphics[width=0.42\linewidth,trim=80 80 80 80,clip]{#3} &
      \raisebox{1.5cm}{\Large$\boldsymbol{\Rightarrow}$} &
      \includegraphics[width=0.42\linewidth,trim=80 80 80 80,clip]{#4}\\[-2pt]
      {\scriptsize Initial State} & & {\scriptsize Final State}
    \end{tabular}\\[3pt]
    \textbf{#1}\\[1pt]
    {\small\textit{#2}}
  \end{minipage}%
}

% ────────────────────────────────────────────────────────────
% BLOCK ALIGNMENT
% ────────────────────────────────────────────────────────────
\begin{figure*}[htbp]
\centering

\benchmarkcard{Base}
  {Align the blocks in a color gradient from left to right on the table}
  {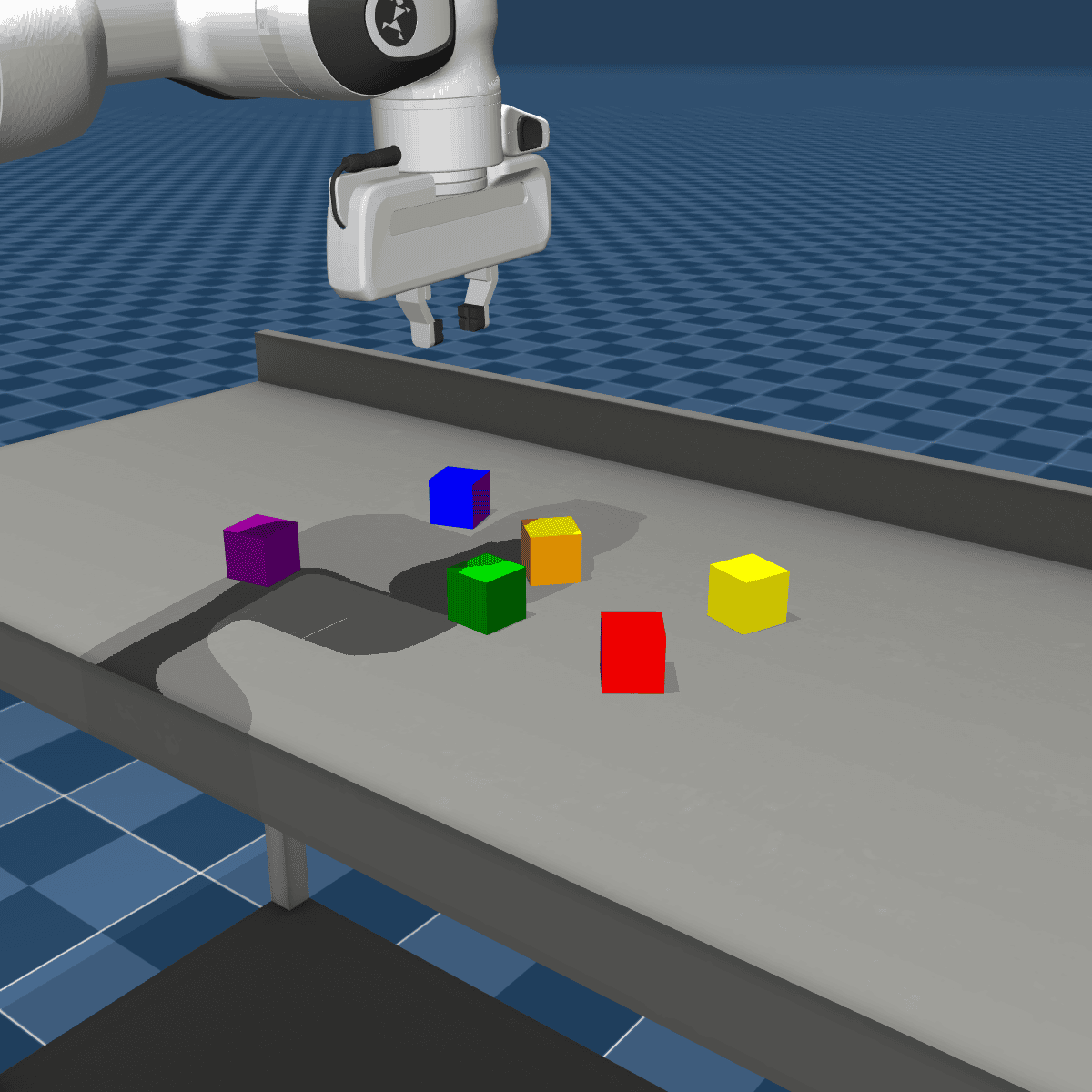}
  {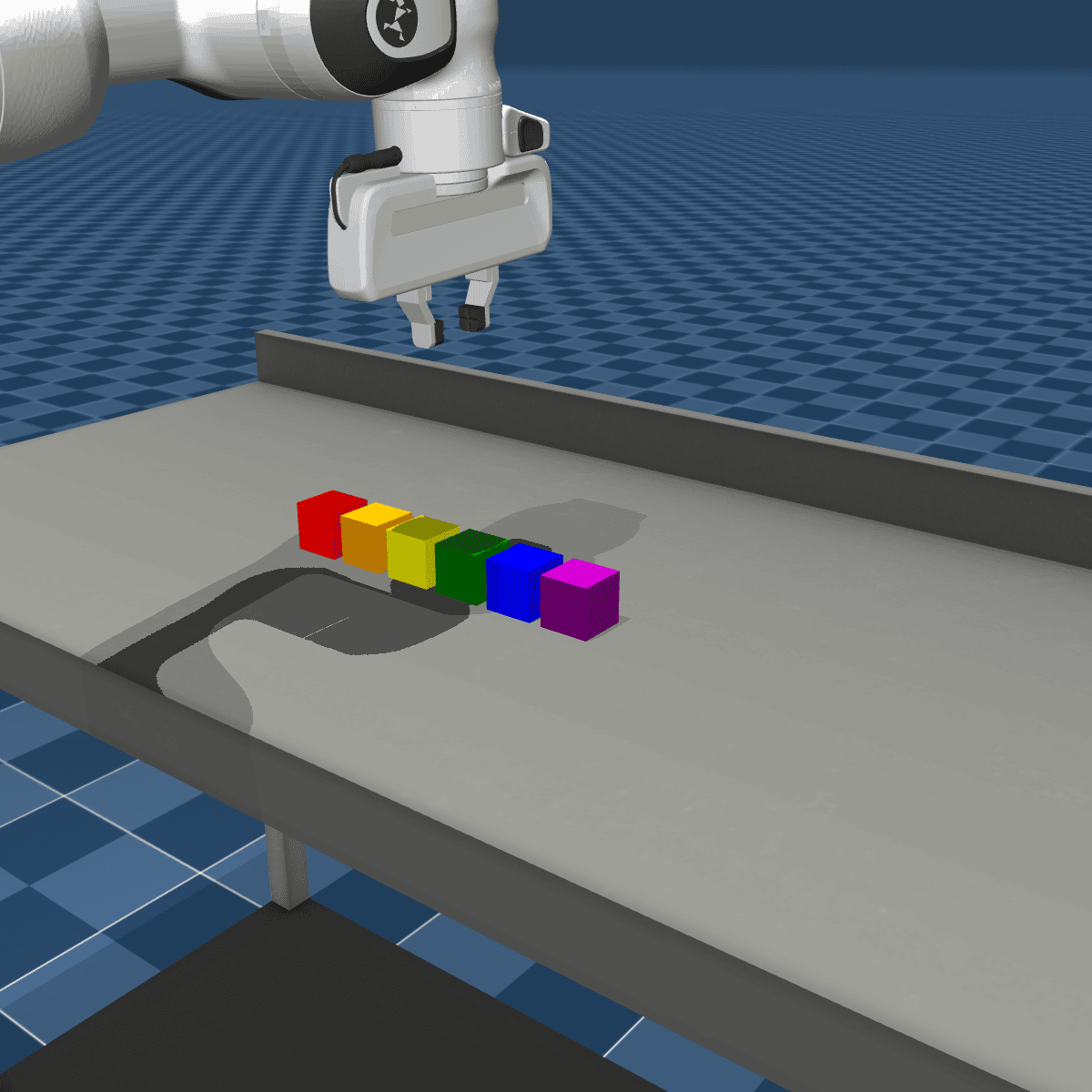}

\caption{\textbf{Block Alignment}}
\label{fig:benchmark_add_5_block_alignment}
\end{figure*}

% ────────────────────────────────────────────────────────────
% BLOCK STACKING TO PYRAMID BUILDING
% ────────────────────────────────────────────────────────────
\begin{figure*}[htbp]
\centering

\benchmarkcard{Base}
  {Stack colored blocks on matching color goal patches where each stack is the same color}
  {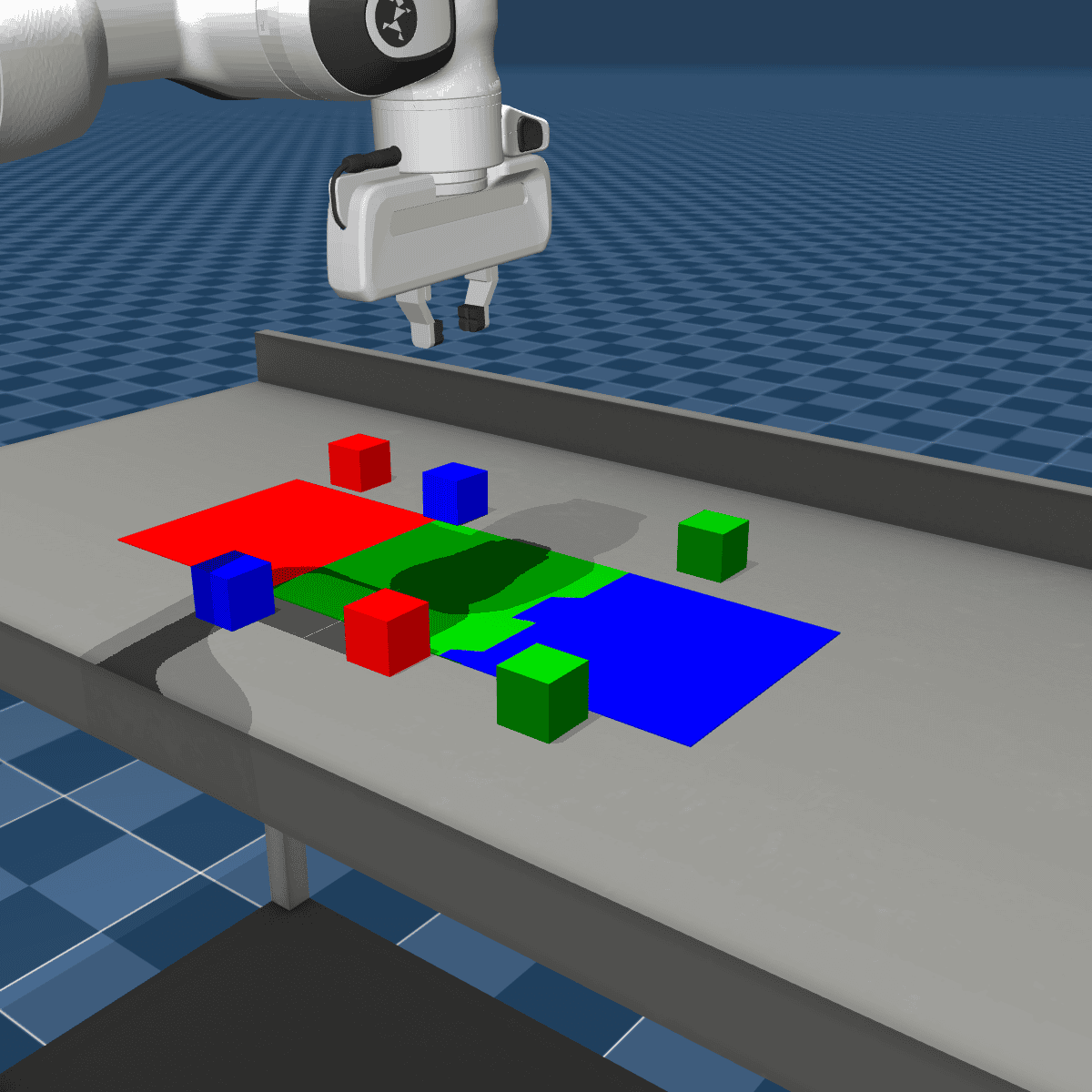}
  {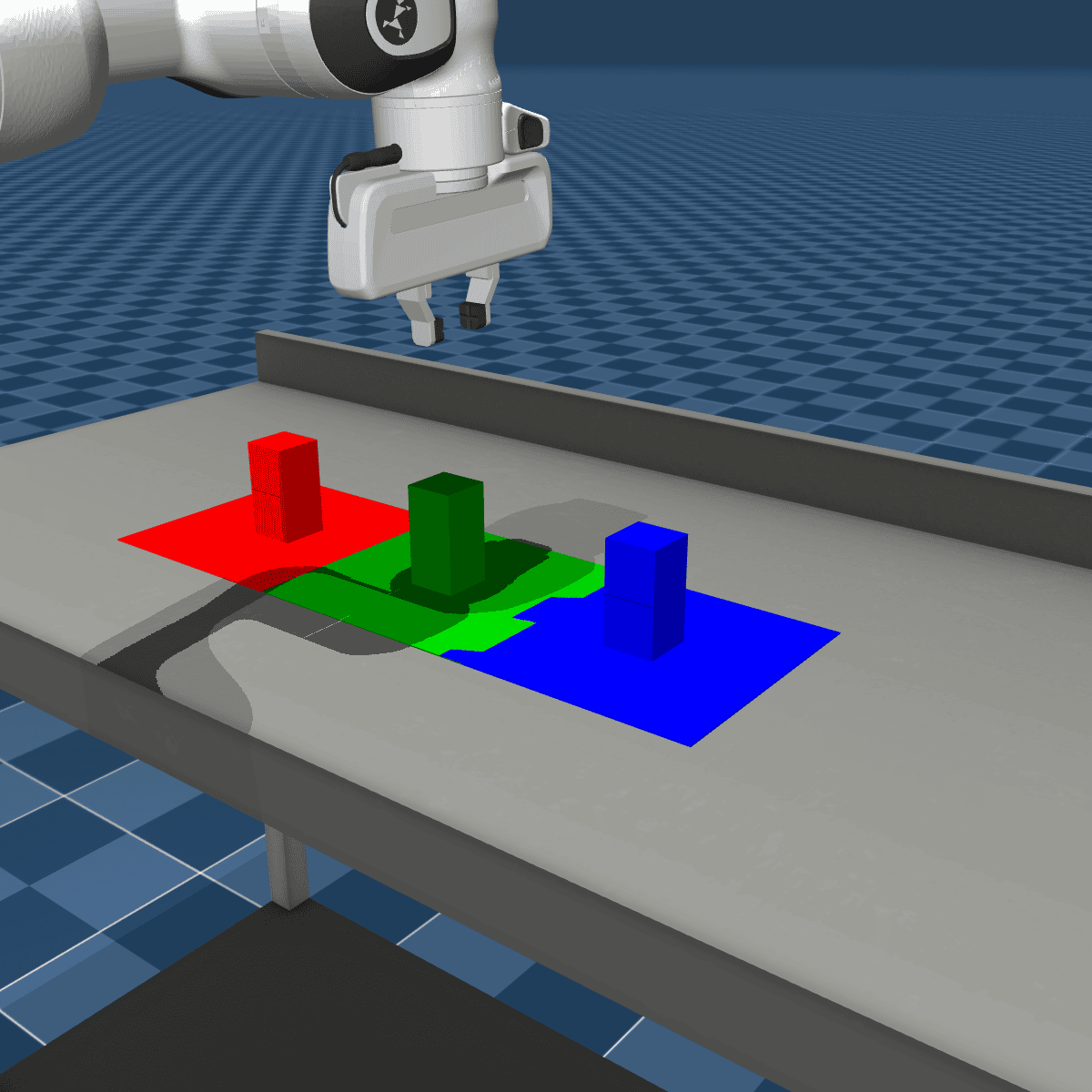}
\hfill
\benchmarkcard{Mod 1}
  {Make a letter pyramid tower with blocks that show letters alphabetically, base is 3 horizontal blocks, two blocks as middle layers and one block on top}
  {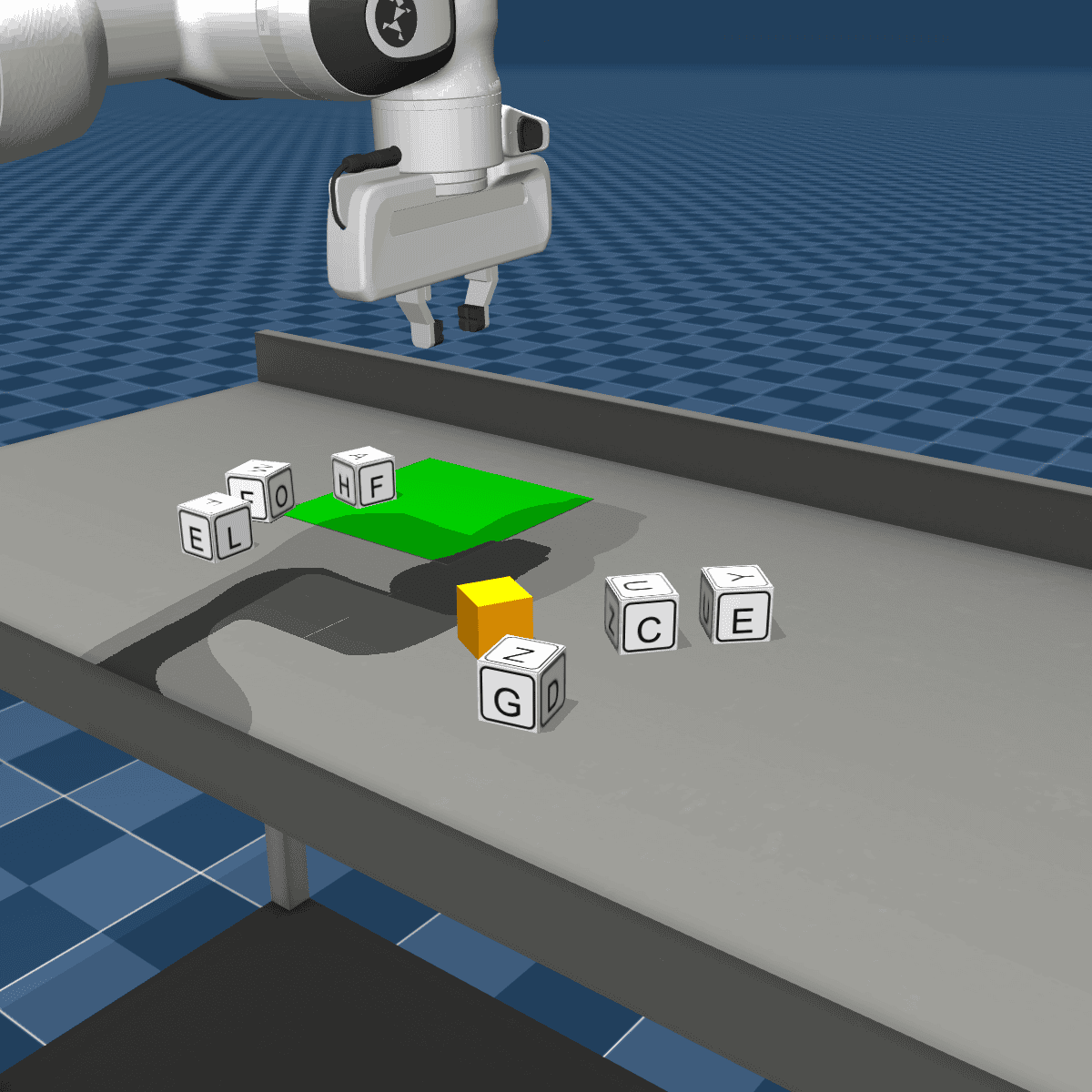}
  {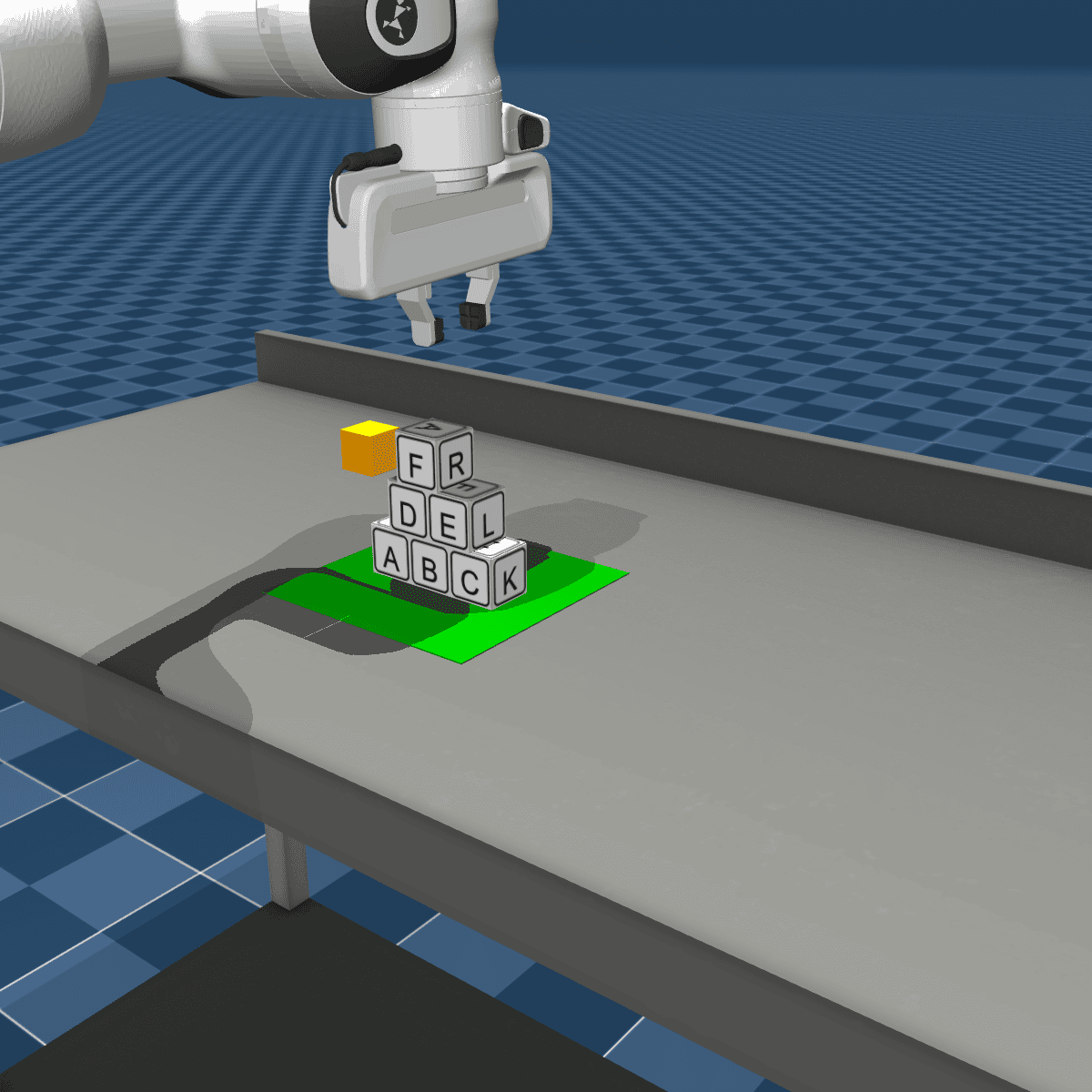}

\caption{\textbf{Block Stacking to Pyramid Building}}
\label{fig:benchmark_add_5_block_stacking_to_pyramid_building}
\end{figure*}

% ────────────────────────────────────────────────────────────
% CIRCLE TEST
% ────────────────────────────────────────────────────────────
\begin{figure*}[htbp]
\centering

\benchmarkcard{Base}
  {Can you build blocks that form a circle shape on the table}
  {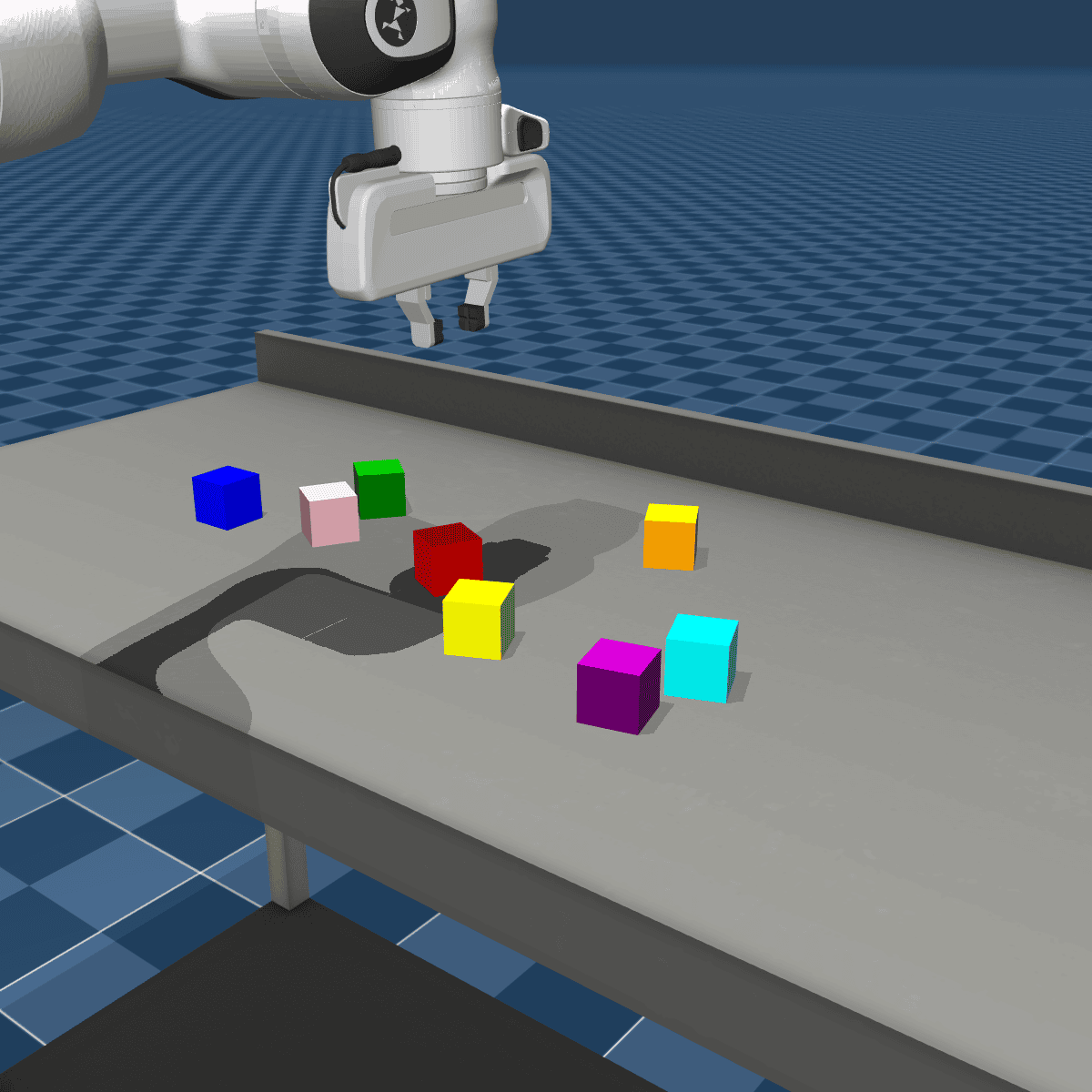}
  {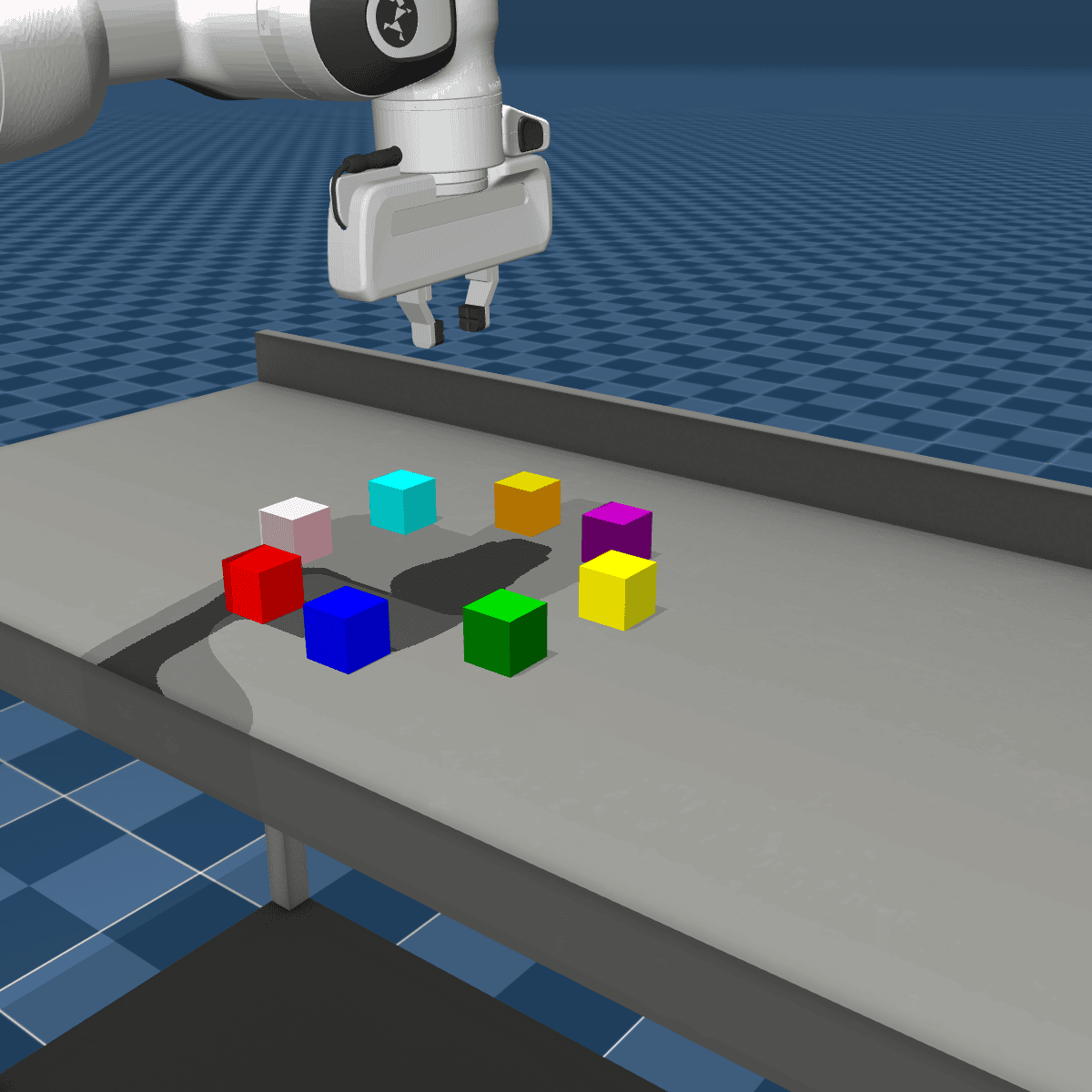}

\caption{\textbf{Circle Test}}
\label{fig:benchmark_add_5_circle_test}
\end{figure*}

% ────────────────────────────────────────────────────────────
% EVOLUTION REVERT AND EXTEND
% ────────────────────────────────────────────────────────────
\begin{figure*}[htbp]
\centering

\benchmarkcard{Base}
  {Make a simple 2-block tower with a red block on bottom and blue on top}
  {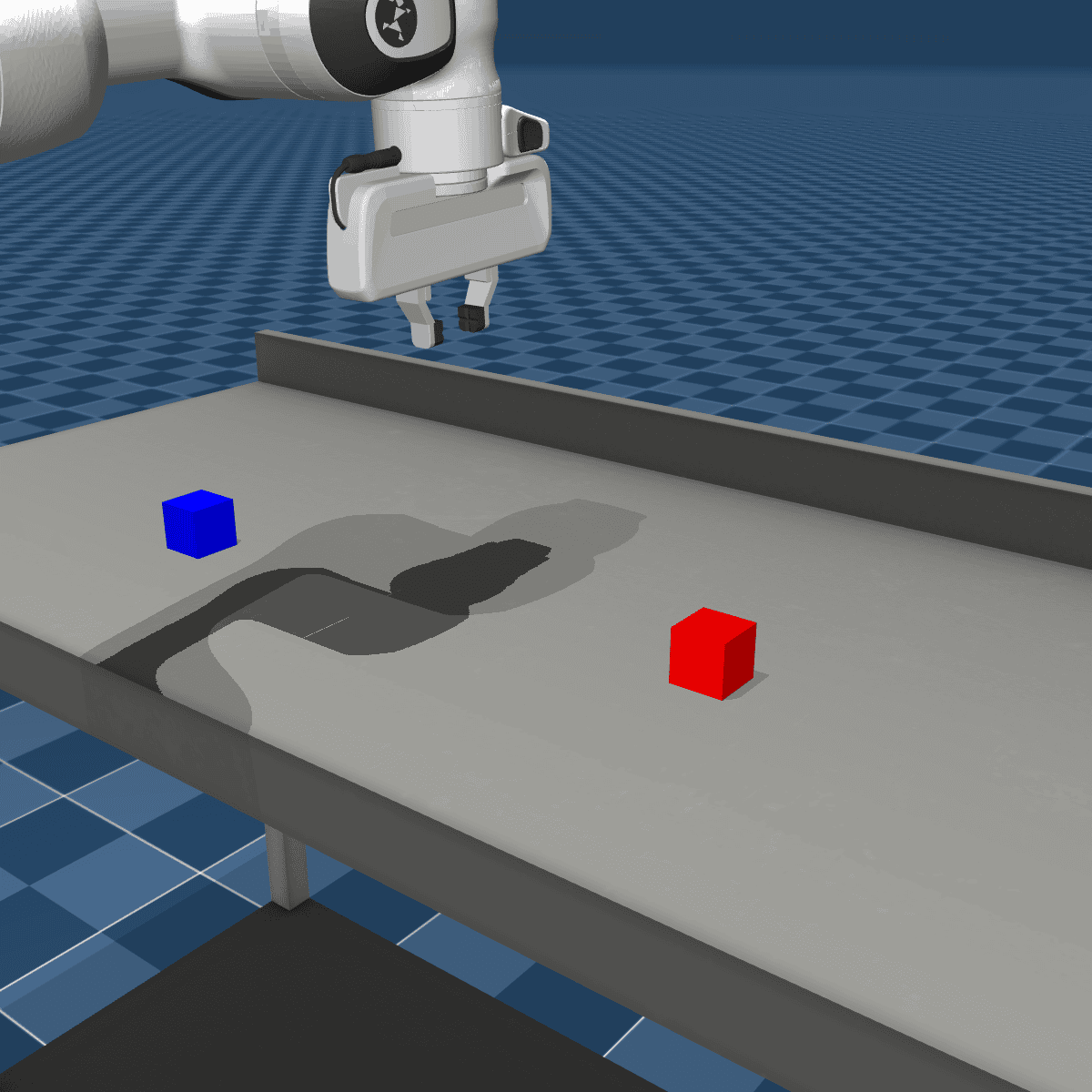}
  {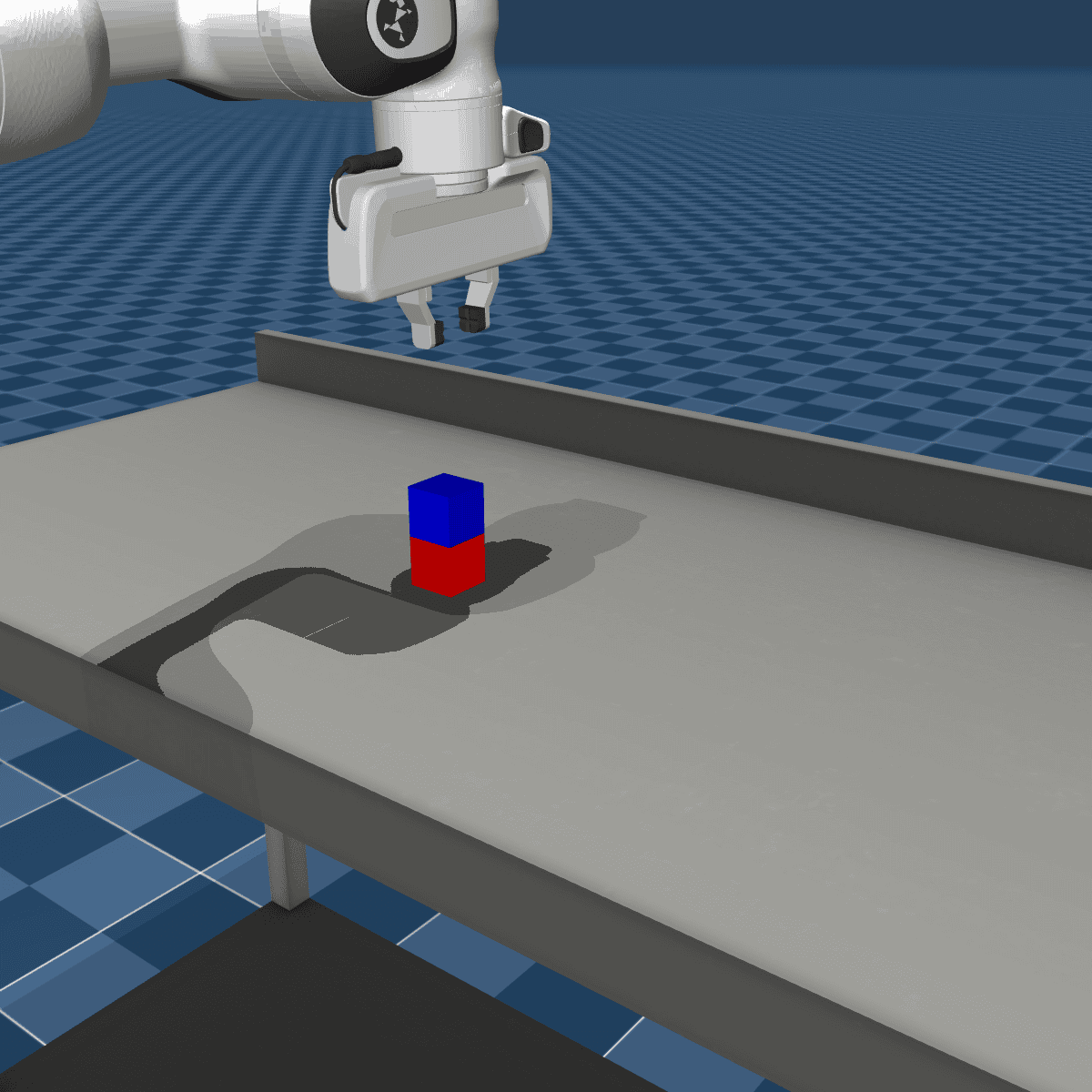}
\hfill
\benchmarkcard{Mod 1}
  {Add a green block on top}
  {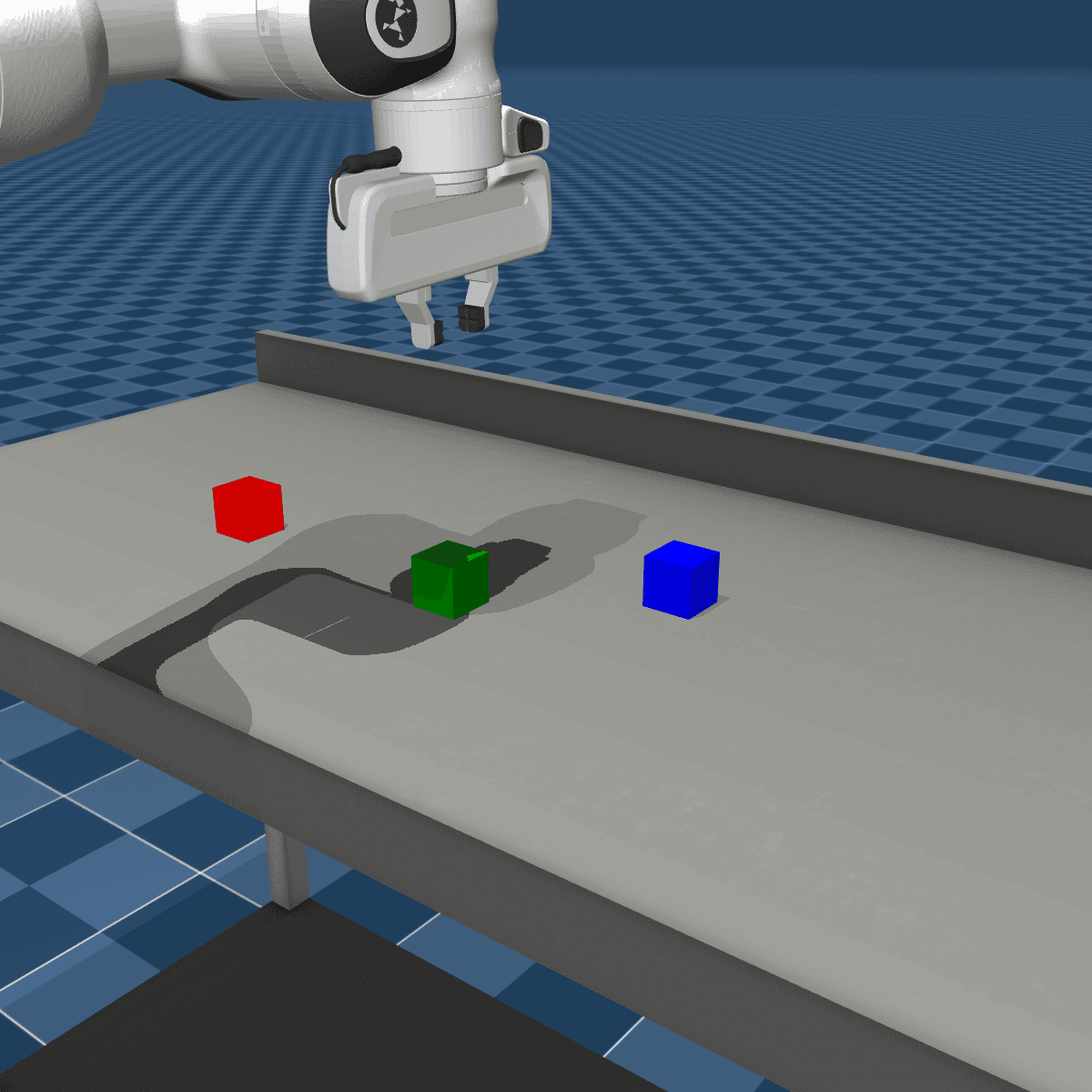}
  {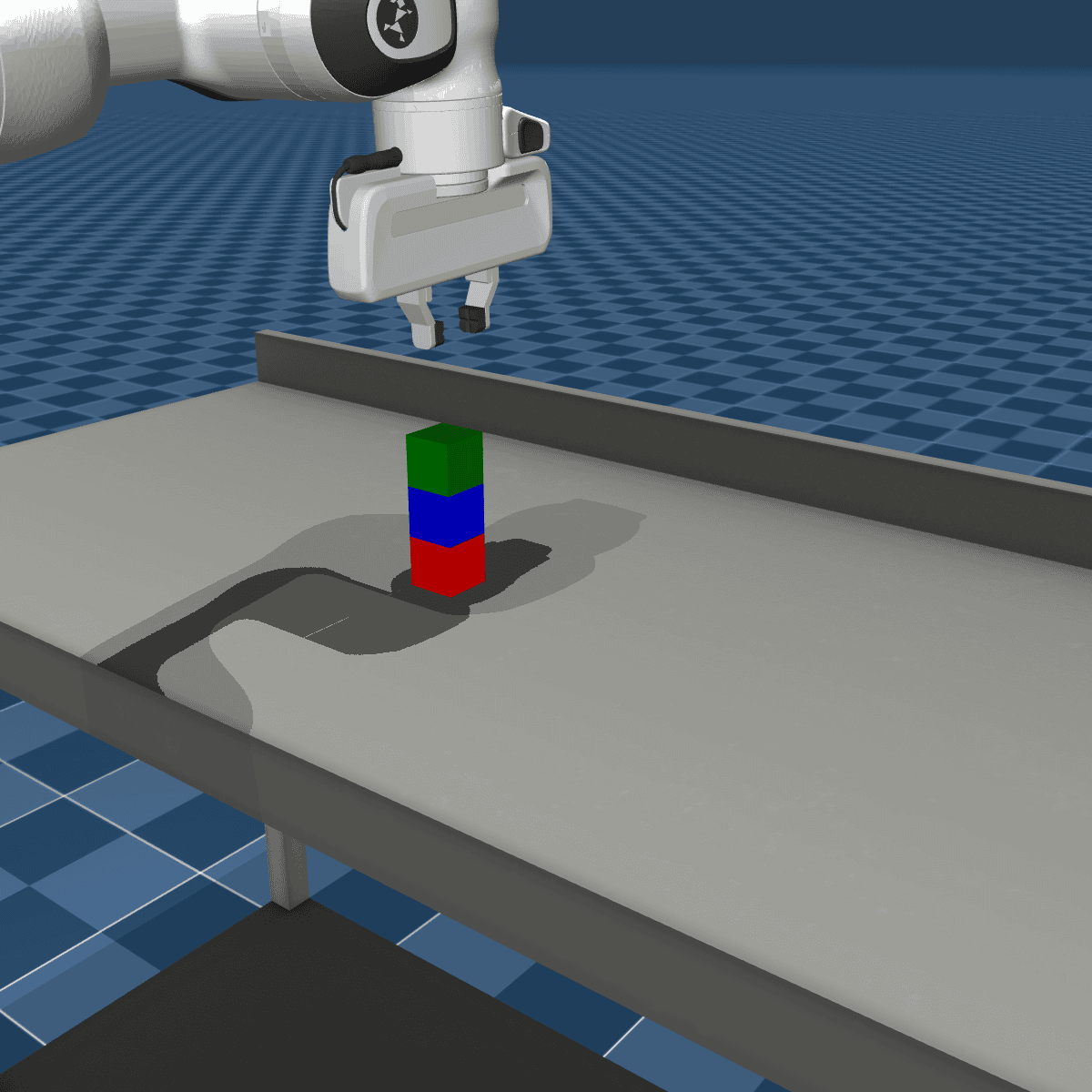}
\vspace{8pt}

\benchmarkcard{Mod 2}
  {Add a yellow block on top of that}
  {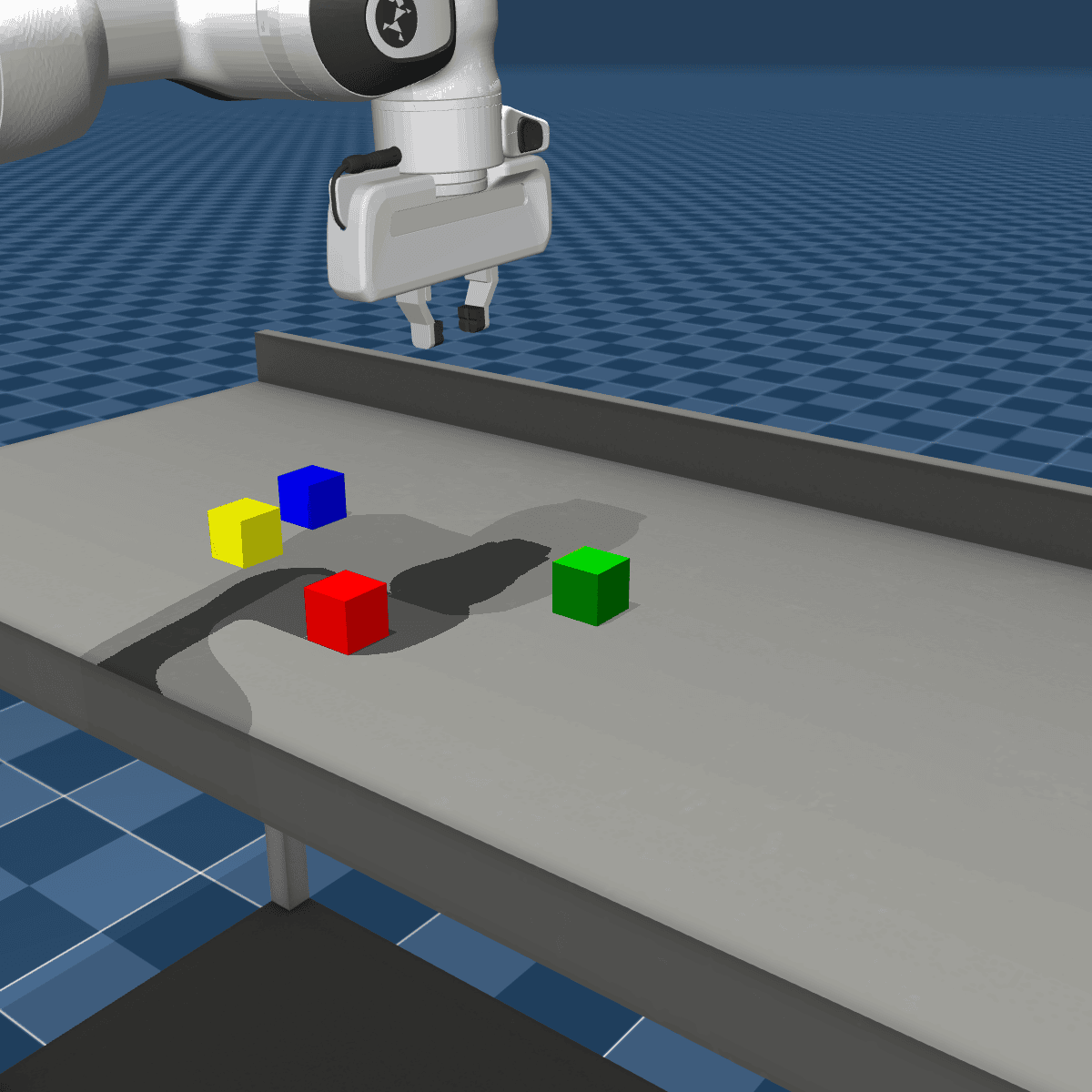}
  {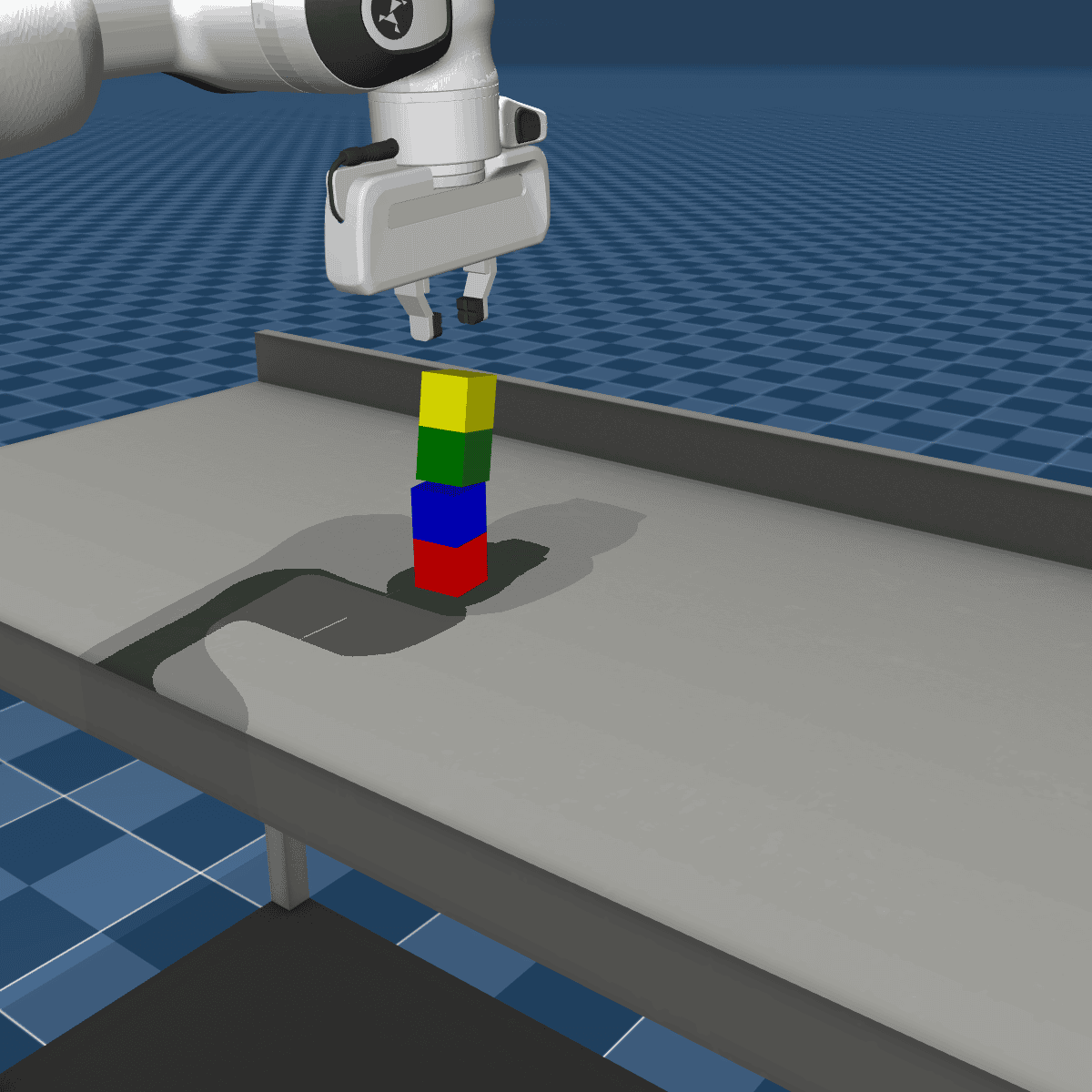}
\hfill
\benchmarkcard{Mod 3}
  {Actually, go back to when it was just red and blue and instead add a purple block}
  {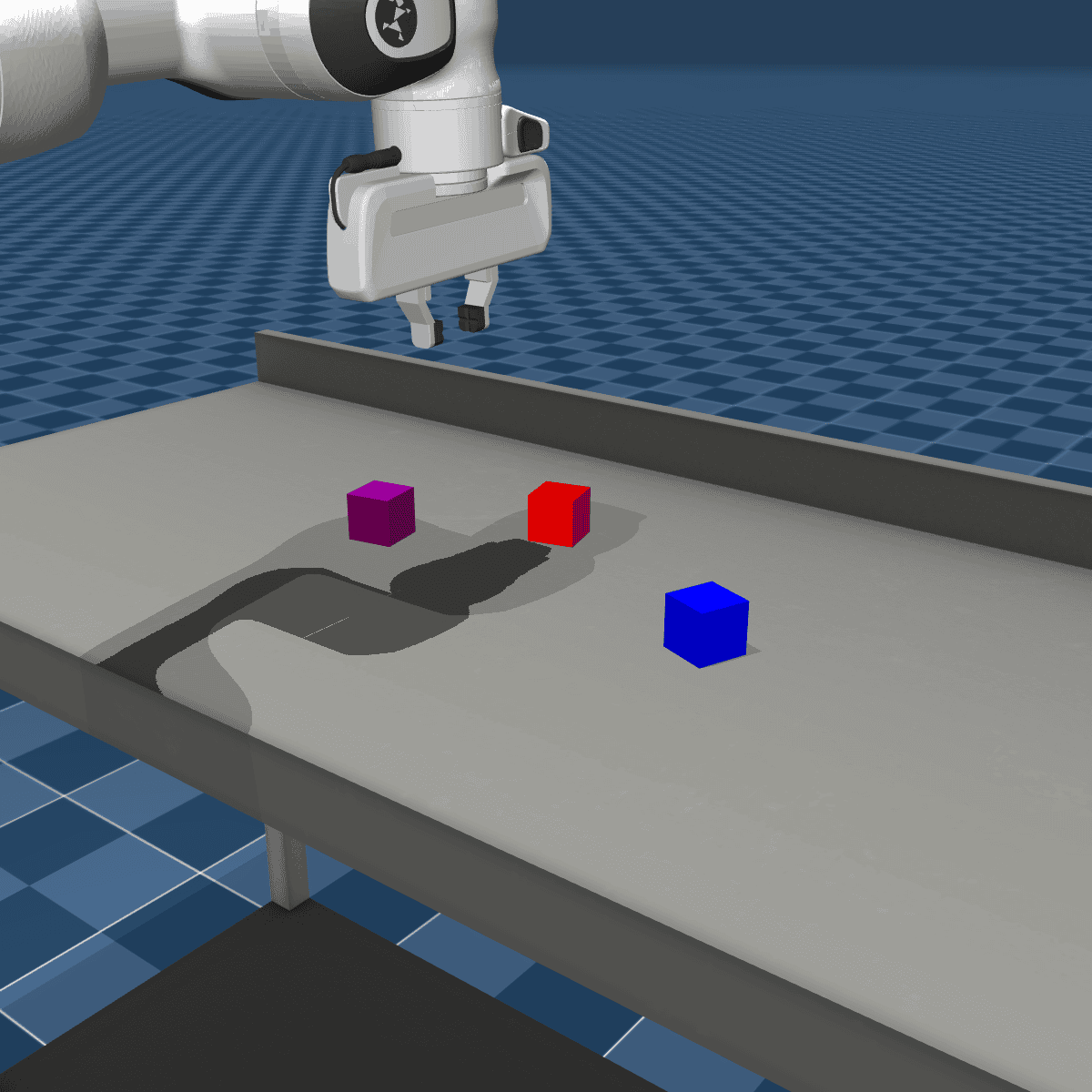}
  {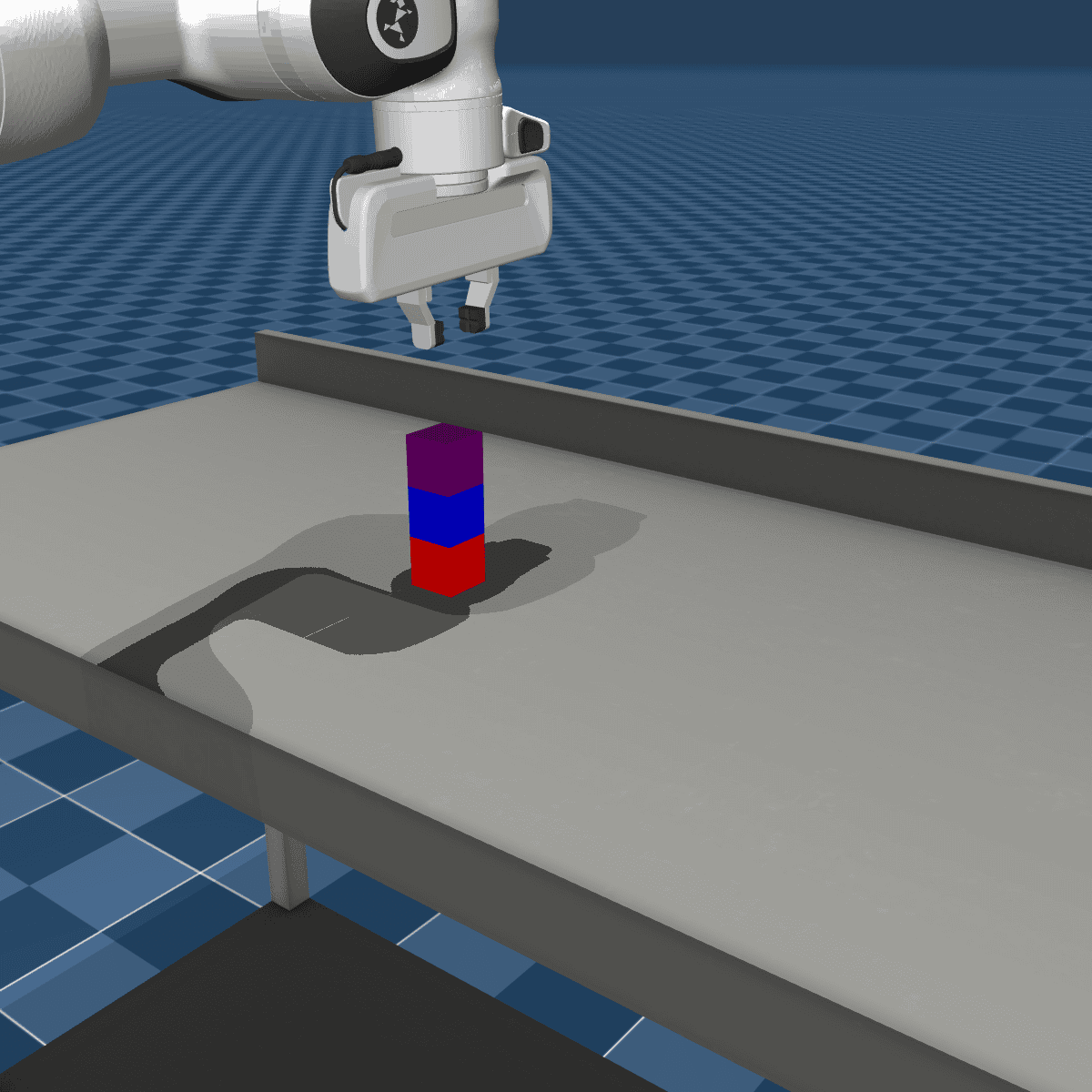}
\vspace{8pt}

\benchmarkcard{Mod 4}
  {Now from this simpler version, add letter cubes instead of colored ones}
  {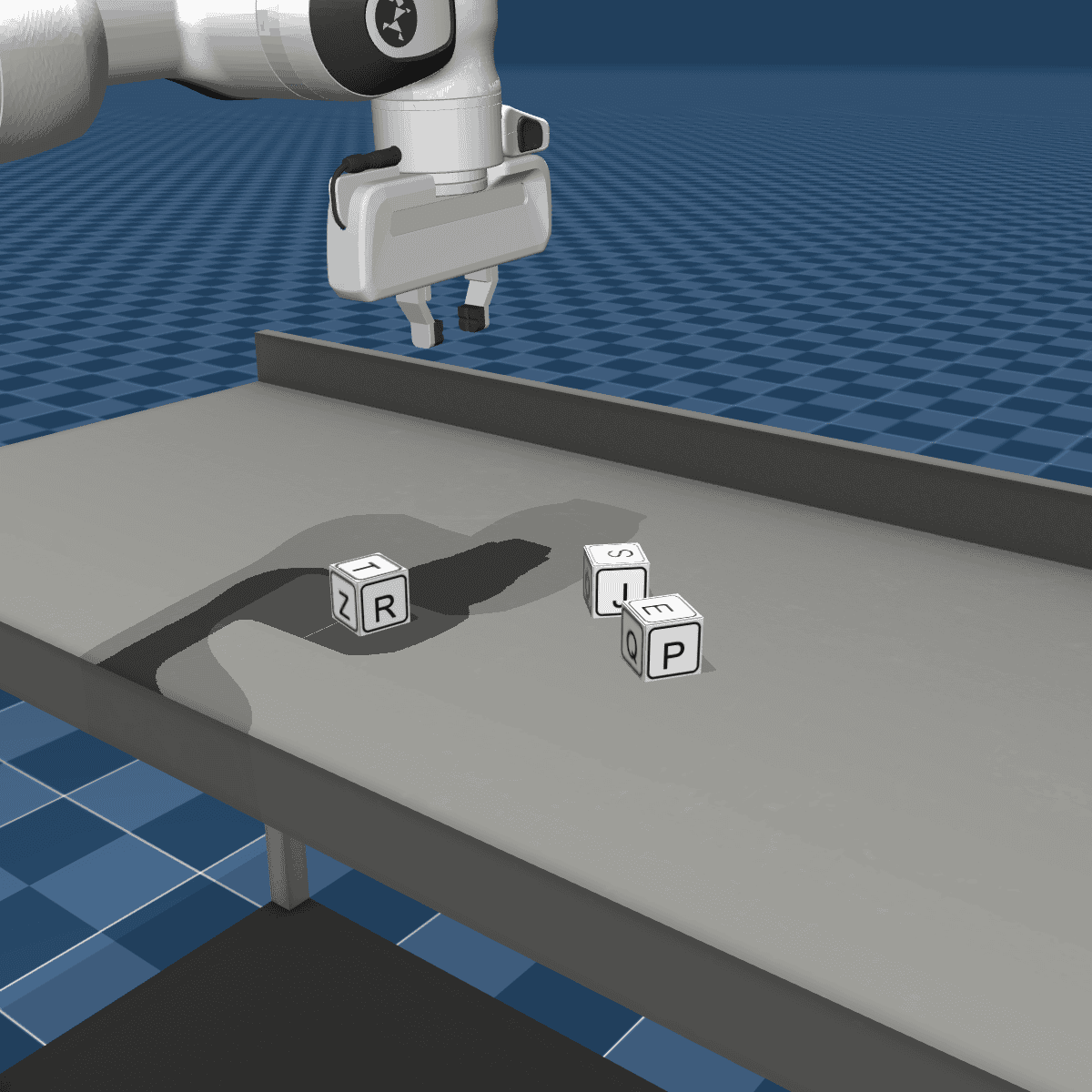}
  {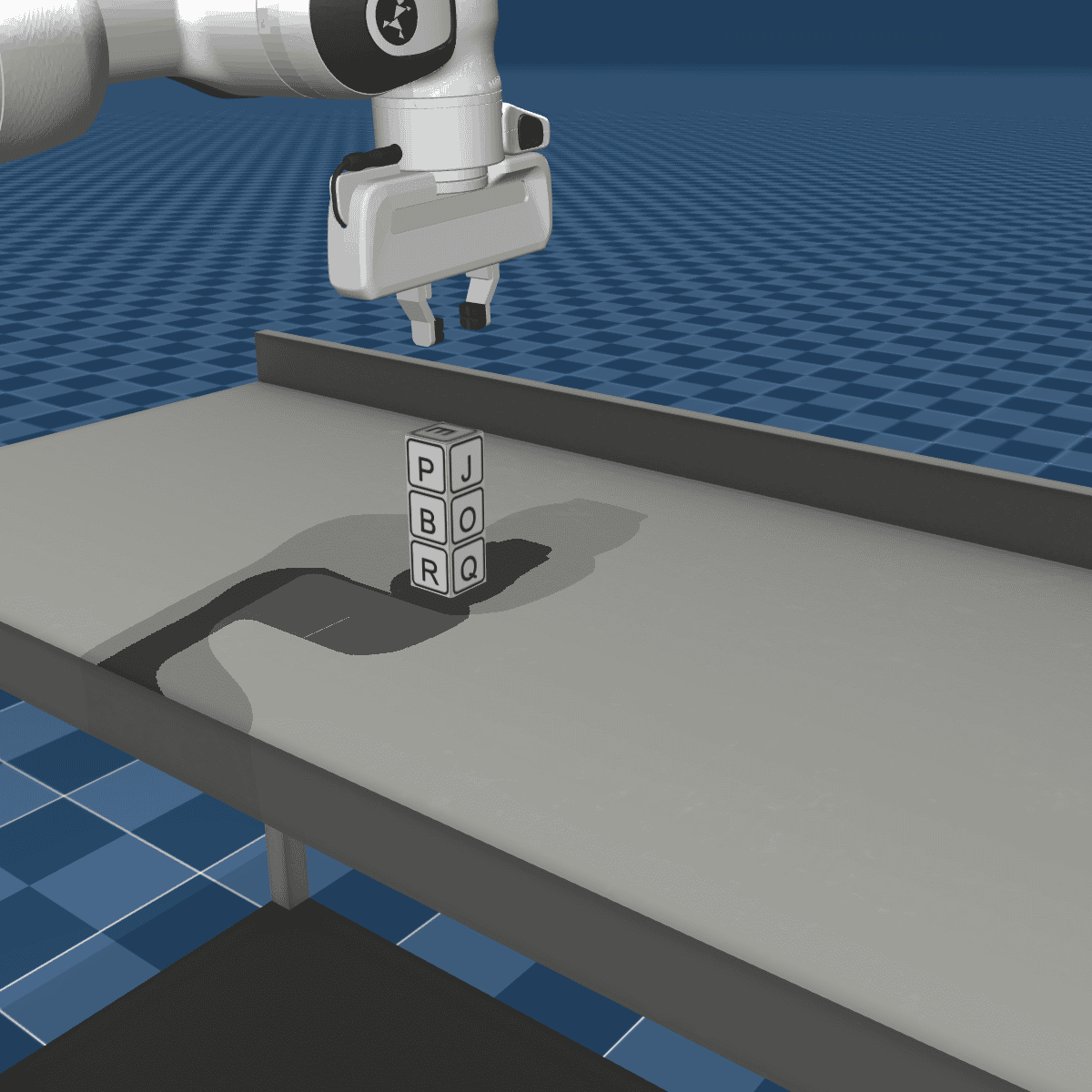}

\caption{\textbf{Evolution Revert and Extend}}
\label{fig:benchmark_add_5_evolution_revert_and_extend}
\end{figure*}

% ────────────────────────────────────────────────────────────
% MULTIPLE STACKS
% ────────────────────────────────────────────────────────────
\begin{figure*}[htbp]
\centering

\benchmarkcard{Base}
  {Stack two towers of blocks on color-specific goal patches on the table}
  {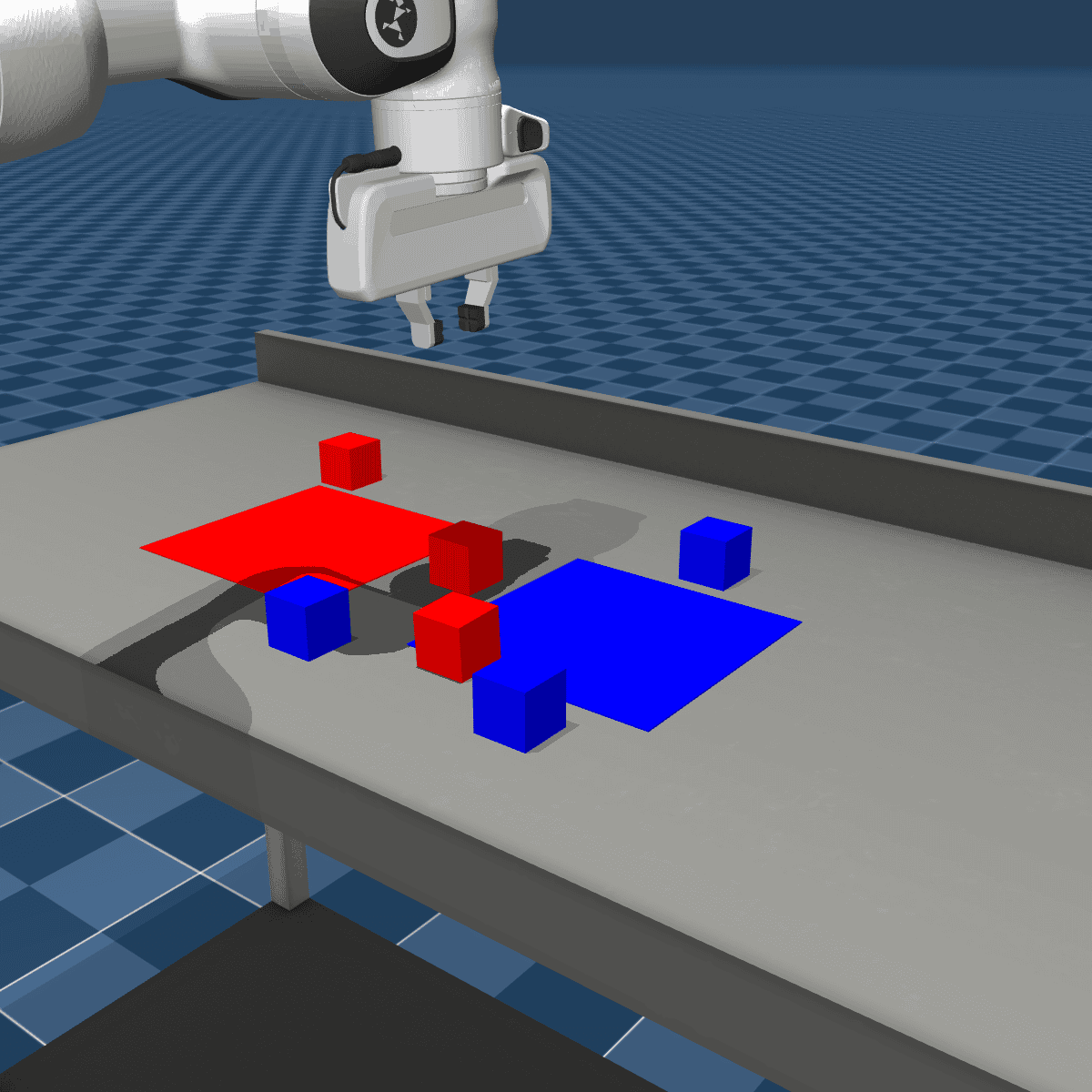}
  {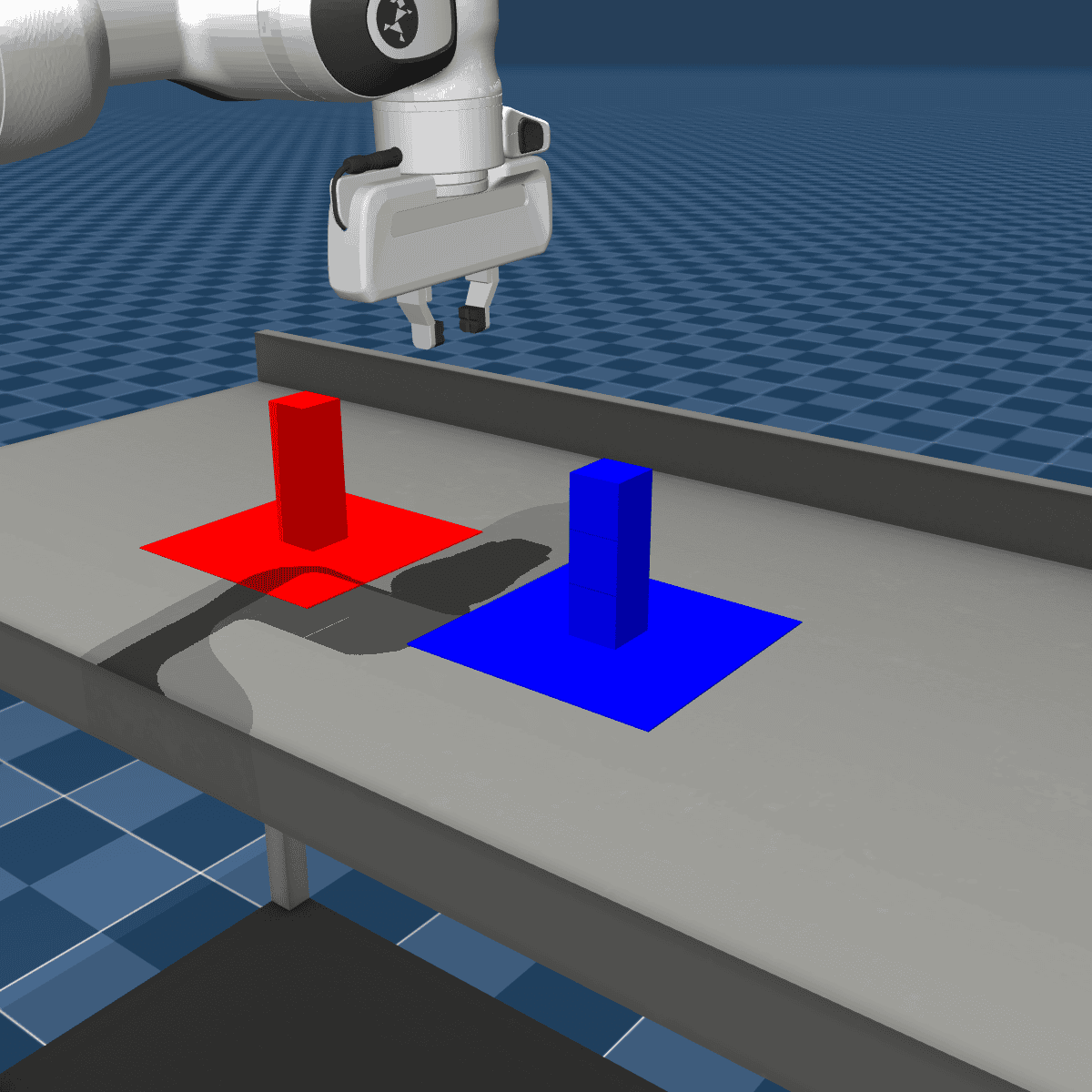}

\caption{\textbf{Multiple Stacks}}
\label{fig:benchmark_add_5_multiple_stacks}
\end{figure*}

% ────────────────────────────────────────────────────────────
% PILE SORTING
% ────────────────────────────────────────────────────────────
\begin{figure*}[htbp]
\centering

\benchmarkcard{Base}
  {Sort piles of blocks by color into separated bunches}
  {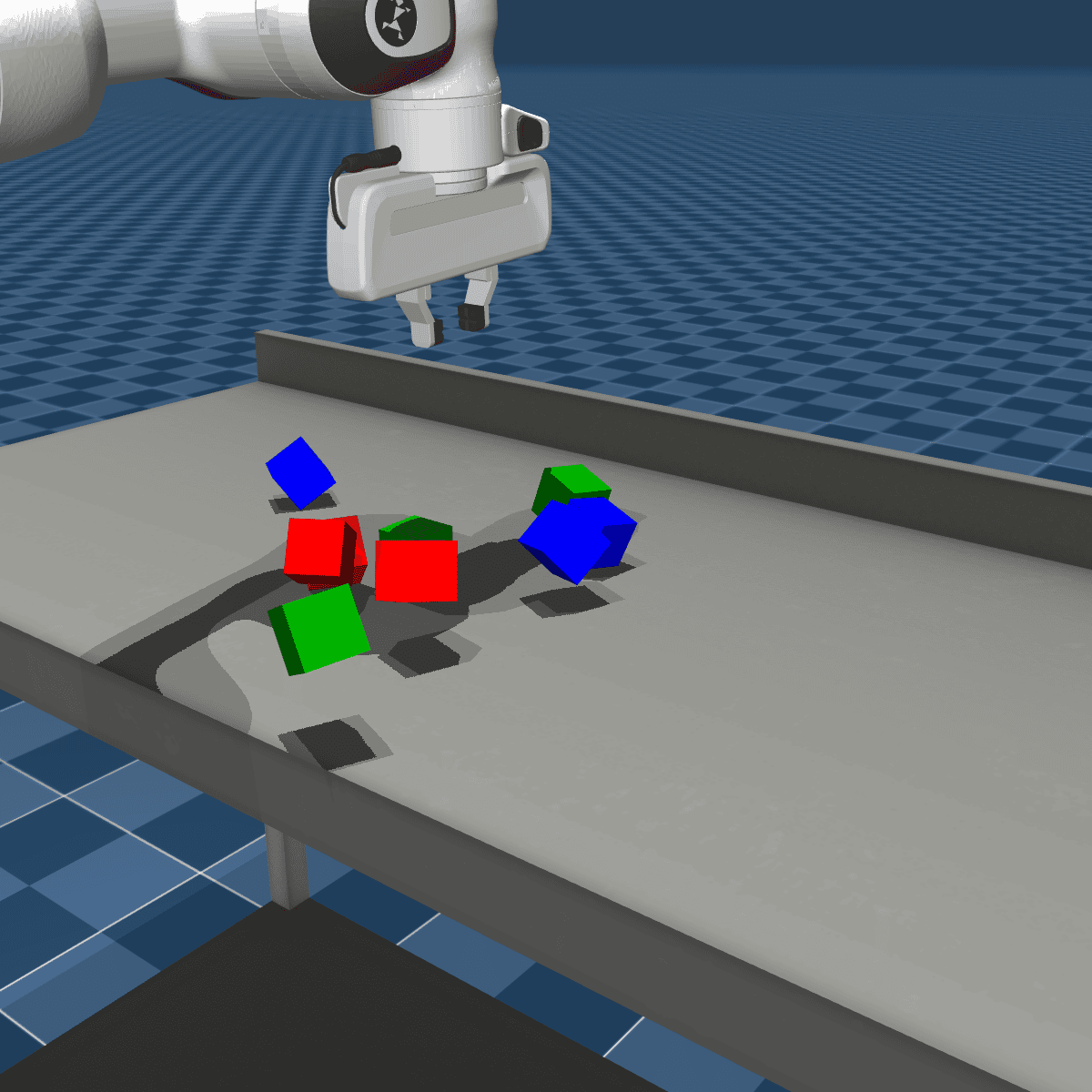}
  {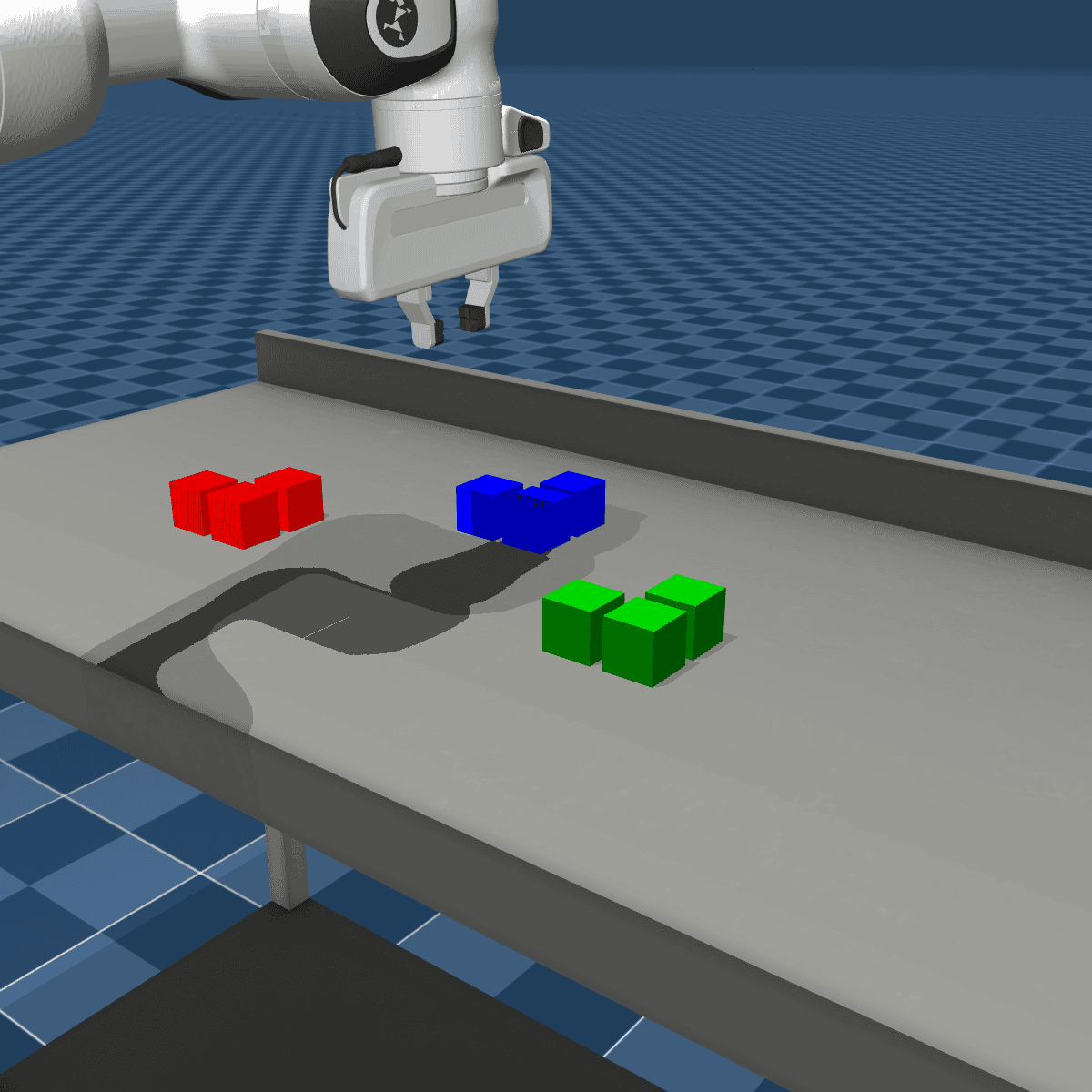}
\hfill
\benchmarkcard{Mod 1}
  {Change blocks to letters, numbers, colors mixed; sort into respective piles}
  {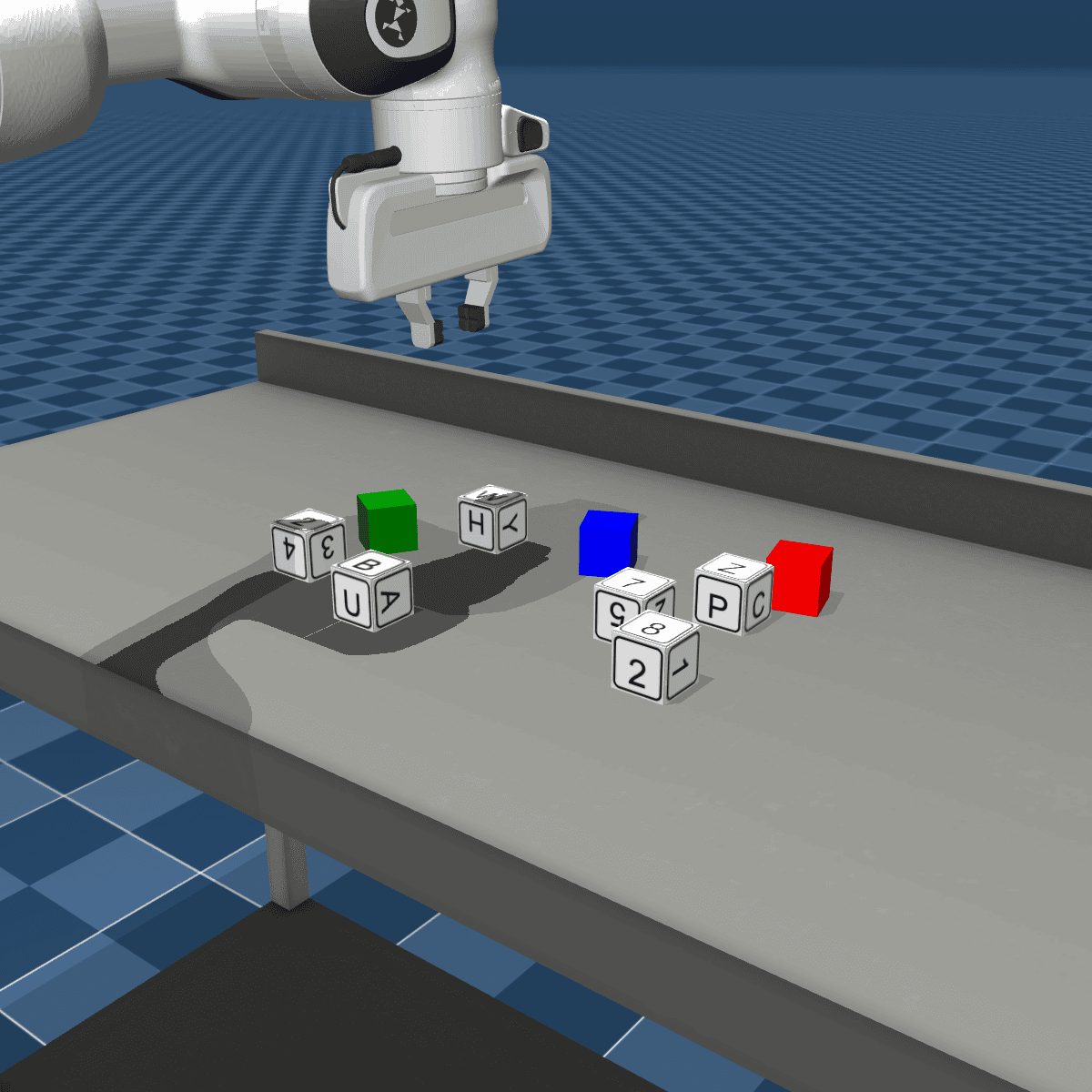}
  {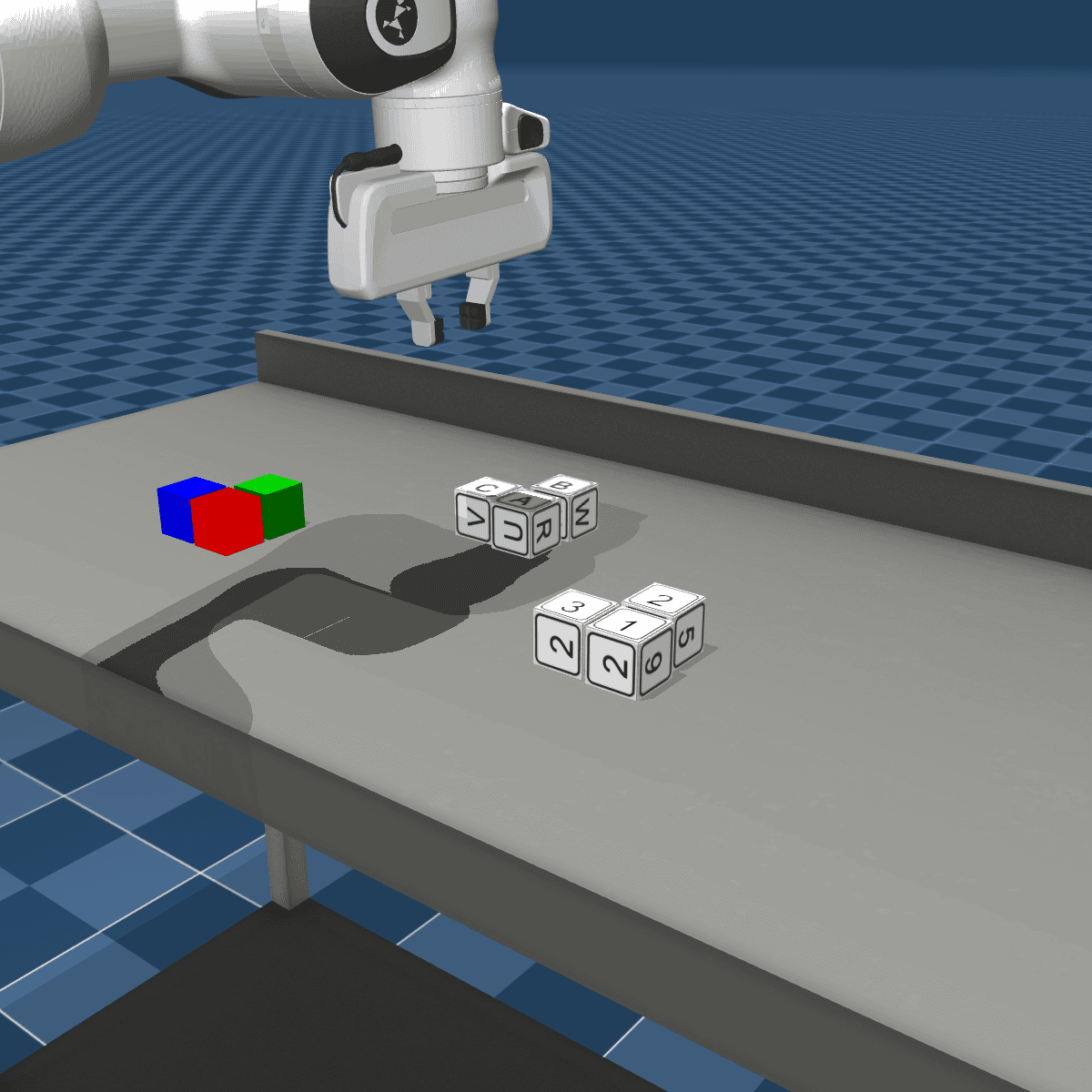}
\vspace{8pt}

\benchmarkcard{Mod 2}
  {Swap one block from each pile into another so each has one outlier}
  {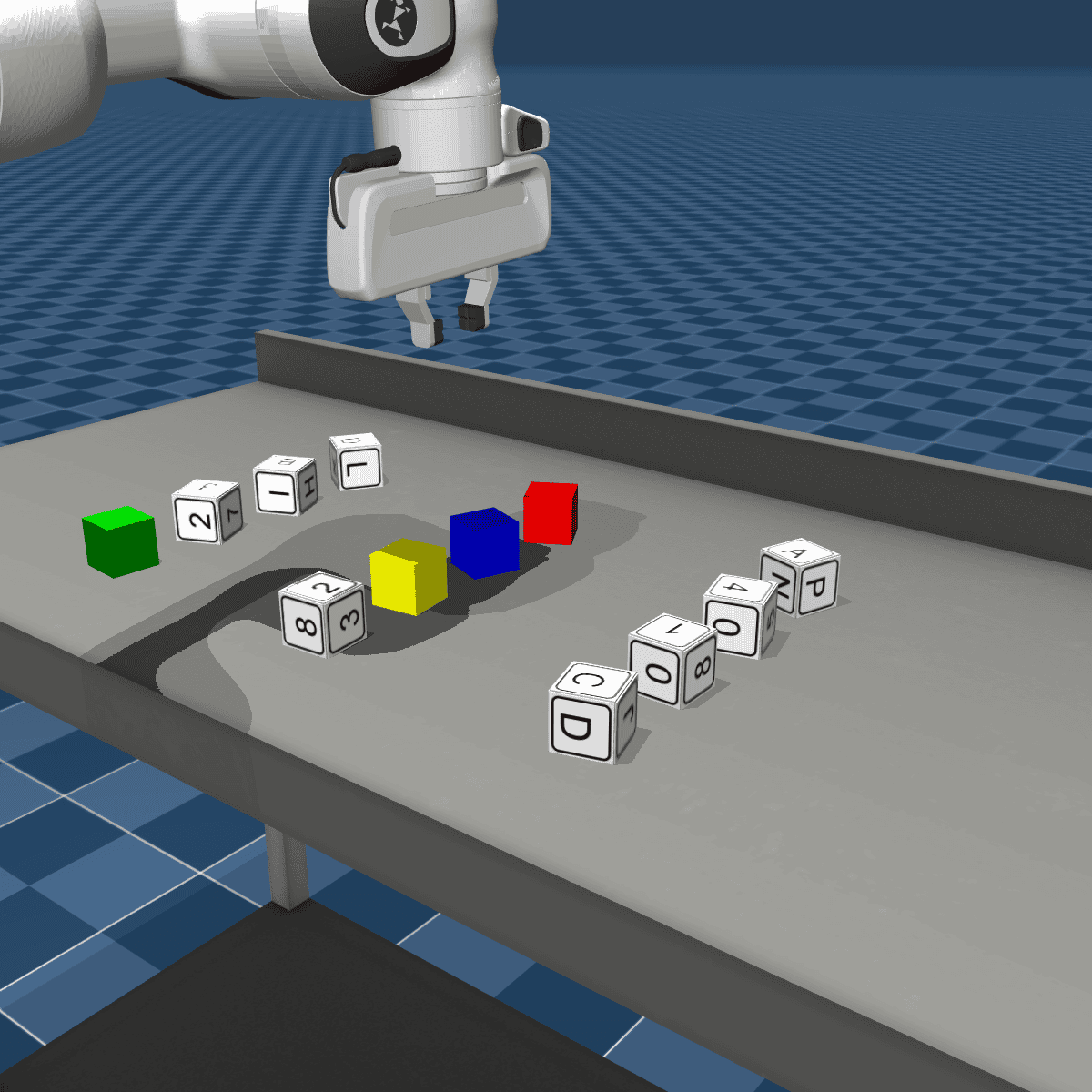}
  {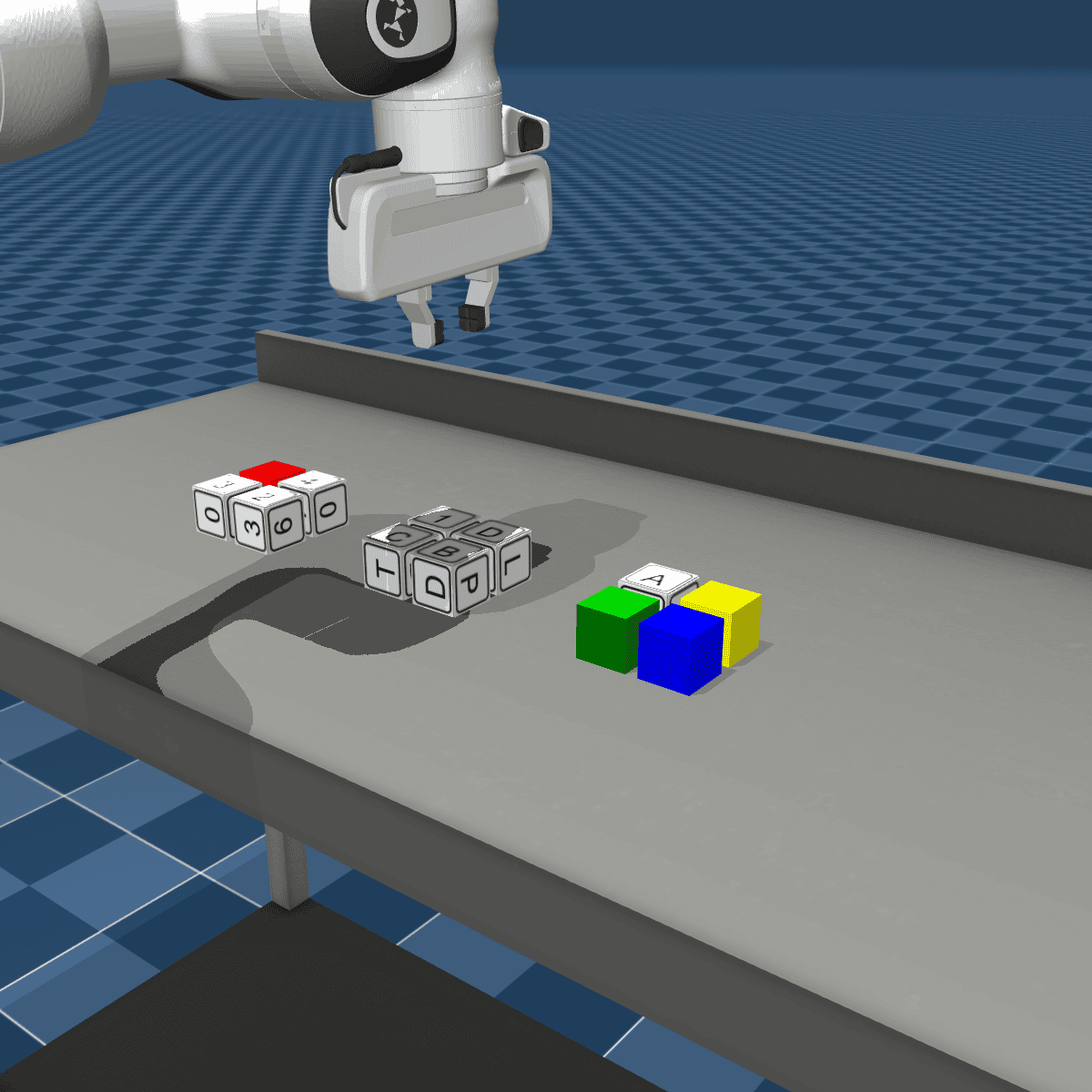}

\caption{\textbf{Pile Sorting}}
\label{fig:benchmark_add_5_pile_sorting}
\end{figure*}

% ────────────────────────────────────────────────────────────
% PROGRESSIVE TOWER
% ────────────────────────────────────────────────────────────
\begin{figure*}[htbp]
\centering

% ── Row: Steps 1-2 ──
\benchmarkcard{Base}
  {Place a blue cube on top of a red cube on the table}
  {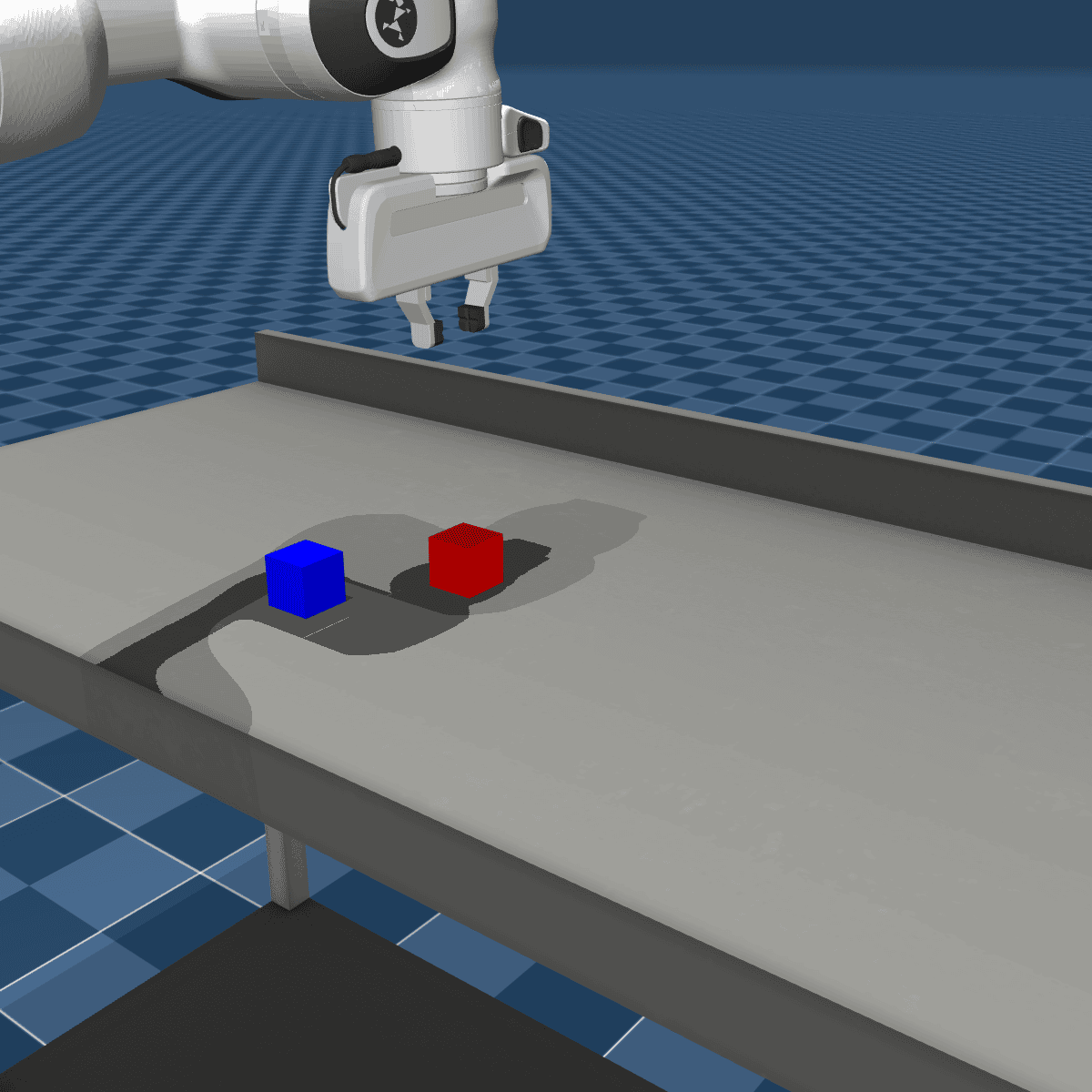}
  {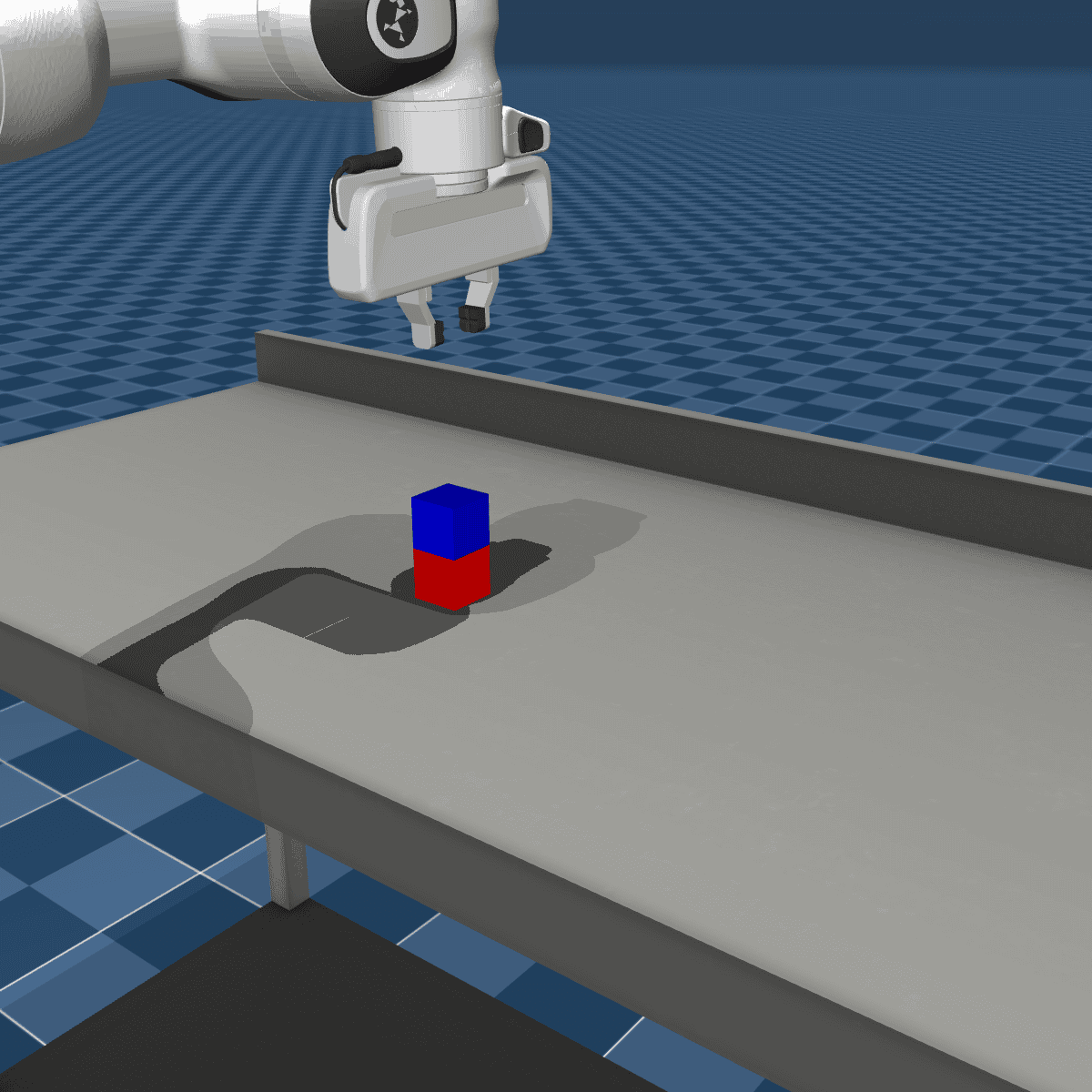}
\hfill
\benchmarkcard{Mod 1}
  {Add a green cube on top of the blue}
  {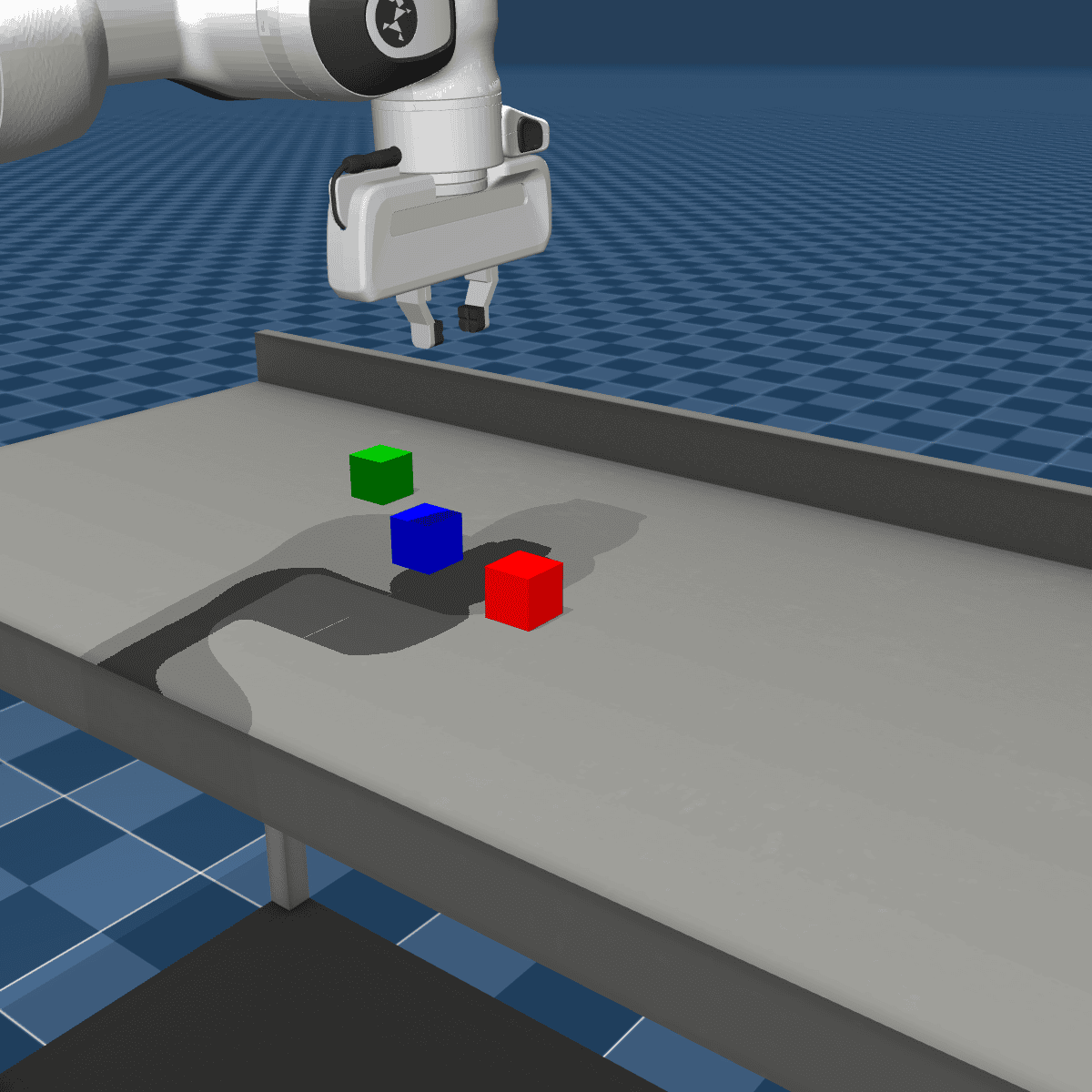}
  {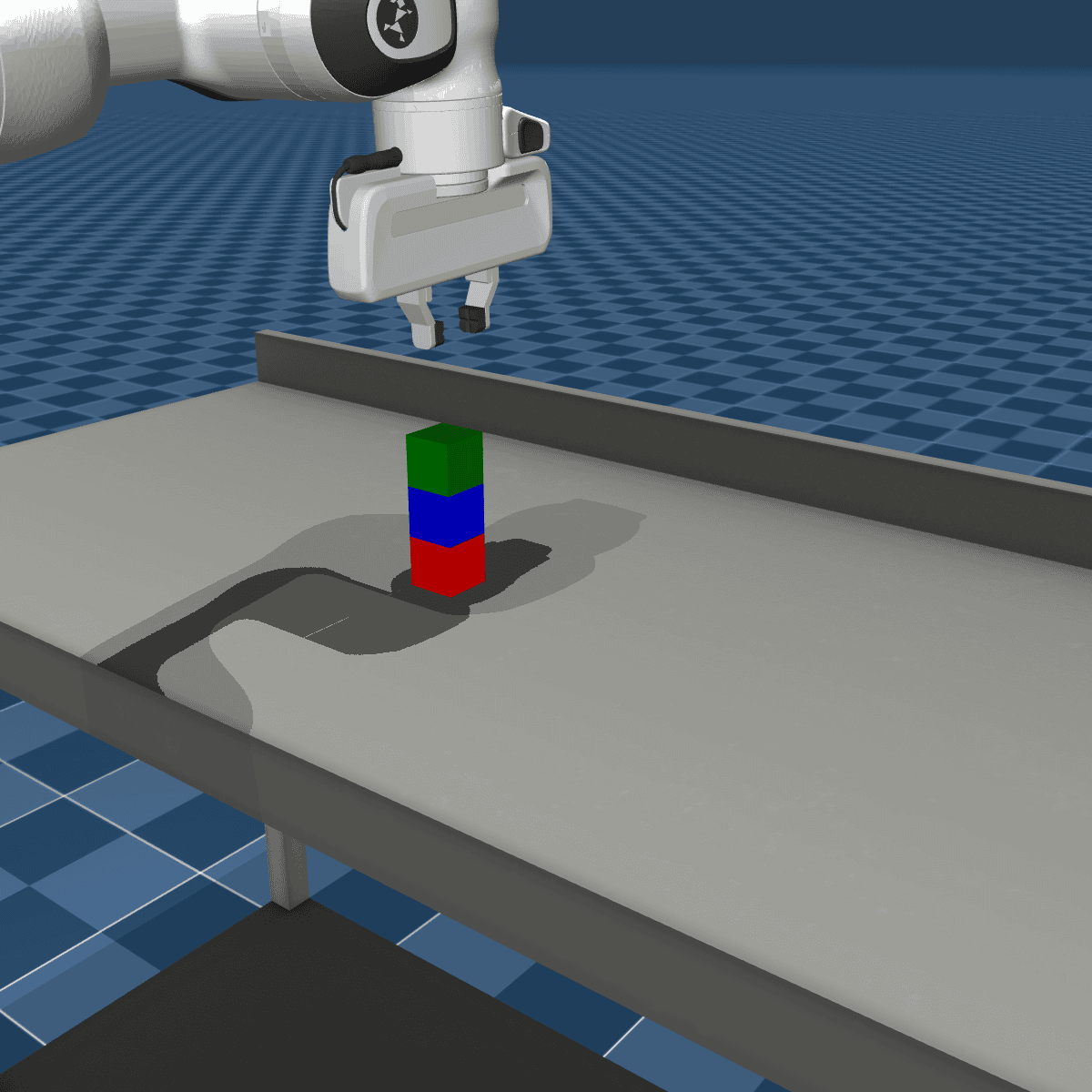}
\vspace{8pt}

% ── Row: Steps 3-4 ──
\benchmarkcard{Mod 2}
  {Replace the green cube with a yellow cube}
  {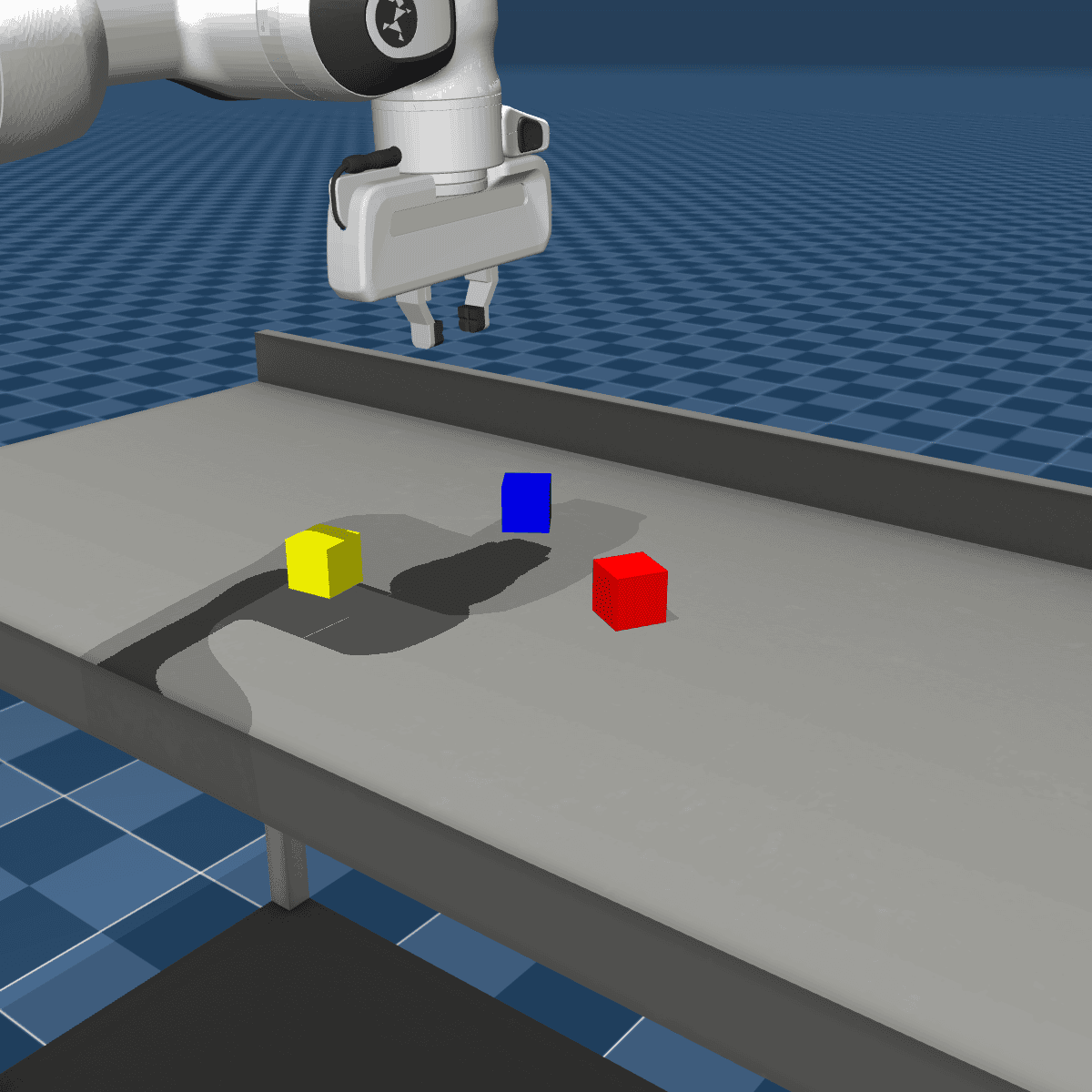}
  {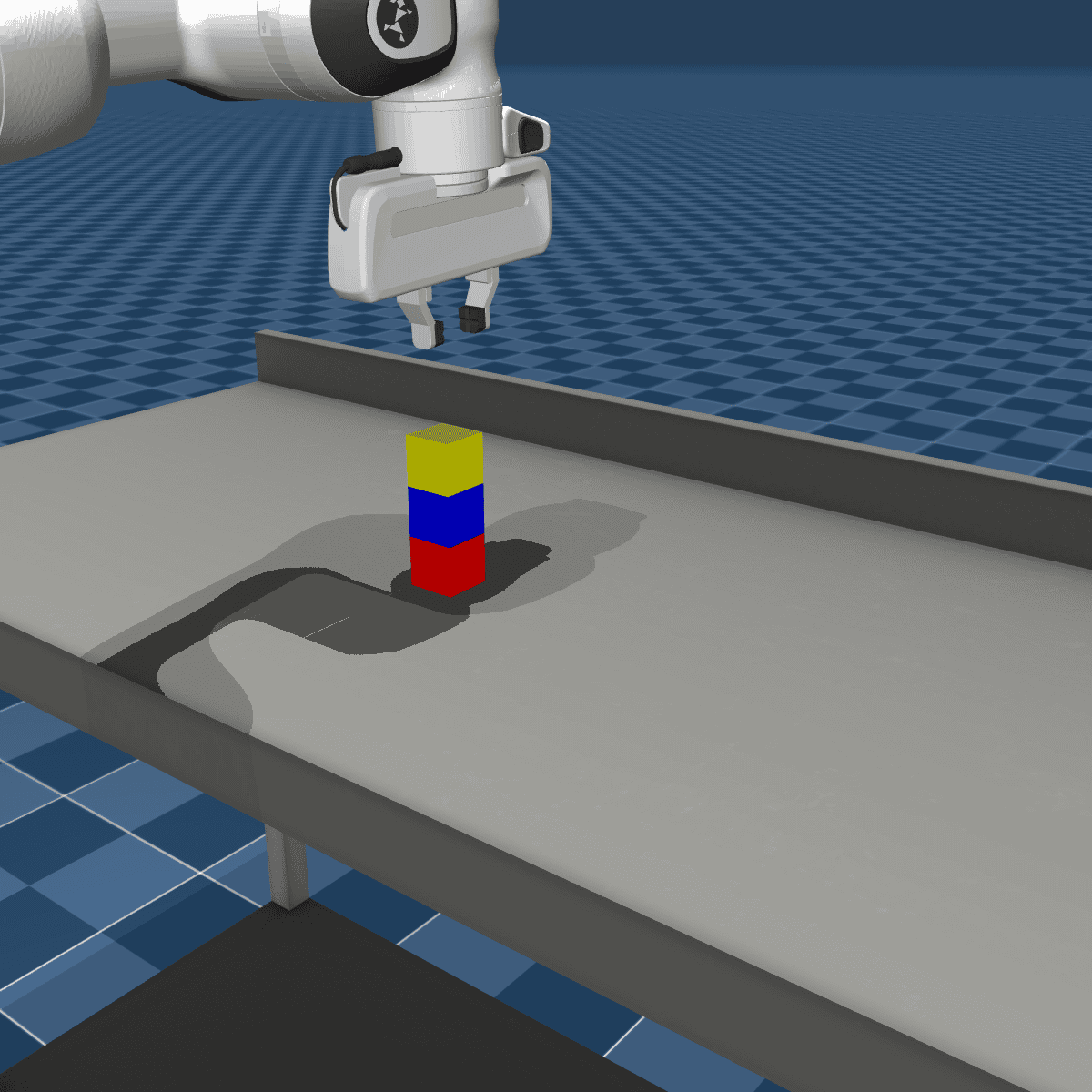}
\hfill
\benchmarkcard{Mod 3}
  {Add letter 'A' semantic cube on top of the tower}
  {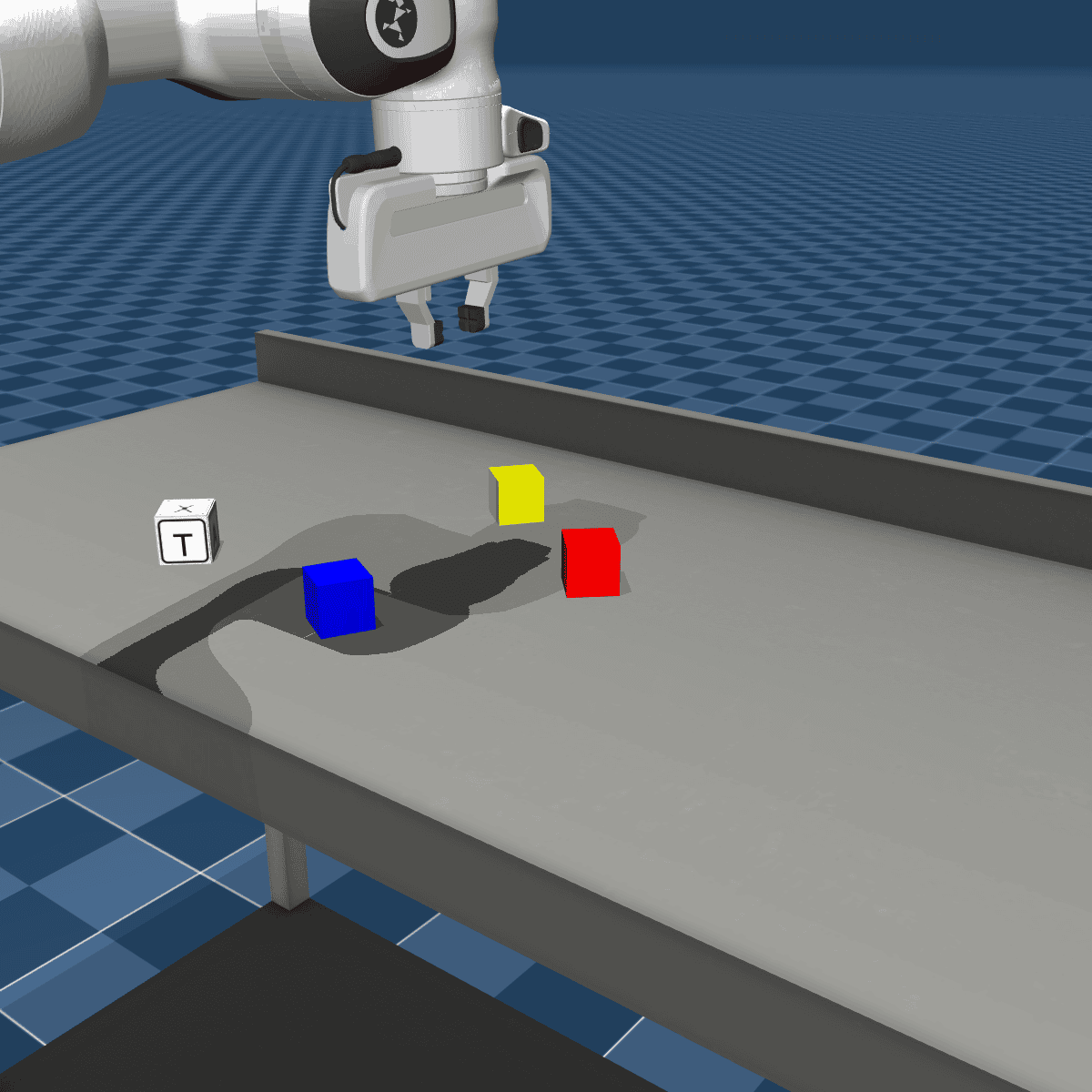}
  {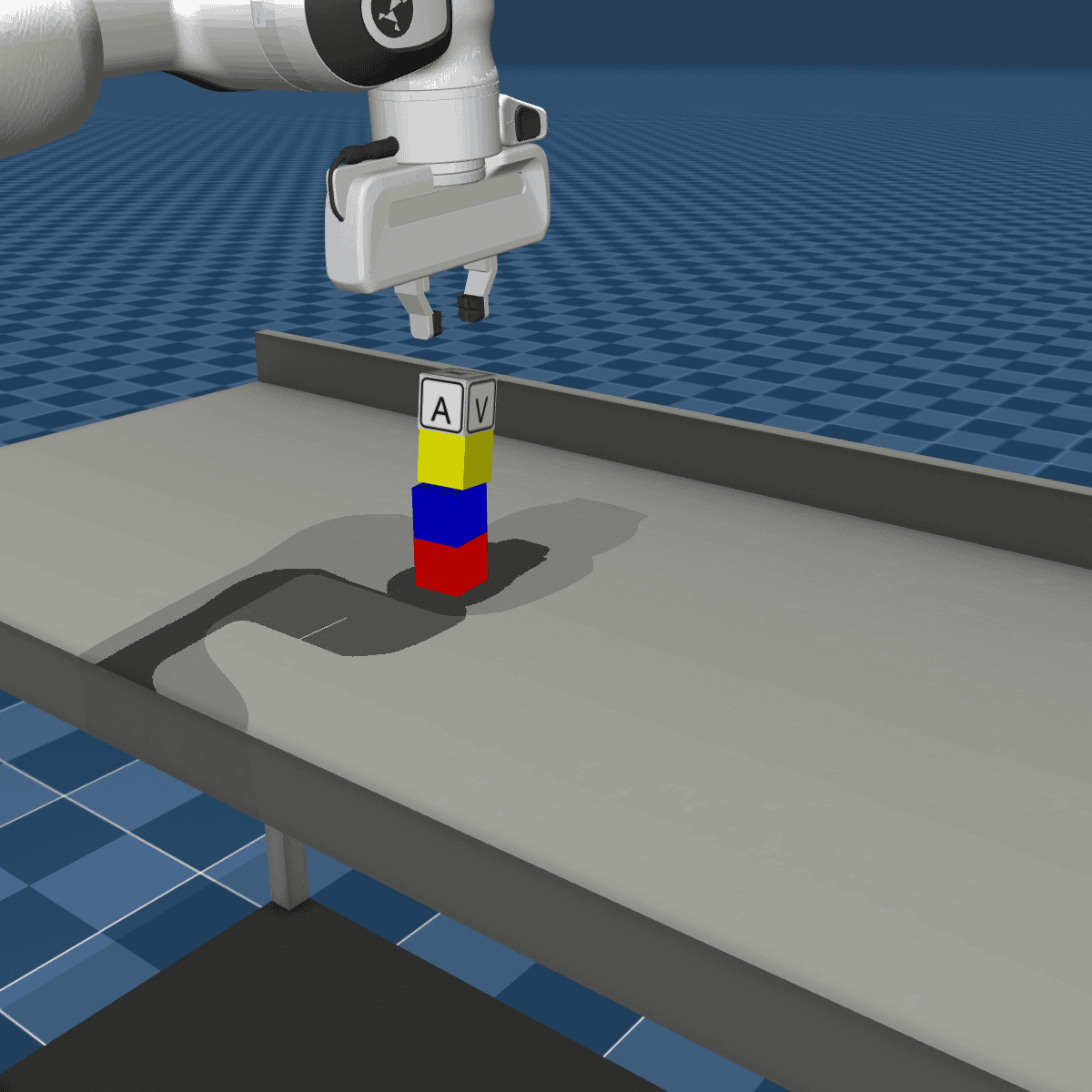}
\vspace{8pt}

% ── Row: Steps 5-6 ──
\benchmarkcard{Mod 4}
  {Flip the color order of the tower but keep the letter 'A' on top}
  {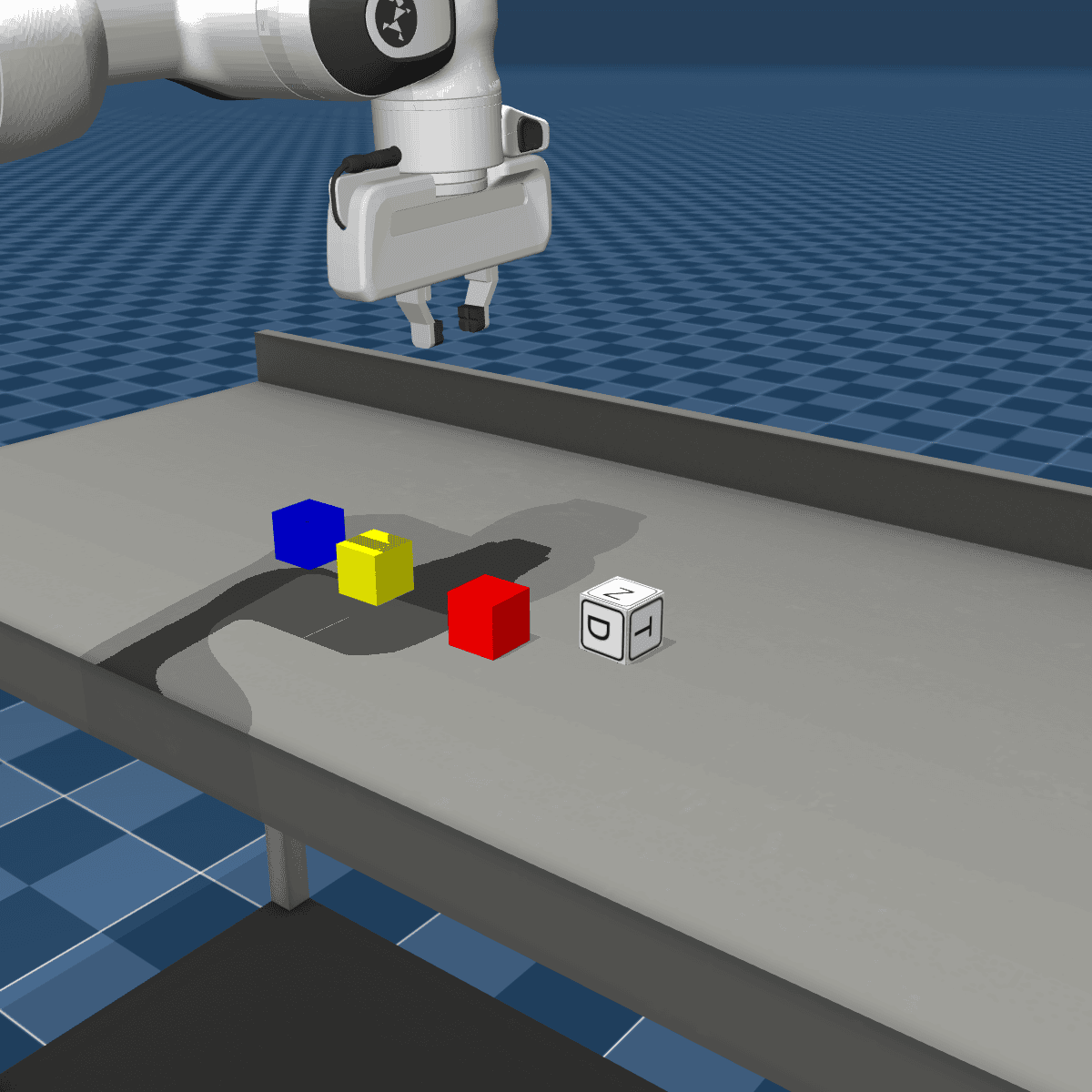}
  {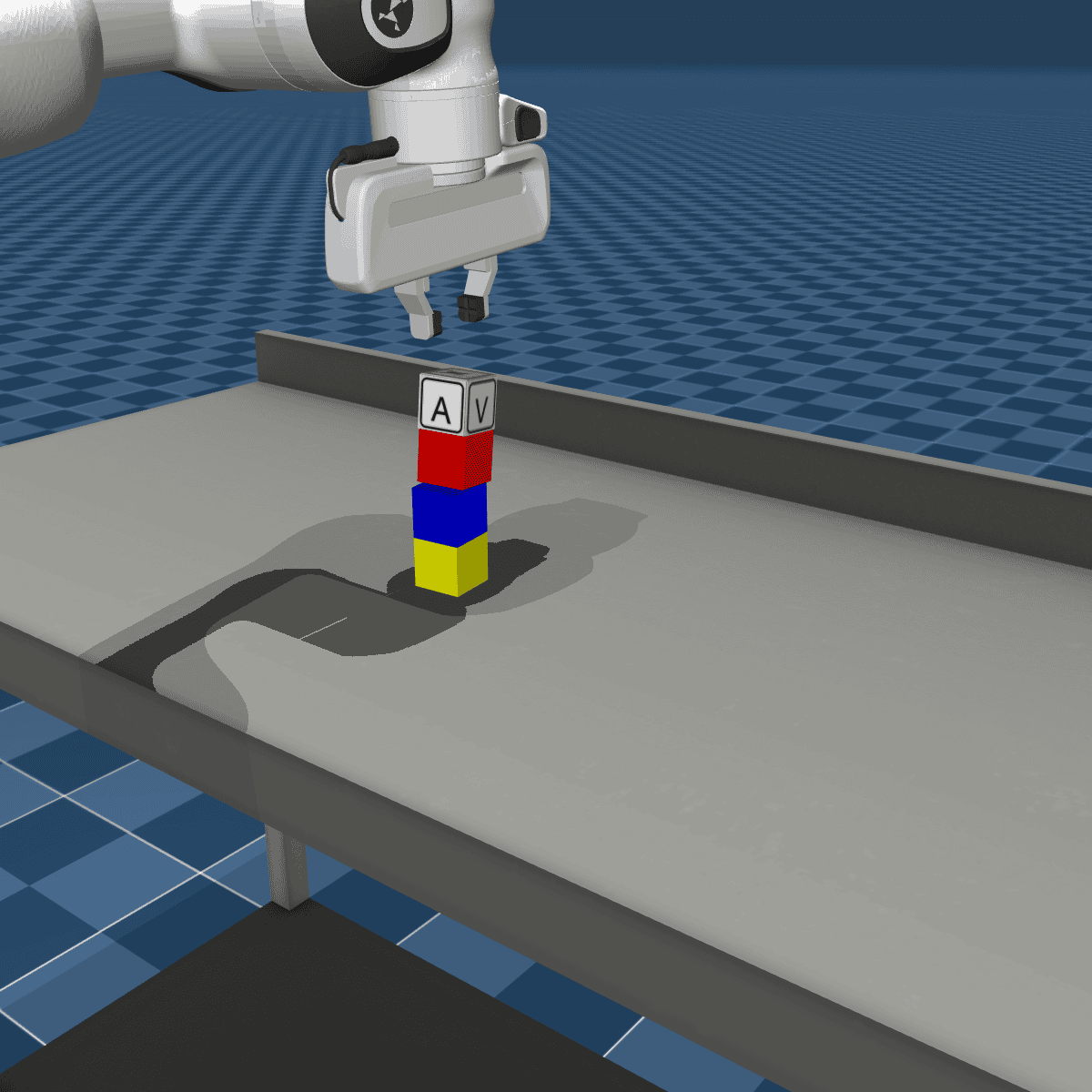}
\hfill
\benchmarkcard{Mod 5}
  {Change the base cube to purple}
  {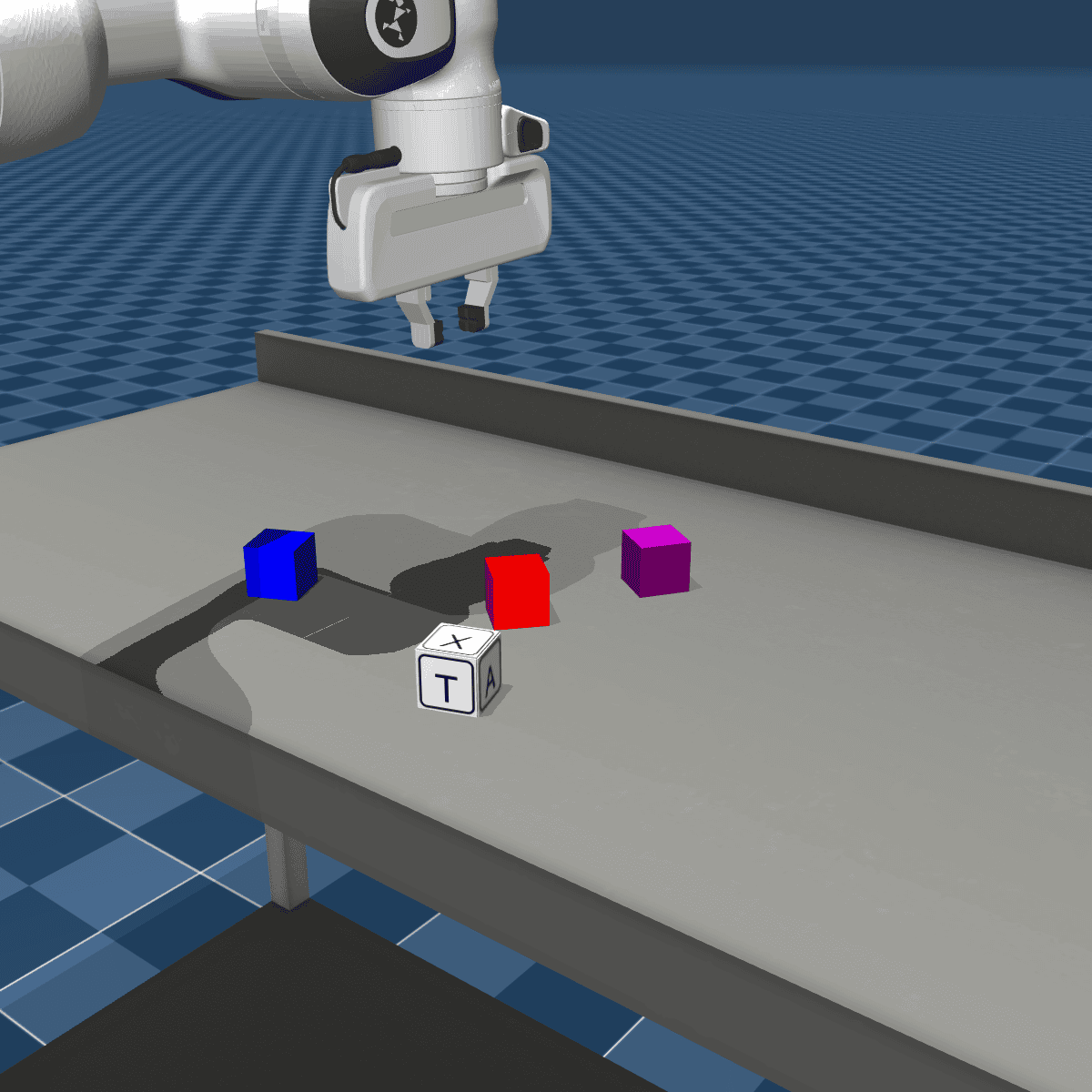}
  {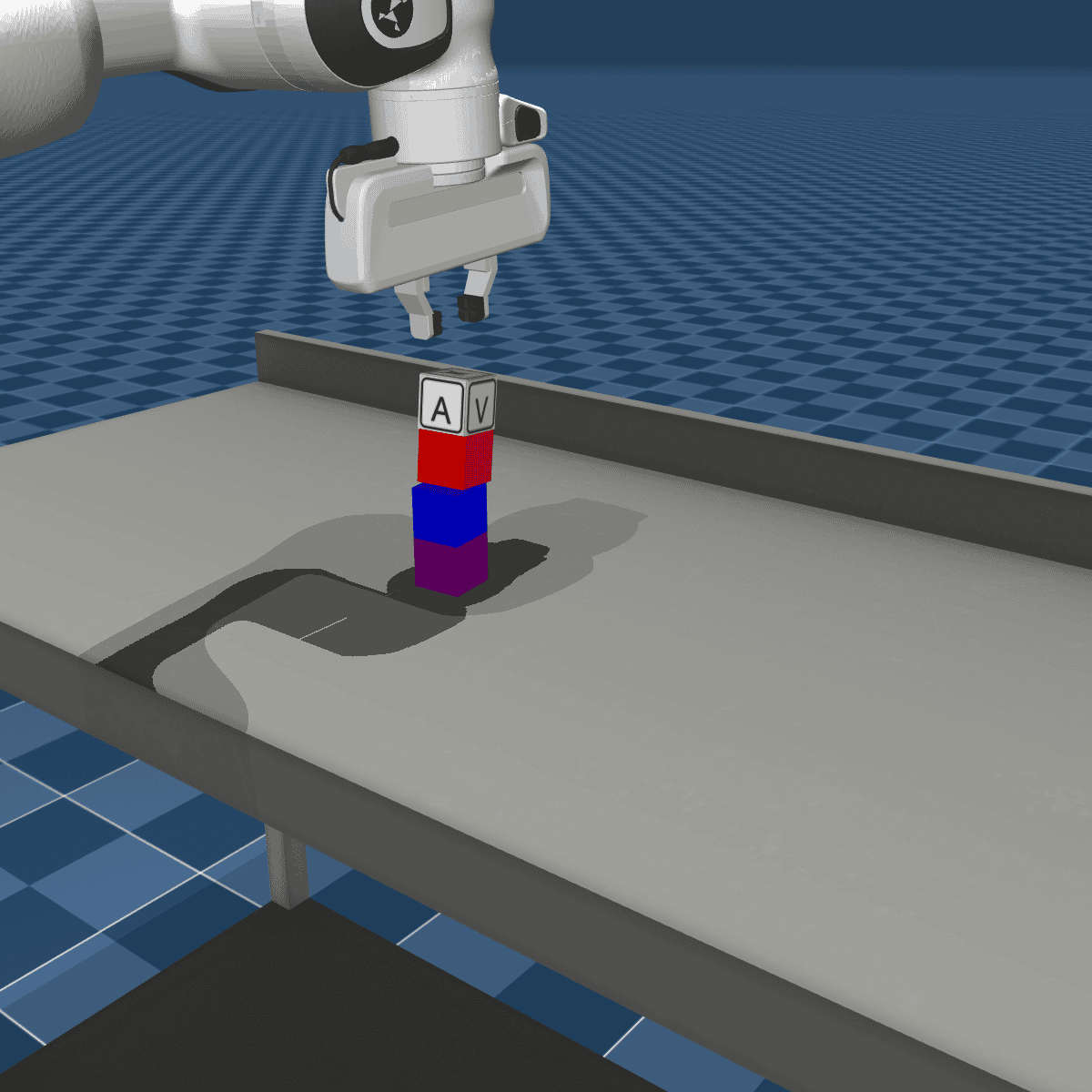}

\caption{\textbf{Progressive Tower} (Tests 5-step progressive modification chain for long-horizon steering). Progression: Base $\rightarrow$ Mod 1 $\rightarrow$ Mod 2 $\rightarrow$ Mod 3 $\rightarrow$ Mod 4 $\rightarrow$ Mod 5. Tests whether the system can follow a long chain of incremental modifications via adding, replacing, reordering, and mixing block types without losing track of the evolving structure. Evaluates long-horizon steering.}
\label{fig:benchmark_add_5_progressive_tower}
\end{figure*}

% ────────────────────────────────────────────────────────────
% PYRAMID CONSTRUCTION
% ────────────────────────────────────────────────────────────
\begin{figure*}[htbp]
\centering

% ── Row: Steps 1-2 ──
\benchmarkcard{Base}
  {Build a pyramid of letter blocks alphabetically with a 3x3 base}
  {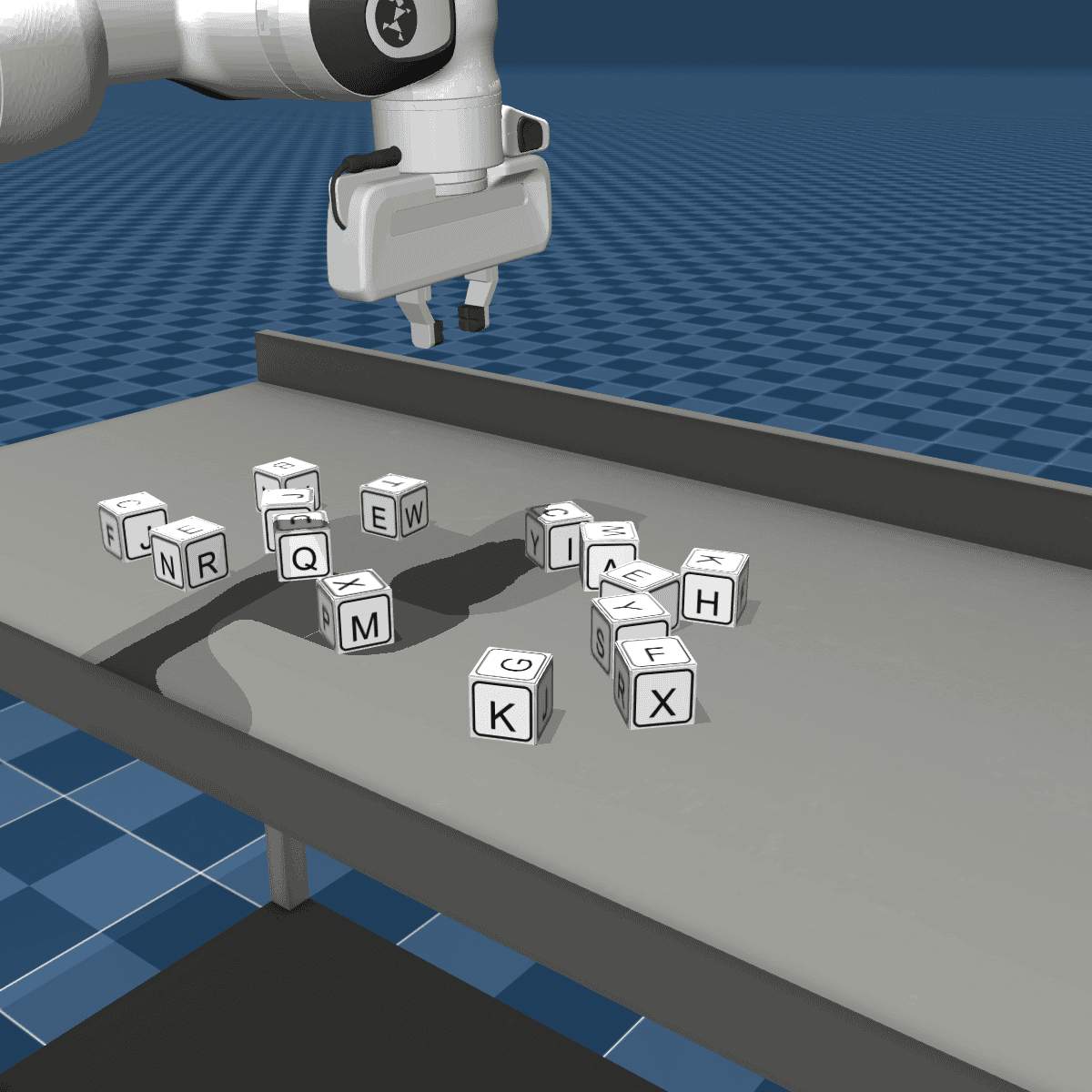}
  {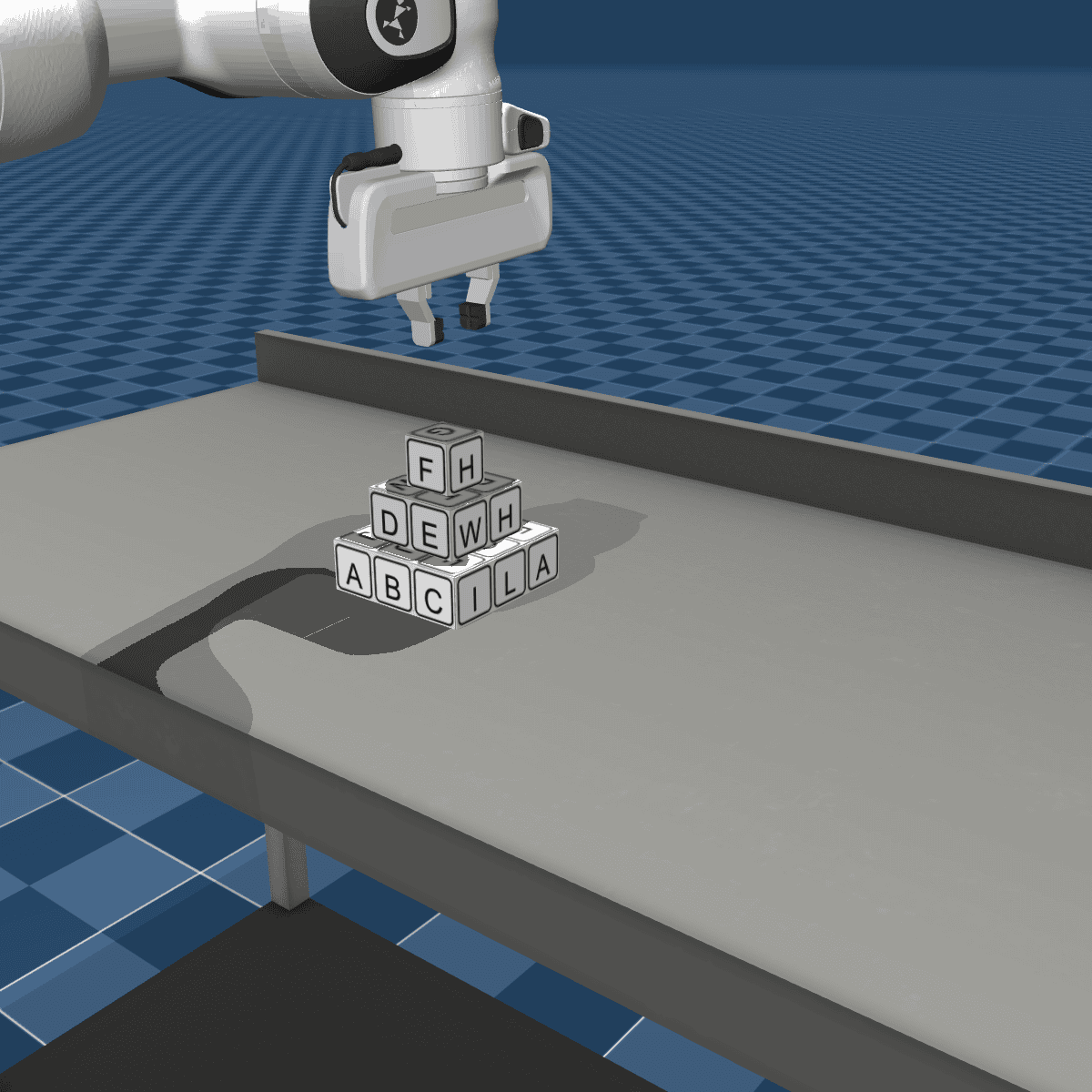}
\hfill
\benchmarkcard{Mod 1}
  {Build a pyramid of blue blocks with a 3x3 base}
  {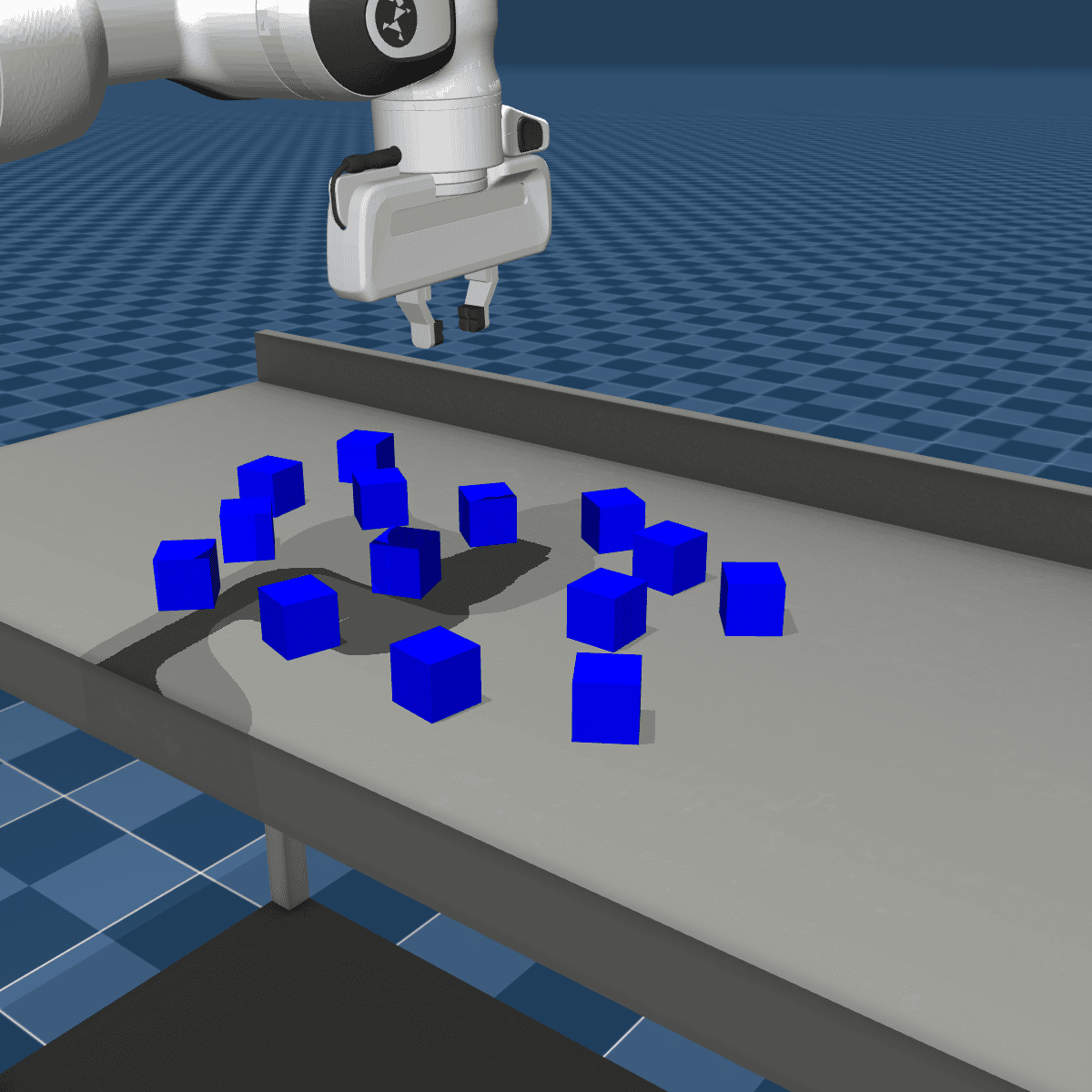}
  {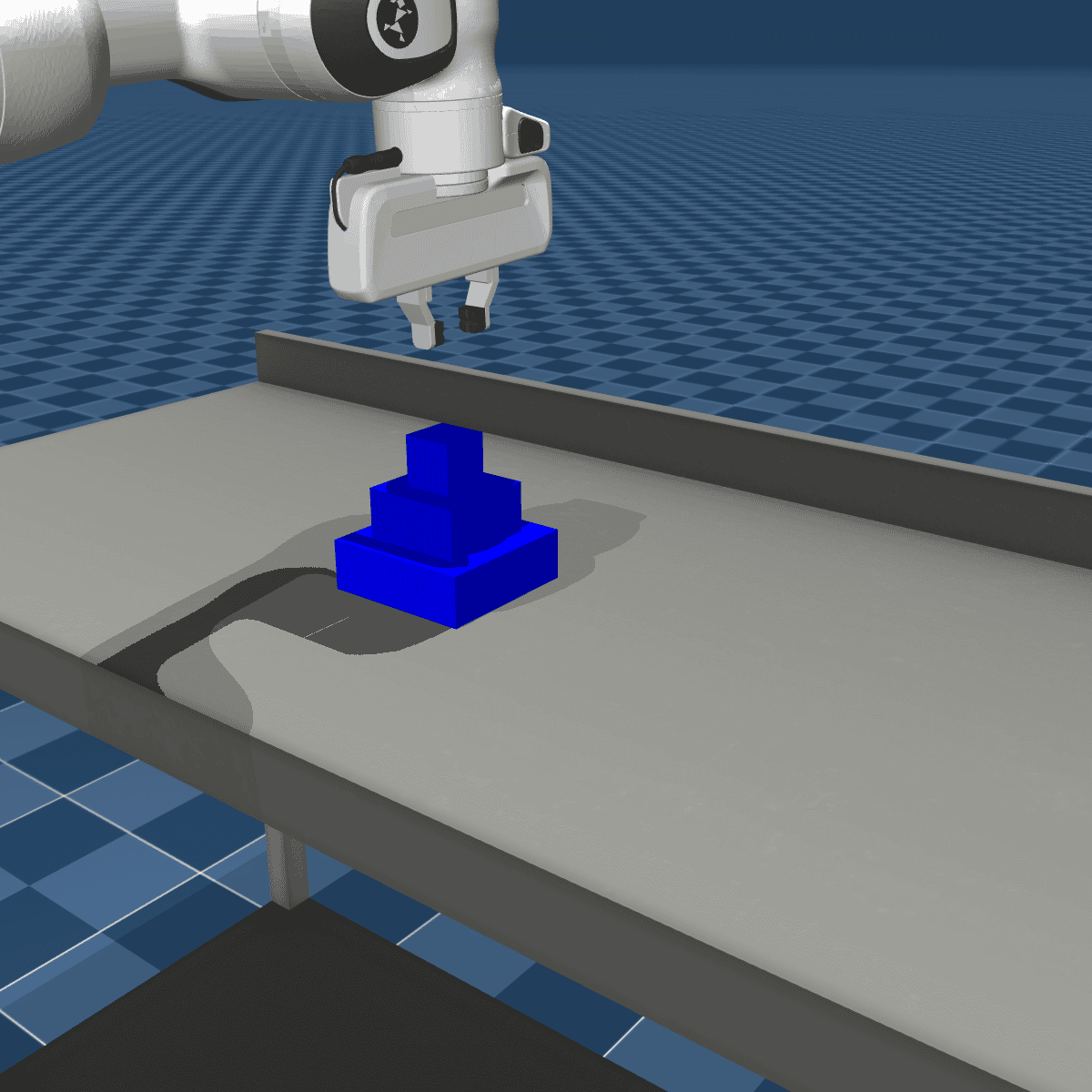}
\vspace{8pt}

% ── Row: Steps 3-4 ──
\benchmarkcard{Mod 2}
  {Now sort them numerically}
  {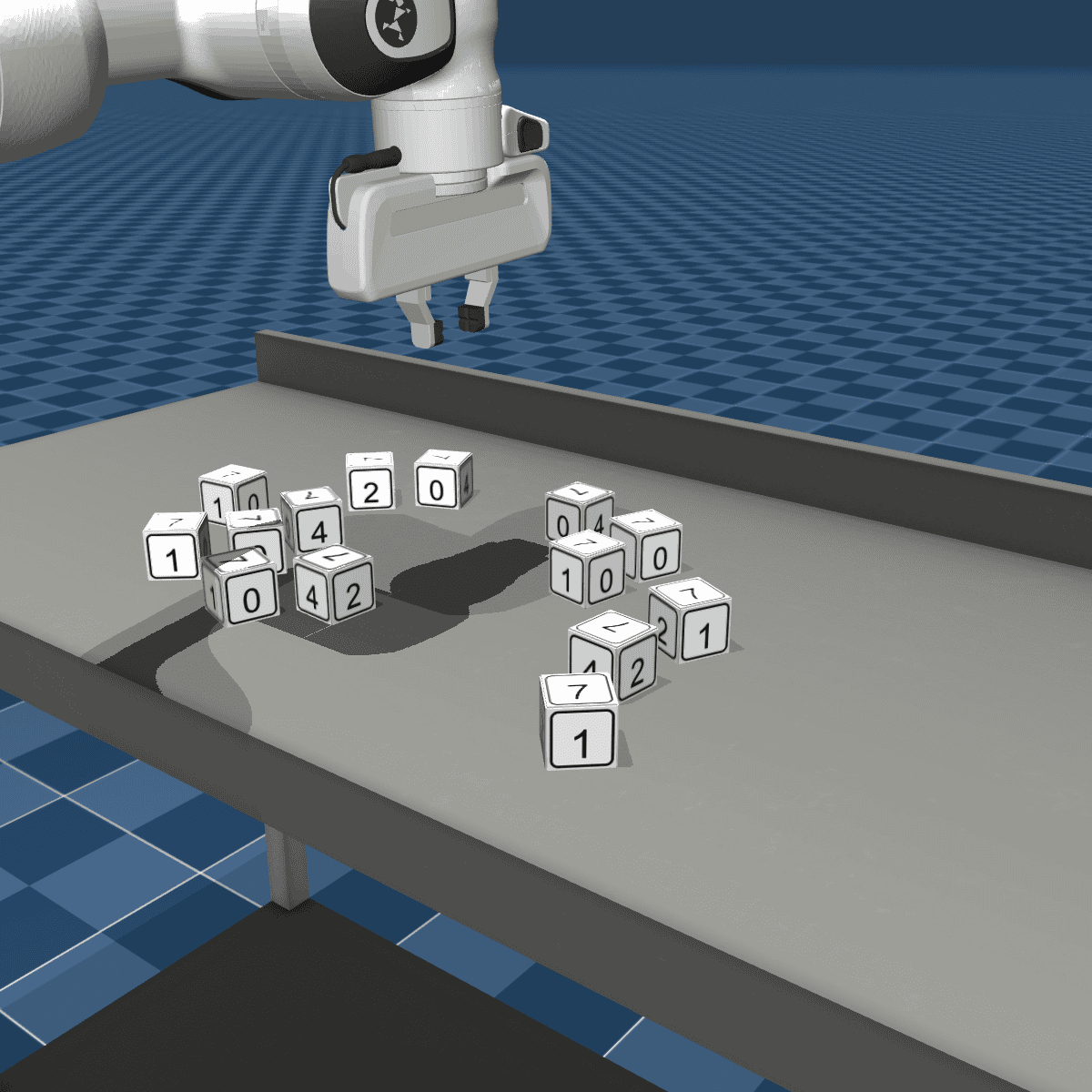}
  {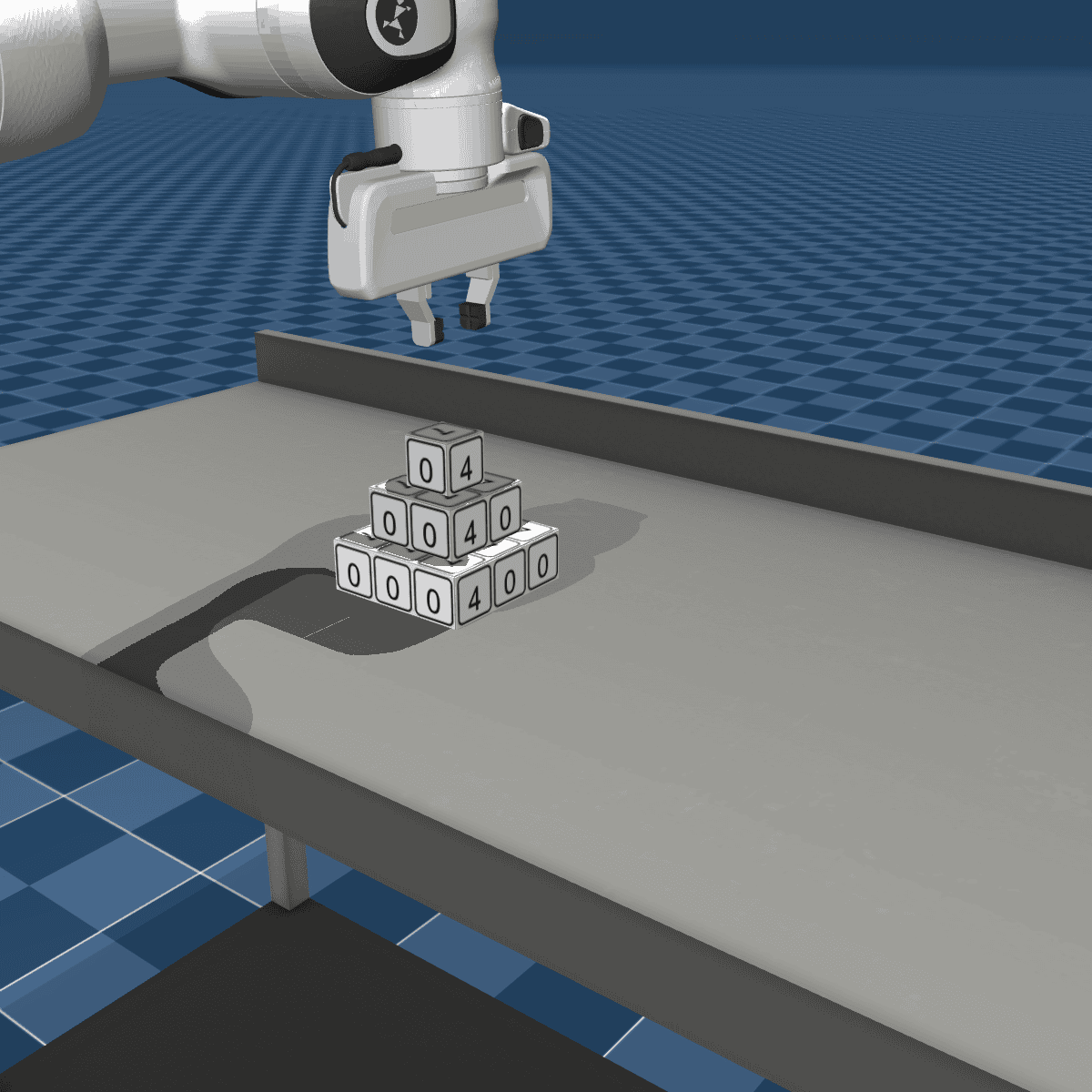}
\hfill
\benchmarkcard{Mod 3}
  {Change back to colored blocks and sort by different colors}
  {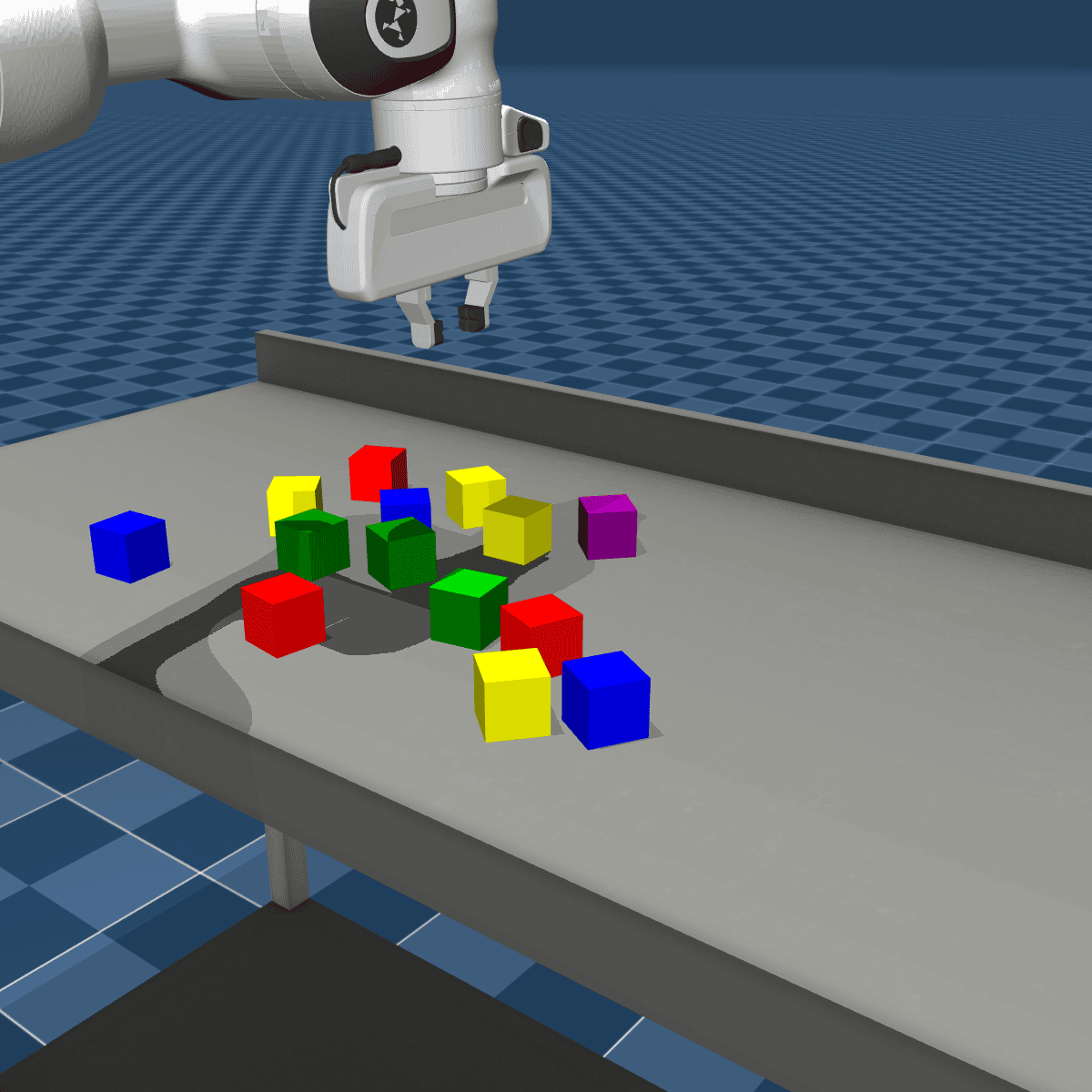}
  {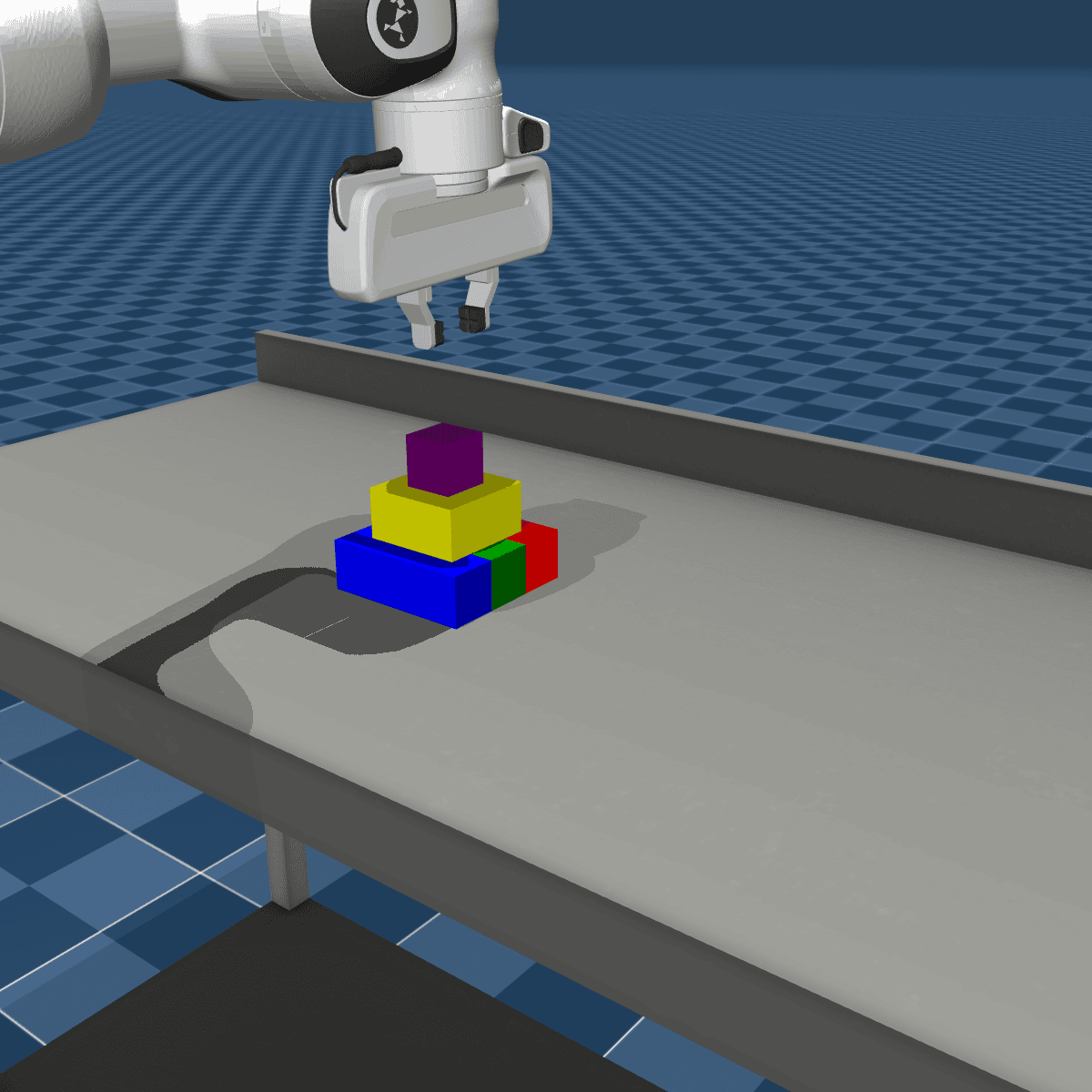}

\caption{\textbf{Pyramid Construction} (Construct a pyramid with steering modifications). Progression: Base $\rightarrow$ Mod 1 $\rightarrow$ Mod 2 $\rightarrow$ Mod 3. Tests the system's ability to build a 3D pyramidal structure and then iteratively transform it by changing block types between letters, uniform colors, numbers, and sorted colors. Evaluates structural understanding and the ability to swap semantic content while preserving geometric form.}
\label{fig:benchmark_add_5_pyramid_construction}
\end{figure*}

% ────────────────────────────────────────────────────────────
% SEMANTIC SPELLING
% ────────────────────────────────────────────────────────────
\begin{figure*}[htbp]
\centering

% ── Row: Steps 1-1 ──
\benchmarkcard{Base}
  {Spell the word 'ROBOT' using semantic cubes arranged in a line}
  {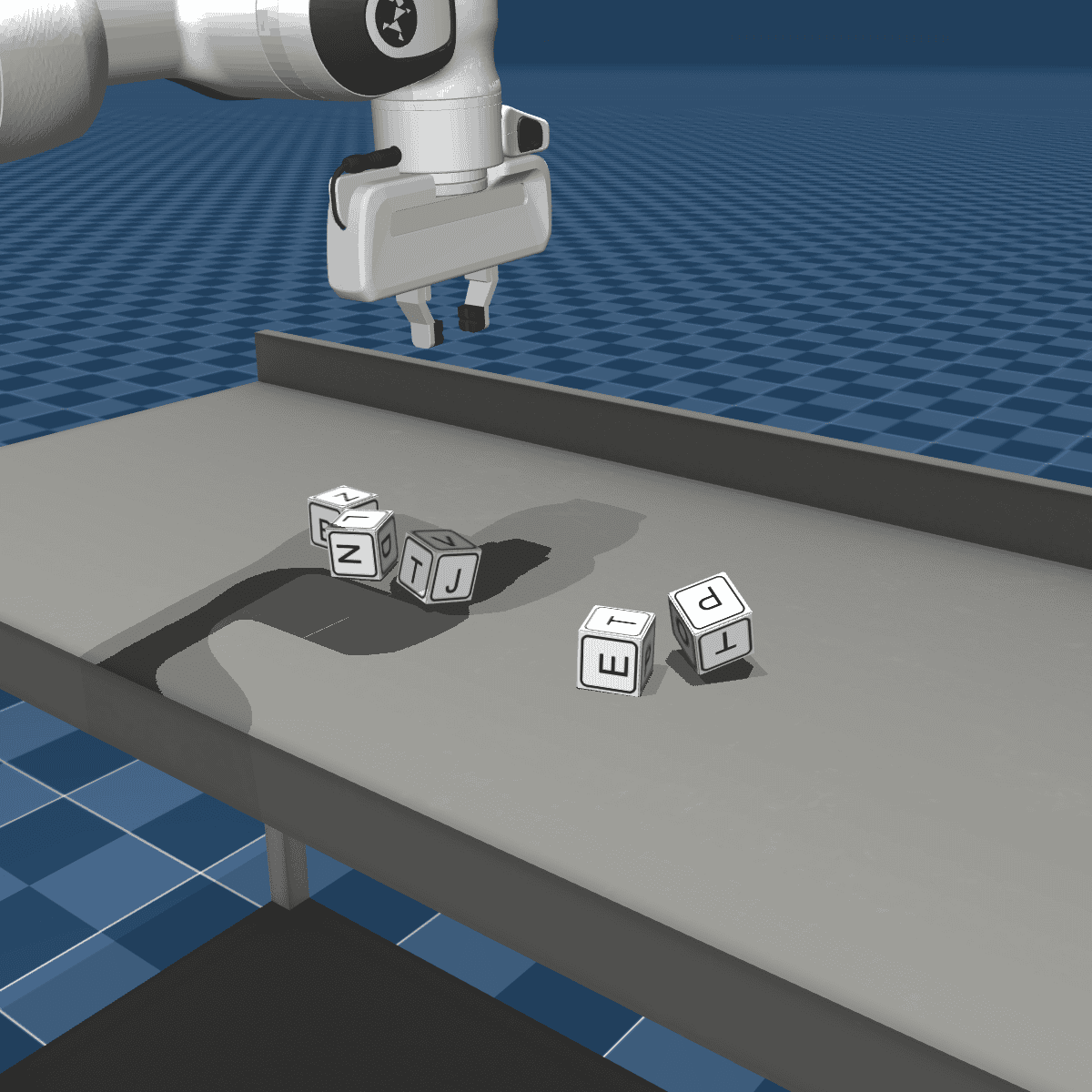}
  {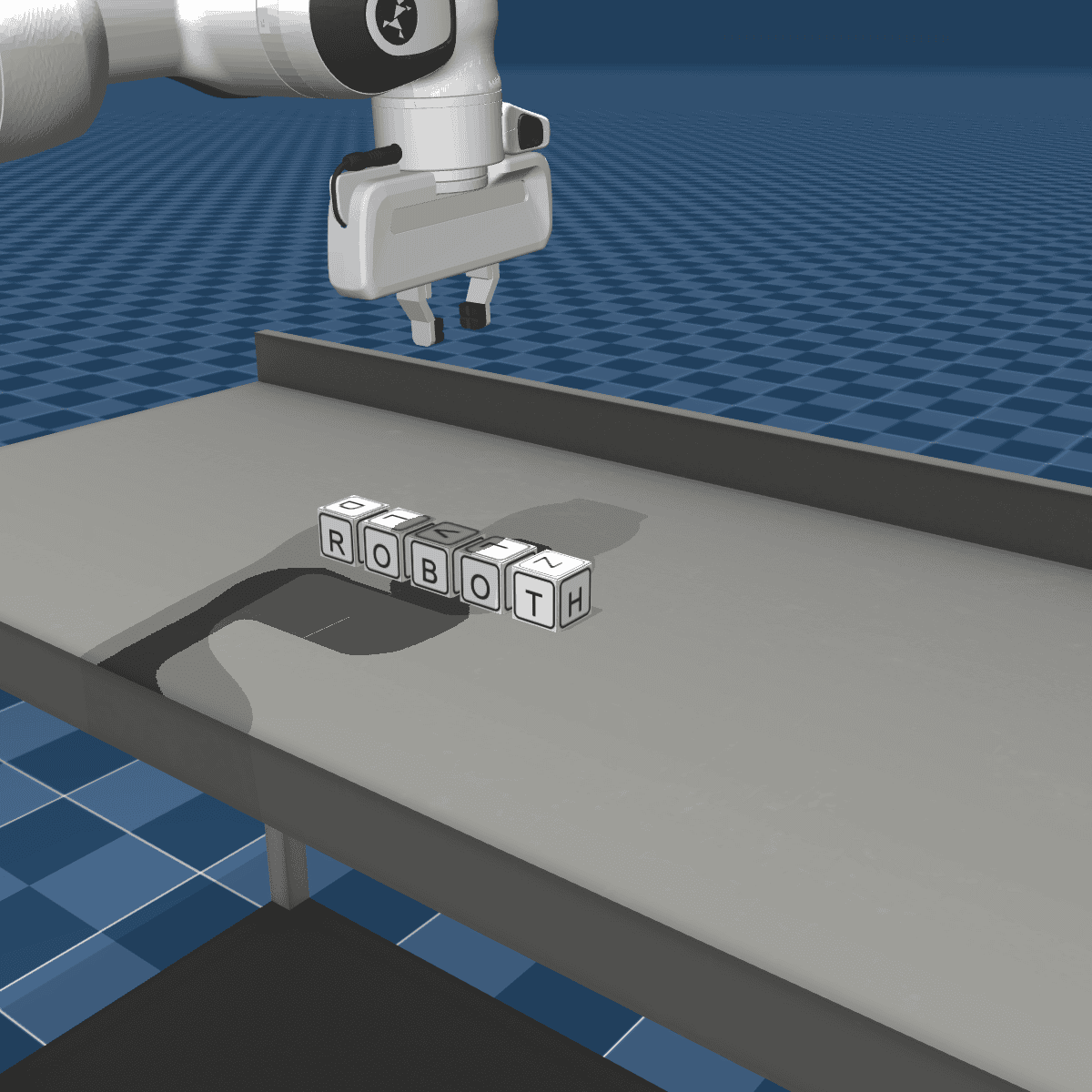}

\caption{\textbf{Semantic Spelling} (Spell a word using semantic cubes). Progression: base prompt only. Tests whether the system can recognize individual letter characters on block faces and arrange them in the correct left-to-right order to spell a word. Evaluates character recognition and sequential semantic reasoning.}
\label{fig:benchmark_add_5_semantic_spelling}
\end{figure*}

% ────────────────────────────────────────────────────────────
% STACK AND MODIFY
% ────────────────────────────────────────────────────────────
\begin{figure*}[htbp]
\centering

% ── Row: Steps 1-2 ──
\benchmarkcard{Base}
  {Make a 3x3 square of red blocks}
  {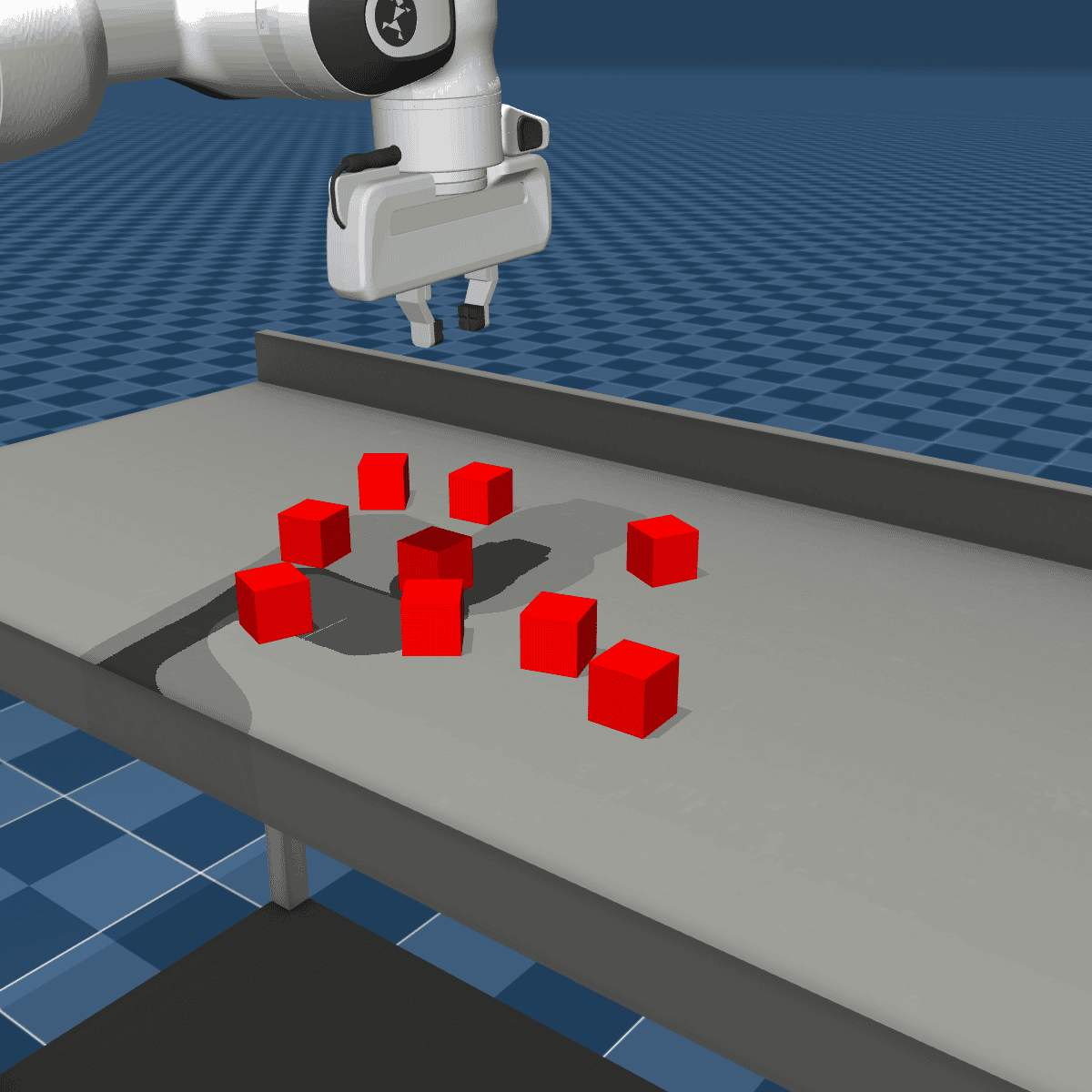}
  {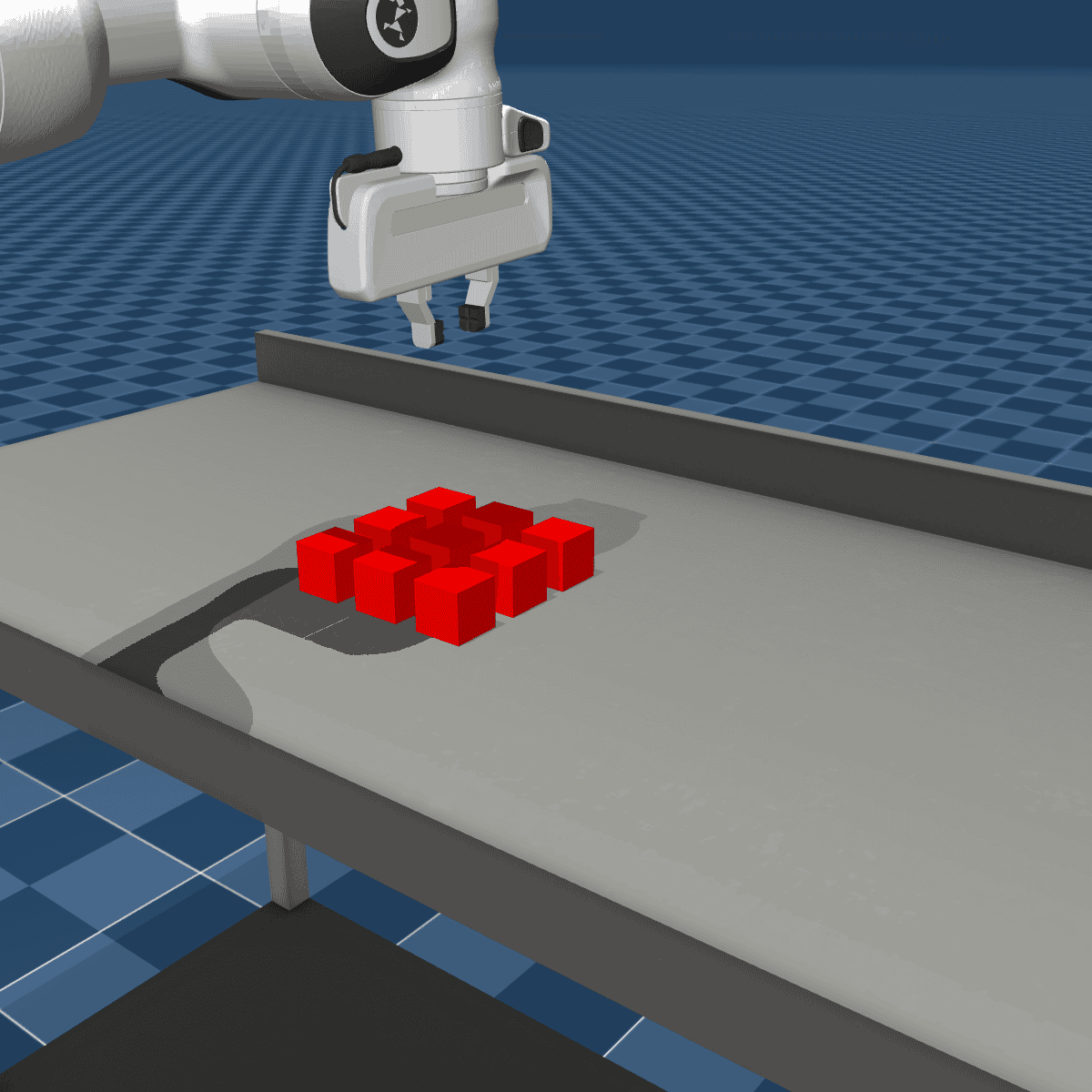}
\hfill
\benchmarkcard{Mod 1}
  {Make a 3x3x3 cube of assorted color blocks where each row is a different color}
  {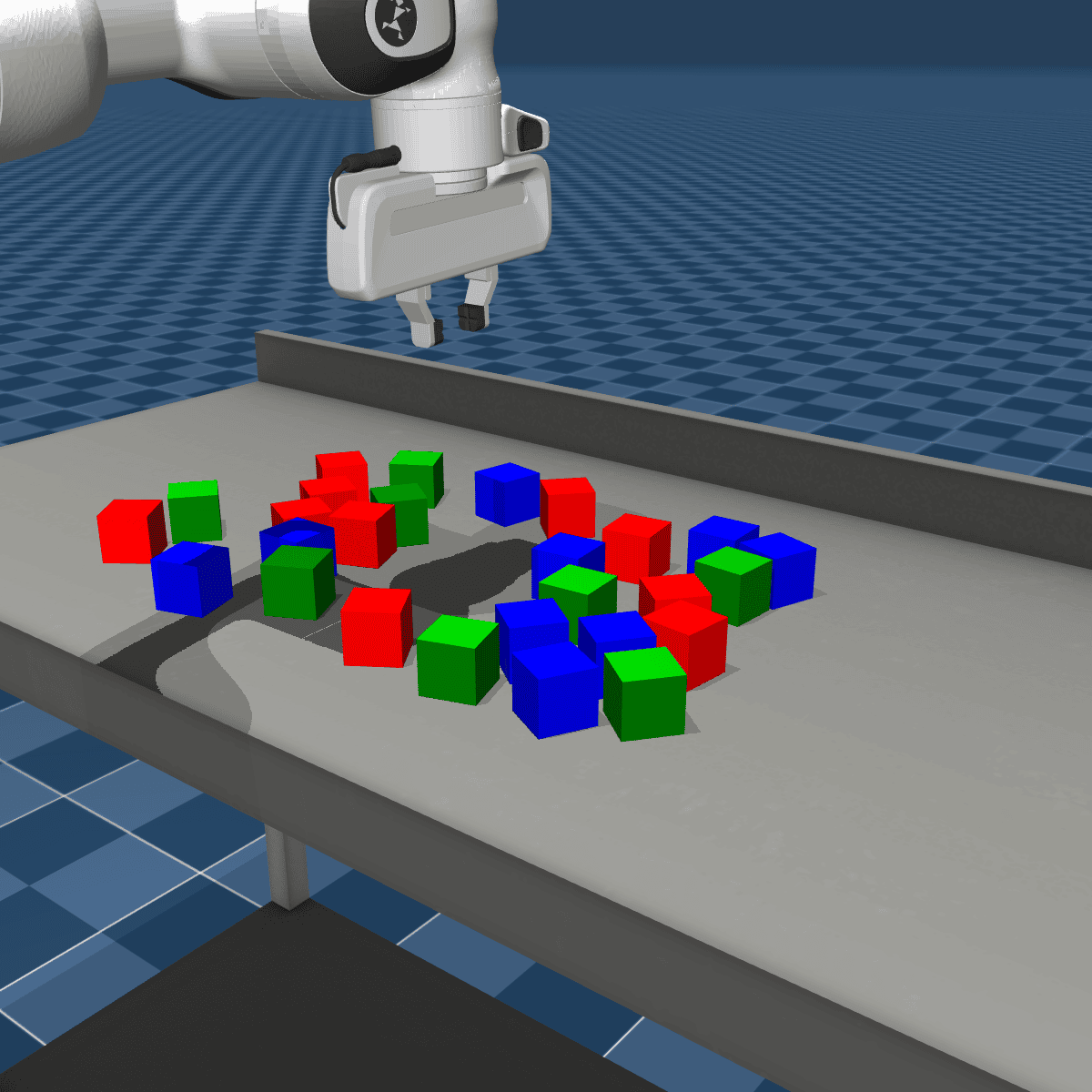}
  {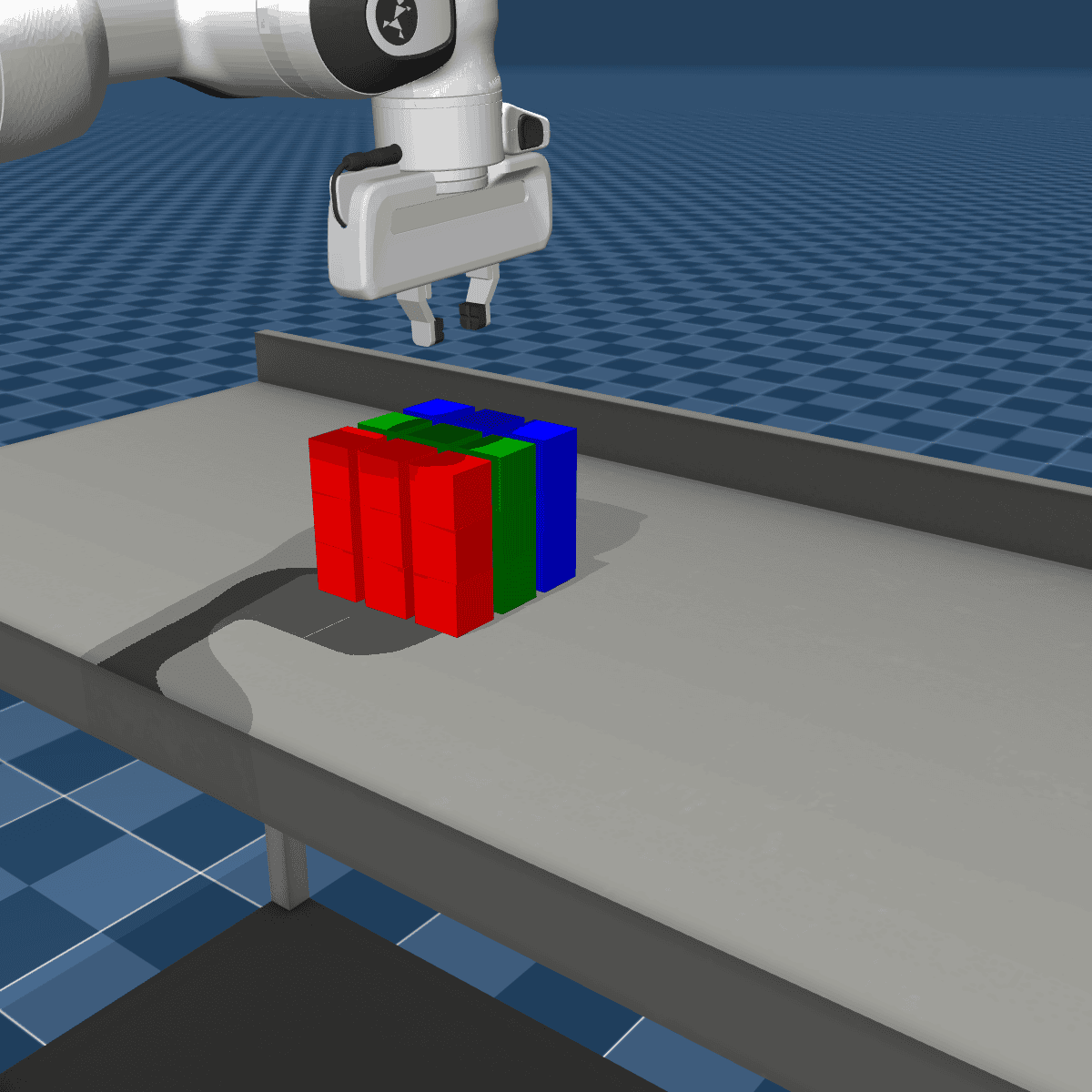}

\caption{\textbf{Stack and Modify} (2D to 3D block structure). Progression: Base $\rightarrow$ Mod 1. Tests the system's ability to transition from a flat 2D arrangement to a full 3D volumetric structure. Evaluates spatial dimensionality understanding.}
\label{fig:benchmark_add_5_stack_and_modify}
\end{figure*}

\clearpage

\end{document}